\documentclass{article} 
\usepackage{arxiv_neutral,times}

\usepackage{amsmath,amsfonts,bm}

\def\eqref#1{equation~\ref{#1}}

\def\1{\bm{1}}

\DeclareMathAlphabet{\mathsfit}{\encodingdefault}{\sfdefault}{m}{sl}
\SetMathAlphabet{\mathsfit}{bold}{\encodingdefault}{\sfdefault}{bx}{n}

\usepackage{hyperref}
\usepackage{url}

\usepackage{graphicx}
\usepackage{xcolor}
\usepackage{booktabs}
\usepackage{arydshln}
\usepackage{multirow}
\usepackage{adjustbox}
\usepackage{array}
\usepackage{tabularx}
\newcolumntype{P}[1]{>{\raggedright\arraybackslash}p{#1}}
\usepackage{float}
\usepackage{wrapfig}
\usepackage{xspace}
\usepackage[section]{placeins}
\usepackage{colortbl}
\usepackage{enumitem}
\usepackage{cleveref}
\usepackage{needspace}
\usepackage{amsmath}
\usepackage{tcolorbox}
\tcbuselibrary{breakable}
\usetikzlibrary{positioning,calc}
\usepackage{fontawesome5}
\usepackage{worldflags}
\usepackage{listings}

\usepackage[utf8]{inputenc}
\usepackage[T1]{fontenc}

\newcommand{\webIcon}{\raisebox{-2pt}{\includegraphics[height=1.15em]{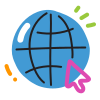}}\xspace}
\newcommand{\huggingface}{\raisebox{-1.5pt}{\includegraphics[height=1.05em]{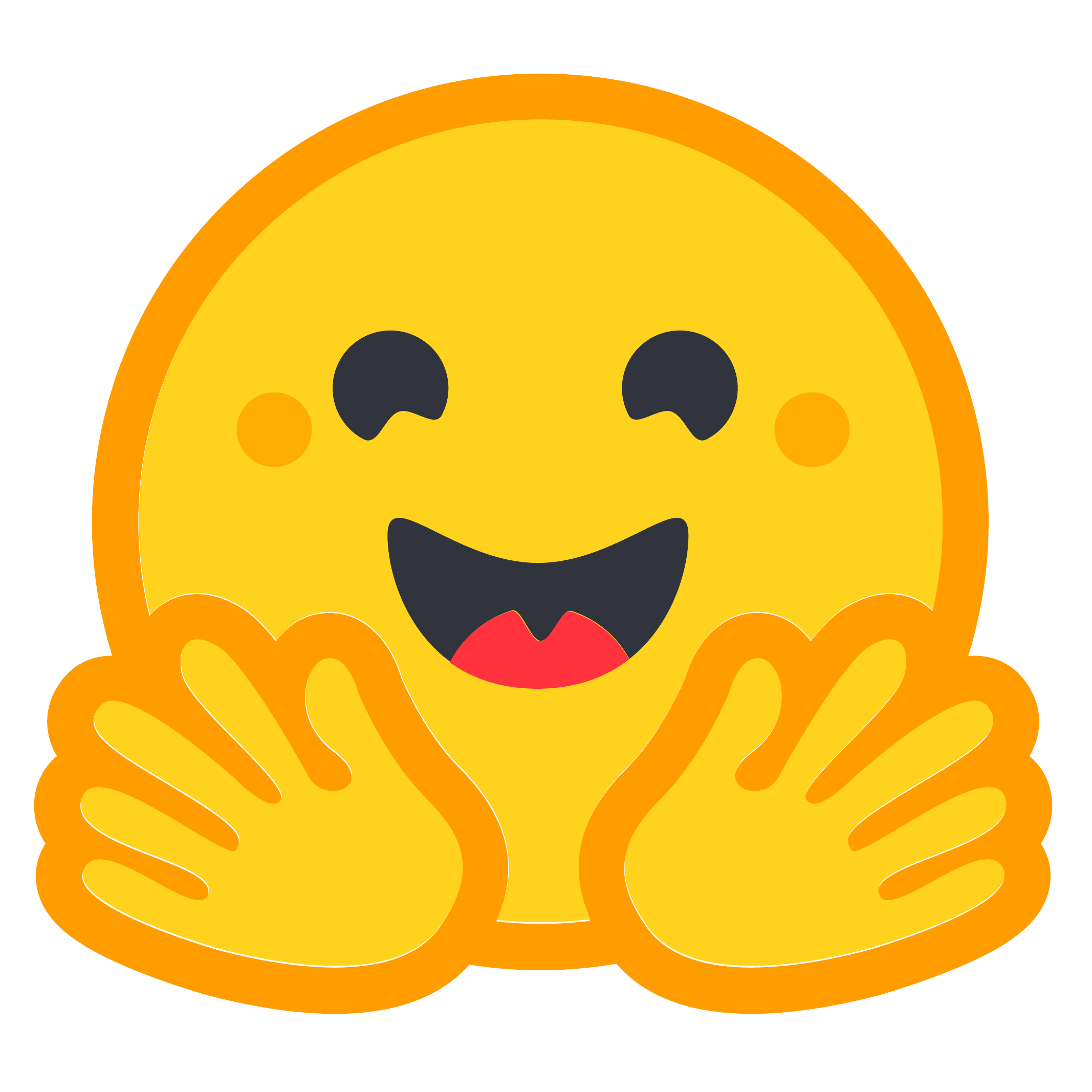}}\xspace}
\newcommand{\github}{\raisebox{-1.5pt}{\includegraphics[height=1.05em]{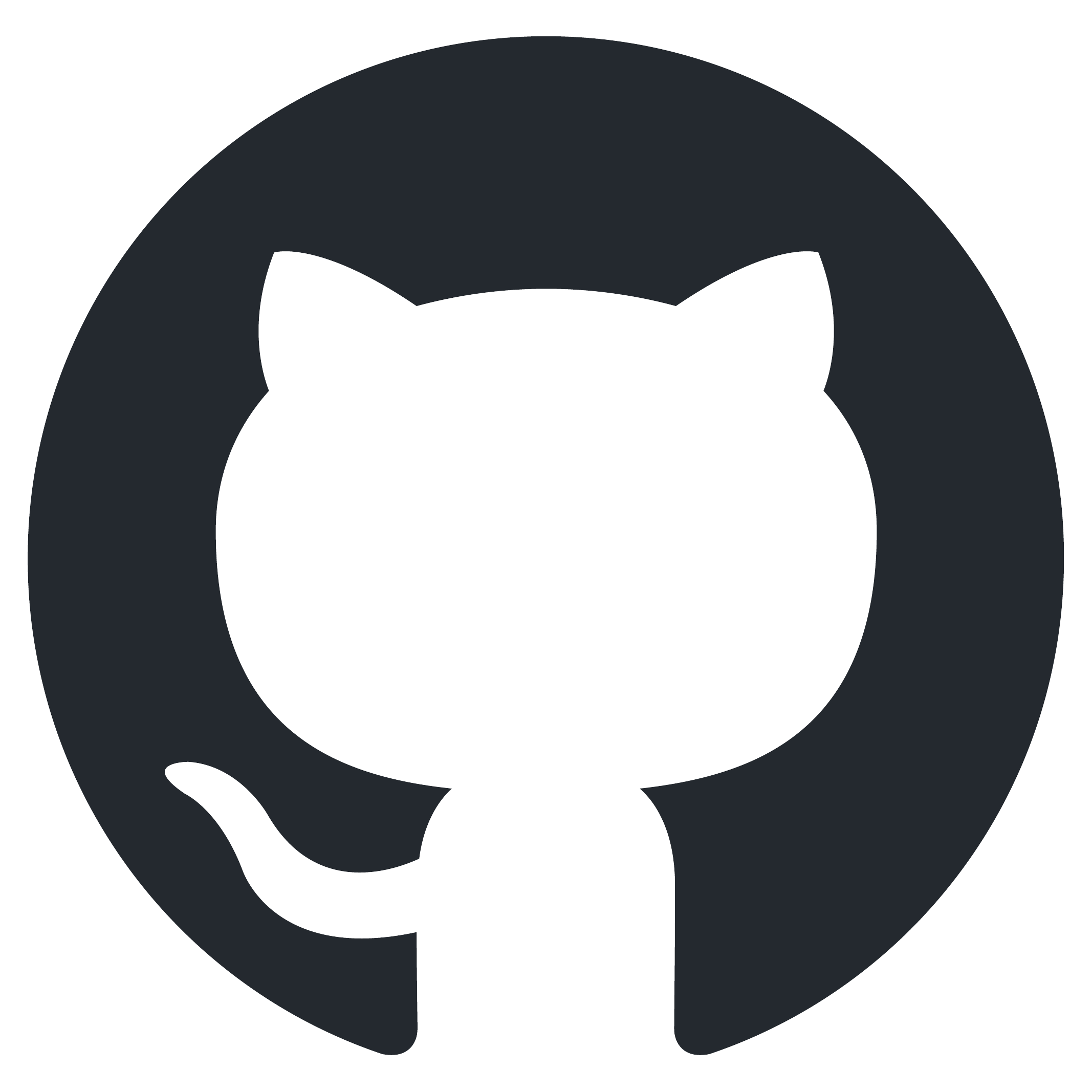}}\xspace}

\definecolor{darkblue}{rgb}{0,0,0.5}
\hypersetup{
    colorlinks=true,
    citecolor=darkblue,
    linkcolor=darkblue,
    urlcolor=darkblue
}

\definecolor{cineNavy}{RGB}{24,45,78}      
\definecolor{cineTeal}{RGB}{18,112,118}    
\definecolor{cineBronze}{RGB}{145,93,24}   
\definecolor{cineGray}{RGB}{92,102,116}    

\definecolor{cineGold}{RGB}{191,143,0}
\definecolor{cineSilver}{RGB}{110,118,130}
\definecolor{cineNarrative}{RGB}{66,104,161}
\definecolor{cineCultural}{RGB}{183,119,46}
\definecolor{cineComposite}{RGB}{104,82,150}
\definecolor{cineDeltaPositive}{RGB}{25,120,70}
\definecolor{cineDeltaNegative}{RGB}{174,55,55}

\newcommand{\cinesubbench}{\textcolor{cineNavy}{Cine}\textcolor{cineTeal}{Sub}\textcolor{cineBronze}{Bench}}
\newcommand{\ENtag}{\textcolor{cineNavy}{\faFlagUsa\,EN}}
\newcommand{\CLtag}{\textcolor{cineTeal}{\faGlobeAmericas\,CL}}
\newcommand{\caseInvent}[1]{\textcolor{cineDeltaNegative}{\bfseries #1}}
\newcommand{\caseResolution}[1]{\textcolor{cineBronze}{\bfseries #1}}
\newcommand{\caseTheme}[1]{\textcolor{cineComposite}{\bfseries #1}}
\newcommand{\caseFlag}[1]{\raisebox{-0.8mm}{\resizebox{6mm}{4mm}{\worldflag{#1}}}}
\newtcolorbox{promptbox}[1]{
  breakable,
  colback=gray!4,
  colframe=cineNavy,
  coltitle=white,
  title=#1,
  fonttitle=\bfseries,
  boxrule=0.55pt,
  arc=1mm,
  left=1.5mm,
  right=1.5mm,
  top=1mm,
  bottom=1mm
}
\newtcolorbox{keyfindings}[1]{
  colback=cineNavy!3,
  colframe=cineNavy!72,
  coltitle=white,
  title={#1},
  fonttitle=\bfseries,
  boxrule=0.65pt,
  arc=1mm,
  left=1.5mm,
  right=1.5mm,
  top=0.8mm,
  bottom=0.8mm,
  before skip=5pt,
  after skip=7pt
}
\newcounter{researchquestion}
\newcommand{\rqlabel}[1]{\refstepcounter{researchquestion}\label{#1}}
\newcounter{pseudocodealgorithm}
\newenvironment{pseudocodealgorithm}[1]{
  \refstepcounter{pseudocodealgorithm}
  \begin{tcolorbox}[
    breakable,
    colback=cineTeal!3,
    colframe=cineTeal,
    coltitle=white,
    title={Algorithm \thepseudocodealgorithm: #1},
    fonttitle=\bfseries,
    boxrule=0.55pt,
    arc=1mm,
    left=1.8mm,
    right=1.8mm,
    top=1mm,
    bottom=1mm
  ]
}{\end{tcolorbox}}
\title{%
\raisebox{-0.12\height}{%
  \includegraphics[
    width=0.60cm,
    trim=0 0 12 0,
    clip
  ]{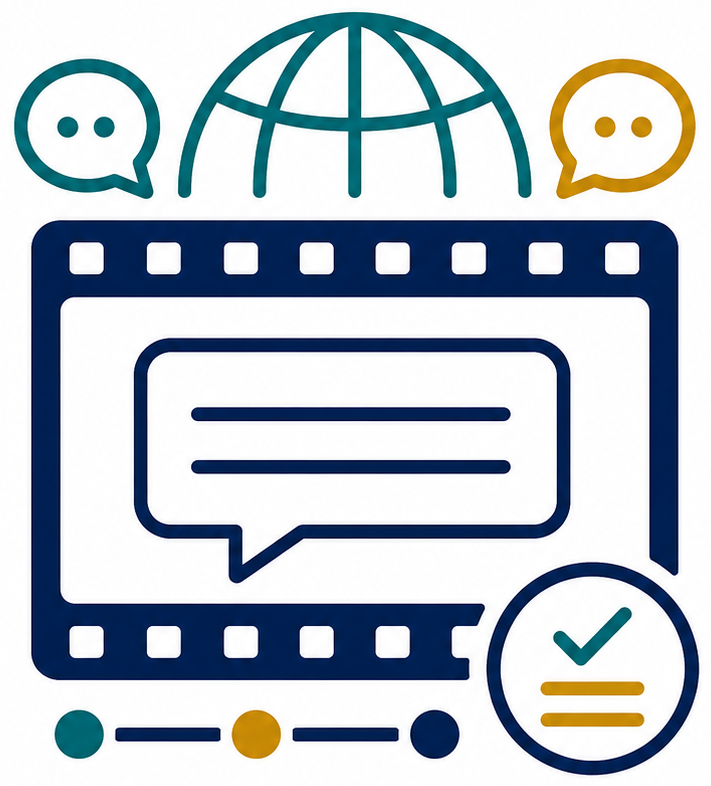}%
}%
\hspace{0.21em}%
\textcolor{cineNavy}{Cine}%
\textcolor{cineTeal}{Sub}%
\textcolor{cineBronze}{Bench}:
Evaluating LLMs on Long-Form Narrative and Cultural Understanding from Multilingual Movie Subtitles
}

\author{
Mir Tafseer Nayeem\textsuperscript{1} \quad
Susmoy Chakraborty\textsuperscript{2} \quad
Davood Rafiei\textsuperscript{1}\\[0.4em]
\textsuperscript{1}University of Alberta \qquad
\textsuperscript{2}Independent Researcher\\[0.3em]
\texttt{\{mnayeem,drafiei\}@ualberta.ca} \qquad
\texttt{susmoy.dip84@gmail.com}
}

\begin{document}

\maketitle


\begin{abstract}
Large language models are increasingly evaluated in specialized domains such
as law, medicine, software engineering, and cybersecurity, yet film remains
comparatively underexplored despite requiring long-form narrative integration,
multilingual interpretation, and culturally situated audience judgments.
We introduce \cinesubbench{}, a benchmark for evaluating long-context film
understanding from multilingual movie subtitles. A subtitle track represents a
film as thousands of short, temporally ordered utterances from which models
must reconstruct characters, relationships, events, causal progression, and
themes without explicit scene or event structure. \cinesubbench{} contains
1,012 films with complete subtitle coverage in six languages, yielding 6,072
tracks and 8.13M timestamped subtitle entries. It provides a matched
\textbf{multi-task, multilingual, and multicultural (MultiX)} evaluation
setting: seven tasks span narrative reconstruction and abstraction, genre
prediction, age suitability, country-specific motion-picture ratings across
ten national classification systems, and subtitle-grounded language safety.
Across nine LLMs, plot premises are recovered more reliably than
event-complete synopses; cross-lingual consistency varies substantially across
models and languages; national rating systems expose distinct calibration
patterns; and strong profanity is far easier to ground than mild obscenity.
\cinesubbench{} establishes film as a long-context LLM evaluation domain and
provides a unified benchmark for measuring narrative, multilingual, cultural,
and evidence-grounding capabilities.\footnote{\webIcon{} \github{} \huggingface{} \textbf{Project website:} \href{https://tafseer-nayeem.github.io/CineSubBench}{\texttt{\path{https://tafseer-nayeem.github.io/CineSubBench}}}}
\end{abstract}


\section{Introduction}
\label{sec:introduction}

Large language models are increasingly evaluated in specialized domains where
general capabilities must transfer to domain-specific forms of evidence,
reasoning, and decision making. Existing benchmarks span legal and clinical
judgment~\citep{guha2023legalbench,singhal2023multimedqa}, finance and
economics~\citep{cao2024financial,guo2024econnli}, software engineering
~\citep{jimenez2024swebench}, scientific and structured-data reasoning
~\citep{zhang2024tablellama,duan2025scigym}, education and journalism \citep{hou2024eeval,li2024newsbench}, and cybersecurity and geospatial
reasoning~\citep{dihan2025mapeval,wang2026cybergym}. Film remains
comparatively underexplored as a domain for LLM evaluation, despite combining
several demanding capabilities: long-form narrative understanding, social
interaction, implicit meaning, multilingual interpretation, and culturally
situated audience judgments.

Film also presents a long-context language problem. Much existing
long-context evaluation centers on articles, papers, books, retrieval
collections, or other documents in which discourse or narrative structure is
largely encoded in the text
~\citep{kryscinski2022booksum,liu2024lost,goldman2024longcontext,
kim2024fables,hamilton2026narrabench}. A film unfolds differently. Its story
develops over hours through dialogue, interactions among characters, and
temporally separated events. Relationships evolve, motivations may remain
implicit, and the significance of an early exchange can become clear only much
later. Film understanding therefore requires a model to integrate evidence
distributed throughout a long context and reconstruct the structure of an
unfolding narrative.

\begin{figure*}[t]
\centering
\includegraphics[width=0.98\textwidth]{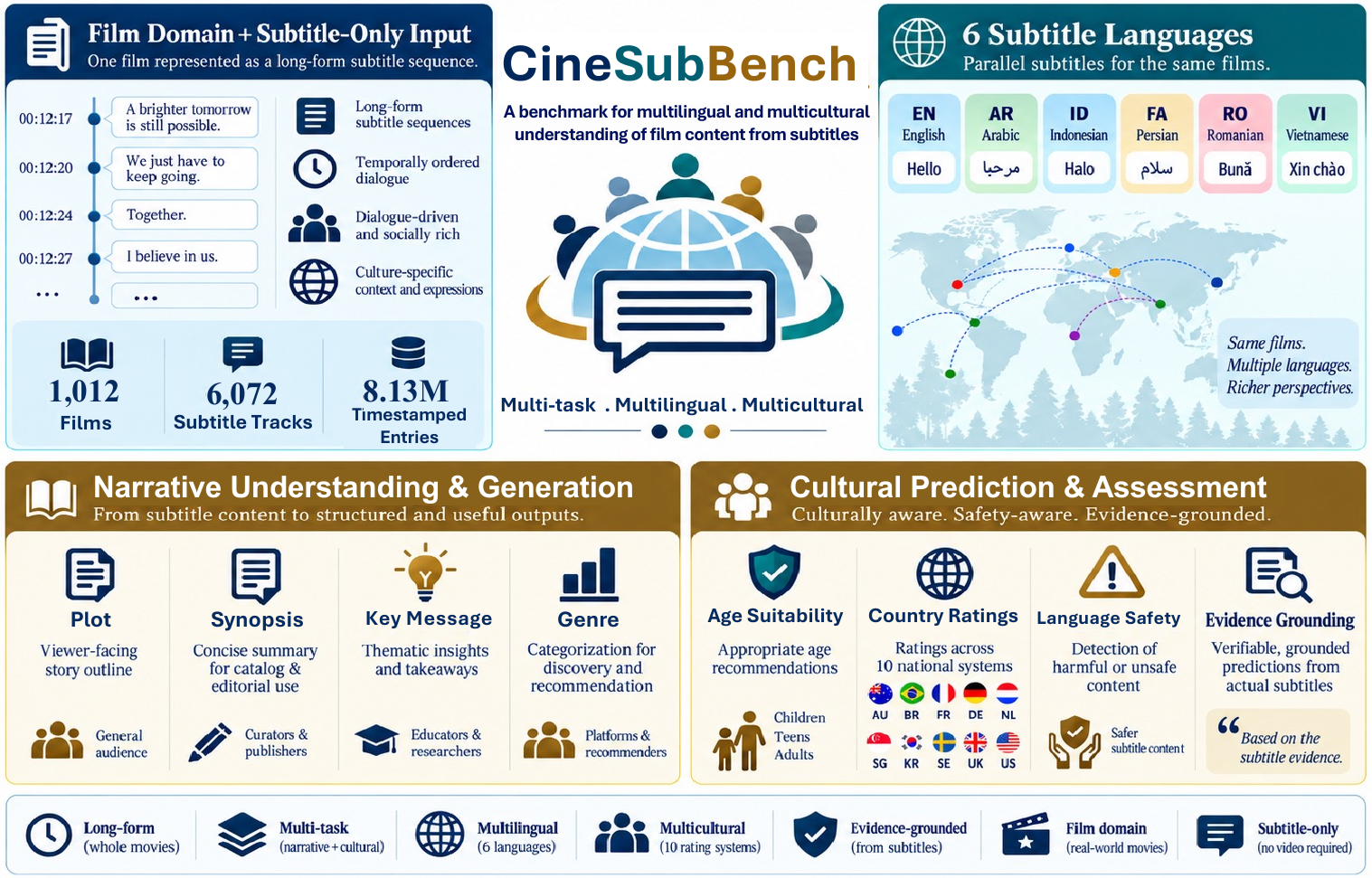}
\vspace{-1mm}
\caption{\small
Overview of \cinesubbench{}, a long-form MultiX benchmark for film understanding from subtitles. Each of 1,012 films has subtitle tracks in English, Arabic, Indonesian, Persian, Romanian, and Vietnamese, yielding 6,072 tracks and 8.13M timestamped subtitle entries. Seven tasks span Narrative Understanding and Generation and Cultural Prediction and Assessment, with ratings across ten national classification systems.
\vspace{-2mm}
}
\label{fig:intro-figure}
\end{figure*}

Movie subtitles provide a distinctive representation of long-context film
understanding. A subtitle track is a film-length sequence of short,
timestamped utterances that preserves the dialogue and temporal progression
through which the narrative develops. Dialogue carries character goals,
relationships, decisions, reactions, conflicts, and consequences, allowing
central narrative arcs and themes to be recovered from evidence across a film.
Yet subtitles do not identify speakers, scene boundaries, narrative events,
importance, motivations, or causal relations explicitly. Models must therefore
reconstruct who is involved, what happened, how events connect, and which
developments matter. This differs from both conventional document summarization
and screenplay summarization: screenplays can provide scene descriptions,
speaker attribution, stage directions, and explicit accounts of actions
~\citep{chen2022summscreen,saxena2024moviesum}, while timeline summarization
can organize events around explicit dates or event markers
~\citep{steen2019timeline}. Subtitle timestamps provide temporal order without
identifying the narrative units themselves, making subtitle-based film
understanding more than compression of an already structured narrative.

Film also creates a natural setting for studying whether understanding remains
consistent across linguistic and cultural contexts. The same story can be
encountered through different subtitle languages, while audience guidance for
that film may vary across national classification systems. These dimensions
intersect in practice: a viewer may encounter a film through translated
subtitles, seek an account of its narrative and themes, and rely on audience
guidance defined within a particular national setting. Evaluating tasks,
languages, and cultural judgments over the same films therefore provides a
coherent basis for examining these dimensions together.

A subtitle-centered setting also enables consistent film-length evaluation
across a broad range of LLMs. Full-video understanding introduces additional
choices and computation for frame or scene selection, video decoding, audio
processing and alignment, temporal selection, and cross-modal inference over
hour-scale inputs. Measured performance can consequently reflect both narrative
reasoning and choices in the audiovisual processing pipeline. Subtitles provide
the same temporally ordered text interface to closed and open-weight language
models, including systems without full-length video input, allowing the
evaluation to focus directly on long-context narrative integration.

We introduce \cinesubbench{}, organized around two challenges:
\textbf{long-form understanding} and \textbf{MultiX evaluation}. The benchmark
contains 1,012 films, 6,072 subtitle tracks, and 8.13M timestamped subtitle
entries, with complete coverage in six subtitle languages and ten national
motion-picture rating systems. Its \textbf{multi-task} dimension spans plot
generation, spoiler-aware synopsis generation, key-message generation,
multi-label genre prediction, age-suitability prediction, country-specific
motion-picture rating prediction, and evidence-grounded language-safety
assessment. Its \textbf{multilingual} dimension presents the same films through
six subtitle languages, while its \textbf{multicultural} dimension retains ten
national classification systems as distinct label spaces~(Figure~\ref{fig:intro-figure}). Cross-source
entity resolution, complete-coverage optimization, subtitle verification, and
task-specific curation support matched and auditable comparisons across these dimensions~(Figure~\ref{fig:data-gen-pipeline}).

Across nine LLMs, \cinesubbench{} reveals capability differences that aggregate
rankings obscure. Models recover broad plot premises more reliably than
event-complete synopses, cross-lingual consistency varies across models and
languages, national rating systems expose distinct calibration behavior, and
evidence-grounded language safety ranges from near-ceiling grounding of strong
profanity to much weaker performance on mild obscenity. These results separate
long-form narrative integration, multilingual consistency, culturally situated
prediction, and evidence grounding as distinct dimensions of current LLM
capability. We make three core contributions:

\begin{enumerate}[label=\arabic*.,leftmargin=*,topsep=2pt,itemsep=1pt]

    \item \textbf{A long-form task setting for LLM film understanding.}
    We formulate subtitle-based film understanding as a long-context language
    problem in which models reconstruct narrative structure from temporally
    ordered, dialogue-centered evidence distributed across an entire film.
    This setting enables evaluation of event selection, chronology, character
    tracking, causal integration, and thematic abstraction from a common
    film-length linguistic representation.

    \item \textbf{A matched MultiX benchmark with broad coverage.}
    \cinesubbench{} contains 1,012 films, 6,072 subtitle tracks, and 8.13M
    timestamped subtitle entries, with complete six-language subtitle coverage
    and ratings from ten national classification systems for every film.
    Seven tasks support \textbf{multi-task, multilingual, and multicultural}
    evaluation over the same film instances.

    \item \textbf{A broad and deep evaluation of nine LLMs.}
    We evaluate nine models using task-specific metrics and matched comparisons
    across narrative tasks, subtitle languages, national rating systems, and
    subtitle-grounded safety evidence. The evaluation further examines
    task-level failure patterns, paired cross-lingual uncertainty,
    country-specific calibration, evidence-grounding errors, and qualitative
    cases that expose behaviors hidden by aggregate scores.

\end{enumerate}


\section{Related Work}
\label{sec:related-work}

\paragraph{Domain-grounded LLM evaluation.}
LLM benchmarks increasingly evaluate whether general-purpose models transfer
to domains with specialized evidence and decision criteria, including law and
medicine~\citep{guha2023legalbench,singhal2023multimedqa}, finance
~\citep{cao2024financial,guo2024econnli}, software engineering
~\citep{jimenez2024swebench}, science and structured data
~\citep{duan2025scigym,zhang2024tablellama}, education and journalism
~\citep{hou2024eeval,li2024newsbench}, cybersecurity and geospatial reasoning
~\citep{dihan2025mapeval,wang2026cybergym}, and multilingual web understanding~\citep{awal2025webmmu}. Film has received far less attention as a comparable
LLM evaluation domain despite combining long-context narrative reasoning,
social interaction, multilingual interpretation, and culturally situated
judgments. \cinesubbench{} addresses this gap with a unified film benchmark
that evaluates these capabilities through its multi-task, multilingual, and
multicultural design.

\paragraph{Long-context narrative and temporal understanding.}
Long input windows do not guarantee effective use of evidence distributed
throughout a context~\citep{liu2024lost,goldman2024longcontext}. Existing
long-form benchmarks study summarization, faithfulness, narrative structure,
and multilingual consistency in books and stories
~\citep{kryscinski2022booksum,kim2024fables,hamilton2026narrabench,
kim2025oneruler}, while culturally specific long-context evaluation shows that
English-centered results need not transfer uniformly
~\citep{arora2025calmqa}. Timeline summarization assumes explicit temporal
events or markers~\citep{steen2019timeline}; subtitle timestamps provide order
without identifying event boundaries, causal structure, or narrative
importance. \cinesubbench{} instead evaluates film-length reconstruction from
dialogue-centered evidence while holding the underlying films fixed across
tasks, subtitle languages, and national classification systems.

\paragraph{Film, subtitle, and audiovisual understanding.}
Prior film and television work has addressed individual problems such as
screenplay summarization
~\citep{chen2022summscreen,saxena2024moviesum}, narrative adaptation from movie
dialogue~\citep{shen2025movieadaptation}, US age-rating~\citep{shafaei2020age}, and content-severity prediction~\citep{zhang2021severity}. These settings do not provide a general LLM
benchmark spanning long-form narrative generation, genre, age suitability,
multiple national rating systems, multilingual subtitle inputs, and
subtitle-indexed safety evidence over the same films. Audiovisual benchmarks
such as MovieQA, LongVideoBench, InfiniBench, MF$^2$, and ShotBench instead target visual grounding, multimodal reasoning, event
understanding, scene interpretation, or cinematographic analysis
~\citep{tapaswi2016movieqa,wu2024longvideobench,ataallah2025infinibench,
zaranis2025mf2,liu2025shotbench,wang2025cinetechbench}. Copyright,
redistribution, and audiovisual-processing requirements can also limit scale
and access, with some benchmarks relying on selected clips, existing video
assets, or separate media acquisition. \cinesubbench{} instead uses publicly
accessible subtitle tracks, enabling matched MultiX comparison across tasks,
languages, national classification systems, and model families, providing a common benchmark for measuring progress.


\section{\cinesubbench{}}
\label{sec:cinesubbench}

\cinesubbench{} is organized around two properties:
\textbf{long-form input} and \textbf{MultiX evaluation}. It contains 1,012
films released between 1977 and 2026, represented by 6,072 subtitle tracks and
8.13M timestamped subtitle entries in English, Arabic, Indonesian, Persian,
Romanian, and Vietnamese. Every film also has ratings from the same ten
national motion-picture classification systems, enabling matched comparisons
across tasks, languages, and countries.

Each subtitle input is film-length: mean prompt-formatted tracks contain roughly
40K--45K tokens across languages, with the longest exceeding 100K. The
difficulty arises from both length and structure, since evidence appears as
short, temporally ordered dialogue rather than continuous exposition. We use
\emph{long-form} for the input and \emph{long-context understanding} for the
capability required to integrate evidence distributed across it. Each subtitle
entry retains its index, time interval, and text, supporting both film-level
understanding and line-level grounding. The benchmark is
\textbf{multi-task}, spanning seven narrative and cultural tasks;
\textbf{multilingual}, with the same films presented in six subtitle languages;
and \textbf{multicultural}, with audience classification evaluated under ten
distinct national rating systems. Appendix~\ref{app:dataset-statistics}
reports corpus, input-length, label, and subtitle-duration statistics.

\subsection{Narrative Understanding and Generation}
\label{sec:narrative-tasks}

The narrative tasks probe levels of understanding from the same
film-length subtitle sequence. We distinguish plot, synopsis, and key message by their function rather than by output length.

\paragraph{Plot Generation.}
The model generates a concise premise identifying the central character or
group, setup, conflict, and driving situation, emphasizing the dominant
narrative over retelling.

\paragraph{Synopsis Generation.}
The model generates a spoiler-aware account of the narrative arc,
including developments, character goals and conflicts, turning points,
climax, and resolution. Compared with plot generation, this requires
event selection, chronology, causality, and character tracking.

\vspace{-2mm}
\paragraph{Key-Message Generation.}
The model generates one concise moral, social, or emotional takeaway, requiring thematic abstraction beyond simply recounting narrative events.

\paragraph{Genre Prediction.}
The model predicts one or more genres from a normalized 22-label space using
evidence from story, character, setting, tone, conflict, and broader thematic and stylistic patterns. Appendix~\ref{app:prompt-structures} provides the complete task instructions and response schemas.

\subsection{Cultural Prediction and Assessment}
\label{sec:cultural-tasks}

These tasks evaluate how models map film-level subtitle evidence to
audience-oriented judgments defined by different national and institutional
target settings and classification frameworks.

\paragraph{Age-Suitability Prediction.}
The model predicts the Common Sense Media minimum recommended age. This
family-oriented suitability judgment reflects perceived developmental
appropriateness and is evaluated separately from formal national
motion-picture classification.

\paragraph{Country-Specific Motion-Picture Rating Prediction.}
For every film, \cinesubbench{} includes ratings from Australia, Brazil,
France, Germany, the Netherlands, Singapore, South Korea, Sweden, the United
Kingdom, and the United States. Each country retains its own ordered label
space because these classifications reflect distinct institutional standards
rather than interchangeable age labels. The task therefore evaluates how the
same film evidence maps into ten national classification systems. Appendix
Table~\ref{tab:rating_spaces} lists the normalized label spaces.

\paragraph{Subtitle-Grounded Language-Safety Assessment.}
Language safety is evaluated from the English subtitle track in four lexical
categories: strong profanity, crude bodily language, mild obscenity, and
religious profanity/exclamation. Models predict category counts and exact
subtitle indices supporting them. Gold labels are constructed from audited
subtitle occurrences, while Kids-in-Mind counts and glossary definitions are
retained as provenance and an audit signal. Requiring evidence indices
separates evidence recovery from category assignment and makes missed or
unsupported judgments directly inspectable. Appendix
\ref{app:language-gold-construction} details the annotation and audit procedure.

\subsection{Construction and Quality Assurance}
\label{sec:construction}

\begin{figure*}[t]
\centering
\includegraphics[width=0.98\linewidth]{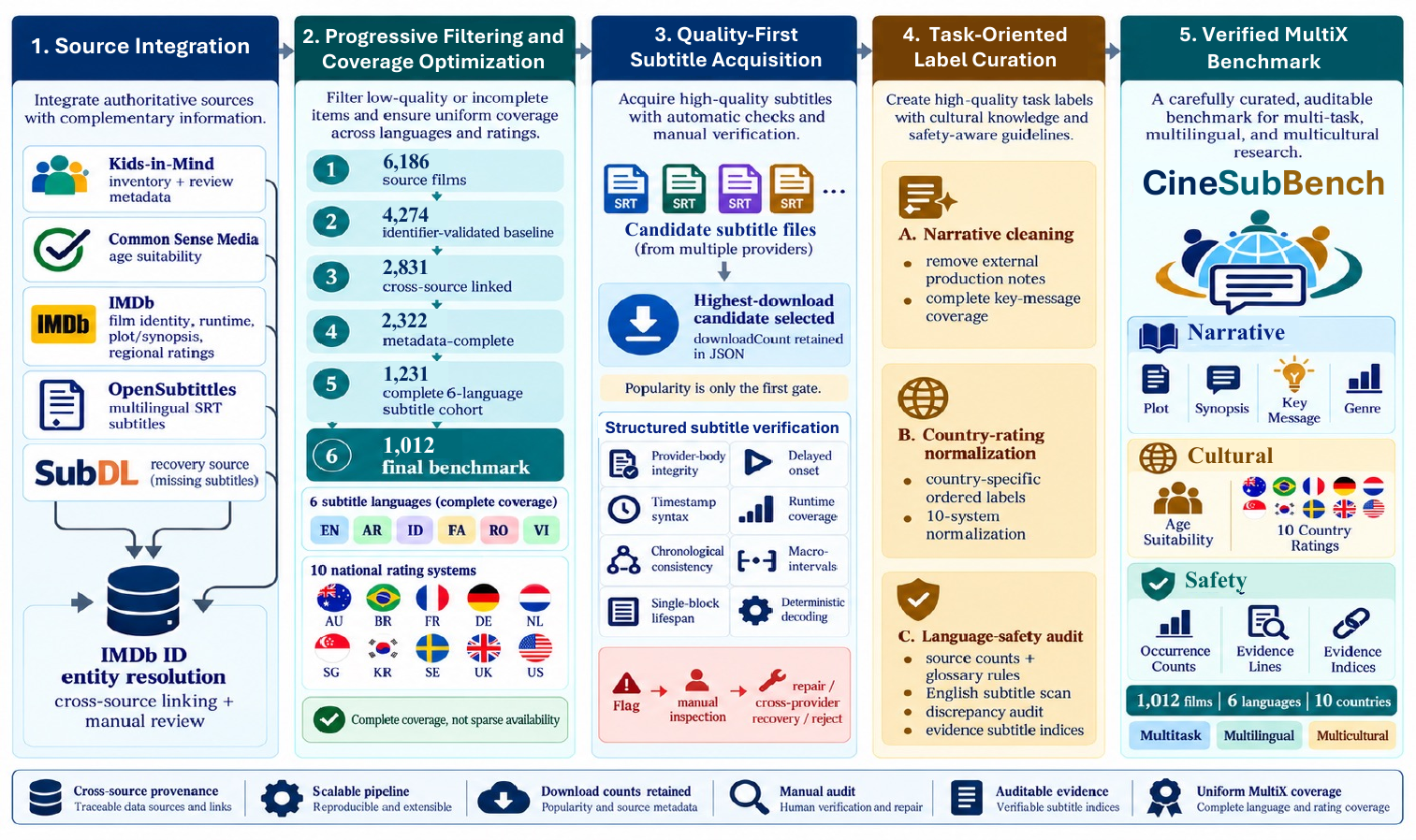}
\vspace{-2mm}
\caption{\small
Construction and quality-assurance pipeline for \cinesubbench{}.
Cross-source entity resolution and filtering reduce the initial
6,186-film inventory to a metadata-complete cohort. Complete-coverage
optimization then selects six subtitle languages and ten national rating
systems. Subtitle candidates undergo structural, temporal, encoding,
language-consistency, and runtime checks with manual inspection and
cross-provider recovery where needed. Task-oriented curation yields the final
matched MultiX benchmark of 1,012 films.}
\vspace{-2mm}
\label{fig:data-gen-pipeline}
\end{figure*}

\cinesubbench{} integrates film metadata, audience guidance, national
classifications, and subtitle assets through IMDb title identifiers.
Kids-in-Mind provides the initial film inventory and language-content metadata,
Common Sense Media provides age-suitability labels and storylines, IMDb
provides narrative metadata, runtimes, and regional ratings, and OpenSubtitles
provides the subtitle source, with SubDL used for recovery.
As Figure~\ref{fig:data-gen-pipeline} shows, cross-source entity resolution and
task-metadata filtering reduce the initial 6,186 films before complete-coverage
optimization selects a six-language cohort of 1,231 films and a final
1,012-film cohort with ratings from ten national systems. Language and
country sets are selected for shared coverage over the same films rather than
marginal frequency alone. Appendix~\ref{app:coverage-optimization} details the
optimization procedure and coverage analysis.

For film-language pairs with multiple subtitle candidates, source-reported
download count provides a selection signal, followed by checks for
provider-error payloads, malformed SRT structure, temporal inconsistencies,
delayed starts, large gaps, insufficient runtime coverage, encoding failures,
and language inconsistencies. Automated flags trigger manual inspection,
repair, or cross-provider recovery where needed. Task references undergo curation: narrative fields are cleaned to retain film-internal
content, age suitability remains distinct from national classification,
country ratings are normalized within their national systems, and
language-safety labels are audited against indexed subtitle evidence.
Appendices~\ref{app:subtitle-qa} and~\ref{app:label-curation} document the
verification and curation procedures.


\section{Evaluation Setup}
\label{sec:evaluation-setup}

\paragraph{Models and input regimes.}
Our main evaluation compares nine models across three access groups: closed
frontier models (GPT-5.6 Sol, Gemini 3.8 Flash, and Claude Haiku 4.5), closed
baselines (GPT-5 Nano and Gemini 3.5 Flash Lite), and open-weight models
(DeepSeek V4 Pro 1.6T, Qwen3 235B, Mistral 4 119B, and Llama 4 Scout 17B).
For the six-language evaluation, the same films are presented through English,
Arabic, Indonesian, Persian, Romanian, and Vietnamese subtitles. Narrative and
cultural predictions are evaluated separately for each language, while
narrative generations are written in English. We use \textsc{CL} for the
equal-weight mean of the five non-English settings; subtitle-grounded
language-safety assessment is English-only because its gold evidence is
localized to the English track. Appendix Tables~\ref{tab:experiment-matrix}
and~\ref{tab:decoding-parameters} report model coverage and generation
settings, and Appendix~\ref{app:scaling-analysis} presents the English-only
within-family scaling analysis.

\paragraph{LLM-as-a-judge evaluation.}
Open-ended plot, synopsis, and key-message generations are evaluated with
GPT-5.6 Luna as a common reference-based LLM judge
~\citep{liu2023geval,gu2024surveyjudge,li2025generationJudgment}.
Because narrative evaluation spans long film contexts across multiple models,
tasks, subtitle languages, and thousands of generations, judge inference is a
substantial component of evaluation cost. We therefore use Luna as a fast,
cost-efficient judge while applying the same references, task-specific rubrics,
and structured scoring protocol to every model.
Appendix~\ref{app:metric-definitions} provides the scoring definitions,
Appendix~\ref{app:prompt-structures} gives the full judge instructions, and
Appendix~\ref{app:independent-judge-agreement} reports agreement with an
independent narrative judge, with 98.4\% within-one-point agreement overall.

\subsection{Evaluation Metrics}
\label{sec:metrics}

We use task-specific metrics. Plot, synopsis, and key-message generations are
scored on a reference-based \textbf{1--5 rubric} with GPT-5.6 Luna; narrative overall is
their equal-weight mean. Genre prediction uses micro-F1. Age suitability uses
MAE in years together with exact and near-match accuracy. Country-specific
ratings use exact match and ordinal MAE computed within each national label
space. Appendix~\ref{app:metric-definitions} provides definitions and
secondary measures, while Appendix~\ref{app:prompt-structures} gives the task
and evaluator instructions. For language safety, category-agnostic evidence F1 measures recovery of relevant
subtitle indices regardless of category, while strict F1 additionally requires
the correct lexical category. In the leaderboard, LC denotes this English-only
language-content assessment.

\paragraph{Cross-task composite.}
We use a composite only as a compact summary across task families; subsequent
analyses retain the original task metrics. Let $S_{\mathrm{nar}}$ denote
narrative overall, $F1_{\mathrm{genre}}$ genre micro-F1,
$\mathrm{MAE}_{\mathrm{age}}$ age MAE,
$\mathrm{EM}_{\mathrm{country}}$ country-rating exact match, and
$F1_{\mathrm{LC}}$ strict language-safety evidence F1. Percentage metrics are
expressed on a 0--100 scale:
\begin{equation}
\resizebox{0.80\linewidth}{!}{$
\begin{aligned}
N_{\mathrm{nar}} &= 25(S_{\mathrm{nar}}-1),
&\qquad
N_{\mathrm{age}} &=
100\left(1-\frac{\min(\mathrm{MAE}_{\mathrm{age}},16)}{16}\right),\\
N_{\mathrm{cult}} &=
\frac{N_{\mathrm{age}}+\mathrm{EM}_{\mathrm{country}}}{2},
&\qquad
S_{\mathrm{overall}} &=
\frac{N_{\mathrm{nar}}+F1_{\mathrm{genre}}+
N_{\mathrm{cult}}+F1_{\mathrm{LC}}}{4}.
\end{aligned}
$}
\label{eq:composite}
\end{equation}
The value 16 is the span of the age-label space from $2+$ to $18+$.
Narrative, genre, age, and country metrics are first averaged equally over the
six subtitle languages, whereas $F1_{\mathrm{LC}}$ is English-only.


\section{Results}
\label{sec:results}

We organize the evaluation around \emph{four} questions: whether aggregate ranking
reflects uniform capability across tasks (RQ1), how reliably understanding
carries across subtitle languages (RQ2), where cultural calibration differs
across national settings and classification systems (RQ3), and whether
language-safety judgments can be grounded in auditable subtitle evidence (RQ4).

\subsection{RQ1: Does a Single Rank Reflect Uniform Capability?}
\rqlabel{rq:model-ranking}

\paragraph{Aggregate ranking hides different capability profiles.}
Table~\ref{tab:composite-leaderboard} summarizes performance across the
benchmark. Gemini 3.8 Flash reaches the highest composite score at 78.04,
followed by GPT-5.6 Sol at 74.70 and DeepSeek V4 Pro at 65.52. Paired film-level comparisons of these leading composite scores are reported in
Appendix~\ref{app:composite-significance}. Their
task-level profiles, however, differ substantially. Sol achieves the strongest
synopsis score (3.40), whereas Gemini leads on plot (4.16), key message
(3.11), genre micro-F1 (84.93\%), country-rating exact match (77.87\%), and
strict language-safety evidence F1 (77.8\%). A similar aggregate position
therefore need not reflect the same strengths in narrative reconstruction,
classification, cultural calibration, and evidence grounding.

\begin{table}[t]
\centering
\small
\setlength{\tabcolsep}{2.25pt}
\renewcommand{\arraystretch}{1.25}
\caption{\small All-language comparison across the nine models with complete benchmark
coverage. Narrative, genre, age, and country metrics are averaged equally over
English, Arabic, Indonesian, Persian, Romanian, and Vietnamese subtitle inputs;
language-content (LC) evidence metrics are English-only. Models are ordered by
the composite score in Equation~\ref{eq:composite}. LC agnostic F1 matches
safety-relevant subtitle evidence regardless of category, while strict F1 also
requires the correct lexical category. Best values are bold and second-best
distinct values are underlined; background shading is normalized within each
metric column.
}
\vspace{2.25pt}
\label{tab:composite-leaderboard}
\resizebox{\textwidth}{!}{%
\begin{tabular}{cllcrrrrrrrrrrrr}
\toprule
Rank & Model & Family & \multicolumn{1}{c}{\textbf{Composite}} & \multicolumn{5}{c}{\textbf{Narrative Understanding and Generation}} & \multicolumn{5}{c}{\textbf{Cultural Prediction and Assessment}} & \multicolumn{2}{c}{\textbf{LC Assessment}} \\
\cmidrule(lr){4-4}\cmidrule(lr){5-9}\cmidrule(lr){10-14}\cmidrule(lr){15-16}
 & & & \shortstack{Overall\\score (0--100) $\uparrow$} & \shortstack{Plot\\$\uparrow$} & \shortstack{Synopsis\\$\uparrow$} & \shortstack{Key\\message $\uparrow$} & \shortstack{Overall\\$\uparrow$} & \shortstack{Genre\\micro-F1 (\%) $\uparrow$} & \shortstack{Age\\exact (\%) $\uparrow$} & \shortstack{Age\\$\pm$1 (\%) $\uparrow$} & \shortstack{Age\\MAE $\downarrow$} & \shortstack{Country\\EM (\%) $\uparrow$} & \shortstack{Country\\MAE $\downarrow$} & \shortstack{LC agnostic\\F1 (\%) $\uparrow$} & \shortstack{LC strict\\F1 (\%) $\uparrow$} \\
\midrule
\textcolor{cineGold}{\faMedal} & \href{https://docs.cloud.google.com/gemini-enterprise-agent-platform/models/gemini/3-8-flash}{Gemini 3.8 Flash} & Closed frontier & \cellcolor{cineComposite!43!white}\textbf{78.04} & \cellcolor{cineNarrative!43!white}\textbf{4.16} & \cellcolor{cineNarrative!42!white}\underline{3.34} & \cellcolor{cineNarrative!43!white}\textbf{3.11} & \cellcolor{cineNarrative!43!white}\textbf{3.54} & \cellcolor{cineNarrative!43!white}\textbf{84.93} & \cellcolor{cineCultural!43!white}\textbf{34.25} & \cellcolor{cineCultural!43!white}\textbf{81.42} & \cellcolor{cineCultural!43!white}\textbf{0.93} & \cellcolor{cineCultural!43!white}\textbf{77.87} & \cellcolor{cineCultural!43!white}\textbf{0.31} & \cellcolor{cineTeal!43!white}\textbf{86.8} & \cellcolor{cineTeal!43!white}\textbf{77.8} \\
\textcolor{cineSilver}{\faMedal} & \href{https://developers.openai.com/api/docs/models/gpt-5.6-sol}{GPT-5.6 Sol} & Closed frontier & \cellcolor{cineComposite!40!white}\underline{74.7} & \cellcolor{cineNarrative!42!white}\underline{4.1} & \cellcolor{cineNarrative!43!white}\textbf{3.4} & \cellcolor{cineNarrative!42!white}\underline{3.08} & \cellcolor{cineNarrative!43!white}\underline{3.52} & \cellcolor{cineNarrative!36!white}81.55 & \cellcolor{cineCultural!36!white}31.08 & \cellcolor{cineCultural!40!white}78.75 & \cellcolor{cineCultural!41!white}\underline{0.98} & \cellcolor{cineCultural!33!white}\underline{65.42} & \cellcolor{cineCultural!35!white}\underline{0.44} & \cellcolor{cineTeal!40!white}\underline{81.7} & \cellcolor{cineTeal!41!white}\underline{74.5} \\
\textcolor{cineBronze}{\faMedal} & \href{https://openrouter.ai/deepseek/deepseek-v4-pro-0813}{DeepSeek V4 Pro 1.6T} & Open-weight & \cellcolor{cineComposite!31!white}65.52 & \cellcolor{cineNarrative!38!white}3.91 & \cellcolor{cineNarrative!33!white}2.93 & \cellcolor{cineNarrative!35!white}2.9 & \cellcolor{cineNarrative!36!white}3.25 & \cellcolor{cineNarrative!36!white}\underline{81.63} & \cellcolor{cineCultural!35!white}30.67 & \cellcolor{cineCultural!41!white}\underline{79.33} & \cellcolor{cineCultural!41!white}\underline{0.98} & \cellcolor{cineCultural!27!white}58.03 & \cellcolor{cineCultural!31!white}0.52 & \cellcolor{cineTeal!27!white}57.5 & \cellcolor{cineTeal!27!white}48.3 \\
\midrule
4 & \href{https://docs.cloud.google.com/gemini-enterprise-agent-platform/models/gemini/3-5-flash-lite}{Gemini 3.5 Flash Lite} & Closed baseline & \cellcolor{cineComposite!30!white}64.7 & \cellcolor{cineNarrative!37!white}3.83 & \cellcolor{cineNarrative!33!white}2.91 & \cellcolor{cineNarrative!38!white}2.99 & \cellcolor{cineNarrative!36!white}3.25 & \cellcolor{cineNarrative!35!white}80.92 & \cellcolor{cineCultural!37!white}\underline{31.83} & \cellcolor{cineCultural!32!white}72.17 & \cellcolor{cineCultural!37!white}1.09 & \cellcolor{cineCultural!32!white}64.65 & \cellcolor{cineCultural!34!white}0.46 & \cellcolor{cineTeal!31!white}65.3 & \cellcolor{cineTeal!25!white}42.8 \\
5 & \href{https://platform.claude.com/docs/en/models/haiku-4-5/overview}{Claude Haiku 4.5} & Closed frontier & \cellcolor{cineComposite!23!white}57.5 & \cellcolor{cineNarrative!30!white}3.45 & \cellcolor{cineNarrative!23!white}2.44 & \cellcolor{cineNarrative!32!white}2.84 & \cellcolor{cineNarrative!28!white}2.91 & \cellcolor{cineNarrative!16!white}71.45 & \cellcolor{cineCultural!27!white}27.42 & \cellcolor{cineCultural!33!white}72.75 & \cellcolor{cineCultural!35!white}1.16 & \cellcolor{cineCultural!24!white}54.32 & \cellcolor{cineCultural!28!white}0.57 & \cellcolor{cineTeal!27!white}58.1 & \cellcolor{cineTeal!22!white}37.2 \\
6 & \href{https://huggingface.co/Qwen/Qwen3-235B-A22B-Instruct-2507}{Qwen3 235B} & Open-weight & \cellcolor{cineComposite!18!white}53.09 & \cellcolor{cineNarrative!28!white}3.36 & \cellcolor{cineNarrative!20!white}2.3 & \cellcolor{cineNarrative!29!white}2.76 & \cellcolor{cineNarrative!26!white}2.8 & \cellcolor{cineNarrative!10!white}68.28 & \cellcolor{cineCultural!15!white}22.0 & \cellcolor{cineCultural!17!white}59.25 & \cellcolor{cineCultural!24!white}1.47 & \cellcolor{cineCultural!18!white}47.88 & \cellcolor{cineCultural!24!white}0.64 & \cellcolor{cineTeal!20!white}45.0 & \cellcolor{cineTeal!18!white}29.6 \\
7 & \href{https://huggingface.co/mistralai/Mistral-Small-4-119B-2603}{Mistral 4 119B} & Open-weight & \cellcolor{cineComposite!16!white}50.67 & \cellcolor{cineNarrative!18!white}2.8 & \cellcolor{cineNarrative!14!white}2.0 & \cellcolor{cineNarrative!20!white}2.52 & \cellcolor{cineNarrative!17!white}2.44 & \cellcolor{cineNarrative!7!white}66.83 & \cellcolor{cineCultural!9!white}19.42 & \cellcolor{cineCultural!12!white}54.5 & \cellcolor{cineCultural!12!white}1.78 & \cellcolor{cineCultural!18!white}47.37 & \cellcolor{cineCultural!21!white}0.68 & \cellcolor{cineTeal!22!white}49.6 & \cellcolor{cineTeal!19!white}31.7 \\
8 & \href{https://developers.openai.com/api/docs/models/gpt-5-nano}{GPT-5 Nano} & Closed baseline & \cellcolor{cineComposite!12!white}46.67 & \cellcolor{cineNarrative!21!white}2.97 & \cellcolor{cineNarrative!12!white}1.9 & \cellcolor{cineNarrative!24!white}2.64 & \cellcolor{cineNarrative!18!white}2.5 & \cellcolor{cineNarrative!18!white}72.43 & \cellcolor{cineCultural!12!white}20.67 & \cellcolor{cineCultural!19!white}60.75 & \cellcolor{cineCultural!20!white}1.56 & \cellcolor{cineCultural!15!white}43.87 & \cellcolor{cineCultural!17!white}0.76 & \cellcolor{cineTeal!7!white}21.7 & \cellcolor{cineTeal!7!white}9.6 \\
9 & \href{https://huggingface.co/meta-llama/Llama-4-Scout-17B-16E-Instruct}{Llama 4 Scout 17B} & Open-weight & \cellcolor{cineComposite!7!white}41.58 & \cellcolor{cineNarrative!7!white}2.21 & \cellcolor{cineNarrative!7!white}1.67 & \cellcolor{cineNarrative!7!white}2.2 & \cellcolor{cineNarrative!7!white}2.03 & \cellcolor{cineNarrative!7!white}66.92 & \cellcolor{cineCultural!7!white}18.58 & \cellcolor{cineCultural!7!white}50.25 & \cellcolor{cineCultural!7!white}1.93 & \cellcolor{cineCultural!7!white}34.12 & \cellcolor{cineCultural!7!white}0.93 & \cellcolor{cineTeal!11!white}28.4 & \cellcolor{cineTeal!9!white}12.7 \\
\bottomrule
\end{tabular}%
}
\vspace{-9.25pt}
\end{table}

\paragraph{Strong performance transfers unevenly across tasks.}
DeepSeek V4 Pro exceeds Sol on genre micro-F1 (81.63\% versus
81.55\%) and age accuracy within one year (79.33\% versus 78.75\%), with both
at 0.98-year MAE. Larger gaps appear in narrative overall (3.25 versus 3.52),
country-rating exact match (58.03\% versus 65.42\%), and
language-safety evidence F1 (48.3\% versus 74.5\%). Its third-place composite
therefore reflects a different capability balance rather than a uniform gap
from the leading models. English-only and five-language-average results appear
in Appendix~\ref{app:rq1-headline-results}.

\subsection{RQ2: How Robust Is Understanding Across Subtitle Languages?}
\rqlabel{rq:crosslingual}

\begin{wrapfigure}{r}{0.52\columnwidth}
\vspace{-18pt}
\centering
\includegraphics[width=\linewidth]
{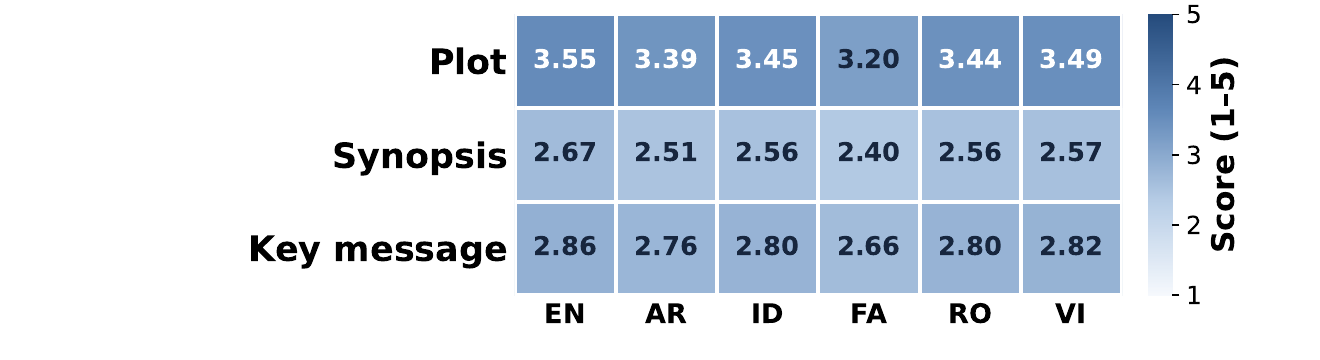}\\[-1mm]
\includegraphics[width=\linewidth]
{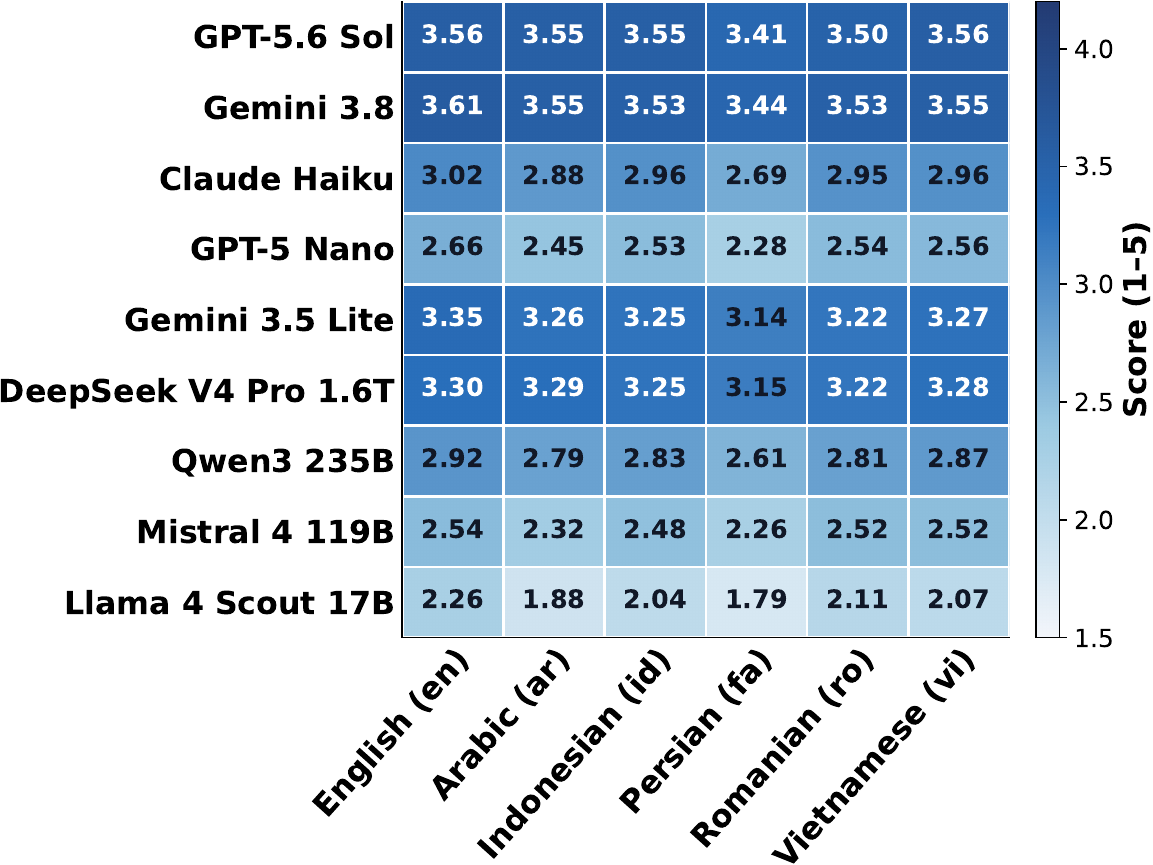}
\vspace{-18pt}
\caption{\small
Narrative performance across subtitle languages. Plot remains more reliably recovered than synopsis, while Persian is the lowest observed narrative setting across the evaluated models.}
\label{fig:narrative-crosslingual-heatmap}
\vspace{-18pt}
\end{wrapfigure}

\paragraph{Cross-lingual consistency varies substantially across models.}
Narrative performance decreases from English to the five-language mean for all
nine models, but the magnitude ranges from $-0.05$ points for GPT-5.6 Sol to
$-0.28$ for Llama 4 Scout on the 1--5 scale. Paired intervals for the smallest
changes include zero, whereas the larger declines are more clearly separated
(Appendix Table~\ref{tab:paired-language-uncertainty-appendix}). The detailed
English-to-cross-lingual comparison across narrative, genre, age, and country
ratings appears in Appendix~\ref{app:rq2-overview}.

\vspace{-3mm}
\paragraph{Persian is the most consistent observed stress setting.}
Figure~\ref{fig:narrative-crosslingual-heatmap} shows that Persian yields the
lowest narrative-overall score for every model, the lowest genre score for
eight of nine, and the lowest country-rating exact match for every model. Sol
drops from 3.56 in English to 3.41 in Persian on narrative overall, while
Llama 4 Scout falls from 2.26 to 1.79; country exact match drops from 67.2\%
to 62.4\% and 33.1\% to 31.8\%, respectively. With films and references held
constant, these results reflect end-to-end consistency across subtitle
languages.

\vspace{-2mm}
\paragraph{Premise recovery is more robust than full narrative reconstruction.}
Across models, English averages are 3.55 for plot, 2.67 for synopsis, and
2.86 for key message; Persian averages are 3.20, 2.40, and 2.66. The gap
persists for strong models: Sol scores 4.10 versus 3.40 and DeepSeek 3.91
versus 2.93 on plot and synopsis. Evaluator annotations show that 79.7\% of
synopses omit a major event and 71.1\% invent one. Full results and error
analyses appear in Appendix~\ref{app:rq2-narrative}.

\subsection{RQ3: Where Does Cultural Calibration Fail?}
\rqlabel{rq:cultural-errors}

\paragraph{Similar age accuracy can conceal opposite calibration tendencies.}
Across six subtitle languages, Gemini 3.8 Flash matches the exact reference age
in 34.2\% of cases and falls within one year in 81.4\%; GPT-5.6 Sol reaches
31.1\% and 78.8\%, respectively. Nearly half of both models' predictions miss
by exactly one year, but in opposite directions: Sol overpredicts more often
than it underpredicts (48.3\% versus 20.6\%), whereas Gemini shows the reverse
(15.2\% versus 50.6\%). Thus, similar near-match accuracy can reflect
substantially different audience-guidance behavior. Appendix~\ref{app:rq3-age}
reports the full error-magnitude and direction analysis.

\begin{figure*}[t]
\centering
\begin{minipage}{0.95\textwidth}
    \centering
    \includegraphics[width=\linewidth]{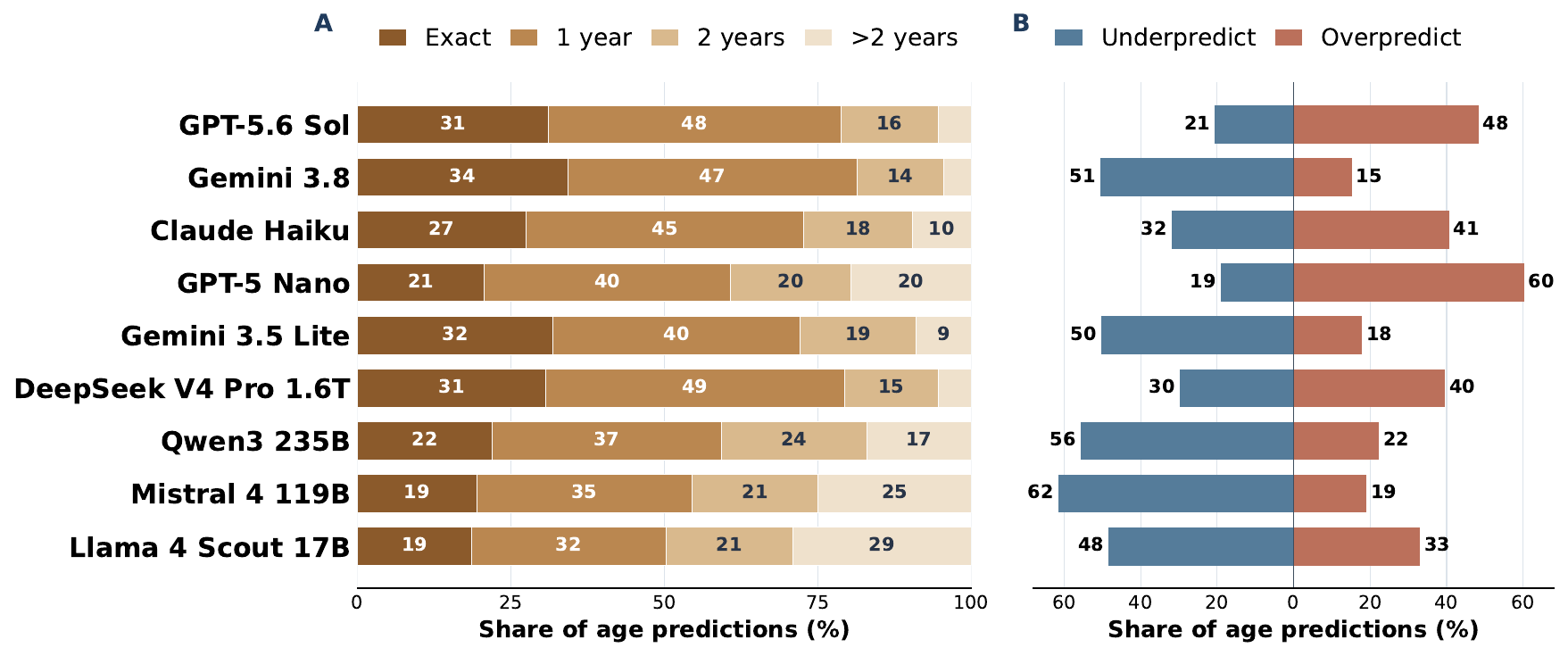}
\end{minipage}

\vspace{1.5mm}

\begin{minipage}{0.95\textwidth}
    \centering
    \includegraphics[width=\linewidth]{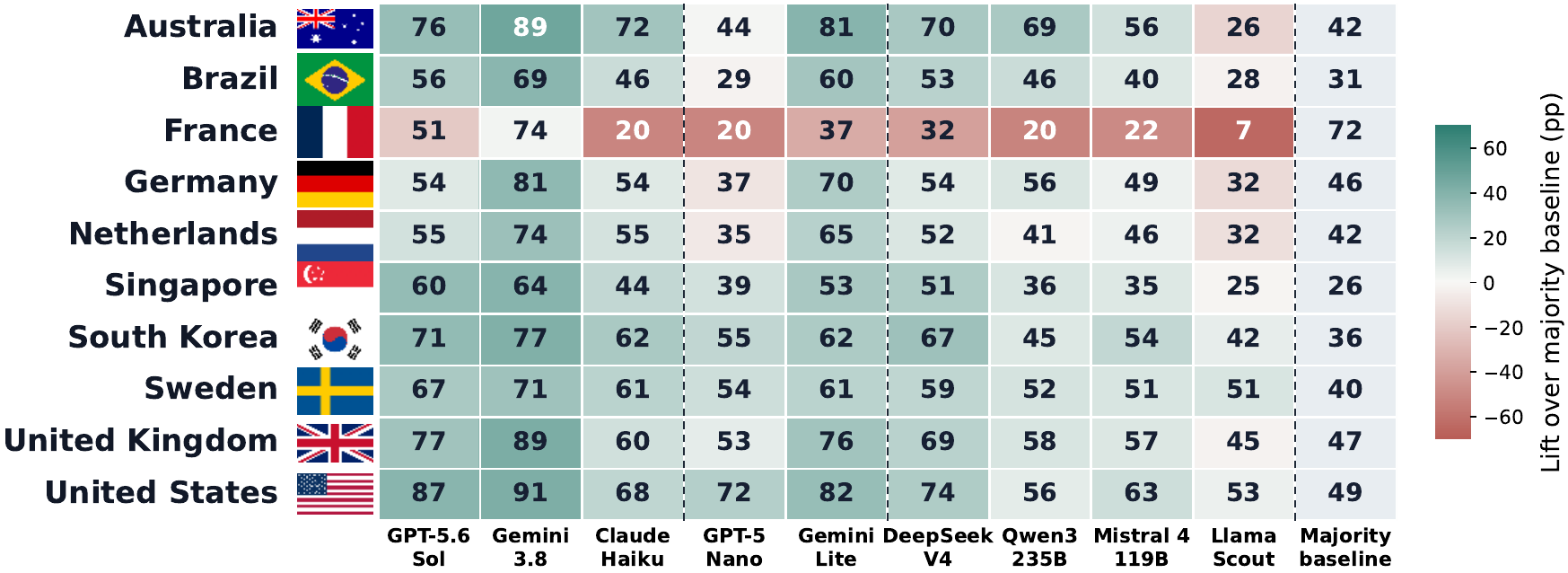}
\end{minipage}
\vspace{-0.5mm}
\caption{\small
Cultural prediction reveals distinct calibration patterns.
\textbf{Top:} Age errors by magnitude and direction.
\textbf{Bottom:} Country-rating exact match across six subtitle languages,
with shading showing lift over each country's majority baseline. France pairs
a high baseline with weak performance for most models, while Brazil shows that
lower raw accuracy can still yield substantial baseline gains.
}
\label{fig:cultural-calibration}
\end{figure*}

\paragraph{National rating systems reveal distinct calibration patterns.}
Figure~\ref{fig:cultural-calibration} shows why country accuracy must be
interpreted within each national label distribution. France has 31.6\% mean
model exact match against a 72.0\% majority baseline; eight of nine models fall
below it, and 95.3\% of nonzero ordinal errors assign a stricter rating.
Gemini reaches 73.8\%, only slightly above baseline. Brazil differs: its largest
class covers 31.0\% of films, while Sol reaches 56.1\%, a 25.1-point gain.
Thus, similar raw accuracies can reflect very different calibration across
national systems. Gemini and Sol achieve six-language country MAE of 0.31 and
0.44, respectively.

\paragraph{Cultural prediction also changes across subtitle languages.}
The direction is not uniform across models and countries. Romanian subtitles
raise country-rating exact match by 3.7 points for Llama 4 Scout and 3.4 for
GPT-5 Nano. Averaged across models, Romanian improves Brazil and Germany by
2.3 and 1.7 points, respectively, while all five non-English settings reduce
US performance. Complete model-language and country-language analyses appear
in Appendix~\ref{app:cultural-country-breakdown}.

\vspace{-1.25mm}
\subsection{RQ4: Can Safety Judgments Be Evidence-Grounded?}
\rqlabel{rq:evidence-grounding}

\begin{figure}[t]
\centering
\includegraphics[width=0.95\linewidth]
{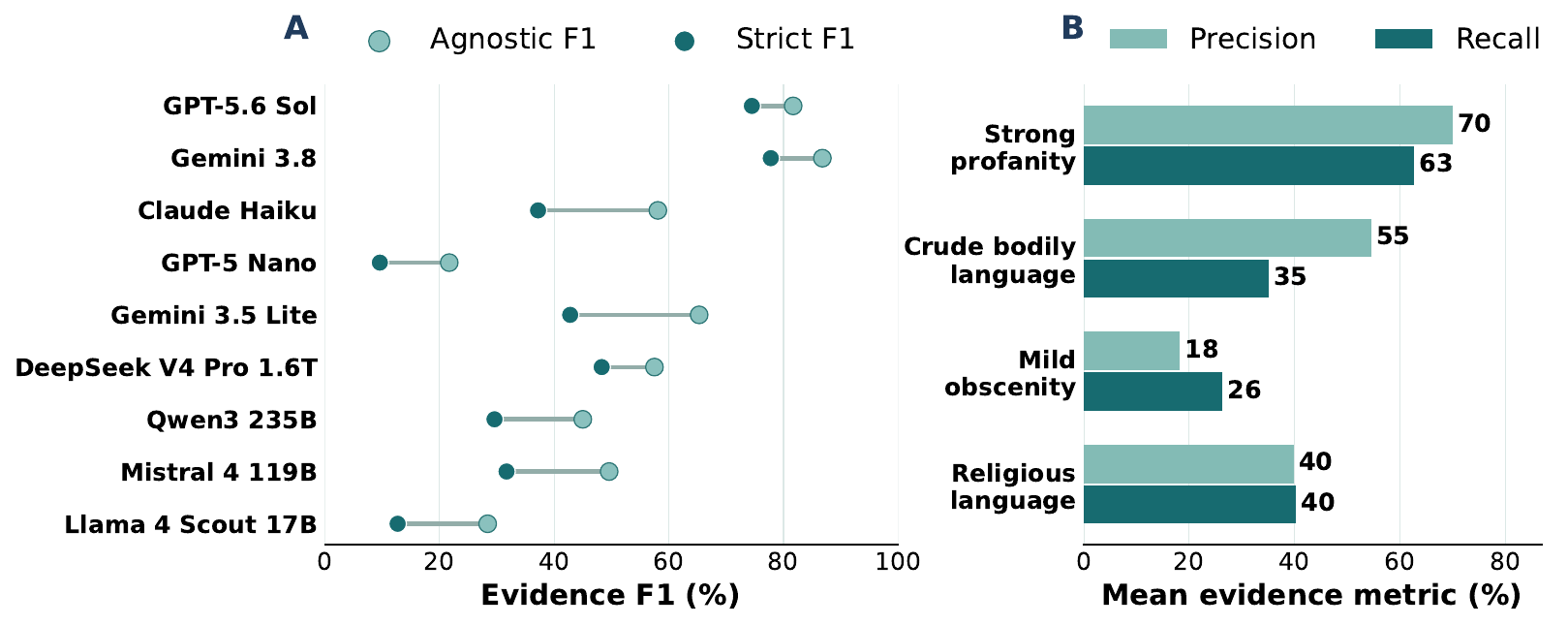}
\vspace{-4.25pt}
\caption{\small
Evidence-grounded language-safety performance.
\textbf{(A)} Category-agnostic F1 measures recovery of relevant subtitle
evidence; strict F1 also requires the correct lexical category.
\textbf{(B)} Mean precision and recall by category across models. Strong
profanity is substantially easier to ground than mild obscenity.}
\vspace{-2.25pt}
\label{fig:language-content-diagnostics}
\end{figure}

\paragraph{Evidence recovery and category assignment are distinct capabilities.}
Figure~\ref{fig:language-content-diagnostics} separates category-agnostic
evidence recovery from strict matching, which also requires the correct lexical
category. Gemini reaches 86.8\% agnostic F1 and 77.8\% strict F1, while Sol
reaches 81.7\% and 74.5\%. This gap shows that models can locate relevant
dialogue yet assign it to the wrong category. Precision and recall reveal
different behaviors: Sol recovers 96.3\% of gold evidence at 70.9\% precision,
whereas DeepSeek reaches 77.8\% precision but only 45.6\% recall.

\paragraph{Grounding difficulty differs sharply by lexical category.}
Strong profanity reaches 99.6\% strict F1 for Sol and 99.1\% for Gemini,
compared with 39.7\% and 52.0\% for mild obscenity. DeepSeek shows the same
ordering at 68.2\% versus 22.6\%. Across nine models, strong profanity
reaches 62.8\% category-level F1 versus 20.3\% for mild obscenity, with 73.7\%
of its gold evidence missed. Precision, recall, count-error,
category-specific, and false-positive analyses appear in
Appendix~\ref{app:language-content-error-analysis}.

\vspace{-3.25pt}
\section{Discussion}
\label{sec:discussion}

\paragraph{Long-form film understanding remains challenging.}
Even frontier models do not uniformly recover a film from its subtitle
sequence. Premises are easier than complete narrative arcs: synopsis scores
remain around 3.4/5 for strong models, with major-event omissions
and invented events. This leaves substantial headroom for applications such as
film summarization, search, catalog enrichment, and narrative analysis, which
require tracking characters, events, and causal progression.

\vspace{-2.25pt}
\paragraph{Multilingual access does not ensure multilingual reliability.}
Changing only the subtitle language produces shifts in model
behavior. Narrative performance declines from English to the non-English
settings for all nine models in point estimate, while Persian repeatedly yields
the weakest results across several tasks. CineSubBench therefore exposes
cross-lingual instability that English-only film evaluation would miss, a
limitation for systems serving multilingual audiences.

\vspace{-2.25pt}
\paragraph{Narrative competence does not guarantee cultural calibration.}
Strong narrative performance can coexist with weak audience-guidance
prediction. In France, mean country-rating exact match is only 31.6\% against a
72.0\% majority baseline, whereas Brazil shows substantial gains over its
weaker baseline. Evaluating the same films across ten national systems reveals
meaningful country-specific calibration differences that a single generic
rating task would obscure.

\vspace{-2.25pt}
\paragraph{Evidence grounding reveals further headroom.}
Film-facing safety assessment requires both a judgment and evidence supporting
it. While strong profanity approaches near-ceiling grounding for the best
models, mild obscenity reaches only 52.0\% strict F1 for Gemini and 39.7\% for
Sol. By jointly evaluating narrative understanding, multilingual consistency,
cultural calibration, and evidence grounding, \cinesubbench{} shows that current
LLMs remain far from uniformly reliable for film understanding and provides a
common benchmark for measuring progress.


\section{Conclusion}
\label{sec:conclusion}

We introduced \cinesubbench{}, a benchmark for long-form film understanding
under a multi-task, multilingual, and multicultural evaluation setting.
Across nine LLMs, premise recovery is more reliable than complete narrative
reconstruction, cross-lingual consistency varies across models and languages,
national classification systems reveal distinct calibration behavior, and
evidence-grounded language safety remains strongly category-dependent. These
findings show substantial headroom for current models across narrative,
linguistic, cultural, and evidence-grounding dimensions of film understanding.
By combining film-length subtitles with matched task, language, and cultural
coverage, \cinesubbench{} provides a unified benchmark for measuring progress
across these capabilities.


\section*{Ethics Statement}
\label{sec:ethics-statement}

\paragraph{Research scope and public data.}
\cinesubbench{} is intended for non-commercial research and benchmarking.
Its construction uses publicly accessible film metadata, audience-guidance
information, national classification labels, and community-contributed subtitle
files. Source roles and provenance are documented in
Appendix~\ref{app:source-roles}, with subtitle acquisition and recovery detailed
in Appendix~\ref{app:subtitle-acquisition}. The benchmark uses these resources
to study model capabilities over film-length linguistic evidence and does not
distribute the underlying audiovisual works.

\paragraph{Subtitle-based film evaluation.}
Film understanding poses a distinctive data-access challenge because the
underlying audiovisual works are copyrighted and costly to store, process, and
redistribute at scale. Existing audiovisual benchmarks such as MovieQA,
LongVideoBench, InfiniBench, MF$^2$, and ShotBench commonly
operate over selected films, clips, frames, or externally obtained video
assets~\citep{tapaswi2016movieqa,wu2024longvideobench,
ataallah2025infinibench,zaranis2025mf2,liu2025shotbench,
wang2025cinetechbench}. \cinesubbench{} instead builds on publicly accessible,
community-contributed subtitle tracks, enabling film-length evaluation over
1,012 films without redistributing movie video. This representation also makes
matched evaluation across six languages, ten national classification systems,
and diverse LLM families practical while retaining temporally ordered
narrative evidence.

\paragraph{Privacy and data sanitization.}
We apply an automated, auditable sanitization procedure to all six subtitle
tracks as part of the quality-assurance pipeline
(Appendix~\ref{app:subtitle-qa}). Across 8,133,088 subtitle entries, 6,594
non-narrative distribution artifacts are removed, including subtitle-team
credits, contact details, social-media handles, advertisements, download
promotions, and external promotional links. In 65 otherwise narrative-relevant
entries containing an email address, the surrounding text is retained and only
the address is replaced with \texttt{[email redacted]}. Film identifiers,
timestamps, language metadata, and original subtitle indices remain unchanged. 

\paragraph{Content and intended use.}
Films naturally contain profanity, mature themes, and other sensitive language
relevant to the benchmark tasks. Language-safety annotations preserve this
task-relevant evidence while localizing it through subtitle indices rather than
reproducing offensive expressions in the label fields;
Appendix~\ref{app:language-gold-construction} documents their construction and
audit. Field-level provenance and the released data structure are described in
Appendix~\ref{app:dataset-schema}. \cinesubbench{} is intended for research on
long-context, multilingual, and culturally situated film understanding.


\bibliography{arxiv_neutral}

@inproceedings{papineni-etal-2002-bleu,
    title = "{B}leu: a Method for Automatic Evaluation of Machine Translation",
    author = "Papineni, Kishore  and
      Roukos, Salim  and
      Ward, Todd  and
      Zhu, Wei-Jing",
    editor = "Isabelle, Pierre  and
      Charniak, Eugene  and
      Lin, Dekang",
    booktitle = "Proceedings of the 40th Annual Meeting of the Association for Computational Linguistics",
    month = jul,
    year = "2002",
    address = "Philadelphia, Pennsylvania, USA",
    publisher = "Association for Computational Linguistics",
    url = "https://aclanthology.org/P02-1040/",
    doi = "10.3115/1073083.1073135",
    pages = "311--318"
}

@inproceedings{lin-2004-rouge,
    title = "{ROUGE}: A Package for Automatic Evaluation of Summaries",
    author = "Lin, Chin-Yew",
    booktitle = "Text Summarization Branches Out",
    month = jul,
    year = "2004",
    address = "Barcelona, Spain",
    publisher = "Association for Computational Linguistics",
    url = "https://aclanthology.org/W04-1013/",
    pages = "74--81"
}

@article{fabbri-etal-2021-summeval,
    title = "{S}umm{E}val: Re-evaluating Summarization Evaluation",
    author = "Fabbri, Alexander R.  and
      Kry{\'s}ci{\'n}ski, Wojciech  and
      McCann, Bryan  and
      Xiong, Caiming  and
      Socher, Richard  and
      Radev, Dragomir",
    editor = "Roark, Brian  and
      Nenkova, Ani",
    journal = "Transactions of the Association for Computational Linguistics",
    volume = "9",
    year = "2021",
    address = "Cambridge, MA",
    publisher = "MIT Press",
    url = "https://aclanthology.org/2021.tacl-1.24/",
    doi = "10.1162/tacl_a_00373",
    pages = "391--409",
}

@inproceedings{zhang-bansal-2021-finding,
    title = "Finding a Balanced Degree of Automation for Summary Evaluation",
    author = "Zhang, Shiyue  and
      Bansal, Mohit",
    editor = "Moens, Marie-Francine  and
      Huang, Xuanjing  and
      Specia, Lucia  and
      Yih, Scott Wen-tau",
    booktitle = "Proceedings of the 2021 Conference on Empirical Methods in Natural Language Processing",
    month = nov,
    year = "2021",
    address = "Online and Punta Cana, Dominican Republic",
    publisher = "Association for Computational Linguistics",
    url = "https://aclanthology.org/2021.emnlp-main.531/",
    doi = "10.18653/v1/2021.emnlp-main.531",
    pages = "6617--6632",
}

@InProceedings{tapaswi2016movieqa,
author = {Tapaswi, Makarand and Zhu, Yukun and Stiefelhagen, Rainer and Torralba, Antonio and Urtasun, Raquel and Fidler, Sanja},
title = {MovieQA: Understanding Stories in Movies Through Question-Answering},
booktitle = {Proceedings of the IEEE Conference on Computer Vision and Pattern Recognition (CVPR)},
month = {June},
year = {2016},
pages = {4631--4640},
url = {https://openaccess.thecvf.com/content_cvpr_2016/html/Tapaswi_MovieQA_Understanding_Stories_CVPR_2016_paper.html}
}

@inproceedings{wu2024longvideobench,
 author = {Wu, Haoning and Li, Dongxu and Chen, Bei and Li, Junnan},
 booktitle = {Advances in Neural Information Processing Systems},
 doi = {10.52202/079017-0907},
 editor = {A. Globerson and L. Mackey and D. Belgrave and A. Fan and U. Paquet and J. Tomczak and C. Zhang},
 pages = {28828--28857},
 publisher = {Curran Associates, Inc.},
 title = {LongVideoBench: A Benchmark for Long-context Interleaved Video-Language Understanding},
 url = {https://proceedings.neurips.cc/paper_files/paper/2024/file/329ad516cf7a6ac306f29882e9c77558-Paper-Datasets_and_Benchmarks_Track.pdf},
 volume = {37},
 year = {2024}
}

@inproceedings{ataallah2025infinibench,
    title = "{I}nfini{B}ench: A Benchmark for Large Multi-Modal Models in Long-Form Movies and {TV} Shows",
    author = "Ataallah, Kirolos  and
      Bakr, Eslam Mohamed  and
      Ahmed, Mahmoud  and
      Gou, Chenhui  and
      Pahwa, Khushbu  and
      Ding, Jian  and
      Elhoseiny, Mohamed",
    editor = "Christodoulopoulos, Christos  and
      Chakraborty, Tanmoy  and
      Rose, Carolyn  and
      Peng, Violet",
    booktitle = "Proceedings of the 2025 Conference on Empirical Methods in Natural Language Processing",
    month = nov,
    year = "2025",
    address = "Suzhou, China",
    publisher = "Association for Computational Linguistics",
    url = "https://aclanthology.org/2025.emnlp-main.984/",
    doi = "10.18653/v1/2025.emnlp-main.984",
    pages = "19485--19512",
    ISBN = "979-8-89176-332-6",
}

@misc{zaranis2025mf2,
      title={Movie Facts and Fibs (MF$^2$): A Benchmark for Long Movie Understanding}, 
      author={Emmanouil Zaranis and António Farinhas and Saul Santos and Beatriz Canaverde and Miguel Moura Ramos and Aditya K Surikuchi and André Viveiros and Baohao Liao and Elena Bueno-Benito and Nithin Sivakumaran and Pavlo Vasylenko and Shoubin Yu and Sonal Sannigrahi and Wafaa Mohammed and Ben Peters and Danae Sánchez Villegas and Elias Stengel-Eskin and Giuseppe Attanasio and Jaehong Yoon and Stella Frank and Alessandro Suglia and Chrysoula Zerva and Desmond Elliott and Mariella Dimiccoli and Mohit Bansal and Oswald Lanz and Raffaella Bernardi and Raquel Fernández and Sandro Pezzelle and Vlad Niculae and André F. T. Martins},
      year={2025},
      eprint={2506.06275},
      archivePrefix={arXiv},
      primaryClass={cs.CV},
      url={https://arxiv.org/abs/2506.06275}, 
}

@inproceedings{liu2025shotbench,
title={ShotBench: Expert-Level Cinematic Understanding in Vision-Language Models},
author={Hongbo Liu and Jingwen He and Yi Jinn and Dian Zheng and Yuhao Dong and Fan Zhang and Ziqi Huang and Yinan He and Weichao Chen and Yu Qiao and Wanli Ouyang and Shengjie Zhao and Ziwei Liu},
booktitle={The Thirty-ninth Annual Conference on Neural Information Processing Systems},
year={2025},
url={https://openreview.net/forum?id=1DgSkx8L63}
}

@inproceedings{wang2025cinetechbench,
title={CineTechBench: A Benchmark for Cinematographic Technique Understanding and Generation},
author={Xinran Wang and Songyu Xu and Shan Xiangxuan and Yuxuan Zhang and Muxi Diao and Xueyan Duan and Yanhua huang and Kongming Liang and Zhanyu Ma},
booktitle={The Thirty-ninth Annual Conference on Neural Information Processing Systems Datasets and Benchmarks Track},
year={2026},
url={https://openreview.net/forum?id=LBKyjz2ESc}
}

@inproceedings{shen2025movieadaptation,
    title = "Adapting Large Language Models for Movie Domain with Narrative Understanding Tasks",
    author = "Shen, Siqi  and
      Garg, Amanmeet",
    editor = "Boleda, Gemma  and
      Roth, Michael",
    booktitle = "Proceedings of the 29th Conference on Computational Natural Language Learning",
    month = jul,
    year = "2025",
    address = "Vienna, Austria",
    publisher = "Association for Computational Linguistics",
    url = "https://aclanthology.org/2025.conll-1.13/",
    doi = "10.18653/v1/2025.conll-1.13",
    pages = "187--200",
    ISBN = "979-8-89176-271-8",
}

@inproceedings{saxena2024moviesum,
    title = "{M}ovie{S}um: An Abstractive Summarization Dataset for Movie Screenplays",
    author = "Saxena, Rohit  and
      Keller, Frank",
    editor = "Ku, Lun-Wei  and
      Martins, Andre  and
      Srikumar, Vivek",
    booktitle = "Findings of the Association for Computational Linguistics: ACL 2024",
    month = aug,
    year = "2024",
    address = "Bangkok, Thailand",
    publisher = "Association for Computational Linguistics",
    url = "https://aclanthology.org/2024.findings-acl.239/",
    doi = "10.18653/v1/2024.findings-acl.239",
    pages = "4043--4050",
}

@inproceedings{shafaei2020age,
    title = "Age Suitability Rating: Predicting the {MPAA} Rating Based on Movie Dialogues",
    author = "Shafaei, Mahsa  and
      Safi Samghabadi, Niloofar  and
      Kar, Sudipta  and
      Solorio, Thamar",
    editor = "Calzolari, Nicoletta  and
      B{\'e}chet, Fr{\'e}d{\'e}ric  and
      Blache, Philippe  and
      Choukri, Khalid  and
      Cieri, Christopher  and
      Declerck, Thierry  and
      Goggi, Sara  and
      Isahara, Hitoshi  and
      Maegaard, Bente  and
      Mariani, Joseph  and
      Mazo, H{\'e}l{\`e}ne  and
      Moreno, Asuncion  and
      Odijk, Jan  and
      Piperidis, Stelios",
    booktitle = "Proceedings of the Twelfth Language Resources and Evaluation Conference",
    month = may,
    year = "2020",
    address = "Marseille, France",
    publisher = "European Language Resources Association",
    url = "https://aclanthology.org/2020.lrec-1.166/",
    pages = "1327--1335",
    language = "eng",
    ISBN = "979-10-95546-34-4",
}

@inproceedings{zhang2021severity,
    title = "From None to Severe: {P}redicting Severity in Movie Scripts",
    author = "Zhang, Yigeng  and
      Shafaei, Mahsa  and
      Gonzalez, Fabio  and
      Solorio, Thamar",
    editor = "Moens, Marie-Francine  and
      Huang, Xuanjing  and
      Specia, Lucia  and
      Yih, Scott Wen-tau",
    booktitle = "Findings of the Association for Computational Linguistics: EMNLP 2021",
    month = nov,
    year = "2021",
    address = "Punta Cana, Dominican Republic",
    publisher = "Association for Computational Linguistics",
    url = "https://aclanthology.org/2021.findings-emnlp.332/",
    doi = "10.18653/v1/2021.findings-emnlp.332",
    pages = "3951--3956",
}

@inproceedings{liu2023geval,
    title = "{G}-Eval: {NLG} Evaluation using Gpt-4 with Better Human Alignment",
    author = "Liu, Yang  and
      Iter, Dan  and
      Xu, Yichong  and
      Wang, Shuohang  and
      Xu, Ruochen  and
      Zhu, Chenguang",
    editor = "Bouamor, Houda  and
      Pino, Juan  and
      Bali, Kalika",
    booktitle = "Proceedings of the 2023 Conference on Empirical Methods in Natural Language Processing",
    month = dec,
    year = "2023",
    address = "Singapore",
    publisher = "Association for Computational Linguistics",
    url = "https://aclanthology.org/2023.emnlp-main.153/",
    doi = "10.18653/v1/2023.emnlp-main.153",
    pages = "2511--2522",
}

@article{gu2024surveyjudge,
title = {A survey on LLM-as-a-judge},
journal = {The Innovation},
volume = {7},
number = {6},
pages = {101253},
year = {2026},
issn = {2666-6758},
doi = {https://doi.org/10.1016/j.xinn.2025.101253},
url = {https://www.sciencedirect.com/science/article/pii/S2666675825004564},
author = {Jiawei Gu and Xuhui Jiang and Zhichao Shi and Hexiang Tan and Xuehao Zhai and Chengjin Xu and Wei Li and Yinghan Shen and Shengjie Ma and Honghao Liu and Saizhuo Wang and Kun Zhang and Zhouchi Lin and Bowen Zhang and Lionel Ni and Wen Gao and Yuanzhuo Wang and Jian Guo}
}

@inproceedings{li2025generationJudgment,
    title = "From Generation to Judgment: Opportunities and Challenges of {LLM}-as-a-judge",
    author = "Li, Dawei  and
      Jiang, Bohan  and
      Huang, Liangjie  and
      Beigi, Alimohammad  and
      Zhao, Chengshuai  and
      Tan, Zhen  and
      Bhattacharjee, Amrita  and
      Jiang, Yuxuan  and
      Chen, Canyu  and
      Wu, Tianhao  and
      Shu, Kai  and
      Cheng, Lu  and
      Liu, Huan",
    editor = "Christodoulopoulos, Christos  and
      Chakraborty, Tanmoy  and
      Rose, Carolyn  and
      Peng, Violet",
    booktitle = "Proceedings of the 2025 Conference on Empirical Methods in Natural Language Processing",
    month = nov,
    year = "2025",
    address = "Suzhou, China",
    publisher = "Association for Computational Linguistics",
    url = "https://aclanthology.org/2025.emnlp-main.138/",
    doi = "10.18653/v1/2025.emnlp-main.138",
    pages = "2757--2791",
    ISBN = "979-8-89176-332-6",
}

@article{liu2024lost,
    title = "Lost in the Middle: How Language Models Use Long Contexts",
    author = "Liu, Nelson F.  and
      Lin, Kevin  and
      Hewitt, John  and
      Paranjape, Ashwin  and
      Bevilacqua, Michele  and
      Petroni, Fabio  and
      Liang, Percy",
    journal = "Transactions of the Association for Computational Linguistics",
    volume = "12",
    year = "2024",
    address = "Cambridge, MA",
    publisher = "MIT Press",
    url = "https://aclanthology.org/2024.tacl-1.9/",
    doi = "10.1162/tacl_a_00638",
    pages = "157--173",
}

@inproceedings{goldman2024longcontext,
    title = "Is It Really Long Context if All You Need Is Retrieval? Towards Genuinely Difficult Long Context {NLP}",
    author = "Goldman, Omer  and
      Jacovi, Alon  and
      Slobodkin, Aviv  and
      Maimon, Aviya  and
      Dagan, Ido  and
      Tsarfaty, Reut",
    editor = "Al-Onaizan, Yaser  and
      Bansal, Mohit  and
      Chen, Yun-Nung",
    booktitle = "Proceedings of the 2024 Conference on Empirical Methods in Natural Language Processing",
    month = nov,
    year = "2024",
    address = "Miami, Florida, USA",
    publisher = "Association for Computational Linguistics",
    url = "https://aclanthology.org/2024.emnlp-main.924/",
    doi = "10.18653/v1/2024.emnlp-main.924",
    pages = "16576--16586",
}

@inproceedings{hamilton2026narrabench,
    title = "{N}arra{B}ench: A Comprehensive Framework for Narrative Benchmarking",
    author = "Hamilton, Sil  and
      Wilkens, Matthew  and
      Piper, Andrew",
    editor = "Demberg, Vera  and
      Inui, Kentaro  and
      Marquez, Llu{\'i}s",
    booktitle = "Proceedings of the 19th Conference of the {E}uropean Chapter of the {A}ssociation for {C}omputational {L}inguistics (Volume 1: Long Papers)",
    month = mar,
    year = "2026",
    address = "Rabat, Morocco",
    publisher = "Association for Computational Linguistics",
    url = "https://aclanthology.org/2026.eacl-long.176/",
    doi = "10.18653/v1/2026.eacl-long.176",
    pages = "3786--3801",
    ISBN = "979-8-89176-380-7",
}

@inproceedings{kryscinski2022booksum,
    title = "{BOOKSUM}: A Collection of Datasets for Long-form Narrative Summarization",
    author = "Kryscinski, Wojciech  and
      Rajani, Nazneen  and
      Agarwal, Divyansh  and
      Xiong, Caiming  and
      Radev, Dragomir",
    editor = "Goldberg, Yoav  and
      Kozareva, Zornitsa  and
      Zhang, Yue",
    booktitle = "Findings of the Association for Computational Linguistics: EMNLP 2022",
    month = dec,
    year = "2022",
    address = "Abu Dhabi, United Arab Emirates",
    publisher = "Association for Computational Linguistics",
    url = "https://aclanthology.org/2022.findings-emnlp.488/",
    doi = "10.18653/v1/2022.findings-emnlp.488",
    pages = "6536--6558",
}

@inproceedings{kim2024fables,
title={{FABLES}: Evaluating faithfulness and content selection in book-length summarization},
author={Yekyung Kim and Yapei Chang and Marzena Karpinska and Aparna Garimella and Varun Manjunatha and Kyle Lo and Tanya Goyal and Mohit Iyyer},
booktitle={First Conference on Language Modeling},
year={2024},
url={https://openreview.net/forum?id=YfHxQSoaWU}
}

@inproceedings{chen2022summscreen,
    title = "{S}umm{S}creen: A Dataset for Abstractive Screenplay Summarization",
    author = "Chen, Mingda  and
      Chu, Zewei  and
      Wiseman, Sam  and
      Gimpel, Kevin",
    editor = "Muresan, Smaranda  and
      Nakov, Preslav  and
      Villavicencio, Aline",
    booktitle = "Proceedings of the 60th Annual Meeting of the Association for Computational Linguistics (Volume 1: Long Papers)",
    month = may,
    year = "2022",
    address = "Dublin, Ireland",
    publisher = "Association for Computational Linguistics",
    url = "https://aclanthology.org/2022.acl-long.589/",
    doi = "10.18653/v1/2022.acl-long.589",
    pages = "8602--8615",
}

@inproceedings{kim2025oneruler,
title={One ruler to measure them all: Benchmarking multilingual long-context language models},
author={Yekyung Kim and Jenna Russell and Marzena Karpinska and Mohit Iyyer},
booktitle={Second Conference on Language Modeling},
year={2025},
url={https://openreview.net/forum?id=3vxxB3Ar9r}
}

@inproceedings{arora2025calmqa,
    title = "{C}a{LMQA}: Exploring culturally specific long-form question answering across 23 languages",
    author = "Arora, Shane  and
      Karpinska, Marzena  and
      Chen, Hung-Ting  and
      Bhattacharjee, Ipsita  and
      Iyyer, Mohit  and
      Choi, Eunsol",
    editor = "Che, Wanxiang  and
      Nabende, Joyce  and
      Shutova, Ekaterina  and
      Pilehvar, Mohammad Taher",
    booktitle = "Proceedings of the 63rd Annual Meeting of the Association for Computational Linguistics (Volume 1: Long Papers)",
    month = jul,
    year = "2025",
    address = "Vienna, Austria",
    publisher = "Association for Computational Linguistics",
    url = "https://aclanthology.org/2025.acl-long.578/",
    doi = "10.18653/v1/2025.acl-long.578",
    pages = "11772--11817",
    ISBN = "979-8-89176-251-0",
}

@inproceedings{steen2019timeline,
    title = "Abstractive Timeline Summarization",
    author = "Steen, Julius  and
      Markert, Katja",
    editor = "Wang, Lu  and
      Cheung, Jackie Chi Kit  and
      Carenini, Giuseppe  and
      Liu, Fei",
    booktitle = "Proceedings of the 2nd Workshop on New Frontiers in Summarization",
    month = nov,
    year = "2019",
    address = "Hong Kong, China",
    publisher = "Association for Computational Linguistics",
    url = "https://aclanthology.org/D19-5403/",
    doi = "10.18653/v1/D19-5403",
    pages = "21--31",
}

@inproceedings{guha2023legalbench,
 author = {Guha, Neel and Nyarko, Julian and Ho, Daniel and R\'{e}, Christopher and Chilton, Adam and K, Aditya and Chohlas-Wood, Alex and Peters, Austin and Waldon, Brandon and Rockmore, Daniel and Zambrano, Diego and Talisman, Dmitry and Hoque, Enam and Surani, Faiz and Fagan, Frank and Sarfaty, Galit and Dickinson, Gregory and Porat, Haggai and Hegland, Jason and Wu, Jessica and Nudell, Joe and Niklaus, Joel and Nay, John and Choi, Jonathan and Tobia, Kevin and Hagan, Margaret and Ma, Megan and Livermore, Michael and Rasumov-Rahe, Nikon and Holzenberger, Nils and Kolt, Noam and Henderson, Peter and Rehaag, Sean and Goel, Sharad and Gao, Shang and Williams, Spencer and Gandhi, Sunny and Zur, Tom and Iyer, Varun and Li, Zehua},
 booktitle = {Advances in Neural Information Processing Systems},
 doi = {10.52202/075280-1915},
 editor = {A. Oh and T. Naumann and A. Globerson and K. Saenko and M. Hardt and S. Levine},
 pages = {44123--44279},
 publisher = {Curran Associates, Inc.},
 title = {LegalBench: A Collaboratively Built Benchmark for Measuring Legal Reasoning in Large Language Models},
 url = {https://proceedings.neurips.cc/paper_files/paper/2023/file/89e44582fd28ddfea1ea4dcb0ebbf4b0-Paper-Datasets_and_Benchmarks.pdf},
 volume = {36},
 year = {2023}
}

@article{singhal2023multimedqa,
  author  = {Singhal, Karan and
             Azizi, Shekoofeh and
             Tu, Tao and
             Mahdavi, S. Sara and
             Wei, Jason and
             Chung, Hyung Won and
             Scales, Nathan and
             Tanwani, Ajay and
             Cole-Lewis, Heather and
             Pfohl, Stephen and
             Payne, Perry and
             Seneviratne, Martin and
             Gamble, Paul and
             Kelly, Chris and
             Babiker, Abubakr and
             Sch{\"a}rli, Nathanael and
             Chowdhery, Aakanksha and
             Mansfield, Philip and
             Demner-Fushman, Dina and
             Ag{\"u}era y Arcas, Blaise and
             Webster, Dale and
             Corrado, Greg S. and
             Matias, Yossi and
             Chou, Katherine and
             Gottweis, Juraj and
             Tomasev, Nenad and
             Liu, Yun and
             Rajkomar, Alvin and
             Barral, Joelle and
             Semturs, Christopher and
             Karthikesalingam, Alan and
             Natarajan, Vivek},
  title   = {Large language models encode clinical knowledge},
  journal = {Nature},
  year    = {2023},
  volume  = {620},
  number  = {7972},
  pages   = {172--180},
  doi     = {10.1038/s41586-023-06291-2},
  url     = {https://doi.org/10.1038/s41586-023-06291-2},
  issn    = {1476-4687}
}

@inproceedings{cao2024financial,
title={Characterizing Multimodal Long-form Summarization: A Case Study on Financial Reports},
author={Tianyu Cao and Natraj Raman and Danial Dervovic and Chenhao Tan},
booktitle={First Conference on Language Modeling},
year={2024},
url={https://openreview.net/forum?id=hDoN0CAy5e}
}

@inproceedings{jimenez2024swebench,
title={{SWE}-bench: Can Language Models Resolve Real-world Github Issues?},
author={Carlos E Jimenez and John Yang and Alexander Wettig and Shunyu Yao and Kexin Pei and Ofir Press and Karthik R Narasimhan},
booktitle={The Twelfth International Conference on Learning Representations},
year={2024},
url={https://openreview.net/forum?id=VTF8yNQM66}
}

@inproceedings{zhang2024tablellama,
    title = "{T}able{L}lama: Towards Open Large Generalist Models for Tables",
    author = "Zhang, Tianshu  and
      Yue, Xiang  and
      Li, Yifei  and
      Sun, Huan",
    editor = "Duh, Kevin  and
      Gomez, Helena  and
      Bethard, Steven",
    booktitle = "Proceedings of the 2024 Conference of the North American Chapter of the Association for Computational Linguistics: Human Language Technologies (Volume 1: Long Papers)",
    month = jun,
    year = "2024",
    address = "Mexico City, Mexico",
    publisher = "Association for Computational Linguistics",
    url = "https://aclanthology.org/2024.naacl-long.335/",
    doi = "10.18653/v1/2024.naacl-long.335",
    pages = "6024--6044",
}

@inproceedings{wang2026cybergym,
title={CyberGym: Evaluating {AI} Agents' Real-World Cybersecurity Capabilities at Scale},
author={Zhun Wang and Tianneng Shi and Jingxuan He and Matthew Cai and Jialin Zhang and Dawn Song},
booktitle={The Fourteenth International Conference on Learning Representations},
year={2026},
url={https://openreview.net/forum?id=2YvbLQEdYt}
}

@InProceedings{dihan2025mapeval,
  title = 	 {{M}ap{E}val: A Map-Based Evaluation of Geo-Spatial Reasoning in Foundation Models},
  author =       {Dihan, Mahir Labib and Hassan, Md Tanvir and Parvez, Md Tanvir and Hasan, Md Hasebul and Alam, Md Almash and Cheema, Muhammad Aamir and Ali, Mohammed Eunus and Parvez, Md Rizwan},
  booktitle = 	 {Proceedings of the 42nd International Conference on Machine Learning},
  pages = 	 {13774--13813},
  year = 	 {2025},
  editor = 	 {Singh, Aarti and Fazel, Maryam and Hsu, Daniel and Lacoste-Julien, Simon and Berkenkamp, Felix and Maharaj, Tegan and Wagstaff, Kiri and Zhu, Jerry},
  volume = 	 {267},
  series = 	 {Proceedings of Machine Learning Research},
  month = 	 {13--19 Jul},
  publisher =    {PMLR},
  url = 	 {https://proceedings.mlr.press/v267/dihan25a.html}
}

@inproceedings{awal2025webmmu,
    title = "{W}eb{MMU}: A Benchmark for Multimodal Multilingual Website Understanding and Code Generation",
    author = "Awal, Rabiul  and
      Massoud, Mahsa  and
      Feizi, Aarash  and
      Li, Zichao  and
      Wang, Suyuchen  and
      Pal, Christopher  and
      Agrawal, Aishwarya  and
      Vazquez, David  and
      Reddy, Siva  and
      Rodriguez, Juan A.  and
      Taslakian, Perouz  and
      Gella, Spandana  and
      Rajeswar, Sai",
    editor = "Christodoulopoulos, Christos  and
      Chakraborty, Tanmoy  and
      Rose, Carolyn  and
      Peng, Violet",
    booktitle = "Proceedings of the 2025 Conference on Empirical Methods in Natural Language Processing",
    month = nov,
    year = "2025",
    address = "Suzhou, China",
    publisher = "Association for Computational Linguistics",
    url = "https://aclanthology.org/2025.emnlp-main.1276/",
    doi = "10.18653/v1/2025.emnlp-main.1276",
    pages = "25118--25145",
    ISBN = "979-8-89176-332-6",
}

@inproceedings{guo2024econnli,
    title = "{E}con{NLI}: Evaluating Large Language Models on Economics Reasoning",
    author = "Guo, Yue  and
      Yang, Yi",
    editor = "Ku, Lun-Wei  and
      Martins, Andre  and
      Srikumar, Vivek",
    booktitle = "Findings of the Association for Computational Linguistics: ACL 2024",
    month = aug,
    year = "2024",
    address = "Bangkok, Thailand",
    publisher = "Association for Computational Linguistics",
    url = "https://aclanthology.org/2024.findings-acl.58/",
    doi = "10.18653/v1/2024.findings-acl.58",
    pages = "982--994",
}

@inproceedings{duan2025scigym,
title={Measuring Scientific Capabilities of Language Models with a Systems Biology Dry Lab},
author={Haonan Duan and Stephen Zhewen Lu and Caitlin Fiona Harrigan and Nishkrit Desai and Jiarui Lu and Micha{\l} Koziarski and Leonardo Cotta and Chris J. Maddison},
booktitle={The Thirty-ninth Annual Conference on Neural Information Processing Systems Datasets and Benchmarks Track},
year={2026},
url={https://openreview.net/forum?id=Cmx6b7w2nk}
}

@inproceedings{hou2024eeval,
    title = "{E}-{EVAL}: A Comprehensive {C}hinese K-12 Education Evaluation Benchmark for Large Language Models",
    author = "Hou, Jinchang  and
      Ao, Chang  and
      Wu, Haihong  and
      Kong, Xiangtao  and
      Zheng, Zhigang  and
      Tang, Daijia  and
      Li, Chengming  and
      Hu, Xiping  and
      Xu, Ruifeng  and
      Ni, Shiwen  and
      Yang, Min",
    editor = "Ku, Lun-Wei  and
      Martins, Andre  and
      Srikumar, Vivek",
    booktitle = "Findings of the Association for Computational Linguistics: ACL 2024",
    month = aug,
    year = "2024",
    address = "Bangkok, Thailand",
    publisher = "Association for Computational Linguistics",
    url = "https://aclanthology.org/2024.findings-acl.462/",
    doi = "10.18653/v1/2024.findings-acl.462",
    pages = "7753--7774",
}

@inproceedings{li2024newsbench,
    title = "{N}ews{B}ench: A Systematic Evaluation Framework for Assessing Editorial Capabilities of Large Language Models in {C}hinese Journalism",
    author = "Li, Miao  and
      Chen, Ming-Bin  and
      Tang, Bo  and
      Hou, Shengbin  and
      Wang, Pengyu  and
      Deng, Haiying  and
      Li, Zhiyu  and
      Xiong, Feiyu  and
      Mao, Keming  and
      Cheng, Peng  and
      Luo, Yi",
    editor = "Ku, Lun-Wei  and
      Martins, Andre  and
      Srikumar, Vivek",
    booktitle = "Proceedings of the 62nd Annual Meeting of the Association for Computational Linguistics (Volume 1: Long Papers)",
    month = aug,
    year = "2024",
    address = "Bangkok, Thailand",
    publisher = "Association for Computational Linguistics",
    url = "https://aclanthology.org/2024.acl-long.538/",
    doi = "10.18653/v1/2024.acl-long.538",
    pages = "9993--10014",
}
\bibliographystyle{arxiv_neutral}


\clearpage
\appendix
\onecolumn
\begin{center}
  {\Large\bfseries Supplementary Material: Appendices}\\[18pt]
\end{center}

\begin{center}
\begin{minipage}{0.92\linewidth}
\noindent{\large\bfseries Appendix Contents}\vspace{7pt}

\noindent\hyperref[app:dataset-docs]{\textbf{A\quad Dataset Construction and Quality Assurance}}\\[-1pt]
\hspace*{1.8em}\hyperref[app:source-roles]{A.1\quad Source roles and construction overview}\\
\hspace*{1.8em}\hyperref[app:entity-filtering]{A.2\quad Entity resolution and progressive filtering}\\
\hspace*{1.8em}\hyperref[app:subtitle-acquisition]{A.3\quad Subtitle acquisition, provenance, and recovery}\\
\hspace*{1.8em}\hyperref[app:coverage-optimization]{A.4\quad Complete-coverage optimization over languages and countries}\\
\hspace*{1.8em}\hyperref[app:subtitle-qa]{A.5\quad Multi-layer subtitle verification and manual audit}\\
\hspace*{1.8em}\hyperref[app:language-gold-construction]{A.6\quad Subtitle-grounded language-safety gold construction and audit}\\
\hspace*{1.8em}\hyperref[app:label-curation]{A.7\quad Task-oriented label curation and normalization}\\
\hspace*{1.8em}\hyperref[app:dataset-schema]{A.8\quad Final film-entity structure and field provenance}\\
\hspace*{1.8em}\hyperref[app:dataset-statistics]{A.9\quad Dataset statistics and coverage audits}\vspace{5pt}

\noindent\hyperref[app:evaluation-design]{\textbf{B\quad Evaluation Design and Reproducibility}}\\[-1pt]
\hspace*{1.8em}\hyperref[app:model-coverage]{B.1\quad Model coverage and generation settings}\\
\hspace*{1.8em}\hyperref[app:metric-definitions]{B.2\quad Metric definitions}\\
\hspace*{1.8em}\hyperref[app:prompt-structures]{B.3\quad Task and evaluator instructions}\\
\hspace*{1.8em}\hyperref[app:independent-judge-agreement]{B.4\quad Independent narrative-judge agreement}\\
\hspace*{1.8em}\hyperref[app:structured-output-recovery-audit]{B.5\quad Response-format reliability}\\
\hspace*{1.8em}\hyperref[app:source-exposure-audit]{B.6\quad Title/year recall and character-name counterfactuals}\vspace{5pt}

\noindent\hyperref[app:rq1-details]{\textbf{C\quad Additional Evidence for RQ~\ref{rq:model-ranking}: Model Capability Profiles}}\\[-1pt]
\hspace*{1.8em}\hyperref[app:rq1-headline-results]{C.1\quad English and cross-lingual headline results}\\
\hspace*{1.8em}\hyperref[app:composite-significance]{C.2\quad Paired comparisons of leading composite scores}\\
\hspace*{1.8em}\hyperref[app:narrative-error-analysis]{C.3\quad Narrative failure patterns}\vspace{5pt}

\noindent\hyperref[app:rq2-details]{\textbf{D\quad Additional Evidence for RQ~\ref{rq:crosslingual}: Cross-Lingual Behavior}}\\[-1pt]
\hspace*{1.8em}\hyperref[app:rq2-overview]{D.1\quad English-to-cross-lingual overview}\\
\hspace*{1.8em}\hyperref[app:rq2-narrative]{D.2\quad Narrative scores by subtitle language}\\
\hspace*{1.8em}\hyperref[app:rq2-genre]{D.3\quad Genre stability and label-specific failures}\\
\hspace*{1.8em}\hyperref[app:rq2-uncertainty]{D.4\quad Paired cross-lingual uncertainty}\vspace{5pt}

\noindent\hyperref[app:rq3-details]{\textbf{E\quad Additional Evidence for RQ~\ref{rq:cultural-errors}: Age and National-Rating Calibration}}\\[-1pt]
\hspace*{1.8em}\hyperref[app:rq3-age]{E.1\quad Age error magnitude and direction}\\
\hspace*{1.8em}\hyperref[app:cultural-country-breakdown]{E.2\quad Country-level calibration and language sensitivity}\vspace{5pt}

\noindent\hyperref[app:rq4-details]{\textbf{F\quad Additional Evidence for RQ~\ref{rq:evidence-grounding}: Auditable Language-Safety Assessment}}\\[-1pt]
\hspace*{1.8em}\hyperref[app:language-content-error-analysis]{F.1\quad Safety-evidence errors by category and model}\vspace{5pt}

\noindent\hyperref[app:scaling-analysis]{\textbf{G\quad Additional Analysis: Within-Family Scaling}}\vspace{5pt}

\noindent\hyperref[app:qualitative-cases]{\textbf{H\quad Qualitative Case Studies}}\vspace{5pt}

\noindent\hyperref[app:limitations]{\textbf{I\quad Limitations, Intended Use, and Data Statement}}
\end{minipage}
\end{center}
\vspace{10pt}
\newpage

\section{Dataset Construction and Quality Assurance}
\label{app:dataset-docs}
\subsection{Source Roles and Construction Overview}
\label{app:source-roles}

\paragraph{Construction sequence.}
Construction proceeds through six stages: source-inventory acquisition, cross-source entity resolution, task-metadata filtering, complete subtitle-language coverage selection, complete national-rating coverage selection, and subtitle verification before final record assembly. Tables~\ref{tab:construction-source-roles} and~\ref{tab:construction-gates} identify the source and retention rule at each stage. Automated subtitle flags initiate manual inspection, and unresolved structural or decoding failures prevent ingestion.

\paragraph{Source roles and provenance.}
Kids-in-Mind provides the starting inventory, Common Sense Media supplies age-suitability judgments, IMDb provides film metadata and the shared title identifier, and subtitle repositories provide the model input. Review metadata is retained for provenance and audit, while models receive subtitle text as their only film evidence.

\begin{table*}[htbp]
\centering
\small
\setlength{\tabcolsep}{2.25pt}
\renewcommand{\arraystretch}{1.55}
\caption{Source roles in CineSubBench construction. Each source contributes a distinct component of the benchmark, while the final record preserves provenance links and exposes only subtitle text to models during evaluation.}
\vspace{5pt}
\label{tab:construction-source-roles}
\begin{tabular}{P{0.20\textwidth}P{0.27\textwidth}P{0.39\textwidth}}
\toprule
Source & Primary role & Quality and provenance treatment \\ 
\midrule
\href{https://kids-in-mind.com/}{Kids-in-Mind} & Initial 6,186-film inventory; review-derived language metadata. & The native identifier supports cross-source traceability; records without a usable IMDb link for subtitle retrieval are excluded. \\
\href{https://www.commonsensemedia.org/}{Common Sense Media} & General age-suitability label, storyline, and review page. & Cross-linked through IMDb ID; anomalous embedded IMDb links receive case-by-case manual adjudication before retention. \\
\href{https://www.imdb.com/}{IMDb} & Stable title identifier; plot synopsis, runtime, film metadata, and regional motion-picture ratings. & Provides the entity key used to link sources and subtitle assets; populated plot-synopsis and rating fields are mandatory for the metadata-complete cohort. \\
Subtitle providers & Timestamped SRT files in six languages, with provider metadata. & \href{https://www.opensubtitles.com/}{OpenSubtitles} is the primary source and \href{https://subdl.com/}{SubDL} supports recovery. The highest-download candidate is selected per film and language when alternatives exist; its count is retained in the JSON. Structural and temporal verification remains independent of this popularity proxy. \\
\bottomrule
\end{tabular}
\end{table*}

\subsection{Cross-Source Entity Resolution and Progressive Filtering}
\label{app:entity-filtering}
\begin{enumerate}[leftmargin=*,itemsep=3pt,topsep=3pt]
\item \textbf{Begin with Kids-in-Mind.} The initial scrape produced 6,186 films. We use the IMDb title identifier to query subtitles and join the other sources. For example, \texttt{tt0465602} identifies an IMDb title independently of how its title or year is written on a review page.
\item \textbf{Validate identifiers.} Among 4,307 films with IMDb links, 33 had malformed links or lacked usable subtitle-search parameters. The remaining 4,274 formed the subtitle-acquisition baseline. A shared title and release year alone did not establish a match, because remakes and namesakes can collide.
\item \textbf{Link Common Sense Media.} We queried it for films in that baseline and matched through IMDb IDs. An anomalous page was reviewed manually: if its embedded IMDb link pointed to a different film, the page was not accepted by URL presence alone. The resolved intersection contained 2,831 films.
\item \textbf{Require IMDb task metadata.} We extracted metadata for those linked films and retained records with populated plot-synopsis and motion-picture-rating fields. This left 2,322 candidates for coverage selection.
\end{enumerate}

\begin{table}[H]
\centering
\footnotesize
\caption{Progressive construction gates in CineSubBench. Counts report the number of films retained after each filtering or coverage decision. The detailed subtitle-verification audit is applied to the resulting 1,231-film multilingual cohort.}
\vspace{5pt}
\label{tab:construction-gates}
\begin{tabular}{@{}rP{0.18\textwidth}P{0.48\textwidth}r@{}}
\toprule
Step & Decision & Retention rule & Films \\
\midrule
1 & Source crawl & Kids-in-Mind films collected & 6,186 \\
2 & IMDb link screen & 4,307 films with links; exclude 33 invalid/search-unusable entries & 4,274 \\
3 & Entity linking & IMDb-ID agreement across Kids-in-Mind and Common Sense Media; anomalous links manually reviewed & 2,831 \\
4 & Metadata completeness & Populated IMDb plot synopsis and motion-picture-rating fields & 2,322 \\
5 & Language coverage & Subtitles in all six selected languages & 1,231 \\
6 & Country coverage & Ratings from all ten selected countries in the final benchmark & 1,012 \\
\bottomrule
\end{tabular}
\end{table}

\subsection{Subtitle Acquisition, Provenance, and Recovery}
\label{app:subtitle-acquisition}
Subtitle collection begins from the 4,274-film identifier-validated baseline, while the full verification protocol in Table~\ref{tab:subtitle-verification-layers} is applied to the selected 1,231-film multilingual cohort. The acquisition and recovery procedure is:
\begin{enumerate}[leftmargin=*,itemsep=3pt,topsep=3pt]
\item \textbf{Retrieve and organize.} We retrieved SRT files through the OpenSubtitles Python interface using IMDb identifiers and organized them by film identifier and subtitle language. This identifier-based structure preserved the cross-source entity links established in the preceding stage.
\item \textbf{Choose among candidate files.} For each film and subtitle language with multiple available SRT files, we chose the candidate with the highest download count reported by the source website. Download count serves only as the initial selection heuristic. The selected file's \texttt{download\_count} is retained in the JSON, and every chosen track remains subject to structural, language-consistency, and temporal verification that can trigger manual review or replacement.
\item \textbf{Recover suspect tracks.} Structural, encoding, language-consistency, or timing concerns led to alternative downloads through SubDL and manual checks of SubDL and OpenSubtitles web pages. A curator inspected every track flagged by the automated verification thresholds before deciding whether the anomaly reflected a genuine asset problem.
\item \textbf{Correct the source asset when possible.} A repairable millisecond separator, such as \texttt{00:09:09:949}, could be changed to the SRT form \texttt{00:09:09,949}; dialogue content was never rewritten as part of this repair. When multiple providers showed the same upstream limitation, the best available native asset could be retained after curator review. Unresolved parsing or decoding failures prevented ingestion.
\end{enumerate}

\subsection{Complete-Coverage Optimization over Languages and Countries}
\label{app:coverage-optimization}
Complete coverage is the mechanism that makes the MultiX comparisons matched: every retained film is evaluated over the same six subtitle languages and, after country selection, the same ten national rating systems.

\begin{figure*}[t]
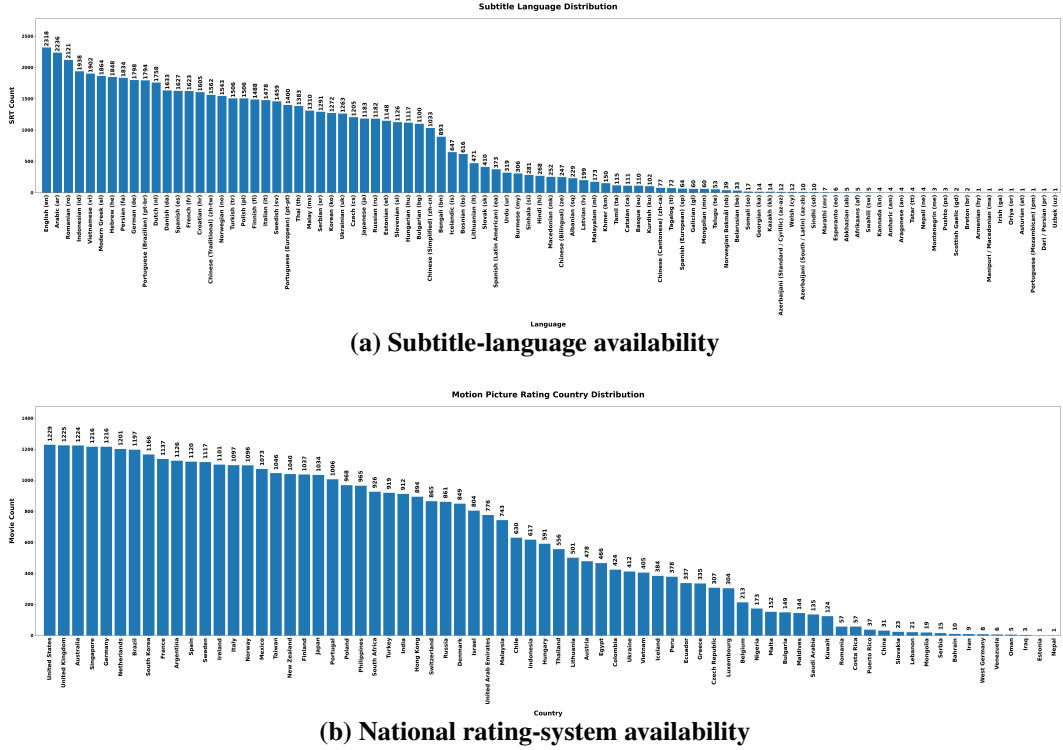

\centering
\begin{minipage}{1.0\textwidth}
  \centering
  \includegraphics[width=\linewidth]{figures/histogram_optimal_lang_selection.png}
  \\[-1mm]\textbf{(a) Subtitle-language availability}
\end{minipage}

\vspace{4mm}

\begin{minipage}{1.0\textwidth}
  \centering
  \includegraphics[width=\linewidth]{figures/histogram_optimal_mpa_country_selection.png}
  \\[-1mm]\textbf{(b) National rating-system availability}
\end{minipage}
\caption{Availability before complete-coverage optimization. Algorithm~\ref{alg:language-set-optimization} is applied to (a) the top 29 subtitle languages over the 2,322 metadata-complete films and (b) the top 21 national rating systems over the 1,231-film multilingual cohort.
\vspace{-3mm}}
\label{fig:coverage-availability}
\end{figure*}

\paragraph{Choosing subtitle languages.} We first apply Algorithm~\ref{alg:language-set-optimization} to the 2,322-film metadata-complete cohort. The availability histogram in Figure~\ref{fig:coverage-availability}a defines a pool of the 29 most frequent subtitle languages; the 29th, Czech, appears for 1,205 films. For a candidate set $S$, let $M(S)$ be the films with an SRT file in \emph{every} language in $S$. We score $S$ by $|S|\,|M(S)|$, the number of subtitle files retained under common film coverage.
\begin{enumerate}[leftmargin=*,itemsep=2pt,topsep=3pt]
\item \textbf{Frequency-prefix baseline (Approach A).} For each set size $K$, we evaluate the top $K$ languages in the frequency ranking. Its best result has five languages and 1,458 films, giving $1{,}458\times5=7{,}290$ files. The top-six prefix has 1,157 films and 6,942 files.
\item \textbf{Combination search (Approach B).} For each $K$, compare candidate $K$-language subsets from the same pool using complete film coverage. Its best five-language set equals the prefix result. At six languages, a different set retains 1,231 films and $1{,}231\times6=7{,}386$ files. This is 444 more files than the six-language prefix and 96 more than the best prefix configuration.
\item \textbf{Selected set.} Arabic (ar), English (en), Indonesian (id), Persian (fa), Romanian (ro), and Vietnamese (vi) form the six-language result. The selected set includes a language ranked eighth in marginal frequency, illustrating why selecting languages one at a time by popularity can miss the best shared-coverage set. At $K\geq7$, complete coverage falls below 1,000 films.
\end{enumerate}

\paragraph{Choosing rating countries.} Starting with those 1,231 films, we first remove a film marked \texttt{Unrated} or \texttt{Not Rated} only if that is its sole motion-picture-rating entry. For example, a film with an unrated tag \emph{and} a valid United Kingdom rating remains eligible. We then apply the same complete-coverage search in Algorithm~\ref{alg:language-set-optimization}, substituting national rating systems for subtitle languages. The candidate pool comprises the 21 most frequent country systems shown in Figure~\ref{fig:coverage-availability}b, and a film belongs to $M(S)$ only when it has a valid rating from every country in $S$. A ten-country set retains approximately one thousand films while supporting a broad national comparison. The selected countries are Australia, Brazil, France, Germany, the Netherlands, Singapore, South Korea, Sweden, the United Kingdom, and the United States. The final benchmark contains 1,012 films with complete ratings in these ten systems.

\begin{table}[htbp]
\centering
\small
\caption{Country-specific ordered motion-picture rating label spaces used in CineSubBench. Labels are normalized within each national classification system and retained as distinct ordinal scales rather than treated as interchangeable across countries.}
\vspace{5pt}
\label{tab:rating_spaces}
\begin{tabular}{ll}
\toprule
Country & Ordered rating labels \\
\midrule
Australia & G, PG, M, MA15+, R18+ \\
Brazil & Livre, 10, 12, 14, 16, 18 \\
France & Tous publics, 12, 16, 18 \\
Germany & 0, 6, 12, 16, 18 \\
Netherlands & AL, 6, 9, 12, 14, 16, 18 \\
Singapore & G, PG, PG13, NC16, M18, R21 \\
South Korea & All, 12, 15, 19 \\
Sweden & Btl, 7, 11, 15 \\
United Kingdom & U, PG, 12, 15, 18 \\
United States & G, PG, PG-13, R, NC-17 \\
\bottomrule
\end{tabular}
\end{table}

\begin{pseudocodealgorithm}{Complete-coverage set selection}
\label{alg:language-set-optimization}
\textbf{Input:} film-by-attribute availability map $A$, candidate pool $\mathcal{X}$, maximum cardinality $K_{\max}$\\
\textbf{Output:} selected attribute set $S^{\star}$ and complete-coverage film set $M(S^{\star})$
\begin{enumerate}[leftmargin=1.65em,label=\textbf{\arabic*.},itemsep=1pt,topsep=3pt]
  \item Set $q^{\star}\gets-\infty$.
  \item For each $K\in\{1,\ldots,K_{\max}\}$, enumerate every $S\subseteq\mathcal{X}$ with $|S|=K$.
  \item For each $S$, compute $M(S)\gets\{i:A_{i,x}=1\ \forall\,x\in S\}$ and $q\gets|S|\times|M(S)|$.
  \item If $q>q^{\star}$, set $(S^{\star},M(S^{\star}),q^{\star})\gets(S,M(S),q)$.
  \item Return $S^{\star}$ and $M(S^{\star})$.
\end{enumerate}
\end{pseudocodealgorithm}

\begin{table}[H]
\centering
\small
\setlength{\tabcolsep}{3.25pt}
\renewcommand{\arraystretch}{1.25}
\caption{Subtitle-level verification layers applied to the optimized six-language cohort. Each automated trigger initiates targeted manual inspection, source recovery, or an ingestion stop, rather than silently removing the affected subtitle track.}
\vspace{5pt}
\label{tab:subtitle-verification-layers}
\begin{tabular}{P{0.23\textwidth}P{0.34\textwidth}P{0.34\textwidth}}
\toprule
Verification layer & Automated rule & Audit purpose and response \\
\midrule
Provider download count & Among alternative SRT files for the same film and language, select the file with the highest source-reported count; retain the selected count in the JSON. & Use popularity only as an initial quality proxy. Subject the selected file to the remaining checks and recover it if verification finds a genuine problem. \\
Provider-body integrity & Scan subtitle bodies for server-error signatures, including Cloudflare and Guru Meditation messages. & Detect error pages returned in place of dialogue; retrieve or inspect a replacement asset. \\
Timestamp syntax & Require valid SRT time ranges with hours, minutes, seconds, millisecond separator, and arrow structure (e.g., \texttt{00:09:09,949 --> 00:09:12,179}). & Catch malformed headers before JSON creation; repair or re-download the source asset. \\
Chronological consistency & Compare the start and end time of every subtitle block. & Flag negative-duration temporal inversions. \\
Single-block lifespan & Flag any subtitle block lasting 15 minutes or more. & Detect frozen or malformed subtitle segments. \\
Delayed onset & Flag a first operational subtitle beginning after five minutes. & Detect missing opening dialogue or late-start tracks. \\
Runtime coverage & Flag a final subtitle end time divided by IMDb movie runtime below two-thirds. & Detect tracks that appear to stop before later acts of the film. \\
Macro-intervals & Flag a 15-minute or larger gap between adjacent subtitle blocks. & Detect missing internal narrative segments or dropped blocks. \\
\bottomrule
\end{tabular}
\end{table}

\subsection{Multi-Layer Subtitle Verification and Manual Audit}
\label{app:subtitle-qa}
The verification layers in Table~\ref{tab:subtitle-verification-layers} target distinct failure modes in the selected 1,231-film multilingual cohort: a provider can return an error page instead of dialogue, a syntactically valid track can cover only part of a film, and an otherwise usable track can contain malformed timestamps. Threshold crossings trigger manual review, after which the curator determines whether the track should be retained, repaired, or replaced.

\paragraph{Examples of audit decisions.}
\begin{itemize}[leftmargin=*,itemsep=2pt,topsep=3pt]
\item \textbf{False subtitle download.} If the file body contains a Cloudflare or ``Guru Meditation'' error message, the payload is a provider response rather than film dialogue; the curator seeks a replacement.
\item \textbf{Short temporal coverage.} For an illustrative 120-minute film, a track whose last subtitle ends at 70 minutes has endpoint coverage $70/120=0.58$, below the $2/3$ trigger. Its source and timeline are inspected before it is accepted or replaced. A silent ending alone is not assumed to be an error.
\item \textbf{Potential missing span.} A first subtitle after five minutes or an internal gap of at least 15 minutes raises a flag. A wordless opening or long action sequence may explain the gap, so a curator checks the track before deciding.
\item \textbf{Malformed block.} A subtitle block whose end precedes its start, or one that stays active for at least 15 minutes, is examined against the source file. Correctable time formatting is repaired there; an unresolved block prevents structured ingestion.
\end{itemize}

\subsubsection{Structural and Encoding Safeguards}
\paragraph{Strict structural ingestion.} Every expected SRT block has three components: a sequential index, a timestamp header, and a text payload. A combined splitter first separates candidate headers and bodies; a header parser extracts the index and time range only when the header satisfies the SRT structure. Thus a file containing plausible dialogue but a malformed header is not partially ingested. A non-matching header is recorded as a structural failure and stops ingestion until the source asset is manually corrected or cleanly re-downloaded. This prevents malformed blocks from being silently omitted.

\paragraph{Deterministic decoding.} Raw subtitle files do not share a universal character encoding, especially across Arabic, Persian, and European-language sources. We therefore attempt the ordered sequence in Table~\ref{tab:decoding-cascade}, without suppressing undecodable bytes. If every tier fails, the asset is rejected until manual normalization or a clean replacement is available. A successful byte-level decode does not itself prove that the text is in the expected language; suspicious language or character content is checked during manual review.

\begin{table*}[htbp]
\centering
\small
\setlength{\tabcolsep}{3.25pt}
\renewcommand{\arraystretch}{1.5}
\caption{Deterministic subtitle-decoding cascade. Decoding proceeds to the next encoding tier only when the preceding one fails to read the source asset; lossy error suppression is never used.}
\vspace{5pt}
\label{tab:decoding-cascade}
\begin{tabular}{clP{0.66\textwidth}}
\toprule
Tier & Encoding & Purpose and representative case \\
\midrule
1 & \texttt{utf-8-sig} & Default modern-web decoding while removing a byte-order mark when present. \\
2 & \texttt{utf-16} & Handles UTF-16 subtitle assets, including files exported by automated transcription or speech-to-text systems. \\
3 & \texttt{cp1256} & Handles legacy Windows Arabic and Persian encodings, preserving right-to-left dialogue instead of dropping unsupported characters. \\
4 & \texttt{cp1252} & Handles legacy Western-European encodings and extended diacritics, such as \texttt{é}, \texttt{ü}, and \texttt{ñ}. \\
Failure & none succeeds & Reject the asset and require manual normalization or a clean replacement before ingestion resumes. \\
\bottomrule
\end{tabular}
\end{table*}

\subsubsection{Subtitle Sanitization and Artifact Removal}
\label{app:subtitle-sanitization}

As part of data preparation, we apply an automated, auditable sanitization procedure to all six subtitle-language tracks. Across 8,133,088 subtitle entries from 1,012 films, the procedure removes 6,594 entries containing non-narrative distribution artifacts, including translator or subtitle-team credits, personal contact details, social-media handles, subtitle-provider advertisements, download promotions, and external promotional links. These elements originate from subtitle-distribution files rather than the films' narrative content.

To preserve meaningful content, entries containing contact information are not removed indiscriminately. In 65 cases where an email address occurs within an otherwise narrative-relevant subtitle entry, such as an on-screen message, the surrounding text is retained and only the address is replaced with \texttt{[email redacted]}. Sanitization preserves film identifiers, timestamps, language-track metadata, and original subtitle indices without renumbering, maintaining temporal alignment and links between language-content annotations and their supporting subtitle evidence.

\subsubsection{Final Dataset Consistency Checks}
After source recovery and JSON creation, the integrated records were checked for:
\begin{itemize}[leftmargin=*,itemsep=2pt,topsep=3pt]
\item unique, stable film identifiers and source links that point to the corresponding film;
\item one named subtitle track for each of the six languages, with indexed entries and start/end timecodes;
\item ten country-rating entries per film in the defined country order, with one normalized movie-rating label per country;
\item populated task fields, consistent field types, and conformance to the dataset schema; and
\item subtitle-grounded language evidence whose indices resolve to actual English subtitle entries.
\end{itemize}
These checks are paired with manual review of narrative fields, rating normalization, and discrepancies between subtitle-derived language counts and Kids-in-Mind summaries; Appendix~\ref{app:language-gold-construction} details the latter audit.

\subsection{Subtitle-Grounded Language-Safety Gold Construction and Audit}
\label{app:language-gold-construction}
The annotation workflow connects the review-level counts in Kids-in-Mind to auditable, line-level evidence in the English subtitle track:
\begin{enumerate}[leftmargin=*,itemsep=2pt,topsep=3pt]
    \item \textbf{Collect source counts and examples.} For each film, we retained the prose language assessment in \texttt{languageContentReference}, including its reported category counts and parenthetical examples. These review-level counts provide targets for comparison but no subtitle locations; the nested \texttt{source} field preserves Kids-in-Mind provenance.
    \item \textbf{Operationalize the glossary.} We followed the definitions and examples in the \href{https://kids-in-mind.com/glossary.htm}{Kids-in-Mind glossary}, together with recurring examples in the film summaries, to construct category-specific wordlists and regular expressions. The benchmark schema keeps these external counts in \texttt{languageContentReference} and the audited subtitle-grounded labels in \texttt{languageContentAssessment}.
    \item \textbf{Normalize and scan the subtitles.} Before matching, we remove HTML-style markup and bracketed or all-capital stage directions. We then scan every English subtitle entry, counting all lexical occurrences and recording the index of each line containing at least one match. Thus occurrence count and evidence-line count remain distinct.
    \item \textbf{Audit discrepancies.} We compare extracted counts with the parsed Kids-in-Mind counts, screen the resulting discrepancy reports, and inspect flagged and large disagreements in subtitle context. We revise a rule only for a broad, defensible lexical gap or false-positive pattern; every adjustment is manually verified against retrieved lines.
    \item \textbf{Preserve evidence-based disagreement.} We do not force the two sources to agree. A difference may reflect a different subtitle version, a censored bleep, a gesture, or reviewed content that is not lexicalized in the evaluated subtitle track. In such cases, the subtitle-derived result is retained.
    \item \textbf{Finalize auditable gold.} The audited subtitle occurrences and evidence indices form the benchmark gold labels. The original Kids-in-Mind summary remains in the record as provenance and an independent audit reference.
\end{enumerate}

\paragraph{Operational categories.}
The policy covers \textbf{strong profanity} (F-word forms and close derivatives), \textbf{crude bodily language} (scatological and crude anatomical language), \textbf{mild obscenity} (lower-intensity obscene or impolite expressions), and \textbf{religious profanity/exclamation} (religious profanity and exclamatory uses, excluding ordinary religious dialogue). Items not recoverable from subtitle text, such as obscene hand gestures, are excluded. Broad derogatory or name-calling categories are likewise excluded because the present rules do not support sufficiently precise and reproducible boundaries.

\paragraph{Stored evidence and auditability.}
For each category, the dataset stores the \texttt{occurrenceCount}, \texttt{evidenceLineCount}, and \texttt{evidenceSubtitleIndices}. The indexed English subtitle entries retain the text and start/end timecodes, so every gold decision can be recovered and inspected without duplicating offensive text in the label field. A separate diagnostic file retains matched spans and category information for reproducibility and future rule refinement. Three malformed source rating values caused by title-number leakage were reconstructed from their structured full-rating fields before that metadata was used for audit.

\paragraph{Examples of audit-informed decisions.}
The mismatch audit identified apostrophe variants such as shortened F-word forms, which were added because they are explicit subtitle evidence. It also identified ordinary uses that must not be counted: a capitalized personal name, food-related uses of ``breast,'' spatial uses such as ``bottom of,'' and ordinary religious dialogue are filtered by contextual rules. Conversely, broad wildcard patterns for partially censored words were rejected after producing unrelated matches, and obscene gestures remain excluded because they cannot be recovered from subtitle text. These decisions make the annotation reproducible without mechanically reproducing a review site's count.

\subsection{Task-Oriented Label Curation and Normalization}
\label{app:label-curation}

\paragraph{Task selection and complete coverage.}
We retained core tasks only when their reference labels could be provided uniformly across the final benchmark and could be meaningfully evaluated from subtitle evidence. This criterion excluded otherwise plausible fields with incomplete coverage. For example, tagline generation was removed because taglines were available for only 946 of the final 1,012 films. The key-message task was retained because its references were made complete across the final cohort and because it captures thematic information that can be inferred from dialogue and narrative progression.

\paragraph{Narrative-reference cleaning.}
Plot, synopsis, and storyline fields were audited for material external to the film narrative, including production history, remake notes, franchise context, and other database-specific editorial text. When such material appeared, the reference was cleaned to retain film-internal narrative content. Raw source text was retained in internal curation records for traceability. A small number of missing key-message entries were completed through curator review so that the final task has full reference coverage.

\paragraph{Age suitability and national-rating normalization.}
General age suitability and formal motion-picture classification are represented as separate targets. Common Sense Media ratings are retained as family-oriented age-suitability labels. National motion-picture ratings are normalized independently within each country's classification system using movie-rating labels rather than television, app-store, or platform-specific categories. Missing, obsolete, or multiple candidate values are resolved to one contemporary ordered label for the corresponding country. For example, South Korean adult ratings are normalized to the current 19+ category, while the United Kingdom's theatrical 12A and video 12 categories are represented by the benchmark label \texttt{12}. Table~\ref{tab:rating_spaces} lists the resulting ordered spaces.

The benchmark keeps these country-specific label spaces as separate ordinal systems. Continuous pseudo-normalized scores are excluded from the benchmark schema, avoiding artificial numerical equivalence between national classifications with different institutional meanings.

\paragraph{Task-level validation and manual audit.}
Curator review supplements the automated dataset checks in Appendix~\ref{app:subtitle-qa}. Narrative references are checked against their source text after cleaning; ambiguous or historical country-rating values are resolved within their national systems; and disagreements between external language-content summaries and subtitle-derived evidence are inspected in the corresponding subtitle context. Final schema validation confirms complete task fields and consistent field types for every retained film.

\subsection{Final Film-Entity Structure and Field Provenance}
\label{app:dataset-schema}
After construction, verification, normalization, and annotation, \cinesubbench{} is represented as a JSON array in which each object corresponds to one film. The schema keeps film metadata, task references, provenance, and six timestamped subtitle tracks in a consistent record. The types below describe the benchmark schema.

\vspace{7pt}

\begin{tcolorbox}[
  breakable,
  colback=cineNavy!2,
  colframe=cineNavy!72,
  coltitle=white,
  title={CineSubBench Film Entity (JSON Object)},
  fonttitle=\bfseries,
  boxrule=0.65pt,
  arc=1mm,
  left=2mm,
  right=2mm,
  top=1.2mm,
  bottom=1.2mm,
  before skip=5pt,
  after skip=7pt
]
\small
\textcolor{cineTeal}{\bfseries Identity and film metadata}

\smallskip
\noindent\texttt{id} \textcolor{cineGray}{\emph{(Integer)}} --- Unique sequential identifier for the film.\par
\noindent\texttt{title} \textcolor{cineGray}{\emph{(String)}} --- Canonical film title.\par
\noindent\texttt{releaseYear} \textcolor{cineGray}{\emph{(String)}} --- Year in which the film was released.\par
\noindent\texttt{duration} \textcolor{cineGray}{\emph{(String)}} --- Human-readable film runtime.\par
\noindent\texttt{imdbRating} \textcolor{cineGray}{\emph{(String)}} --- IMDb user-rating value recorded during collection.\par
\noindent\texttt{countriesOfOrigin} \textcolor{cineGray}{\emph{(String)}} --- Country or countries associated with the production.\par
\noindent\texttt{originalLanguages} \textcolor{cineGray}{\emph{(String)}} --- Film's original spoken language or languages.\par
\noindent\texttt{directors} \textcolor{cineGray}{\emph{(Array of Strings)}} --- Credited director names.\par
\noindent\texttt{writers} \textcolor{cineGray}{\emph{(Array of Strings)}} --- Credited writer names.\par
\noindent\texttt{topCast} \textcolor{cineGray}{\emph{(Array of Objects)}} --- Principal cast entries, each containing an \texttt{actor} and primary \texttt{character}.\par
\noindent\texttt{fullCast} \textcolor{cineGray}{\emph{(Array of Objects)}} --- Expanded cast entries with \texttt{actor}, \texttt{characters}, and any \texttt{additionalTexts}.\par

\medskip
\textcolor{cineTeal}{\bfseries Narrative and classification references}

\smallskip
\noindent\texttt{plot} \textcolor{cineGray}{\emph{(String)}} --- Concise premise-level narrative reference.\par
\noindent\texttt{synopsis} \textcolor{cineGray}{\emph{(String)}} --- Longer, spoiler-aware account of the narrative arc.\par
\noindent\texttt{storyline} \textcolor{cineGray}{\emph{(String)}} --- Short source-provided storyline description.\par
\noindent\texttt{keyMessage} \textcolor{cineGray}{\emph{(String)}} --- Reference statement of the film's principal theme or lesson.\par
\noindent\texttt{genres} \textcolor{cineGray}{\emph{(Array of Strings)}} --- Gold multi-label genre set.\par
\noindent\texttt{topics} \textcolor{cineGray}{\emph{(String)}} --- Source-provided topical descriptors.\par
\noindent\texttt{ageSuitabilityRating} \textcolor{cineGray}{\emph{(String)}} --- General age-suitability label, expressed as an age threshold such as \texttt{13+}.\par
\noindent\texttt{motionPictureRatings} \textcolor{cineGray}{\emph{(Array of Objects)}} --- Ratings for the ten selected national systems. Each entry stores \texttt{country} (String), \texttt{label} (String), \texttt{minimumAge} (Integer), and \texttt{ordinal} (Integer).\par

\medskip
\textcolor{cineTeal}{\bfseries Language-content assessment}

\smallskip
\noindent\texttt{languageContentReference} \textcolor{cineGray}{\emph{(Object)}} --- External review information retained for provenance and audit: \texttt{rating} (Integer), \texttt{summary} (String), and \texttt{source} (String).\par
\noindent\texttt{languageContentAssessment} \textcolor{cineGray}{\emph{(Object)}} --- Audited labels derived from the English subtitle track. It records \texttt{inputLanguage} (String) and a \texttt{categories} object containing \texttt{strongProfanity}, \texttt{crudeBodilyLanguage}, \texttt{mildObscenity}, and \texttt{religiousProfanityAndExclamation}.\par
\noindent\texttt{occurrenceCount} \textcolor{cineGray}{\emph{(Integer)}} --- Number of matched expressions for one language-content category.\par
\noindent\texttt{evidenceLineCount} \textcolor{cineGray}{\emph{(Integer)}} --- Number of distinct subtitle entries containing category evidence.\par
\noindent\texttt{evidenceSubtitleIndices} \textcolor{cineGray}{\emph{(Array of Integers)}} --- Indices that resolve each category's evidence to the English subtitle entries.\par

\medskip
\textcolor{cineTeal}{\bfseries Provenance and multilingual subtitle input}

\smallskip
\noindent\texttt{sources} \textcolor{cineGray}{\emph{(Object)}} --- Source URLs stored as \texttt{imdbLink}, \texttt{commonSenseLink}, and \texttt{kidsInMindLink}.\par
\noindent\texttt{subtitles} \textcolor{cineGray}{\emph{(Array of Objects)}} --- Six subtitle tracks, one per benchmark language. Each track contains \texttt{language} (String), \texttt{downloadCount} (Integer), and \texttt{entries} (Array of Objects).\par
\noindent\texttt{entries} \textcolor{cineGray}{\emph{(Array of Objects)}} --- Ordered subtitle units containing \texttt{index} (Integer), \texttt{timeframe} (String), and \texttt{content} (String).\par
\end{tcolorbox}

Table~\ref{tab:dataset-field-sources} makes the field-level provenance explicit. The mapping names the originating website rather than the later cleaning or normalization stage applied during benchmark construction.

\begin{table}[H]
\centering
\small
\caption{Mapping from public CineSubBench fields to their source or construction process. Website names link to the corresponding source. The subtitle-derived \texttt{languageContentAssessment} follows Kids-in-Mind definitions and uses its reference counts for auditing, while the final labels and evidence are constructed directly from the English subtitle track.}
\vspace{5pt}
\label{tab:dataset-field-sources}
\begin{tabular}{P{0.22\textwidth}P{0.69\textwidth}}
\toprule
Source website & Public dataset fields \\
\midrule
\href{https://www.commonsensemedia.org/}{Common Sense Media}
& \texttt{ageSuitabilityRating}; \texttt{storyline}. \\
\addlinespace[2pt]
\href{https://kids-in-mind.com/}{Kids-in-Mind}
& \texttt{keyMessage}; \texttt{languageContentReference}. \\
\addlinespace[2pt]
\href{https://www.imdb.com/}{IMDb}
& \texttt{title}; \texttt{releaseYear}; \texttt{duration}; \texttt{imdbRating}; \texttt{countriesOfOrigin}; \texttt{originalLanguages}; \texttt{genres}; \texttt{topics}; \texttt{plot}; \texttt{synopsis}; \texttt{motionPictureRatings}; \texttt{directors}; \texttt{writers}; \texttt{topCast}; \texttt{fullCast}. \\
\addlinespace[2pt]
\href{https://www.opensubtitles.com/}{OpenSubtitles} and
\href{https://subdl.com/}{SubDL}
& \texttt{subtitles}, including each track's \texttt{language}, \texttt{downloadCount}, and timestamped \texttt{entries} (\texttt{index}, \texttt{timeframe}, and \texttt{content}). \\
\addlinespace[2pt]
\cinesubbench{} annotation pipeline
& \texttt{languageContentAssessment}, containing manually audited, subtitle-localized category counts and evidence indices derived from the English subtitle track. \\
\bottomrule
\end{tabular}

\vspace{2pt}
\parbox{0.93\textwidth}{\footnotesize\textit{Schema note.} \texttt{id} is assigned by CineSubBench, while \texttt{sources} stores the IMDb, Common Sense Media, and Kids-in-Mind provenance URLs for each film.}
\end{table}

\begin{wraptable}{r}{0.50\textwidth}
\centering
\small
\vspace{-27pt}
\caption{Core \cinesubbench{} dataset statistics.}
\vspace{5pt}
\label{tab:dataset-core-statistics}
\begin{tabular}{lr}
\toprule
Statistic & Value \\
\midrule
Films & 1,012 \\
Subtitle tracks & 6,072 \\
Subtitle entries & 8.13M \\
Release years & 1977--2026 \\
Runtime, mean / median & 114.6 / 113.0 min \\
Runtime, 95th pct. / max & 149.4 / 242.0 min \\
IMDb rating, mean / median & 6.80 / 6.80 \\
IMDb rating, min / max & 3.5 / 8.8 \\
Raw genre labels & 149 \\
Core genre labels & 22 \\
Raw genres per film, mean / median & 4.08 / 4.0 \\
\bottomrule
\end{tabular}
\end{wraptable}

\begin{table}[t]
\centering
\small
\caption{Subtitle input size by language. Token counts are computed from prompt-formatted subtitle inputs using the tokenizer adopted for prompt-length accounting; they reflect tokenization-dependent input length rather than language-neutral semantic length.}
\vspace{5pt}
\label{tab:subtitle-token-statistics}
\begin{tabular}{lrrrrrr}
\toprule
Language & Mean tok. & Median tok. & 95th pct. & Max tok. & Median entries & Median ratio \\
\midrule
English (en) & 43,123 & 41,759 & 68,833 & 144,268 & 1,418 & 1.00 \\
Arabic (ar) & 39,867 & 38,536 & 65,023 & 98,701 & 1,214 & 0.95 \\
Indonesian (id) & 40,641 & 39,739 & 64,760 & 120,034 & 1,308 & 0.97 \\
Persian (fa) & 45,162 & 43,916 & 73,731 & 112,920 & 1,325 & 1.08 \\
Romanian (ro) & 41,010 & 39,676 & 65,542 & 129,799 & 1,244 & 0.99 \\
Vietnamese (vi) & 43,613 & 42,374 & 70,575 & 108,761 & 1,322 & 1.05 \\
\bottomrule
\end{tabular}
\end{table}

\subsection{Dataset Statistics and Coverage Audits}
\label{app:dataset-statistics}
\suppressfloats[t]
The tables and figures in this section describe all 1,012 films in the benchmark. Core counts are presented compactly, while task distributions and temporal-coverage diagnostics receive full-width treatment. Table~\ref{tab:reference-output-lengths} reports reference-output lengths for narrative tasks; Tables~\ref{tab:core-genre-distribution}, \ref{tab:commonsense-age-distribution}, and \ref{tab:language-content-statistics} report the supporting label distributions.

\begin{figure}[htbp]
\centering
\includegraphics[width=1.0\linewidth]{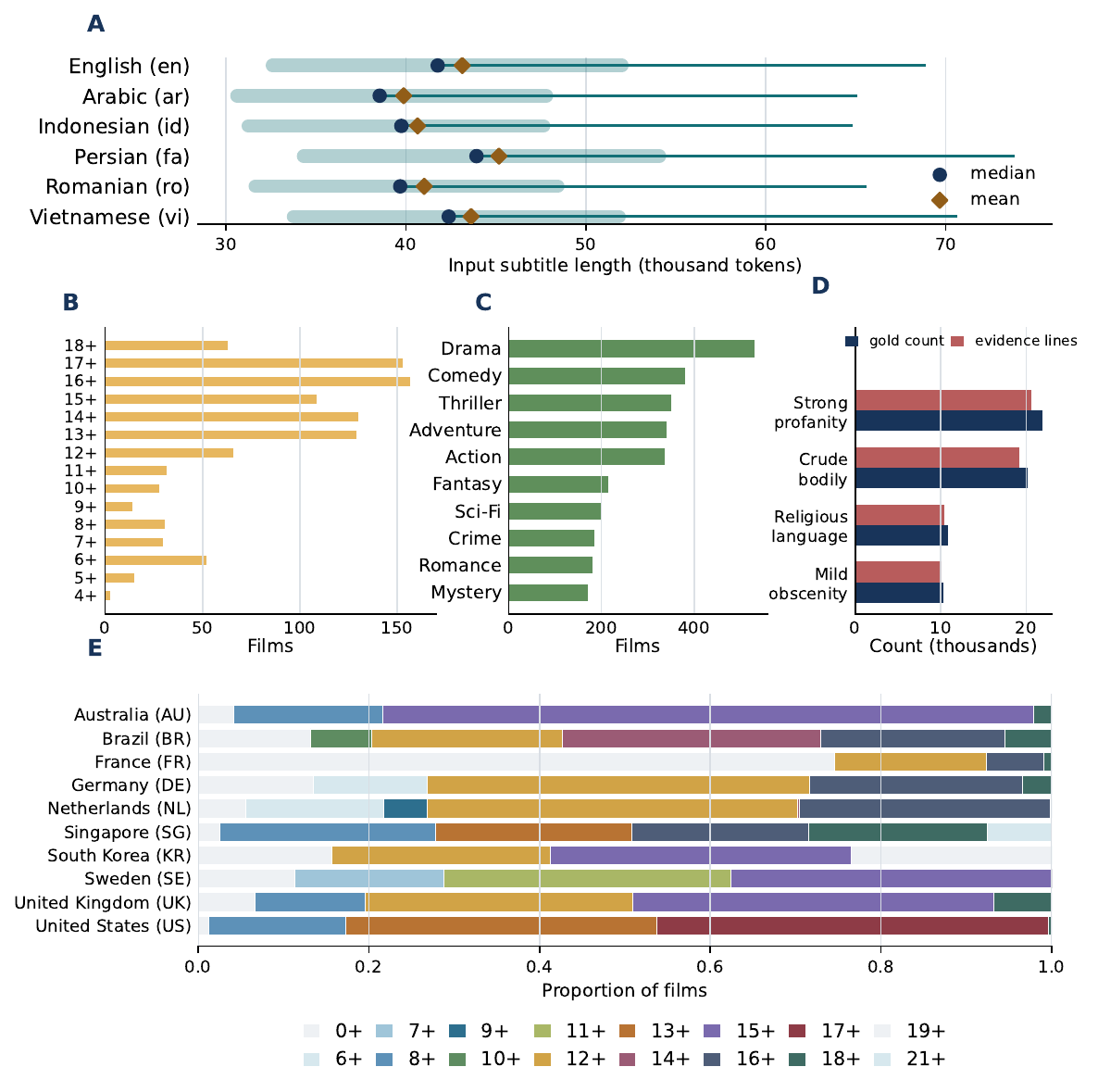}
\caption{Task-relevant statistics for all 1,012 \cinesubbench{} films. Panels show (A) subtitle input length by language, (B) age-suitability label distributions, (C) normalized genre frequencies, (D) subtitle-grounded language-safety counts and corresponding evidence-line counts, and (E) national motion-picture rating distributions represented by minimum-age values. Token counts are computed from prompt-formatted subtitle inputs using the \texttt{o200k\_base} tokenizer.}
\label{fig:dataset-overview}
\end{figure}

\subsubsection{Subtitle-Duration Coverage Audit}
We audit subtitle duration and density relative to the corresponding film runtime and English track. The audit measures temporal span/runtime, subtitle-interval density/runtime, token ratio to English, and entry ratio to English. Tracks are flagged when span/runtime is below $0.65$, token ratio is below $0.60$, entry ratio is below $0.70$, or temporal span differs unusually from English. Because subtitle activity need not span every visual moment, these thresholds initiate inspection rather than automatic deletion.

\paragraph{Arabic duration and density audit.}
Table~\ref{tab:subtitle-duration-audit} reports a mean Arabic temporal span/runtime of $0.931$ and a fifth percentile of $0.868$, with no track below the short-runtime threshold. Arabic nevertheless has lower subtitle density than English: 27 tracks fall below the token-ratio threshold and 120 below the entry-ratio threshold. A supplementary correlation audit finds little association between these density indicators and Arabic narrative quality, indicating that duration-level truncation alone does not account for the observed model gap.

\begin{table}[htbp]
\centering
\begin{minipage}[t]{0.49\linewidth}
\centering
\footnotesize
\setlength{\tabcolsep}{2.5pt}
\caption{Reference output length statistics for narrative tasks.}
\vspace{5pt}
\label{tab:reference-output-lengths}
\begin{tabular}{lrrrr}
\toprule
Reference field & Mean & Median & P95 & Max \\
\midrule
Plot & 33 & 33 & 50 & 66 \\
Synopsis & 1,448 & 1,262 & 2,802 & 7,151 \\
Storyline & 156 & 154 & 242 & 437 \\
Key message & 14 & 12 & 26 & 55 \\
\bottomrule
\end{tabular}
\end{minipage}\hfill
\begin{minipage}[t]{0.49\linewidth}
\centering
\scriptsize
\setlength{\tabcolsep}{2.5pt}
\caption{Subtitle-grounded language-content statistics. Evidence is defined over the English subtitle track.}
\vspace{5pt}
\label{tab:language-content-statistics}
\begin{tabularx}{\linewidth}{>{\raggedright\arraybackslash}Xrr}
\toprule
Category & Gold count & Evidence lines \\
\midrule
Strong profanity & 21,892 & 20,602 \\
Crude bodily language & 20,131 & 19,149 \\
Religious profanity/exclamation & 10,861 & 10,447 \\
Mild obscenity & 10,360 & 10,088 \\
\bottomrule
\end{tabularx}
\end{minipage}
\end{table}

\begin{table}[htbp]
\centering
\scriptsize
\setlength{\tabcolsep}{3.4pt}
\caption{Subtitle-duration coverage audit by language. Span/runtime measures the fraction of the film timeline covered by each subtitle track, while token and entry ratios compare each track with the English subtitle track for the same film. Low-token and low-entry counts use thresholds of $<0.60$ and $<0.70$ relative to English, respectively.}
\vspace{5pt}
\label{tab:subtitle-duration-audit}
\begin{tabular}{lrrrrrrr}
\toprule
Language & Flagged & Span/run. & Span p05 & Tok. ratio & Entry ratio & Low tok. & Low entry \\
\midrule
English (en) & 9 & 0.95 & 0.89 & 1.00 & 1.00 & 0 & 0 \\
Arabic (ar) & 142 & 0.93 & 0.87 & 0.95 & 0.88 & 27 & 120 \\
Indonesian (id) & 112 & 0.95 & 0.89 & 0.97 & 0.93 & 28 & 84 \\
Persian (fa) & 123 & 0.96 & 0.88 & 1.08 & 0.95 & 23 & 77 \\
Romanian (ro) & 155 & 0.95 & 0.89 & 0.98 & 0.89 & 27 & 125 \\
Vietnamese (vi) & 93 & 0.96 & 0.90 & 1.04 & 0.95 & 20 & 57 \\
\bottomrule
\end{tabular}
\end{table}

\begin{table}[htbp]
\centering
\begin{minipage}[t]{0.46\linewidth}
\centering
\small
\caption{Most frequent normalized core genres.}
\label{tab:core-genre-distribution}
\begin{tabular}{lr}
\toprule
Genre & Films \\
\midrule
Drama & 531 \\
Comedy & 380 \\
Thriller & 350 \\
Adventure & 340 \\
Action & 336 \\
Fantasy & 215 \\
Sci-Fi & 199 \\
Crime & 185 \\
Romance & 181 \\
Mystery & 171 \\
Family & 158 \\
Animation & 135 \\
Horror & 114 \\
Biography & 89 \\
Music & 78 \\
\bottomrule
\end{tabular}
\end{minipage}\hfill
\begin{minipage}[t]{0.46\linewidth}
\centering
\small
\caption{Common Sense age-suitability labels.}
\label{tab:commonsense-age-distribution}
\begin{tabular}{lr}
\toprule
Age label & Films \\
\midrule
16+ & 157 \\
17+ & 153 \\
14+ & 130 \\
13+ & 129 \\
15+ & 109 \\
12+ & 66 \\
18+ & 63 \\
6+ & 52 \\
11+ & 32 \\
8+ & 31 \\
7+ & 30 \\
10+ & 28 \\
5+ & 15 \\
9+ & 14 \\
4+ & 3 \\
\bottomrule
\end{tabular}
\end{minipage}
\end{table}

\section{Evaluation Design and Reproducibility}
\label{app:evaluation-design}
\subsection{Model Coverage and Generation Settings}
\label{app:model-coverage}
Table~\ref{tab:experiment-matrix} reports model access, subtitle languages, and task coverage for every comparison. The main six-language evaluation contains nine models spanning closed frontier, closed baseline, and open-weight models. A separate Ministral 3B, 8B, and 14B experiment provides a controlled English-only within-family scaling comparison. GPT-5.6 Luna serves as the common reference-based evaluator for narrative generation. Narrative, genre, age-suitability, and country-rating analyses use the matched six-language film set; subtitle-grounded language-safety assessment remains English-only because its gold evidence is indexed to the English track.

\begin{figure*}[t]
\centering
\begin{minipage}[t]{0.49\textwidth}
  \centering
  \includegraphics[width=\linewidth]{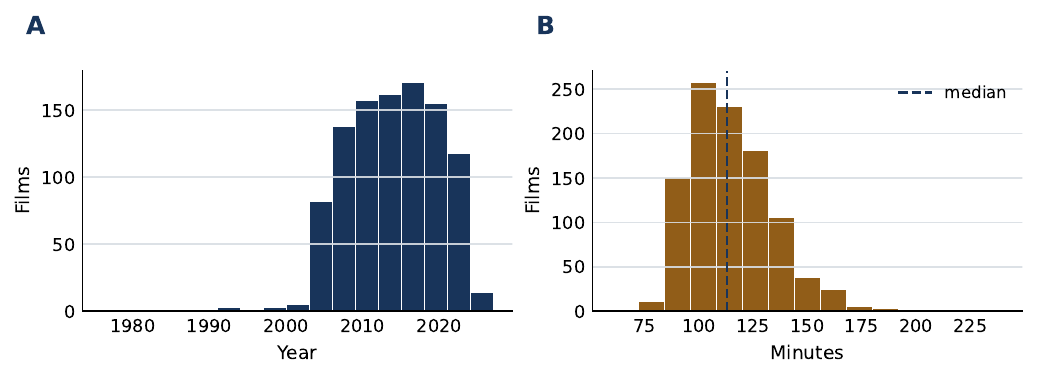}
  \textbf{(a) Release year and runtime}
\end{minipage}\hfill
\begin{minipage}[t]{0.49\textwidth}
  \centering
  \includegraphics[width=\linewidth]{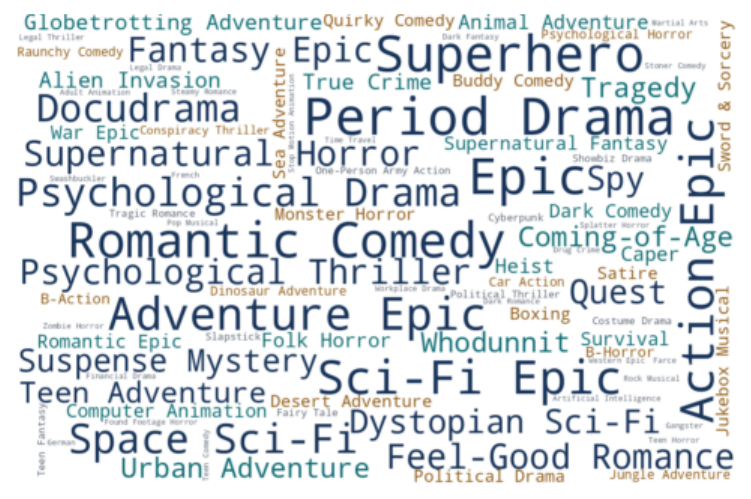}
  \textbf{(b) Topical coverage}
\end{minipage}
\caption{Additional corpus context. (a) Release-year and runtime coverage. (b) Topic frequency after splitting comma-separated tags and removing generic core-genre labels; font size indicates frequency.}
\label{fig:dataset-context}
\end{figure*}

All API models use deterministic decoding with temperature $0$, while local models use greedy decoding. We apply the same task-specific output ceilings across providers and validate responses against the corresponding task schema. Table~\ref{tab:decoding-parameters} gives the complete settings and provider parameter mappings.
\begin{table*}[htbp]
\centering
\small
\setlength{\tabcolsep}{4pt}
\caption{Models and task coverage in \cinesubbench{}. All models receive subtitle text as their only film evidence and generate English outputs for narrative generation tasks. Six-language comparisons use the same matched film set, while evidence-grounded language-safety assessment is evaluated only on the English subtitle track.}
\vspace{5pt}
\label{tab:experiment-matrix}
\begin{tabular}{@{}P{.12\textwidth}P{.22\textwidth}P{.18\textwidth}P{.20\textwidth}P{.18\textwidth}@{}}
\toprule
Evaluation role & Models & Model access & Subtitle inputs & Tasks \\
\midrule
Closed frontier &
\href{https://developers.openai.com/api/docs/models/gpt-5.6-sol}{GPT-5.6 Sol}; \href{https://docs.cloud.google.com/gemini-enterprise-agent-platform/models/gemini/3-8-flash}{Gemini 3.8 Flash}; \href{https://platform.claude.com/docs/en/models/haiku-4-5/overview}{Claude Haiku 4.5} &
OpenAI, Gemini, and Anthropic APIs &
English + Arabic, Indonesian, Persian, Romanian, Vietnamese &
Narrative and cultural; English language safety where available \\
\addlinespace
Closed baseline &
\href{https://developers.openai.com/api/docs/models/gpt-5-nano}{GPT-5 Nano}; \href{https://docs.cloud.google.com/gemini-enterprise-agent-platform/models/gemini/3-5-flash-lite}{Gemini 3.5 Flash Lite} &
OpenAI and Gemini APIs &
English + Arabic, Indonesian, Persian, Romanian, Vietnamese &
Narrative and cultural; English language safety where available \\
\addlinespace
Open-weight &
\href{https://openrouter.ai/deepseek/deepseek-v4-pro-0813}{DeepSeek V4 Pro 1.6T}; \href{https://huggingface.co/Qwen/Qwen3-235B-A22B-Instruct-2507}{Qwen3 235B}; \href{https://huggingface.co/mistralai/Mistral-Small-4-119B-2603}{Mistral 4 119B}; \href{https://huggingface.co/meta-llama/Llama-4-Scout-17B-16E-Instruct}{Llama 4 Scout 17B} &
OpenRouter API &
English + Arabic, Indonesian, Persian, Romanian, Vietnamese &
Narrative and cultural; language safety for completed English lanes \\
Local scaling &
\href{https://huggingface.co/mistralai/Ministral-3-3B-Instruct-2512}{Ministral 3B}, \href{https://huggingface.co/mistralai/Ministral-3-8B-Instruct-2512}{8B}, and \href{https://huggingface.co/mistralai/Ministral-3-14B-Instruct-2512}{14B} &
Local GPU inference &
English &
Controlled English scaling analysis \\
\addlinespace
Narrative evaluator &
GPT-5.6 Luna &
OpenAI API &
Narrative outputs from all compared models &
Reference-based narrative scoring and structured error annotations \\
\bottomrule
\end{tabular}
\end{table*}

\begin{table*}[htbp]
\centering
\small
\setlength{\tabcolsep}{5pt}
\caption{Standardized decoding settings across model families. API models use deterministic decoding with temperature $0$, while local models use greedy decoding. Task-specific output ceilings are shared across providers: 1,200 tokens for narrative generation, 1,400 for cultural prediction, 2,500 for LC Assessment, and 4,096 for narrative judging. Provider adapters map these settings to the corresponding API parameters (e.g., \texttt{max\_completion\_tokens}, \texttt{max\_output\_tokens}, \texttt{max\_tokens}, or \texttt{max\_new\_tokens}) without changing the numerical limits. All outputs are validated against the corresponding task schema.}
\vspace{5pt}
\label{tab:decoding-parameters}
\begin{adjustbox}{max width=\textwidth}
\begin{tabular}{@{}P{.20\textwidth}P{.42\textwidth}P{.15\textwidth}P{.16\textwidth}@{}}
\toprule
Model access & Evaluated models & Decoding & Task-specific output ceilings \\
\midrule
OpenAI API &
\href{https://developers.openai.com/api/docs/models/gpt-5.6-sol}{GPT-5.6 Sol}, \href{https://developers.openai.com/api/docs/models/gpt-5-nano}{GPT-5 Nano}; \href{https://developers.openai.com/api/docs/models/gpt-5.6-luna}{GPT-5.6 Luna} as narrative evaluator &
Temperature 0 &
1,200 / 1,400 / 2,500 / 4,096 \\
\addlinespace
Gemini API &
\href{https://docs.cloud.google.com/gemini-enterprise-agent-platform/models/gemini/3-8-flash}{Gemini 3.8 Flash}, \href{https://docs.cloud.google.com/gemini-enterprise-agent-platform/models/gemini/3-5-flash-lite}{Gemini 3.5 Flash Lite} &
Temperature 0 &
1,200 / 1,400 / 2,500 / 4,096 \\
\addlinespace
Anthropic API &
\href{https://platform.claude.com/docs/en/models/haiku-4-5/overview}{Claude Haiku 4.5} &
Temperature 0 &
1,200 / 1,400 / 2,500 / 4,096 \\
\addlinespace
OpenRouter API &
\href{https://openrouter.ai/deepseek/deepseek-v4-pro-0813}{DeepSeek V4 Pro 1.6T}, \href{https://huggingface.co/Qwen/Qwen3-235B-A22B-Instruct-2507}{Qwen3 235B}, \href{https://huggingface.co/mistralai/Mistral-Small-4-119B-2603}{Mistral 4 119B}, \href{https://huggingface.co/meta-llama/Llama-4-Scout-17B-16E-Instruct}{Llama 4 Scout 17B} &
Temperature 0 &
1,200 / 1,400 / 2,500 / 4,096 \\
\addlinespace
Local inference &
\href{https://huggingface.co/mistralai/Ministral-3-3B-Instruct-2512}{Ministral 3B}, \href{https://huggingface.co/mistralai/Ministral-3-8B-Instruct-2512}{8B}, and \href{https://huggingface.co/mistralai/Ministral-3-14B-Instruct-2512}{14B} &
Greedy &
1,200 / 1,400 / 2,500 / 4,096 \\
\bottomrule
\end{tabular}
\end{adjustbox}
\end{table*}

\subsection{Metric Definitions}
\label{app:metric-definitions}

The main paper reports one headline measure for each task; this section provides the complete definitions and secondary diagnostics. Within each comparison, all metrics are computed over the same evaluated films.

\paragraph{Narrative generation.}
Plot, synopsis, and key-message generations admit substantial valid paraphrase,
making surface-overlap metrics such as BLEU~\citep{papineni-etal-2002-bleu} and ROUGE~\citep{lin-2004-rouge} poorly suited to
capturing narrative quality; prior work likewise finds that automatic metrics
can correlate weakly with human judgments for open-ended generation
~\citep{fabbri-etal-2021-summeval,zhang-bansal-2021-finding,liu2023geval}.
Human evaluation provides richer semantic judgments but is costly,
time-consuming, and difficult to scale across the benchmark's models, tasks,
languages, and long-form inputs~\citep{zhang-bansal-2021-finding}.
We therefore use GPT-5.6 Luna as a reference-based LLM judge with an explicit
task-specific rubric, following recent work on structured model-based
evaluation~\citep{liu2023geval,li2025generationJudgment,gu2024surveyjudge}.
The judge first identifies concrete narrative errors and then assigns a score.
To ensure comparability, every model is evaluated with the same judge,
references, rubric, input ordering, and structured response format. Complete
judge instructions appear in Appendix~\ref{app:prompt-structures}.

For film $i$ and narrative task $t\in\{\mathrm{plot},\mathrm{synopsis},\mathrm{key}\}$, let $s_{i,t}\in\{1,\ldots,5\}$ be the evaluator's score. The aggregate narrative score is
\begin{equation}
S_{\mathrm{nar}}
=
\frac{1}{|\mathcal{D}|}
\sum_{i\in\mathcal{D}}
\frac{
s_{i,\mathrm{plot}}+
s_{i,\mathrm{synopsis}}+
s_{i,\mathrm{key}}
}{3}.
\label{eq:app-narrative-score}
\end{equation}
We additionally report plot, synopsis, and key-message scores separately and analyze the evaluator's task-specific error annotations.

\paragraph{Genre prediction.}
Genre prediction is multi-label. For true label set $G_i$ and predicted set $\hat{G}_i$, the headline metric is micro-F1:
\begin{equation}
\mathrm{F1}_{\mathrm{micro}}
=
\frac{
2\sum_{\ell}\mathrm{TP}_{\ell}
}{
2\sum_{\ell}\mathrm{TP}_{\ell}
+
\sum_{\ell}\mathrm{FP}_{\ell}
+
\sum_{\ell}\mathrm{FN}_{\ell}
}.
\label{eq:app-genre-micro-f1}
\end{equation}
We additionally report macro-F1, exact set match, and mean Jaccard similarity. Exact set match requires the complete predicted genre set to equal the gold set, while Jaccard similarity measures partial overlap.

\paragraph{Age suitability.}
Age labels are converted to their minimum ages in years. For gold age $a_i$ and predicted age $\hat{a}_i$, mean absolute error is
\begin{equation}
\mathrm{MAE}_{\mathrm{age}}
=
\frac{1}{|\mathcal{D}|}
\sum_{i\in\mathcal{D}}
|\hat{a}_i-a_i|.
\label{eq:app-age-mae}
\end{equation}
We also report exact match, accuracy within one year, and accuracy within two years. The error analysis further separates underprediction from overprediction.

\paragraph{Country-specific motion-picture ratings.}
Each country $c$ retains its own ordered label space $\mathcal{R}_c$. For gold rating $r_{i,c}$ and predicted rating $\hat{r}_{i,c}$, country exact match is
\begin{equation}
\mathrm{EM}_{\mathrm{country}}
=
\frac{1}{|\mathcal{D}||\mathcal{C}|}
\sum_{i\in\mathcal{D}}
\sum_{c\in\mathcal{C}}
\mathbb{1}\{\hat{r}_{i,c}=r_{i,c}\}.
\label{eq:app-country-em}
\end{equation}
We additionally report country-specific exact match, ordinal MAE, valid-prediction percentage, and performance relative to each country's majority-label baseline. Ordinal error is computed within each country's ordered label space rather than across a shared global scale.

\paragraph{Subtitle-grounded language safety.}
For film $i$ and lexical category $k$, let $E_{i,k}$ and $\hat{E}_{i,k}$ denote the gold and predicted subtitle-index evidence sets. Strict-category evidence precision and recall are
\begin{equation}
P_{\mathrm{strict}}
=
\frac{
\sum_{i,k}|\hat{E}_{i,k}\cap E_{i,k}|
}{
\sum_{i,k}|\hat{E}_{i,k}|
},
\qquad
R_{\mathrm{strict}}
=
\frac{
\sum_{i,k}|\hat{E}_{i,k}\cap E_{i,k}|
}{
\sum_{i,k}|E_{i,k}|
},
\label{eq:app-lc-pr}
\end{equation}
with
\begin{equation}
F1_{\mathrm{strict}}
=
\frac{
2P_{\mathrm{strict}}R_{\mathrm{strict}}
}{
P_{\mathrm{strict}}+R_{\mathrm{strict}}
}.
\label{eq:app-lc-f1}
\end{equation}
Strict matching requires both the correct subtitle index and the correct lexical category. Category-agnostic evidence metrics pool evidence indices across categories before matching, separating evidence localization from category assignment. We additionally report precision and recall separately, count MAE, per-category F1, gold-evidence miss rate, false-positive rate, and cross-category false-positive rate.

\paragraph{Cross-task composite.}
For models with complete comparable coverage, narrative, genre, age, and country metrics are first averaged equally over the six subtitle-input languages. Language-safety strict F1 remains English-only because the evidence annotations are defined over the English subtitle track. We normalize
\begin{equation}
N_{\mathrm{nar}} = 25(S_{\mathrm{nar}}-1),
\qquad
N_{\mathrm{age}}
=
100\left(
1-
\frac{\min(\mathrm{MAE}_{\mathrm{age}},16)}{16}
\right),
\label{eq:app-composite-normalization}
\end{equation}
where 16 is the span of the declared Common Sense age-label space from $2+$ to $18+$. The cultural component and overall score are
\begin{equation}
N_{\mathrm{cult}}
=
\frac{
N_{\mathrm{age}}+\mathrm{EM}_{\mathrm{country}}
}{2},
\qquad
S_{\mathrm{overall}}
=
\frac{
N_{\mathrm{nar}}
+
F1_{\mathrm{genre}}
+
N_{\mathrm{cult}}
+
F1_{\mathrm{LC}}
}{4}.
\label{eq:app-composite}
\end{equation}
The task-level tables retain every component in its original unit.

\subsection{Task and Evaluator Instructions}
\label{app:prompt-structures}
The following boxes reproduce the model-facing task instructions and narrative-evaluation criteria used in the experiments. Line breaks separate input constraints, requested outputs, and scoring rules for readability without changing the instruction content.
\begin{promptbox}{Narrative Understanding and Generation Prompt}
\small\raggedright
\textbf{Identity.}
You are a professional film analyst, story editor, and metadata curator. You write concise, faithful film descriptions from subtitles only.

\medskip
\textbf{Task.}
Given subtitle entries for one film, produce all outputs for a narrative understanding and generation task group in one English JSON object:
plot generation, synopsis generation, key-message generation, and genre prediction.

\medskip
\textbf{Input constraint.}
Use only the provided subtitles. The subtitles may be English or translated subtitles in another language, but every generated output must be written in English. Do not rely on outside knowledge, the film title, cast, franchise knowledge, reviews, or memorized plot information. If a detail is not inferable from the subtitles, omit it. The user message specifies the subtitle language as \texttt{inputLanguage}; treat that language label as authoritative.

\medskip
\textbf{Writing standards.}
\begin{itemize}[leftmargin=*,itemsep=2pt,topsep=3pt]
\item \textbf{Plot (45--80 words):} a present-tense, third-person premise naming the central character or group, setup, main conflict, and driving situation.
\item \textbf{Synopsis (180--300 words):} a present-tense, third-person, spoiler-aware account of the beginning, major developments, escalation, character goals or conflicts, and resolution when inferable. Connect events causally.
\item \textbf{Key message (8--25 words):} one moral, social, emotional, or positive thematic takeaway, not another plot summary.
\item \textbf{Genres (1--4 labels):} select exact strings from the allowed set using the narrative arc, characters, setting, tone, and dominant conflict; do not invent or substitute labels.
\end{itemize}

\medskip
\textbf{Allowed genre labels.}
\begin{quote}
\small\raggedright
Action \textperiodcentered{} Adventure \textperiodcentered{} Animation \textperiodcentered{} Biography \textperiodcentered{} Comedy \textperiodcentered{} Crime \textperiodcentered{} Documentary \textperiodcentered{} Drama \textperiodcentered{} Family \textperiodcentered{} Fantasy \textperiodcentered{} Film-Noir \textperiodcentered{} History \textperiodcentered{} Horror \textperiodcentered{} Music \textperiodcentered{} Musical \textperiodcentered{} Mystery \textperiodcentered{} Romance \textperiodcentered{} Sci-Fi \textperiodcentered{} Sport \textperiodcentered{} Thriller \textperiodcentered{} War \textperiodcentered{} Western.
\end{quote}

\medskip
\textbf{Output rules.}
Return only valid JSON that conforms to the provided schema. Do not include Markdown, commentary, confidence scores, explanations, wrapper keys, or snake-case aliases.

\begin{lstlisting}
{
  "id": 0,
  "plot": "...",
  "synopsis": "...",
  "keyMessage": "...",
  "genres": ["Drama"]
}
\end{lstlisting}
\end{promptbox}

\begin{promptbox}{Subtitle-Grounded Language-Safety Assessment Prompt}
\small\raggedright
\textbf{Identity.}
You are a subtitle evidence auditor for film language safety. You identify subtitle-grounded language-content categories using English subtitle lines.

\medskip
\textbf{Task framing.}
This is an exhaustive lexical extraction task, not a summary, representative-assessment, or severity-only task. Given English subtitle entries for one film or one non-overlapping subtitle chunk, scan all provided subtitle lines and return language-content counts and evidence indices in one compact JSON object. Long evidence lists are expected when many subtitle lines match.

\medskip
\textbf{Input constraint.}
Use only the provided English subtitles. Do not use the film title, cast, reviews, ratings databases, or outside knowledge. Do not output raw offensive wording. Evidence must be returned only as subtitle indices.

\medskip
\textbf{Category priority rules.}
Assign each matched expression to the most specific applicable category:
\begin{itemize}[leftmargin=*,itemsep=1pt,topsep=3pt]
\item \texttt{strongProfanity}: F-word forms and close derivatives.
\item \texttt{crudeBodilyLanguage}: S-word forms, scatological terms, and crude anatomical language.
\item \texttt{religiousProfanityAndExclamation}: religious profanity or exclamations.
\item \texttt{mildObscenity}: lower-intensity obscene, impolite, softened, or euphemistic expressions; never a catch-all for stronger categories.
\end{itemize}
One subtitle line may appear under multiple categories if it contains multiple category types.

\medskip
\textbf{Counting and evidence rules.}
The category count is the number of matched expressions, not necessarily the number of evidence lines. Include every index containing a matching expression for that category. For example, two same-category expressions in one line count twice but yield one evidence index. If uncertain, prefer not to include an index.

\medskip
\textbf{Output rules.}
Return only valid JSON with flat schema keys. Do not include matched text, nested evidence objects, time intervals, Markdown, commentary, or wrapper keys.

\smallskip
\textbf{Required flat JSON fields.}
\begin{center}
\footnotesize
\begin{adjustbox}{max width=0.96\linewidth}
\begin{tabular}{@{}ll@{}}
\toprule
\textbf{Count key} & \textbf{Evidence-index key} \\
\midrule
\texttt{strongProfanityCount} & \texttt{strongProfanityIndices} \\
\texttt{crudeBodilyLanguageCount} & \texttt{crudeBodilyLanguageIndices} \\
\texttt{mildObscenityCount} & \texttt{mildObscenityIndices} \\
\texttt{religiousProfanityAndExclamationCount} & \texttt{religiousProfanityAndExclamationIndices} \\
\bottomrule
\end{tabular}
\end{adjustbox}
\end{center}
\end{promptbox}

\begin{promptbox}{Reference-Based Narrative LLM-as-Judge Prompt Structure}
\small\raggedright
\textbf{Identity.}
The judge is instructed to act as a senior film critic, story analyst, and metadata quality auditor. The prompt uses a two-stage reference-based procedure to reduce impressionistic scoring.

\medskip
\textbf{Scope and inputs.}
For each sample, the judge receives the sample id, gold reference plot, gold reference synopsis, gold reference key message, and the model-generated plot, synopsis, and key message. The judge evaluates only the three narrative generation tasks. Genre prediction is excluded from the judge prompt and evaluated separately with automatic multi-label metrics.

\medskip
\textbf{Evaluation anchor.}
References are the evaluation anchor. The judge does not require exact wording, but penalizes missing central conflicts, missing resolutions, invented events, wrong relationships, weak causality, generic key messages, and outputs that are locally plausible but globally mis-centered. Each task is judged against its task-matched reference, while the other reference fields may be used only to disambiguate the film's central story or theme.

For plot scoring, the evaluator gives priority to the reference plot and uses the reference synopsis only to clarify factual context. It checks whether the response preserves the distinctive premise, central characters, setup, conflict, and dominant emphasis. A fluent but generic description does not receive a high score if it omits the film-specific trigger or shifts the story's center of gravity.

\medskip
\textbf{Stage 0: centrality audit.}
Before scoring, the judge internally identifies central characters or groups, relationships, motivations, conflicts, major events, turning points, end states, and dominant theme. This inventory is not output; it is used to avoid rewarding outputs that over-focus on secondary subplots or local subtitle-visible details.

\medskip
\textbf{Stage 1: standardized error identification.}
For each task, the judge first lists concrete errors with a type, severity (\texttt{minor}, \texttt{moderate}, or \texttt{major}), and short explanation. The type inventories are task-specific:
\begin{itemize}[leftmargin=*,itemsep=2pt,topsep=3pt]
\item \textbf{Plot:} \texttt{missingCentralConflict}, \texttt{missingProtagonistOrGroup}, \texttt{wrongSetup}, \texttt{wrongRelationship}, \texttt{inventedEvent}, \texttt{wrongEmphasis}, \texttt{tooVague}, \texttt{tooDetailed}, \texttt{styleOrLengthIssue}.
\item \textbf{Synopsis:} \texttt{missingMajorEvent}, \texttt{missingResolution}, \texttt{wrongResolution}, \texttt{weakCausality}, \texttt{missingCharacterMotivation}, \texttt{wrongRelationship}, \texttt{inventedEvent}, \texttt{wrongEventOrder}, \texttt{tooVague}, \texttt{tooDetailed}, \texttt{styleOrLengthIssue}.
\item \textbf{Key message:} \texttt{semanticMismatch}, \texttt{tooGeneric}, \texttt{tooPlotLike}, \texttt{missesPositiveMessage}, \texttt{contradictedByReference}, \texttt{unsupportedTheme}, \texttt{styleOrLengthIssue}.
\end{itemize}

\medskip
\textbf{Stage 2: score assignment.}
After identifying errors, the judge assigns dimension scores and an overall score on a 1--5 scale:
\begin{center}
\small
\begin{tabularx}{0.96\linewidth}{@{}cX@{}}
\toprule
\textbf{Score} & \textbf{Interpretation} \\
\midrule
5 & Exceptionally faithful, specific, and well-centered. \\
4 & Strong, with only limited omissions or drift. \\
3 & Recognizable, but incomplete or materially mis-centered. \\
2 & Substantial omissions, major factual errors, invented events, or a weak/generic response. \\
1 & Mostly wrong, contradictory, hallucinated, or non-responsive. \\
\bottomrule
\end{tabularx}
\end{center}
The plot rubric applies these anchors consistently: a strong score requires the distinctive premise, central conflict, and dominant emphasis to be substantially preserved.

\medskip
\textbf{Output contract.}
Return one JSON object with \texttt{id}, \texttt{plotEvaluation}, \texttt{synopsisEvaluation}, and \texttt{keyMessageEvaluation}. Each task record contains typed, severity-tagged \texttt{errors}, task-specific 1--5 dimension scores and an \texttt{overall} score, and a concise \texttt{overallRationale}. No Markdown or commentary appears outside the schema; the complete field-level schema is retained in the experiment artifacts.
\end{promptbox}

\begin{promptbox}{Cultural Prediction and Assessment Prompt}
\small\raggedright
\textbf{Identity.}
You are a professional film classification analyst. You infer audience suitability and country-specific motion-picture classifications from subtitles only.

\medskip
\textbf{Task.}
Given subtitle entries for one film, predict the cultural content labels in one English JSON object: age-suitability prediction and country-specific motion-picture rating prediction.

\medskip
\textbf{Input constraint.}
Use only the provided subtitles. The subtitles may be English or translated subtitles in another language, but every output field and rationale must be written in English. The user message specifies \texttt{inputLanguage}; treat that language label as authoritative. Do not rely on film title, cast, franchise knowledge, reviews, ratings databases, trailers, images, or memorized outside knowledge. If a content issue is not inferable from subtitles, do not invent it.

\medskip
\textbf{Classification principles.}
Treat ratings as audience-suitability classifications. Consider subtitle-visible evidence such as profanity, threats, fear, violence described in dialogue, sexual references, substance references, mature themes, emotional intensity, crime, death, and disturbing situations. Be conservative when visual evidence would be required but is absent from subtitles.

Country ratings are culturally situated. Predict each country independently using that country's allowed label set and ordered scale. Do not collapse countries into a universal rating.

\medskip
\textbf{Allowed labels.}
\textbf{Age suitability:} 2+, 3+, 4+, 5+, 6+, 7+, 8+, 9+, 10+, 11+, 12+, 13+, 14+, 15+, 16+, 17+, 18+.

\smallskip
\textbf{Country-specific motion-picture labels:}
\begin{center}
\small
\begin{tabularx}{0.98\linewidth}{@{}lX@{}}
\toprule
\textbf{Country} & \textbf{Allowed labels} \\
\midrule
Australia & G, PG, M, MA15+, R18+ \\
Brazil & Livre, 10, 12, 14, 16, 18 \\
France & Tous publics, 12, 16, 18 \\
Germany & 0, 6, 12, 16, 18 \\
Netherlands & AL, 6, 9, 12, 14, 16, 18 \\
Singapore & G, PG, PG13, NC16, M18, R21 \\
South Korea & All, 12, 15, 19 \\
Sweden & Btl, 7, 11, 15 \\
United Kingdom & U, PG, 12, 15, 18 \\
United States & G, PG, PG-13, R, NC-17 \\
\bottomrule
\end{tabularx}
\end{center}

\medskip
\textbf{Output rules.}
Return only valid JSON that conforms to the provided schema. Predict exactly one age-suitability label and exactly one rating for each of the ten countries. Use concise rationales that reference subtitle-visible evidence generally, not raw offensive wording. Do not include Markdown, commentary, confidence scores, wrapper keys, or snake-case aliases.

\smallskip
\emph{Abbreviated schema illustration: the actual response includes one country object for each of the ten countries.}

\begin{lstlisting}
{
  "id": 0,
  "ageSuitabilityPrediction": {
    "ageRating": "13+",
    "rationale": "Concise subtitle-visible rationale."
  },
  "countryMotionPictureRatingPrediction": [
    {
      "country": "Australia",
      "rating": "M",
      "rationale": "Concise subtitle-visible rationale."
    }
  ]
}
\end{lstlisting}
\end{promptbox}

\subsection{Independent Narrative-Judge Agreement}
\label{app:independent-judge-agreement}

We assess agreement between Claude Sonnet~5\footnote{\url{https://platform.claude.com/docs/en/models/sonnet-5/}} and GPT-5.6 Luna on the same
model-generated narrative outputs using shared references and the same two-stage
evaluation rubric. The audit samples 50 films and nine evaluated models,
yielding 450 model--film cases. Claude Sonnet~5 uses its default settings.
Complete plot, synopsis, and key-message scores are available for every case,
giving 1,350 paired task ratings.

Table~\ref{tab:independent-judge-agreement} shows high agreement across the
pooled ratings. Pearson correlation is $r=0.808$, Spearman correlation is
$\rho=0.815$, and quadratic weighted $\kappa=0.778$. Film-clustered bootstrap
95\% confidence intervals are $[0.780,0.833]$, $[0.785,0.839]$, and
$[0.749,0.803]$, respectively. Exact agreement is 56.8\%, while 98.4\% of
paired ratings differ by at most one point. Claude Sonnet~5 assigns scores
0.19 points lower on average than GPT-5.6 Luna
(95\% CI $[-0.253,-0.122]$). Thus, the two judges show closely aligned
relative scoring and near-universal agreement within one point, alongside a
small systematic difference in score level.

\begin{table}[htbp]
\centering
\small
\caption{Agreement between Claude Sonnet~5 and GPT-5.6 Luna on narrative-task
scores in the 50-film audit. The analysis covers all 450 model--film cases
from nine evaluated models, yielding 1,350 paired task ratings across plot,
synopsis, and key message.}
\vspace{2pt}
\label{tab:independent-judge-agreement}
\resizebox{\linewidth}{!} 
{
\begin{tabular}{lrrrrrrr}
\toprule
Task & Paired scores & Pearson $r$ & Spearman $\rho$ & Weighted $\kappa$
& Exact (\%) & Within 1 (\%) & Mean $\Delta$ \\
\midrule
All three tasks & 1350 & 0.808 & 0.815 & 0.778 & 56.8 & 98.4 & $-0.192$ \\
Plot            & 450  & 0.761 & 0.710 & 0.675 & 43.4 & 96.9 & $-0.438$ \\
Synopsis        & 450  & 0.750 & 0.751 & 0.729 & 61.5 & 98.2 & $-0.084$ \\
Key message     & 450  & 0.766 & 0.776 & 0.763 & 65.5 & 100.0 & $-0.053$ \\
\bottomrule
\end{tabular}
}
\vspace{2pt}
\begin{minipage}{0.98\linewidth}\footnotesize
\emph{Note.} Each task is scored on a 1--5 scale. Mean $\Delta$ is
Claude Sonnet~5 minus GPT-5.6 Luna, so negative values indicate lower
Claude Sonnet~5 ratings. Weighted $\kappa$ denotes quadratic weighted
$\kappa$. For the pooled Pearson correlation, Spearman correlation, weighted
$\kappa$, and mean difference, 95\% confidence intervals use 5,000
film-clustered bootstrap resamples. Task-specific rows report point estimates.
\end{minipage}
\end{table}

\subsection{Response-Format Reliability}
\label{app:structured-output-recovery-audit}
A response is scored only after it satisfies the task schema. If an identifiable prediction is wrapped in Markdown or contains a strictly syntactic JSON defect, deterministic repair may correct the serialization before validation is repeated. A local model may first receive one formatting-only retry with the same evidence. Neither operation supplies a gold label or changes generated text, predicted labels, ratings, categories, or evidence indices.

\begin{figure}[htbp]
\centering
\includegraphics[width=0.90\linewidth]{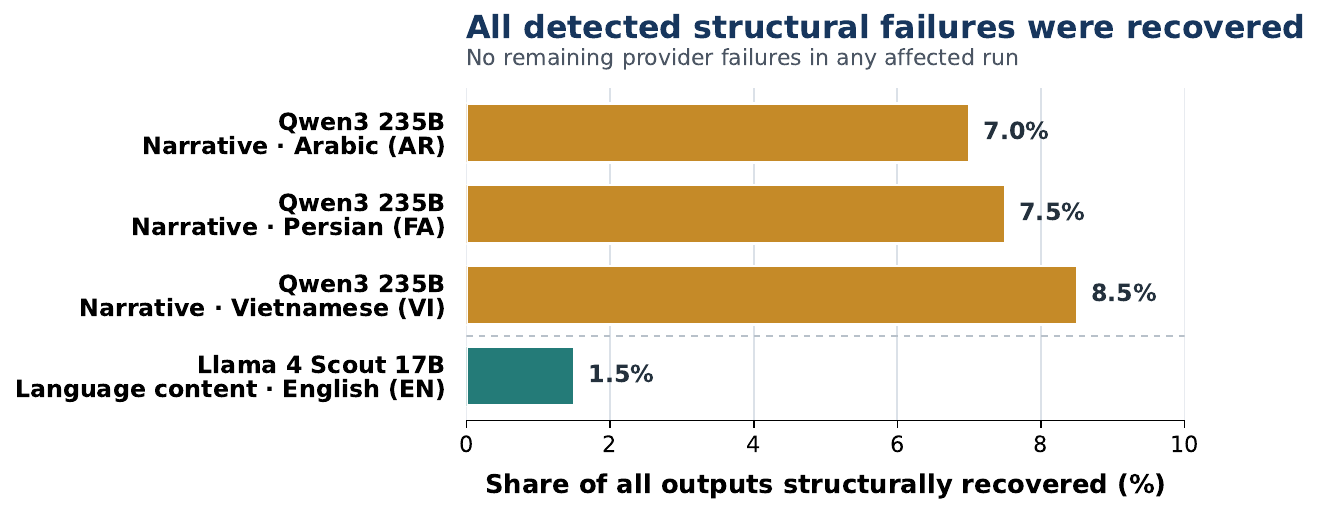}
\caption{Response-format recovery in affected runs. Repairs modify serialization only; they do not change task content or scores.}
\label{fig:structural-recovery-audit}
\end{figure}

Responses with no usable prediction, provider errors, or substantial truncation remain failures. Table~\ref{tab:structural-recovery-audit} and Figure~\ref{fig:structural-recovery-audit} isolate this reliability layer from semantic task accuracy.
\begin{table}[htbp]
\centering
\small
\setlength{\tabcolsep}{5pt}
\caption{Structured-output recovery audit. AR, FA, VI, and EN denote Arabic, Persian, Vietnamese, and English, respectively. Arrows indicate the preferred direction: failure rates are lower-is-better ($\downarrow$), while successful recovery is higher-is-better ($\uparrow$). Percentages summarize serialization and provider-cleanup outcomes for saved raw model outputs. Recovery is restricted to structural parsing and schema validation; semantic labels, predictions, evidence, and scores are never altered.}
\vspace{5pt}
\label{tab:structural-recovery-audit}
\begin{adjustbox}{max width=\textwidth}
\begin{tabular}{lllrrr}
\toprule
Model & Task & Language & Initial structural failure (\%) $\downarrow$ & Structurally recovered (\%) $\uparrow$ & Remaining provider failure (\%) $\downarrow$ \\
\midrule
\href{https://huggingface.co/Qwen/Qwen3-235B-A22B-Instruct-2507}{Qwen3 235B} & Narrative & AR & 7.0 & 7.0 & 0.0 \\
\href{https://huggingface.co/Qwen/Qwen3-235B-A22B-Instruct-2507}{Qwen3 235B} & Narrative & FA & 7.5 & 7.5 & 0.0 \\
\href{https://huggingface.co/Qwen/Qwen3-235B-A22B-Instruct-2507}{Qwen3 235B} & Narrative & VI & 8.5 & 8.5 & 0.0 \\
\href{https://huggingface.co/meta-llama/Llama-4-Scout-17B-16E-Instruct}{Llama 4 Scout 17B} & Language-Safety & EN & 1.5 & 1.5 & 0.0 \\
\bottomrule
\end{tabular}%
\end{adjustbox}
\end{table}

\subsection{Title/Year Recall and Character-Name Counterfactuals}
\label{app:source-exposure-audit}

We conduct two targeted probes to separate information recoverable from film
identity cues from information used during subtitle-conditioned generation.
The first measures how much task performance can be recovered from title and
release year alone. The second measures how narrative generation responds when
central-character names are systematically altered while the surrounding
subtitle sequence and corresponding references are preserved.

\paragraph{Experimental conditions.}
The standard benchmark provides subtitle text, the input-language identifier,
and task instructions/schema, but not film titles, release years, cast,
reference narratives, or gold labels. For the title/year probe, we select 50 films evaluated with English subtitles evenly across five strata of narrative-judge
difficulty. Each model receives only the film title and release year and
generates a plot, synopsis, key message, and genre labels. We compare these
outputs with the standard subtitle-only outputs for the same 50 films.

For the character-name counterfactual, we select 20 films, four from each
difficulty stratum, and exchange two manually screened central-character
aliases throughout each film's English subtitles and narrative references.
Models then generate from the transformed subtitles without access to the film
title or release year. We compare these generations with the standard
subtitle-only outputs for the same films and score them against the
correspondingly transformed references. Genre labels are unchanged by the
transformation and are therefore excluded from the paired counterfactual
narrative comparison.

\paragraph{Scoring and uncertainty.}
GPT-5.6 Luna applies the same two-stage reference-based rubric used for the main
narrative evaluation. Existing subtitle-only scores are reused, and all 350 new
outputs are collected successfully: 50 title/year and 20 counterfactual
generations for each of five models.
Tables~\ref{tab:contamination-title-year}
and~\ref{tab:contamination-counterfactual} report per-task means and paired
changes in the mean of plot, synopsis, and key-message scores on the 1--5
scale. Uncertainty is estimated using paired film-level 95\% bootstrap
intervals. Table~\ref{tab:contamination-genre} reports multi-label genre
micro-F1 and exact match for the paired 50-film subtitle-only and title/year
conditions.

\begin{table}[htbp]
\centering
\small
\caption{Narrative performance with title and year only versus subtitles. Task
entries are mean rubric scores on a 1--5 scale, shown as subtitle-only/title-year-only
for paired films. The final column is the paired change in the per-film mean
across the three tasks, with a 95\% film-level bootstrap interval.}
\vspace{5pt}
\label{tab:contamination-title-year}
\begin{tabular}{lcccc}
\toprule
Model & Plot (S/T) & Synopsis (S/T) & Key message (S/T) & Mean change [95\% CI] \\
\midrule
GPT-5.6 Sol      & 4.58 / 4.68 & 3.38 / 3.54 & 3.00 / 2.96 & $+0.07$ [$-0.05$, $+0.19$] \\
Gemini 3.8 Flash & 4.58 / 4.68 & 3.58 / 3.36 & 3.14 / 3.20 & $-0.02$ [$-0.13$, $+0.09$] \\
Claude Haiku 4.5 & 4.12 / 3.94 & 2.48 / 2.24 & 2.94 / 2.70 & $-0.22$ [$-0.40$, $-0.05$] \\
DeepSeek V4 Pro  & 4.38 / 4.52 & 3.30 / 2.96 & 2.98 / 3.06 & $-0.04$ [$-0.20$, $+0.13$] \\
Llama 4 Scout    & 2.82 / 2.60 & 1.88 / 1.60 & 2.50 / 2.30 & $-0.23$ [$-0.47$, $-0.01$] \\
\bottomrule
\end{tabular}

\vspace{2pt}
\begin{minipage}{0.98\linewidth}
\footnotesize
\emph{Note.} S/T denotes subtitle-only/title-year-only. $n=50$ paired films
for every model. A positive change favors title/year-only generation; a
negative change favors subtitle-conditioned generation.
\end{minipage}
\end{table}
\begin{table}[htbp]
\centering
\small
\caption{Narrative performance on the character-name counterfactual subset.
Task entries are mean rubric scores on a 1--5 scale, shown as
subtitle-only/counterfactual-subtitle for paired films. The final column is the
paired change in the per-film mean across the three tasks, with a 95\% film-level
bootstrap interval.}
\vspace{5pt}
\label{tab:contamination-counterfactual}
\begin{tabular}{lcccc}
\toprule
Model & Plot (S/C) & Synopsis (S/C) & Key message (S/C) & Mean change [95\% CI] \\
\midrule
GPT-5.6 Sol      & 4.70 / 3.80 & 3.30 / 2.70 & 3.00 / 3.05 & $-0.48$ [$-0.82$, $-0.17$] \\
Gemini 3.8 Flash & 4.65 / 4.50 & 3.55 / 2.80 & 3.15 / 3.00 & $-0.35$ [$-0.57$, $-0.15$] \\
Claude Haiku 4.5 & 4.10 / 3.95 & 2.40 / 2.35 & 2.90 / 2.75 & $-0.12$ [$-0.30$, $+0.07$] \\
DeepSeek V4 Pro  & 4.45 / 4.25 & 3.20 / 2.60 & 2.95 / 3.15 & $-0.20$ [$-0.45$, $+0.03$] \\
Llama 4 Scout    & 2.80 / 2.65 & 1.90 / 1.70 & 2.55 / 2.25 & $-0.22$ [$-0.53$, $+0.10$] \\
\bottomrule
\end{tabular}

\vspace{2pt}
\begin{minipage}{0.98\linewidth}
\footnotesize
\emph{Note.} S/C denotes subtitle-only/counterfactual-subtitle. $n=20$
paired films (four sampled from each of five narrative-difficulty strata).
In the counterfactual condition, two character aliases are exchanged in the
subtitles and narrative references; the film title and year are withheld.
Negative changes indicate lower rubric scores under the counterfactual
condition.
\end{minipage}
\end{table}
\begin{table}[htbp]
\centering
\small
\caption{Genre prediction from subtitles versus film title and year on the paired
50-film sample. Micro-F1 summarizes multi-label decisions across all genre
labels; exact match requires the complete predicted genre set to match the reference.}
\vspace{5pt}
\label{tab:contamination-genre}
\begin{tabular}{lrrrrr}
\toprule
Model & Micro-F1 (S) & Micro-F1 (T) & Change (pp) & Exact (S, \%) & Exact (T, \%) \\
\midrule
GPT-5.6 Sol      & 75.7 & 82.3 & $+6.6$ & 20.0 & 30.0 \\
Gemini 3.8 Flash & 79.8 & 84.7 & $+4.9$ & 24.0 & 40.0 \\
Claude Haiku 4.5 & 67.3 & 72.4 & $+5.1$ & 14.0 & 8.0 \\
DeepSeek V4 Pro  & 79.3 & 81.8 & $+2.5$ & 30.0 & 34.0 \\
Llama 4 Scout    & 61.2 & 72.2 & $+11.0$ & 2.0 & 18.0 \\
\bottomrule
\end{tabular}

\vspace{2pt}
\begin{minipage}{0.98\linewidth}
\footnotesize
\emph{Note.} S/T denotes subtitle-only/title-year-only; both conditions use
the same 50 films and references. Micro-F1 values and changes are percentages
and percentage points, respectively. Higher is better.
\end{minipage}
\end{table}

\paragraph{Film-identity cues carry substantial task signal.}
Title/year-only narrative performance remains close to the subtitle-only
baseline for several models: paired mean changes range from $-0.23$ to
$+0.07$. The intervals exclude zero for Claude Haiku 4.5 and Llama 4 Scout,
both favoring subtitle-conditioned generation. Genre prediction shows a
stronger identity-cue effect: all five models achieve higher micro-F1 from
title and year alone, with gains of 2.5--11.0 percentage points. This contrast
shows that film identity alone provides substantial information for genre
recognition, while detailed narrative reconstruction benefits more unevenly
from subtitle evidence across models.

\paragraph{Narrative generation is sensitive to character-name perturbations.}
The counterfactual transformation lowers the paired narrative mean for all five
models. The largest decreases occur for GPT-5.6 Sol ($-0.48$) and Gemini 3.8
Flash ($-0.35$), whose bootstrap intervals exclude zero; intervals for the
other three models include zero. The consistent direction across models shows
that altering central-character names can materially change
subtitle-conditioned narrative generation. Because a name substitution can
simultaneously change lexical form, gender cues, cultural associations,
frequency, tokenization, and character familiarity, the intervention identifies
sensitivity to the substitution as a whole rather than to any single property
of the names. Together, the two probes show that benchmark performance reflects
both information available from film identity and the model's response to
evidence supplied in the subtitle context.

\section{Additional Evidence for RQ~\ref{rq:model-ranking}: Model Capability Profiles}
\label{app:rq1-details}
\subsection{English and Cross-Lingual Headline Results}
\label{app:rq1-headline-results}
Tables~\ref{tab:english-headline-results} and~\ref{tab:crosslingual-headline-results} separate the benchmark's task families under English subtitle input and the equal-weight average of Arabic, Indonesian, Persian, Romanian, and Vietnamese. Reporting the original metric units keeps task-level differences visible rather than collapsing them into the cross-task composite.

\paragraph{Task strengths already diverge under English input.}
DeepSeek V4 Pro approaches the two leading frontier models on English genre micro-F1 (83.2\%) and age accuracy within one year (82.5\%), while its country-rating exact match is 58.5\%, compared with 78.8\% for Gemini 3.8 Flash. GPT-5.6 Sol and Gemini are nearly tied on English synopsis generation (3.42 and 3.43), yet their country exact-match rates differ by 11.6 percentage points. Narrative quality and cultural prediction therefore need not move together even under the same subtitle language.

\paragraph{Model access does not determine a uniform capability profile.}
Across the six-language averages, DeepSeek V4 Pro exceeds Claude Haiku 4.5 on narrative overall, genre micro-F1, age prediction, country-rating exact match, and strict language-safety evidence F1. Its gaps from GPT-5.6 Sol vary considerably by task, from near-equal genre and age performance to larger differences in narrative reconstruction, country ratings, and evidence grounding.

\paragraph{Task-specific differences persist under non-English subtitle input.}
Gemini 3.8 Flash retains the highest genre micro-F1, age exact match, and country-rating exact match in both the English and five-language-average results. GPT-5.6 Sol achieves the highest cross-lingual synopsis score at 3.39/5, while the highest cross-lingual age exact-match rate is 33.8\% for Gemini. Language-safety results are not included in this comparison because the evidence-grounding task is evaluated over English subtitle indices.

\begin{table*}[htbp]
\centering
\small
\setlength{\tabcolsep}{2.5pt}
\renewcommand{\arraystretch}{1.25}
\caption{English-only results. Arrows indicate the preferred direction: $\uparrow$ is higher-is-better and $\downarrow$ is lower-is-better. Narrative metrics are reference-based scores on a 1--5 scale; genre, age-threshold, and country exact-match metrics are percentages, while MAE metrics report error. Within each metric column, shading is normalized across the displayed models in the preferred direction: muted blue denotes narrative metrics and muted bronze denotes cultural metrics. Shade intensity is therefore not comparable across columns. Best values are bold, and second-best distinct values are underlined at the displayed precision; ties share the same emphasis.}
\vspace{5pt}
\label{tab:english-headline-results}
\begin{adjustbox}{max width=\textwidth}
\begin{tabular}{llrrrrrrrrrr}
\toprule
Model & Family & \multicolumn{5}{c}{\textbf{Narrative Understanding and Generation}} & \multicolumn{5}{c}{\textbf{Cultural Prediction and Assessment}} \\
\cmidrule(lr){3-7}\cmidrule(lr){8-12}
 & & \shortstack{Plot\\$\uparrow$} & \shortstack{Synopsis\\$\uparrow$} & \shortstack{Key\\message $\uparrow$} & \shortstack{Overall\\$\uparrow$} & \shortstack{Genre\\micro-F1 $\uparrow$} & \shortstack{Age\\exact $\uparrow$} & \shortstack{Age\\$\pm$1 $\uparrow$} & \shortstack{Age\\MAE $\downarrow$} & \shortstack{Country\\EM $\uparrow$} & \shortstack{Country\\MAE $\downarrow$} \\
\midrule
\href{https://developers.openai.com/api/docs/models/gpt-5.6-sol}{GPT-5.6 Sol} & Closed frontier & \cellcolor{cineNarrative!43!white}\textbf{4.17} & \cellcolor{cineNarrative!43!white}\underline{3.42} & \cellcolor{cineNarrative!37!white}\underline{3.10} & \cellcolor{cineNarrative!42!white}\underline{3.56} & \cellcolor{cineNarrative!38!white}82.2 & \cellcolor{cineCultural!38!white}33.5 & \cellcolor{cineCultural!42!white}82.0 & \cellcolor{cineCultural!42!white}\underline{0.88} & \cellcolor{cineCultural!34!white}\underline{67.2} & \cellcolor{cineCultural!36!white}\underline{0.42} \\
\href{https://docs.cloud.google.com/gemini-enterprise-agent-platform/models/gemini/3-8-flash}{Gemini 3.8 Flash} & Closed frontier & \cellcolor{cineNarrative!43!white}\textbf{4.17} & \cellcolor{cineNarrative!43!white}\textbf{3.43} & \cellcolor{cineNarrative!43!white}\textbf{3.23} & \cellcolor{cineNarrative!43!white}\textbf{3.61} & \cellcolor{cineNarrative!43!white}\textbf{84.7} & \cellcolor{cineCultural!43!white}\textbf{36.5} & \cellcolor{cineCultural!43!white}\textbf{83.0} & \cellcolor{cineCultural!43!white}\textbf{0.85} & \cellcolor{cineCultural!43!white}\textbf{78.8} & \cellcolor{cineCultural!43!white}\textbf{0.29} \\
\href{https://platform.claude.com/docs/en/models/haiku-4-5/overview}{Claude Haiku 4.5} & Closed frontier & \cellcolor{cineNarrative!31!white}3.62 & \cellcolor{cineNarrative!22!white}2.54 & \cellcolor{cineNarrative!29!white}2.92 & \cellcolor{cineNarrative!27!white}3.02 & \cellcolor{cineNarrative!16!white}72.2 & \cellcolor{cineCultural!30!white}29.5 & \cellcolor{cineCultural!38!white}78.0 & \cellcolor{cineCultural!39!white}0.99 & \cellcolor{cineCultural!27!white}58.0 & \cellcolor{cineCultural!29!white}0.53 \\
\midrule
\href{https://developers.openai.com/api/docs/models/gpt-5-nano}{GPT-5 Nano} & Closed baseline & \cellcolor{cineNarrative!22!white}3.21 & \cellcolor{cineNarrative!11!white}2.06 & \cellcolor{cineNarrative!20!white}2.71 & \cellcolor{cineNarrative!18!white}2.66 & \cellcolor{cineNarrative!21!white}74.8 & \cellcolor{cineCultural!12!white}19.5 & \cellcolor{cineCultural!22!white}60.5 & \cellcolor{cineCultural!24!white}1.55 & \cellcolor{cineCultural!15!white}43.8 & \cellcolor{cineCultural!17!white}0.75 \\
\href{https://docs.cloud.google.com/gemini-enterprise-agent-platform/models/gemini/3-5-flash-lite}{Gemini 3.5 Flash Lite} & Closed baseline & \cellcolor{cineNarrative!39!white}\underline{3.97} & \cellcolor{cineNarrative!34!white}3.02 & \cellcolor{cineNarrative!36!white}3.06 & \cellcolor{cineNarrative!36!white}3.35 & \cellcolor{cineNarrative!37!white}81.8 & \cellcolor{cineCultural!36!white}32.5 & \cellcolor{cineCultural!36!white}75.5 & \cellcolor{cineCultural!39!white}0.99 & \cellcolor{cineCultural!33!white}66.3 & \cellcolor{cineCultural!34!white}0.44 \\
\midrule
\href{https://openrouter.ai/deepseek/deepseek-v4-pro-0813}{DeepSeek V4 Pro 1.6T} & Open-weight & \cellcolor{cineNarrative!37!white}3.91 & \cellcolor{cineNarrative!35!white}3.08 & \cellcolor{cineNarrative!28!white}2.90 & \cellcolor{cineNarrative!35!white}3.30 & \cellcolor{cineNarrative!40!white}\underline{83.2} & \cellcolor{cineCultural!39!white}\underline{34.5} & \cellcolor{cineCultural!43!white}\underline{82.5} & \cellcolor{cineCultural!42!white}\underline{0.88} & \cellcolor{cineCultural!27!white}58.5 & \cellcolor{cineCultural!31!white}0.51 \\
\href{https://huggingface.co/Qwen/Qwen3-235B-A22B-Instruct-2507}{Qwen3 235B} & Open-weight & \cellcolor{cineNarrative!27!white}3.43 & \cellcolor{cineNarrative!20!white}2.46 & \cellcolor{cineNarrative!28!white}2.88 & \cellcolor{cineNarrative!25!white}2.92 & \cellcolor{cineNarrative!11!white}69.9 & \cellcolor{cineCultural!16!white}21.5 & \cellcolor{cineCultural!25!white}63.5 & \cellcolor{cineCultural!30!white}1.36 & \cellcolor{cineCultural!22!white}51.7 & \cellcolor{cineCultural!25!white}0.60 \\
\href{https://huggingface.co/mistralai/Mistral-Small-4-119B-2603}{Mistral 4 119B} & Open-weight & \cellcolor{cineNarrative!17!white}2.98 & \cellcolor{cineNarrative!12!white}2.12 & \cellcolor{cineNarrative!12!white}2.52 & \cellcolor{cineNarrative!14!white}2.54 & \cellcolor{cineNarrative!11!white}70.0 & \cellcolor{cineCultural!9!white}17.5 & \cellcolor{cineCultural!18!white}56.0 & \cellcolor{cineCultural!18!white}1.77 & \cellcolor{cineCultural!20!white}49.4 & \cellcolor{cineCultural!22!white}0.65 \\
\href{https://huggingface.co/meta-llama/Llama-4-Scout-17B-16E-Instruct}{Llama 4 Scout 17B} & Open-weight & \cellcolor{cineNarrative!7!white}2.50 & \cellcolor{cineNarrative!7!white}1.89 & \cellcolor{cineNarrative!7!white}2.41 & \cellcolor{cineNarrative!7!white}2.26 & \cellcolor{cineNarrative!7!white}68.3 & \cellcolor{cineCultural!7!white}16.5 & \cellcolor{cineCultural!7!white}44.0 & \cellcolor{cineCultural!7!white}2.21 & \cellcolor{cineCultural!7!white}33.1 & \cellcolor{cineCultural!7!white}0.92 \\
\bottomrule
\end{tabular}%
\end{adjustbox}
\end{table*}

\begin{table*}[htbp]
\centering
\small
\setlength{\tabcolsep}{2.5pt}
\renewcommand{\arraystretch}{1.25}
\caption{Cross-lingual results averaged over Arabic, Indonesian, Persian, Romanian, and Vietnamese subtitle inputs; English is excluded. Arrows indicate the preferred direction: $\uparrow$ is higher-is-better and $\downarrow$ is lower-is-better. Narrative metrics are reference-based scores on a 1--5 scale; genre, age-threshold, and country exact-match metrics are percentages, while MAE metrics report error. Within each metric column, shading is normalized across the displayed models in the preferred direction: muted blue denotes narrative metrics and muted bronze denotes cultural metrics. Shade intensity is therefore not comparable across columns. Best values are bold, and second-best distinct values are underlined at the displayed precision; ties share the same emphasis.}
\vspace{5pt}
\label{tab:crosslingual-headline-results}
\begin{adjustbox}{max width=\textwidth}
\begin{tabular}{llrrrrrrrrrr}
\toprule
Model & Family & \multicolumn{5}{c}{\textbf{Narrative Understanding and Generation}} & \multicolumn{5}{c}{\textbf{Cultural Prediction and Assessment}} \\
\cmidrule(lr){3-7}\cmidrule(lr){8-12}
 & & \shortstack{Plot\\$\uparrow$} & \shortstack{Synopsis\\$\uparrow$} & \shortstack{Key\\message $\uparrow$} & \shortstack{Overall\\$\uparrow$} & \shortstack{Genre\\micro-F1 $\uparrow$} & \shortstack{Age\\exact $\uparrow$} & \shortstack{Age\\$\pm$1 $\uparrow$} & \shortstack{Age\\MAE $\downarrow$} & \shortstack{Country\\EM $\uparrow$} & \shortstack{Country\\MAE $\downarrow$} \\
\midrule
\href{https://developers.openai.com/api/docs/models/gpt-5.6-sol}{GPT-5.6 Sol} & Closed frontier & \cellcolor{cineNarrative!42!white}\underline{4.08} & \cellcolor{cineNarrative!43!white}\textbf{3.39} & \cellcolor{cineNarrative!43!white}\textbf{3.08} & \cellcolor{cineNarrative!43!white}\textbf{3.52} & \cellcolor{cineNarrative!36!white}\underline{81.4} & \cellcolor{cineCultural!35!white}30.6 & \cellcolor{cineCultural!39!white}78.1 & \cellcolor{cineCultural!41!white}\underline{1.01} & \cellcolor{cineCultural!33!white}\underline{65.1} & \cellcolor{cineCultural!35!white}\underline{0.45} \\
\href{https://docs.cloud.google.com/gemini-enterprise-agent-platform/models/gemini/3-8-flash}{Gemini 3.8 Flash} & Closed frontier & \cellcolor{cineNarrative!43!white}\textbf{4.16} & \cellcolor{cineNarrative!42!white}\underline{3.32} & \cellcolor{cineNarrative!43!white}\textbf{3.08} & \cellcolor{cineNarrative!43!white}\textbf{3.52} & \cellcolor{cineNarrative!43!white}\textbf{85.0} & \cellcolor{cineCultural!43!white}\textbf{33.8} & \cellcolor{cineCultural!43!white}\textbf{81.1} & \cellcolor{cineCultural!43!white}\textbf{0.95} & \cellcolor{cineCultural!43!white}\textbf{77.7} & \cellcolor{cineCultural!43!white}\textbf{0.31} \\
\href{https://platform.claude.com/docs/en/models/haiku-4-5/overview}{Claude Haiku 4.5} & Closed frontier & \cellcolor{cineNarrative!30!white}3.42 & \cellcolor{cineNarrative!23!white}2.42 & \cellcolor{cineNarrative!33!white}2.82 & \cellcolor{cineNarrative!28!white}2.89 & \cellcolor{cineNarrative!17!white}71.3 & \cellcolor{cineCultural!26!white}27.0 & \cellcolor{cineCultural!32!white}71.7 & \cellcolor{cineCultural!34!white}1.19 & \cellcolor{cineCultural!23!white}53.6 & \cellcolor{cineCultural!27!white}0.58 \\
\midrule
\href{https://developers.openai.com/api/docs/models/gpt-5-nano}{GPT-5 Nano} & Closed baseline & \cellcolor{cineNarrative!21!white}2.92 & \cellcolor{cineNarrative!12!white}1.87 & \cellcolor{cineNarrative!25!white}2.62 & \cellcolor{cineNarrative!19!white}2.47 & \cellcolor{cineNarrative!18!white}72.0 & \cellcolor{cineCultural!12!white}20.9 & \cellcolor{cineCultural!18!white}60.8 & \cellcolor{cineCultural!19!white}1.56 & \cellcolor{cineCultural!15!white}43.9 & \cellcolor{cineCultural!17!white}0.77 \\
\href{https://docs.cloud.google.com/gemini-enterprise-agent-platform/models/gemini/3-5-flash-lite}{Gemini 3.5 Flash Lite} & Closed baseline & \cellcolor{cineNarrative!37!white}3.81 & \cellcolor{cineNarrative!33!white}2.89 & \cellcolor{cineNarrative!39!white}\underline{2.98} & \cellcolor{cineNarrative!36!white}3.23 & \cellcolor{cineNarrative!35!white}80.7 & \cellcolor{cineCultural!38!white}\underline{31.7} & \cellcolor{cineCultural!31!white}71.5 & \cellcolor{cineCultural!36!white}1.11 & \cellcolor{cineCultural!32!white}64.3 & \cellcolor{cineCultural!34!white}0.47 \\
\midrule
\href{https://openrouter.ai/deepseek/deepseek-v4-pro-0813}{DeepSeek V4 Pro 1.6T} & Open-weight & \cellcolor{cineNarrative!39!white}3.91 & \cellcolor{cineNarrative!33!white}2.90 & \cellcolor{cineNarrative!36!white}2.90 & \cellcolor{cineNarrative!36!white}\underline{3.24} & \cellcolor{cineNarrative!36!white}81.3 & \cellcolor{cineCultural!34!white}29.9 & \cellcolor{cineCultural!40!white}\underline{78.7} & \cellcolor{cineCultural!41!white}\underline{1.01} & \cellcolor{cineCultural!27!white}57.9 & \cellcolor{cineCultural!31!white}0.53 \\
\href{https://huggingface.co/Qwen/Qwen3-235B-A22B-Instruct-2507}{Qwen3 235B} & Open-weight & \cellcolor{cineNarrative!28!white}3.34 & \cellcolor{cineNarrative!20!white}2.27 & \cellcolor{cineNarrative!29!white}2.73 & \cellcolor{cineNarrative!26!white}2.78 & \cellcolor{cineNarrative!10!white}68.0 & \cellcolor{cineCultural!15!white}22.1 & \cellcolor{cineCultural!15!white}58.4 & \cellcolor{cineCultural!22!white}1.49 & \cellcolor{cineCultural!18!white}47.1 & \cellcolor{cineCultural!23!white}0.65 \\
\href{https://huggingface.co/mistralai/Mistral-Small-4-119B-2603}{Mistral 4 119B} & Open-weight & \cellcolor{cineNarrative!18!white}2.77 & \cellcolor{cineNarrative!14!white}1.97 & \cellcolor{cineNarrative!21!white}2.52 & \cellcolor{cineNarrative!17!white}2.42 & \cellcolor{cineNarrative!7!white}66.2 & \cellcolor{cineCultural!9!white}19.8 & \cellcolor{cineCultural!10!white}54.2 & \cellcolor{cineCultural!11!white}1.78 & \cellcolor{cineCultural!17!white}47.0 & \cellcolor{cineCultural!21!white}0.69 \\
\href{https://huggingface.co/meta-llama/Llama-4-Scout-17B-16E-Instruct}{Llama 4 Scout 17B} & Open-weight & \cellcolor{cineNarrative!7!white}2.15 & \cellcolor{cineNarrative!7!white}1.63 & \cellcolor{cineNarrative!7!white}2.16 & \cellcolor{cineNarrative!7!white}1.98 & \cellcolor{cineNarrative!8!white}66.6 & \cellcolor{cineCultural!7!white}19.0 & \cellcolor{cineCultural!7!white}51.5 & \cellcolor{cineCultural!7!white}1.87 & \cellcolor{cineCultural!7!white}34.3 & \cellcolor{cineCultural!7!white}0.93 \\
\bottomrule
\end{tabular}%
\end{adjustbox}
\end{table*}

\begin{table}[htbp]
\centering
\small
\caption{Narrative generation scores averaged across all six subtitle languages. All models are evaluated with the same reference-based 1--5 rubric; higher is better ($\uparrow$). Within each metric column, blue shading is normalized across the displayed models. Best values are bold, and second-best distinct values are underlined at the displayed precision; ties share the same emphasis.}
\vspace{5pt}
\label{tab:narrative-scores-appendix}
\begin{adjustbox}{max width=\textwidth}
\begin{tabular}{lrrrr}
\toprule
Model & Plot $\uparrow$ & Synopsis $\uparrow$ & Key message $\uparrow$ & Overall $\uparrow$ \\
\midrule
\href{https://developers.openai.com/api/docs/models/gpt-5.6-sol}{GPT-5.6 Sol} & \cellcolor{cineNarrative!42!white}\underline{4.10} & \cellcolor{cineNarrative!43!white}\textbf{3.40} & \cellcolor{cineNarrative!42!white}\underline{3.08} & \cellcolor{cineNarrative!43!white}\underline{3.52} \\
\href{https://docs.cloud.google.com/gemini-enterprise-agent-platform/models/gemini/3-8-flash}{Gemini 3.8 Flash} & \cellcolor{cineNarrative!43!white}\textbf{4.16} & \cellcolor{cineNarrative!42!white}\underline{3.34} & \cellcolor{cineNarrative!43!white}\textbf{3.11} & \cellcolor{cineNarrative!43!white}\textbf{3.54} \\
\href{https://platform.claude.com/docs/en/models/haiku-4-5/overview}{Claude Haiku 4.5} & \cellcolor{cineNarrative!30!white}3.45 & \cellcolor{cineNarrative!23!white}2.44 & \cellcolor{cineNarrative!32!white}2.84 & \cellcolor{cineNarrative!28!white}2.91 \\
\href{https://developers.openai.com/api/docs/models/gpt-5-nano}{GPT-5 Nano} & \cellcolor{cineNarrative!21!white}2.97 & \cellcolor{cineNarrative!12!white}1.90 & \cellcolor{cineNarrative!24!white}2.64 & \cellcolor{cineNarrative!18!white}2.50 \\
\href{https://docs.cloud.google.com/gemini-enterprise-agent-platform/models/gemini/3-5-flash-lite}{Gemini 3.5 Flash Lite} & \cellcolor{cineNarrative!37!white}3.83 & \cellcolor{cineNarrative!33!white}2.91 & \cellcolor{cineNarrative!38!white}2.99 & \cellcolor{cineNarrative!36!white}3.25 \\
\href{https://openrouter.ai/deepseek/deepseek-v4-pro-0813}{DeepSeek V4 Pro 1.6T} & \cellcolor{cineNarrative!38!white}3.91 & \cellcolor{cineNarrative!33!white}2.93 & \cellcolor{cineNarrative!35!white}2.90 & \cellcolor{cineNarrative!36!white}3.25 \\
\href{https://huggingface.co/Qwen/Qwen3-235B-A22B-Instruct-2507}{Qwen3 235B} & \cellcolor{cineNarrative!28!white}3.36 & \cellcolor{cineNarrative!20!white}2.30 & \cellcolor{cineNarrative!29!white}2.76 & \cellcolor{cineNarrative!26!white}2.80 \\
\href{https://huggingface.co/mistralai/Mistral-Small-4-119B-2603}{Mistral 4 119B} & \cellcolor{cineNarrative!18!white}2.80 & \cellcolor{cineNarrative!14!white}2.00 & \cellcolor{cineNarrative!20!white}2.52 & \cellcolor{cineNarrative!17!white}2.44 \\
\href{https://huggingface.co/meta-llama/Llama-4-Scout-17B-16E-Instruct}{Llama 4 Scout 17B} & \cellcolor{cineNarrative!7!white}2.21 & \cellcolor{cineNarrative!7!white}1.67 & \cellcolor{cineNarrative!7!white}2.20 & \cellcolor{cineNarrative!7!white}2.03 \\
\bottomrule
\end{tabular}%
\end{adjustbox}
\end{table}

\subsection{Paired Comparisons of Leading Composite Scores}
\label{app:composite-significance}

We assess uncertainty in the composite-score gaps among the three leading
models. Gemini 3.8 Flash, the highest-scoring model, is compared with
GPT-5.6 Sol and with DeepSeek V4 Pro 1.6T, the highest-scoring open-weight
model. For each comparison, films are resampled as paired clusters while
retaining all subtitle-language conditions and the English-only
language-content component. The composite score is then recomputed from its
four components. We report percentile 95\% confidence intervals from 10,000
paired film-level bootstrap resamples.

We additionally conduct two-sided paired randomization tests by swapping the
two models' complete outputs within each film cluster and recomputing the
composite difference over 9,999 random swaps. Holm correction is applied
across the two planned comparisons.

Gemini's composite score exceeds both comparators. Using comparator minus
Gemini as the paired difference, GPT-5.6 Sol differs by $-3.34$ points
(95\% CI $[-3.86,-2.84]$), while DeepSeek V4 Pro 1.6T differs by
$-12.52$ points (95\% CI $[-13.61,-11.44]$). Both intervals lie below zero,
and both Holm-adjusted randomization-test $p$-values are $<0.001$
(Table~\ref{tab:composite-top3-significance}). These estimates characterize
film-level sampling uncertainty in the present evaluation and do not include
run-to-run generation variability.

\begin{table}[t]
\centering
\small
\caption{Paired film-level comparisons of the leading composite scores.
Differences are computed as comparator minus Gemini 3.8 Flash, so negative values indicate a higher Gemini score.}
\vspace{3pt}
\label{tab:composite-top3-significance}

\begin{tabular}{lrrr}
\toprule
Model & Composite score & Difference from Gemini (95\% CI) & Holm-adjusted $p$ \\
\midrule
Gemini 3.8 Flash       & 78.04 & Reference                         & --- \\
GPT-5.6 Sol            & 74.70 & $-3.34\;[-3.86,-2.84]$           & $<0.001$ \\
DeepSeek V4 Pro 1.6T   & 65.52 & $-12.52\;[-13.61,-11.44]$        & $<0.001$ \\
\bottomrule
\end{tabular}

\vspace{2pt}
\begin{minipage}{0.98\linewidth}\footnotesize
\emph{Note.} Confidence intervals are percentile intervals from 10,000 paired
film-level bootstrap resamples. Two-sided paired randomization tests use 9,999
random within-film swaps of the compared models' complete outputs, with
$p$-values Holm-adjusted across the two planned comparisons. The composite is
recomputed from its four components in every resample.
\end{minipage}
\end{table}

\subsection{Narrative Failure Patterns}
\label{app:narrative-error-analysis}
The structured evaluator assigns task-specific error tags before scoring every judged narrative output. We aggregate those tags here to reveal recurring failure mechanisms; they are evaluator diagnostics rather than independent human annotations.

\paragraph{Synopsis failures combine omission and invention.}
The evaluator marks a missing major event in 79.7\% and an invented event in 71.1\% of judged synopses. Plot errors instead most often concern wrong emphasis (54.7\%), whereas key-message errors most often concern semantic mismatch (80.3\%). These are output-level diagnostic incidences, not independent human labels.
\begin{table}[htbp]
\centering
\small
\setlength{\tabcolsep}{5pt}
\caption{Narrative error patterns identified by the reference-based evaluator across the comparable model suite. \emph{Share of errors} is the proportion of all error annotations within a narrative task assigned to each error type, while \emph{outputs with error} is the percentage of judged outputs containing that error at least once. The final column reports the mean number of annotations of that type per output.}
\vspace{5pt}
\label{tab:narrative-error-types}
\begin{adjustbox}{max width=\textwidth}
\begin{tabular}{llrrr}
\toprule
Task & Error type & Share of errors (\%) & Outputs with error (\%) & Avg. annotations/output \\
\midrule
Plot & Wrong Emphasis & 27.0 & 54.7 & 0.55 \\
 & Wrong Setup & 18.5 & 37.5 & 0.38 \\
 & Missing Central Conflict & 18.2 & 36.9 & 0.37 \\
 & Invented Event & 13.4 & 27.0 & 0.27 \\
 & Too Vague & 10.3 & 20.9 & 0.21 \\
 & Missing Protagonist Or Group & 9.1 & 18.5 & 0.19 \\
 & Wrong Relationship & 3.3 & 6.7 & 0.07 \\
 & Style Or Length Issue & 0.1 & 0.3 & 0.00 \\
\midrule
Synopsis & Missing Major Event & 23.4 & 79.7 & 0.82 \\
 & Invented Event & 22.9 & 71.1 & 0.80 \\
 & Wrong Resolution & 13.0 & 44.6 & 0.46 \\
 & Wrong Relationship & 10.0 & 33.5 & 0.35 \\
 & Weak Causality & 9.1 & 31.8 & 0.32 \\
 & Missing Resolution & 8.0 & 27.9 & 0.28 \\
 & Missing Character Motivation & 7.1 & 24.6 & 0.25 \\
 & Wrong Event Order & 3.0 & 10.4 & 0.11 \\
 & Too Vague & 2.2 & 7.7 & 0.08 \\
 & Style Or Length Issue & 0.8 & 2.6 & 0.03 \\
 & Too Detailed & 0.5 & 1.6 & 0.02 \\
\midrule
Key Message & Semantic Mismatch & 44.1 & 80.3 & 0.80 \\
 & Too Generic & 31.9 & 58.2 & 0.58 \\
 & Unsupported Theme & 19.9 & 36.3 & 0.36 \\
 & Misses Positive Message & 3.4 & 6.3 & 0.06 \\
 & Contradicted By Reference & 0.3 & 0.5 & 0.00 \\
 & Style Or Length Issue & 0.2 & 0.4 & 0.00 \\
 & Too Plot Like & 0.2 & 0.4 & 0.00 \\
\bottomrule
\end{tabular}%
\end{adjustbox}
\end{table}

\begin{figure}[htbp]
\centering
\includegraphics[width=1.0\linewidth]{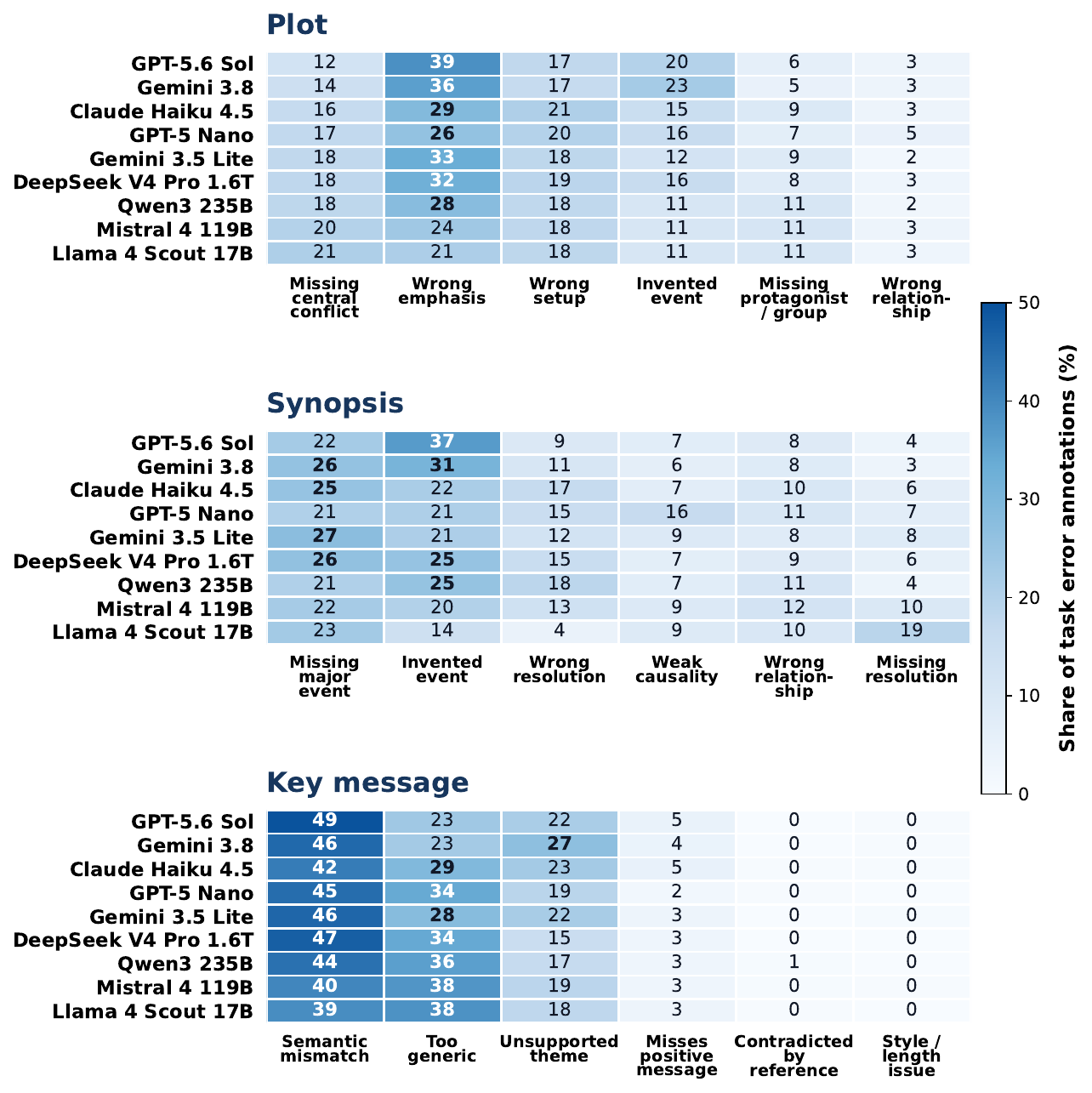}
\caption{Narrative error profiles by model and subtask. Each cell reports the share of that subtask's structured error annotations assigned to the indicated type. Synopsis errors concentrate in missing and invented events; key-message errors concentrate in semantic mismatch and over-generic themes.}
\label{fig:narrative-error-profile-by-model}
\end{figure}

\section{Additional Evidence for RQ~\ref{rq:crosslingual}: Cross-Lingual Behavior}
\label{app:rq2-details}

\subsection{English-to-Cross-Lingual Overview}
\label{app:rq2-overview}
Table~\ref{tab:english-crosslingual-gap} compares English input with the equal-weight average of the five non-English subtitle settings on the same films. Narrative overall decreases for all nine models, although the magnitude varies considerably. Other tasks are less uniform: some model-language combinations improve on genre, age suitability, or country ratings even when narrative performance declines. The aggregate view therefore motivates the language-specific analyses that follow.

\begin{table*}[htbp]
\centering
\small
\setlength{\tabcolsep}{2.7pt}
\renewcommand{\arraystretch}{1.20}
\caption{English-to-cross-lingual comparison across headline tasks. \ENtag{} denotes English, while \CLtag{} is the equal-weight mean over Arabic, Indonesian, Persian, Romanian, and Vietnamese; $\Delta=\mathrm{CL}-\mathrm{EN}$. Positive and negative markers indicate the direction of the observed change. Paired film-level uncertainty intervals are reported in Table~\ref{tab:paired-language-uncertainty-appendix}.}
\vspace{5pt}
\label{tab:english-crosslingual-gap}
\resizebox{\textwidth}{!}{%
\begin{tabular}{llrrrrrrrrrrrr}
\toprule
Model & Family & \multicolumn{6}{c}{\textbf{Narrative Understanding and Generation}} & \multicolumn{6}{c}{\textbf{Cultural Prediction and Assessment}} \\
\cmidrule(lr){3-8}\cmidrule(lr){9-14}
 & & \multicolumn{3}{c}{\shortstack{Overall\\(1--5)}} & \multicolumn{3}{c}{\shortstack{Genre\\micro-F1 (\%)}} & \multicolumn{3}{c}{\shortstack{Age\\exact (\%)}} & \multicolumn{3}{c}{\shortstack{Country\\EM (\%)}} \\
\cmidrule(lr){3-5}\cmidrule(lr){6-8}\cmidrule(lr){9-11}\cmidrule(lr){12-14}
 & & \ENtag & \CLtag & $\Delta$ & \ENtag & \CLtag & $\Delta$ & \ENtag & \CLtag & $\Delta$ & \ENtag & \CLtag & $\Delta$ \\
\midrule
\href{https://developers.openai.com/api/docs/models/gpt-5.6-sol}{GPT-5.6 Sol} & Closed frontier & \cellcolor{cineNarrative!42!white}3.56 & \cellcolor{cineNarrative!41!white}3.52 & \textcolor{cineDeltaNegative}{$\downarrow$\,-0.05} & \cellcolor{cineNarrative!38!white}82.2 & \cellcolor{cineNarrative!36!white}81.4 & \textcolor{cineDeltaNegative}{$\downarrow$\,-0.8} & \cellcolor{cineCultural!38!white}33.5 & \cellcolor{cineCultural!32!white}30.6 & \textcolor{cineDeltaNegative}{$\downarrow$\,-2.9} & \cellcolor{cineCultural!34!white}67.2 & \cellcolor{cineCultural!32!white}65.1 & \textcolor{cineDeltaNegative}{$\downarrow$\,-2.1} \\
\href{https://docs.cloud.google.com/gemini-enterprise-agent-platform/models/gemini/3-8-flash}{Gemini 3.8 Flash} & Closed frontier & \cellcolor{cineNarrative!43!white}3.61 & \cellcolor{cineNarrative!41!white}3.52 & \textcolor{cineDeltaNegative}{$\downarrow$\,-0.09} & \cellcolor{cineNarrative!42!white}84.7 & \cellcolor{cineNarrative!43!white}85.0 & \textcolor{cineDeltaPositive}{$\uparrow$\,+0.3} & \cellcolor{cineCultural!43!white}36.5 & \cellcolor{cineCultural!38!white}33.8 & \textcolor{cineDeltaNegative}{$\downarrow$\,-2.7} & \cellcolor{cineCultural!43!white}78.8 & \cellcolor{cineCultural!42!white}77.7 & \textcolor{cineDeltaNegative}{$\downarrow$\,-1.1} \\
\href{https://platform.claude.com/docs/en/models/haiku-4-5/overview}{Claude Haiku 4.5} & Closed frontier & \cellcolor{cineNarrative!30!white}3.02 & \cellcolor{cineNarrative!27!white}2.89 & \textcolor{cineDeltaNegative}{$\downarrow$\,-0.14} & \cellcolor{cineNarrative!19!white}72.2 & \cellcolor{cineNarrative!17!white}71.3 & \textcolor{cineDeltaNegative}{$\downarrow$\,-0.9} & \cellcolor{cineCultural!30!white}29.5 & \cellcolor{cineCultural!26!white}27.0 & \textcolor{cineDeltaNegative}{$\downarrow$\,-2.5} & \cellcolor{cineCultural!27!white}58.0 & \cellcolor{cineCultural!23!white}53.6 & \textcolor{cineDeltaNegative}{$\downarrow$\,-4.4} \\
\midrule
\href{https://developers.openai.com/api/docs/models/gpt-5-nano}{GPT-5 Nano} & Closed baseline & \cellcolor{cineNarrative!22!white}2.66 & \cellcolor{cineNarrative!18!white}2.47 & \textcolor{cineDeltaNegative}{$\downarrow$\,-0.19} & \cellcolor{cineNarrative!23!white}74.8 & \cellcolor{cineNarrative!18!white}72.0 & \textcolor{cineDeltaNegative}{$\downarrow$\,-2.8} & \cellcolor{cineCultural!12!white}19.5 & \cellcolor{cineCultural!15!white}20.9 & \textcolor{cineDeltaPositive}{$\uparrow$\,+1.4} & \cellcolor{cineCultural!15!white}43.8 & \cellcolor{cineCultural!15!white}43.9 & \textcolor{cineDeltaPositive}{$\uparrow$\,+0.1} \\
\href{https://docs.cloud.google.com/gemini-enterprise-agent-platform/models/gemini/3-5-flash-lite}{Gemini 3.5 Flash Lite} & Closed baseline & \cellcolor{cineNarrative!37!white}3.35 & \cellcolor{cineNarrative!34!white}3.23 & \textcolor{cineDeltaNegative}{$\downarrow$\,-0.13} & \cellcolor{cineNarrative!37!white}81.8 & \cellcolor{cineNarrative!35!white}80.7 & \textcolor{cineDeltaNegative}{$\downarrow$\,-1.1} & \cellcolor{cineCultural!36!white}32.5 & \cellcolor{cineCultural!34!white}31.7 & \textcolor{cineDeltaNegative}{$\downarrow$\,-0.8} & \cellcolor{cineCultural!33!white}66.3 & \cellcolor{cineCultural!32!white}64.3 & \textcolor{cineDeltaNegative}{$\downarrow$\,-2.0} \\
\midrule
\href{https://openrouter.ai/deepseek/deepseek-v4-pro-0813}{DeepSeek V4 Pro 1.6T} & Open-weight/API & \cellcolor{cineNarrative!36!white}3.3 & \cellcolor{cineNarrative!35!white}3.24 & \textcolor{cineDeltaNegative}{$\downarrow$\,-0.06} & \cellcolor{cineNarrative!40!white}83.2 & \cellcolor{cineNarrative!36!white}81.3 & \textcolor{cineDeltaNegative}{$\downarrow$\,-1.9} & \cellcolor{cineCultural!39!white}34.5 & \cellcolor{cineCultural!31!white}29.9 & \textcolor{cineDeltaNegative}{$\downarrow$\,-4.6} & \cellcolor{cineCultural!27!white}58.5 & \cellcolor{cineCultural!27!white}57.9 & \textcolor{cineDeltaNegative}{$\downarrow$\,-0.6} \\
\href{https://huggingface.co/Qwen/Qwen3-235B-A22B-Instruct-2507}{Qwen3 235B} & Open-weight/API & \cellcolor{cineNarrative!28!white}2.92 & \cellcolor{cineNarrative!25!white}2.78 & \textcolor{cineDeltaNegative}{$\downarrow$\,-0.14} & \cellcolor{cineNarrative!14!white}69.9 & \cellcolor{cineNarrative!10!white}68.0 & \textcolor{cineDeltaNegative}{$\downarrow$\,-1.9} & \cellcolor{cineCultural!16!white}21.5 & \cellcolor{cineCultural!17!white}22.1 & \textcolor{cineDeltaPositive}{$\uparrow$\,+0.6} & \cellcolor{cineCultural!22!white}51.7 & \cellcolor{cineCultural!18!white}47.1 & \textcolor{cineDeltaNegative}{$\downarrow$\,-4.6} \\
\href{https://huggingface.co/mistralai/Mistral-Small-4-119B-2603}{Mistral 4 119B} & Open-weight/API & \cellcolor{cineNarrative!19!white}2.54 & \cellcolor{cineNarrative!17!white}2.42 & \textcolor{cineDeltaNegative}{$\downarrow$\,-0.12} & \cellcolor{cineNarrative!14!white}70.0 & \cellcolor{cineNarrative!7!white}66.2 & \textcolor{cineDeltaNegative}{$\downarrow$\,-3.8} & \cellcolor{cineCultural!9!white}17.5 & \cellcolor{cineCultural!13!white}19.8 & \textcolor{cineDeltaPositive}{$\uparrow$\,+2.3} & \cellcolor{cineCultural!20!white}49.4 & \cellcolor{cineCultural!18!white}47.0 & \textcolor{cineDeltaNegative}{$\downarrow$\,-2.4} \\
\href{https://huggingface.co/meta-llama/Llama-4-Scout-17B-16E-Instruct}{Llama 4 Scout 17B} & Open-weight/API & \cellcolor{cineNarrative!13!white}2.26 & \cellcolor{cineNarrative!7!white}1.98 & \textcolor{cineDeltaNegative}{$\downarrow$\,-0.28} & \cellcolor{cineNarrative!11!white}68.3 & \cellcolor{cineNarrative!8!white}66.6 & \textcolor{cineDeltaNegative}{$\downarrow$\,-1.7} & \cellcolor{cineCultural!7!white}16.5 & \cellcolor{cineCultural!12!white}19.0 & \textcolor{cineDeltaPositive}{$\uparrow$\,+2.5} & \cellcolor{cineCultural!7!white}33.1 & \cellcolor{cineCultural!8!white}34.3 & \textcolor{cineDeltaPositive}{$\uparrow$\,+1.2} \\
\bottomrule
\end{tabular}%
}
\end{table*}

\subsection{Narrative Scores by Subtitle Language}
\label{app:rq2-narrative}
\paragraph{Plot recovery remains more reliable than event-complete reconstruction.}
The tables separate plot, synopsis, key-message, and their overall score for every model and subtitle language. Across models, the English means are 3.55 for plot, 2.67 for synopsis, and 2.86 for key message; under Persian input they are 3.20, 2.40, and 2.66. The plot--synopsis separation therefore persists within languages rather than arising only from the English setting.

\begin{table*}[htbp]
\centering
\small
\renewcommand{\arraystretch}{1.13}
\caption{Plot-generation scores by model and subtitle-input language. Values use the same reference-based 1--5 rubric across models and languages; higher is better ($\uparrow$). EN, AR, ID, FA, RO, and VI denote English, Arabic, Indonesian, Persian, Romanian, and Vietnamese, respectively. Within each language column, muted blue shading is normalized across the displayed models in the preferred direction. Best values are bold, and second-best distinct values are underlined at the displayed precision.}
\vspace{5pt}
\label{tab:narrative-plot-luna-1to5-by-language-appendix}
\begin{adjustbox}{max width=\textwidth}
\begin{tabular}{lrrrrrr}
\toprule
Model & EN & AR & ID & FA & RO & VI \\
\midrule
\href{https://developers.openai.com/api/docs/models/gpt-5.6-sol}{GPT-5.6 Sol} & \cellcolor{cineNarrative!43!white}\textbf{4.17} & \cellcolor{cineNarrative!41!white}\underline{4.13} & \cellcolor{cineNarrative!42!white}\underline{4.10} & \cellcolor{cineNarrative!41!white}\underline{3.97} & \cellcolor{cineNarrative!41!white}\underline{4.05} & \cellcolor{cineNarrative!43!white}\textbf{4.17} \\
\href{https://docs.cloud.google.com/gemini-enterprise-agent-platform/models/gemini/3-8-flash}{Gemini 3.8 Flash} & \cellcolor{cineNarrative!43!white}\textbf{4.17} & \cellcolor{cineNarrative!43!white}\textbf{4.23} & \cellcolor{cineNarrative!43!white}\textbf{4.13} & \cellcolor{cineNarrative!43!white}\textbf{4.08} & \cellcolor{cineNarrative!43!white}\textbf{4.17} & \cellcolor{cineNarrative!43!white}\textbf{4.17} \\
\href{https://platform.claude.com/docs/en/models/haiku-4-5/overview}{Claude Haiku 4.5} & \cellcolor{cineNarrative!31!white}3.62 & \cellcolor{cineNarrative!29!white}3.38 & \cellcolor{cineNarrative!31!white}3.53 & \cellcolor{cineNarrative!28!white}3.13 & \cellcolor{cineNarrative!29!white}3.48 & \cellcolor{cineNarrative!32!white}3.58 \\
\midrule
\href{https://developers.openai.com/api/docs/models/gpt-5-nano}{GPT-5 Nano} & \cellcolor{cineNarrative!22!white}3.21 & \cellcolor{cineNarrative!22!white}2.93 & \cellcolor{cineNarrative!22!white}3.03 & \cellcolor{cineNarrative!20!white}2.64 & \cellcolor{cineNarrative!20!white}2.98 & \cellcolor{cineNarrative!21!white}3.02 \\
\href{https://docs.cloud.google.com/gemini-enterprise-agent-platform/models/gemini/3-5-flash-lite}{Gemini 3.5 Flash Lite} & \cellcolor{cineNarrative!39!white}\underline{3.97} & \cellcolor{cineNarrative!36!white}3.83 & \cellcolor{cineNarrative!38!white}3.88 & \cellcolor{cineNarrative!37!white}3.69 & \cellcolor{cineNarrative!35!white}3.79 & \cellcolor{cineNarrative!37!white}3.85 \\
\midrule
\href{https://openrouter.ai/deepseek/deepseek-v4-pro-0813}{DeepSeek V4 Pro 1.6T} & \cellcolor{cineNarrative!37!white}3.91 & \cellcolor{cineNarrative!39!white}3.96 & \cellcolor{cineNarrative!39!white}3.94 & \cellcolor{cineNarrative!37!white}3.75 & \cellcolor{cineNarrative!37!white}3.88 & \cellcolor{cineNarrative!40!white}\underline{4.03} \\
\href{https://huggingface.co/Qwen/Qwen3-235B-A22B-Instruct-2507}{Qwen3 235B} & \cellcolor{cineNarrative!27!white}3.43 & \cellcolor{cineNarrative!29!white}3.38 & \cellcolor{cineNarrative!29!white}3.39 & \cellcolor{cineNarrative!27!white}3.12 & \cellcolor{cineNarrative!28!white}3.38 & \cellcolor{cineNarrative!29!white}3.44 \\
\href{https://huggingface.co/mistralai/Mistral-Small-4-119B-2603}{Mistral 4 119B} & \cellcolor{cineNarrative!17!white}2.98 & \cellcolor{cineNarrative!17!white}2.65 & \cellcolor{cineNarrative!17!white}2.81 & \cellcolor{cineNarrative!18!white}2.56 & \cellcolor{cineNarrative!19!white}2.92 & \cellcolor{cineNarrative!19!white}2.90 \\
\href{https://huggingface.co/meta-llama/Llama-4-Scout-17B-16E-Instruct}{Llama 4 Scout 17B} & \cellcolor{cineNarrative!7!white}2.50 & \cellcolor{cineNarrative!7!white}2.02 & \cellcolor{cineNarrative!7!white}2.27 & \cellcolor{cineNarrative!7!white}1.86 & \cellcolor{cineNarrative!7!white}2.33 & \cellcolor{cineNarrative!7!white}2.29 \\
\bottomrule
\end{tabular}%
\end{adjustbox}
\end{table*}

\begin{table*}[htbp]
\centering
\small
\renewcommand{\arraystretch}{1.13}
\caption{Synopsis-generation scores by model and subtitle-input language. Values use the same reference-based 1--5 rubric across models and languages; higher is better ($\uparrow$). EN, AR, ID, FA, RO, and VI denote English, Arabic, Indonesian, Persian, Romanian, and Vietnamese, respectively. Within each language column, muted blue shading is normalized across the displayed models in the preferred direction. Best values are bold, and second-best distinct values are underlined at the displayed precision.}
\label{tab:narrative-synopsis-luna-1to5-by-language-appendix}
\begin{adjustbox}{max width=\textwidth}
\begin{tabular}{lrrrrrr}
\toprule
Model & EN & AR & ID & FA & RO & VI \\
\midrule
\href{https://developers.openai.com/api/docs/models/gpt-5.6-sol}{GPT-5.6 Sol} & \cellcolor{cineNarrative!43!white}\underline{3.42} & \cellcolor{cineNarrative!43!white}\textbf{3.39} & \cellcolor{cineNarrative!43!white}\textbf{3.45} & \cellcolor{cineNarrative!43!white}\textbf{3.28} & \cellcolor{cineNarrative!43!white}\textbf{3.40} & \cellcolor{cineNarrative!43!white}\textbf{3.43} \\
\href{https://docs.cloud.google.com/gemini-enterprise-agent-platform/models/gemini/3-8-flash}{Gemini 3.8 Flash} & \cellcolor{cineNarrative!43!white}\textbf{3.43} & \cellcolor{cineNarrative!42!white}\underline{3.35} & \cellcolor{cineNarrative!41!white}\underline{3.35} & \cellcolor{cineNarrative!42!white}\underline{3.23} & \cellcolor{cineNarrative!41!white}\underline{3.31} & \cellcolor{cineNarrative!42!white}\underline{3.36} \\
\href{https://platform.claude.com/docs/en/models/haiku-4-5/overview}{Claude Haiku 4.5} & \cellcolor{cineNarrative!22!white}2.54 & \cellcolor{cineNarrative!24!white}2.41 & \cellcolor{cineNarrative!24!white}2.50 & \cellcolor{cineNarrative!23!white}2.28 & \cellcolor{cineNarrative!24!white}2.50 & \cellcolor{cineNarrative!22!white}2.44 \\
\midrule
\href{https://developers.openai.com/api/docs/models/gpt-5-nano}{GPT-5 Nano} & \cellcolor{cineNarrative!11!white}2.06 & \cellcolor{cineNarrative!12!white}1.81 & \cellcolor{cineNarrative!12!white}1.92 & \cellcolor{cineNarrative!12!white}1.75 & \cellcolor{cineNarrative!12!white}1.94 & \cellcolor{cineNarrative!12!white}1.95 \\
\href{https://docs.cloud.google.com/gemini-enterprise-agent-platform/models/gemini/3-5-flash-lite}{Gemini 3.5 Flash Lite} & \cellcolor{cineNarrative!34!white}3.02 & \cellcolor{cineNarrative!33!white}2.89 & \cellcolor{cineNarrative!32!white}2.90 & \cellcolor{cineNarrative!34!white}2.83 & \cellcolor{cineNarrative!32!white}2.90 & \cellcolor{cineNarrative!33!white}2.94 \\
\midrule
\href{https://openrouter.ai/deepseek/deepseek-v4-pro-0813}{DeepSeek V4 Pro 1.6T} & \cellcolor{cineNarrative!35!white}3.08 & \cellcolor{cineNarrative!35!white}3.00 & \cellcolor{cineNarrative!32!white}2.88 & \cellcolor{cineNarrative!35!white}2.87 & \cellcolor{cineNarrative!33!white}2.91 & \cellcolor{cineNarrative!31!white}2.86 \\
\href{https://huggingface.co/Qwen/Qwen3-235B-A22B-Instruct-2507}{Qwen3 235B} & \cellcolor{cineNarrative!20!white}2.46 & \cellcolor{cineNarrative!21!white}2.27 & \cellcolor{cineNarrative!21!white}2.33 & \cellcolor{cineNarrative!19!white}2.10 & \cellcolor{cineNarrative!19!white}2.29 & \cellcolor{cineNarrative!21!white}2.37 \\
\href{https://huggingface.co/mistralai/Mistral-Small-4-119B-2603}{Mistral 4 119B} & \cellcolor{cineNarrative!12!white}2.12 & \cellcolor{cineNarrative!13!white}1.90 & \cellcolor{cineNarrative!15!white}2.04 & \cellcolor{cineNarrative!14!white}1.83 & \cellcolor{cineNarrative!14!white}2.05 & \cellcolor{cineNarrative!14!white}2.04 \\
\href{https://huggingface.co/meta-llama/Llama-4-Scout-17B-16E-Instruct}{Llama 4 Scout 17B} & \cellcolor{cineNarrative!7!white}1.89 & \cellcolor{cineNarrative!7!white}1.57 & \cellcolor{cineNarrative!7!white}1.64 & \cellcolor{cineNarrative!7!white}1.49 & \cellcolor{cineNarrative!7!white}1.72 & \cellcolor{cineNarrative!7!white}1.72 \\
\bottomrule
\end{tabular}%
\end{adjustbox}
\end{table*}

\begin{table*}[htbp]
\centering
\small
\renewcommand{\arraystretch}{1.13}
\caption{Key-message generation scores by model and subtitle-input language. Values use the same reference-based 1--5 rubric across models and languages; higher is better ($\uparrow$). EN, AR, ID, FA, RO, and VI denote English, Arabic, Indonesian, Persian, Romanian, and Vietnamese, respectively. Within each language column, muted blue shading is normalized across the displayed models in the preferred direction. Best values are bold, and second-best distinct values are underlined at the displayed precision.}
\label{tab:narrative-key-message-luna-1to5-by-language-appendix}
\begin{adjustbox}{max width=\textwidth}
\begin{tabular}{lrrrrrr}
\toprule
Model & EN & AR & ID & FA & RO & VI \\
\midrule
\href{https://developers.openai.com/api/docs/models/gpt-5.6-sol}{GPT-5.6 Sol} & \cellcolor{cineNarrative!37!white}\underline{3.10} & \cellcolor{cineNarrative!43!white}\textbf{3.13} & \cellcolor{cineNarrative!43!white}\textbf{3.11} & \cellcolor{cineNarrative!42!white}\underline{2.99} & \cellcolor{cineNarrative!41!white}\underline{3.06} & \cellcolor{cineNarrative!41!white}\underline{3.08} \\
\href{https://docs.cloud.google.com/gemini-enterprise-agent-platform/models/gemini/3-8-flash}{Gemini 3.8 Flash} & \cellcolor{cineNarrative!43!white}\textbf{3.23} & \cellcolor{cineNarrative!41!white}\underline{3.06} & \cellcolor{cineNarrative!43!white}\textbf{3.11} & \cellcolor{cineNarrative!43!white}\textbf{3.02} & \cellcolor{cineNarrative!43!white}\textbf{3.10} & \cellcolor{cineNarrative!43!white}\textbf{3.12} \\
\href{https://platform.claude.com/docs/en/models/haiku-4-5/overview}{Claude Haiku 4.5} & \cellcolor{cineNarrative!29!white}2.92 & \cellcolor{cineNarrative!33!white}2.85 & \cellcolor{cineNarrative!33!white}2.87 & \cellcolor{cineNarrative!31!white}2.67 & \cellcolor{cineNarrative!33!white}2.87 & \cellcolor{cineNarrative!32!white}2.85 \\
\midrule
\href{https://developers.openai.com/api/docs/models/gpt-5-nano}{GPT-5 Nano} & \cellcolor{cineNarrative!20!white}2.71 & \cellcolor{cineNarrative!25!white}2.60 & \cellcolor{cineNarrative!24!white}2.65 & \cellcolor{cineNarrative!23!white}2.45 & \cellcolor{cineNarrative!25!white}2.71 & \cellcolor{cineNarrative!26!white}2.70 \\
\href{https://docs.cloud.google.com/gemini-enterprise-agent-platform/models/gemini/3-5-flash-lite}{Gemini 3.5 Flash Lite} & \cellcolor{cineNarrative!36!white}3.06 & \cellcolor{cineNarrative!40!white}3.05 & \cellcolor{cineNarrative!37!white}\underline{2.96} & \cellcolor{cineNarrative!39!white}2.90 & \cellcolor{cineNarrative!37!white}2.97 & \cellcolor{cineNarrative!39!white}3.02 \\
\midrule
\href{https://openrouter.ai/deepseek/deepseek-v4-pro-0813}{DeepSeek V4 Pro 1.6T} & \cellcolor{cineNarrative!28!white}2.90 & \cellcolor{cineNarrative!36!white}2.92 & \cellcolor{cineNarrative!35!white}2.92 & \cellcolor{cineNarrative!37!white}2.84 & \cellcolor{cineNarrative!33!white}2.88 & \cellcolor{cineNarrative!36!white}2.95 \\
\href{https://huggingface.co/Qwen/Qwen3-235B-A22B-Instruct-2507}{Qwen3 235B} & \cellcolor{cineNarrative!28!white}2.88 & \cellcolor{cineNarrative!30!white}2.73 & \cellcolor{cineNarrative!29!white}2.76 & \cellcolor{cineNarrative!28!white}2.62 & \cellcolor{cineNarrative!27!white}2.75 & \cellcolor{cineNarrative!30!white}2.79 \\
\href{https://huggingface.co/mistralai/Mistral-Small-4-119B-2603}{Mistral 4 119B} & \cellcolor{cineNarrative!12!white}2.52 & \cellcolor{cineNarrative!19!white}2.42 & \cellcolor{cineNarrative!22!white}2.60 & \cellcolor{cineNarrative!21!white}2.40 & \cellcolor{cineNarrative!20!white}2.58 & \cellcolor{cineNarrative!23!white}2.62 \\
\href{https://huggingface.co/meta-llama/Llama-4-Scout-17B-16E-Instruct}{Llama 4 Scout 17B} & \cellcolor{cineNarrative!7!white}2.41 & \cellcolor{cineNarrative!7!white}2.04 & \cellcolor{cineNarrative!7!white}2.23 & \cellcolor{cineNarrative!7!white}2.02 & \cellcolor{cineNarrative!7!white}2.29 & \cellcolor{cineNarrative!7!white}2.21 \\
\bottomrule
\end{tabular}%
\end{adjustbox}
\end{table*}

\begin{table*}[htbp]
\centering
\small
\renewcommand{\arraystretch}{1.13}
\caption{Narrative-overall scores by model and subtitle-input language. Values use the same reference-based 1--5 rubric across models and languages; higher is better ($\uparrow$). EN, AR, ID, FA, RO, and VI denote English, Arabic, Indonesian, Persian, Romanian, and Vietnamese, respectively. Within each language column, muted blue shading is normalized across the displayed models in the preferred direction. Best values are bold, and second-best distinct values are underlined at the displayed precision.}
\label{tab:narrative-overall-luna-1to5-by-language-appendix}
\begin{adjustbox}{max width=\textwidth}
\begin{tabular}{lrrrrrr}
\toprule
Model & EN & AR & ID & FA & RO & VI \\
\midrule
\href{https://developers.openai.com/api/docs/models/gpt-5.6-sol}{GPT-5.6 Sol} & \cellcolor{cineNarrative!42!white}\underline{3.56} & \cellcolor{cineNarrative!43!white}\textbf{3.55} & \cellcolor{cineNarrative!43!white}\textbf{3.55} & \cellcolor{cineNarrative!42!white}\underline{3.41} & \cellcolor{cineNarrative!42!white}\underline{3.50} & \cellcolor{cineNarrative!43!white}\textbf{3.56} \\
\href{https://docs.cloud.google.com/gemini-enterprise-agent-platform/models/gemini/3-8-flash}{Gemini 3.8 Flash} & \cellcolor{cineNarrative!43!white}\textbf{3.61} & \cellcolor{cineNarrative!43!white}\textbf{3.55} & \cellcolor{cineNarrative!42!white}\underline{3.53} & \cellcolor{cineNarrative!43!white}\textbf{3.44} & \cellcolor{cineNarrative!43!white}\textbf{3.53} & \cellcolor{cineNarrative!43!white}\underline{3.55} \\
\href{https://platform.claude.com/docs/en/models/haiku-4-5/overview}{Claude Haiku 4.5} & \cellcolor{cineNarrative!27!white}3.02 & \cellcolor{cineNarrative!29!white}2.88 & \cellcolor{cineNarrative!29!white}2.96 & \cellcolor{cineNarrative!27!white}2.69 & \cellcolor{cineNarrative!28!white}2.95 & \cellcolor{cineNarrative!29!white}2.96 \\
\midrule
\href{https://developers.openai.com/api/docs/models/gpt-5-nano}{GPT-5 Nano} & \cellcolor{cineNarrative!18!white}2.66 & \cellcolor{cineNarrative!19!white}2.45 & \cellcolor{cineNarrative!19!white}2.53 & \cellcolor{cineNarrative!18!white}2.28 & \cellcolor{cineNarrative!18!white}2.54 & \cellcolor{cineNarrative!19!white}2.56 \\
\href{https://docs.cloud.google.com/gemini-enterprise-agent-platform/models/gemini/3-5-flash-lite}{Gemini 3.5 Flash Lite} & \cellcolor{cineNarrative!36!white}3.35 & \cellcolor{cineNarrative!37!white}3.26 & \cellcolor{cineNarrative!36!white}3.25 & \cellcolor{cineNarrative!36!white}3.14 & \cellcolor{cineNarrative!35!white}3.22 & \cellcolor{cineNarrative!36!white}3.27 \\
\midrule
\href{https://openrouter.ai/deepseek/deepseek-v4-pro-0813}{DeepSeek V4 Pro 1.6T} & \cellcolor{cineNarrative!35!white}3.30 & \cellcolor{cineNarrative!37!white}\underline{3.29} & \cellcolor{cineNarrative!36!white}3.25 & \cellcolor{cineNarrative!37!white}3.15 & \cellcolor{cineNarrative!35!white}3.22 & \cellcolor{cineNarrative!36!white}3.28 \\
\href{https://huggingface.co/Qwen/Qwen3-235B-A22B-Instruct-2507}{Qwen3 235B} & \cellcolor{cineNarrative!25!white}2.92 & \cellcolor{cineNarrative!27!white}2.79 & \cellcolor{cineNarrative!26!white}2.83 & \cellcolor{cineNarrative!25!white}2.61 & \cellcolor{cineNarrative!25!white}2.81 & \cellcolor{cineNarrative!26!white}2.87 \\
\href{https://huggingface.co/mistralai/Mistral-Small-4-119B-2603}{Mistral 4 119B} & \cellcolor{cineNarrative!14!white}2.54 & \cellcolor{cineNarrative!17!white}2.32 & \cellcolor{cineNarrative!17!white}2.48 & \cellcolor{cineNarrative!17!white}2.26 & \cellcolor{cineNarrative!17!white}2.52 & \cellcolor{cineNarrative!18!white}2.52 \\
\href{https://huggingface.co/meta-llama/Llama-4-Scout-17B-16E-Instruct}{Llama 4 Scout 17B} & \cellcolor{cineNarrative!7!white}2.26 & \cellcolor{cineNarrative!7!white}1.88 & \cellcolor{cineNarrative!7!white}2.04 & \cellcolor{cineNarrative!7!white}1.79 & \cellcolor{cineNarrative!7!white}2.11 & \cellcolor{cineNarrative!7!white}2.07 \\
\bottomrule
\end{tabular}%
\end{adjustbox}
\end{table*}

\begin{figure}[htbp]
\centering
\includegraphics[width=1.0\linewidth]{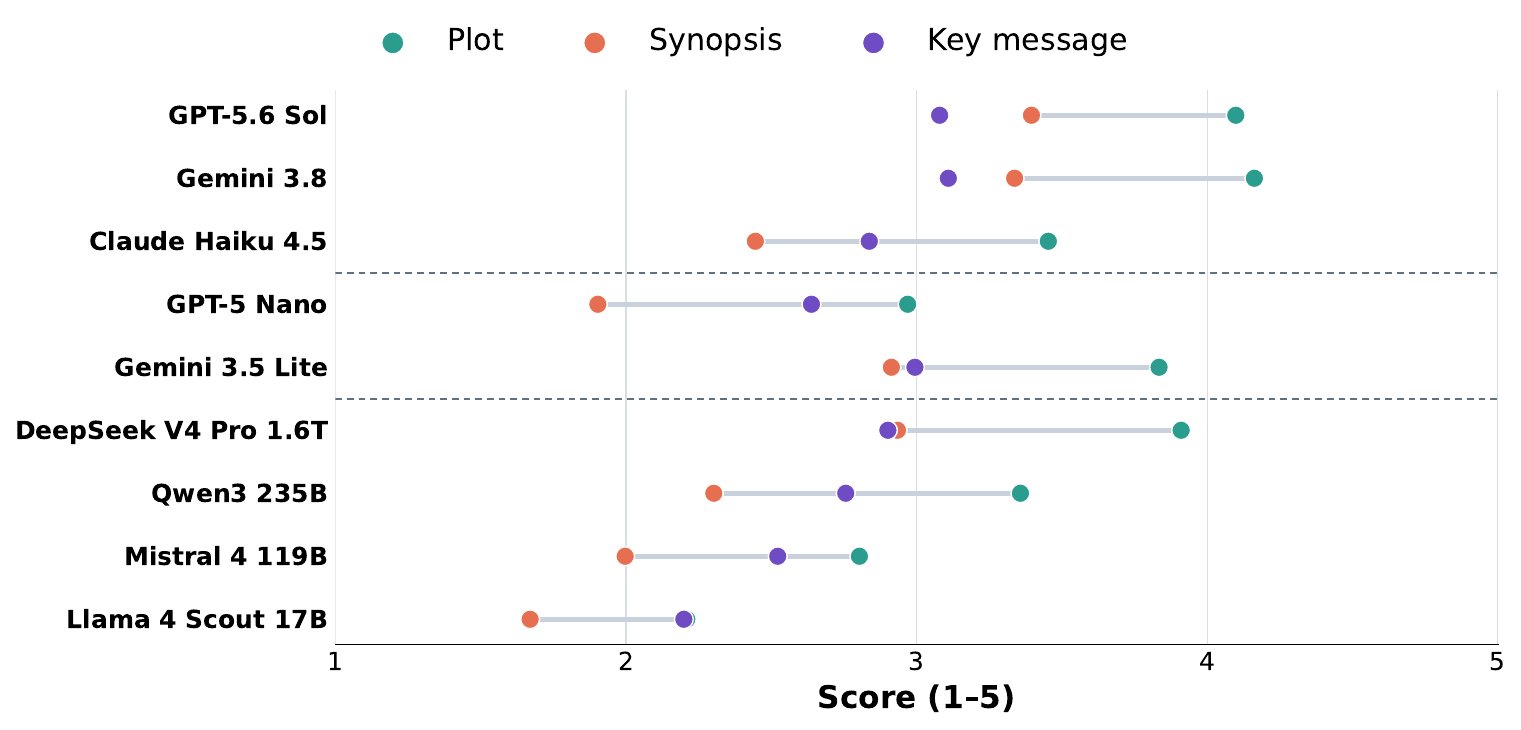}
\caption{Six-language-average scores for plot, synopsis, and key-message generation. Each row represents one model, and the connecting segment highlights the separation between premise-level plot recovery and fuller synopsis reconstruction.}
\label{fig:narrative-task-gradient}
\end{figure}

\subsection{Genre Stability and Label-Specific Failures}
\label{app:rq2-genre}

\paragraph{Cross-lingual genre changes are model dependent.}
From English to the five-language average, Mistral 4 119B loses 3.8 percentage points of genre micro-F1, while Gemini 3.8 Flash changes by only $+0.3$ points; Gemini's paired interval includes zero. These differences are small for some models and more pronounced for others, reinforcing the need for model-specific cross-lingual analysis.

\paragraph{High overlap does not imply exact label-set recovery.}
Gemini 3.8 reaches 84.9\% six-language micro-F1 but 40.8\% exact match; DeepSeek reaches 81.6\% and 34.0\%. A model may therefore recognize several relevant genres while adding or omitting another.

\paragraph{Agreement is strongest on several recognizable genres.}
Sol and Gemini differ by only 0.7 points on Comedy and 0.4 on Animation, and both reach 100.0\% on Sport and Documentary. Comedy is common (38.0\% of evaluated films), making its agreement more persuasive than Sport (2.0\%) or Documentary (0.5\%). Film-Noir is omitted because the evaluated set contains no positive examples.

\paragraph{Genre errors are label-selective rather than uniform.}
DeepSeek nearly matches Gemini on Science Fiction (85.5\% versus 85.9\%) and Animation (90.7\% versus 91.7\%), and exceeds Sol on Biography (86.6\% versus 82.5\%). By contrast, Qwen3 and Mistral 4 score 5.1\% and 6.8\% on Biography after missing 111 and 110 of 114 positive film--language instances with no false positives: a systematic omission hidden by aggregate F1.
\begin{table}[htbp]
\centering
\small
\caption{Genre-prediction metrics averaged across all six subtitle languages; higher is better ($\uparrow$) for every metric. Genre prediction is multi-label, and exact match requires the complete predicted label set to equal the reference set. Within each metric column, blue shading is normalized across the displayed models. Best values are bold, and second-best distinct values are underlined at the displayed precision; ties share the same emphasis.}
\vspace{5pt}
\label{tab:genre-results-appendix}
\begin{adjustbox}{max width=\textwidth}
\begin{tabular}{lrrrr}
\toprule
Model & Exact (\%) $\uparrow$ & Jaccard (\%) $\uparrow$ & Micro-F1 (\%) $\uparrow$ & Macro-F1 (\%) $\uparrow$ \\
\midrule
\href{https://developers.openai.com/api/docs/models/gpt-5.6-sol}{GPT-5.6 Sol} & \cellcolor{cineNarrative!29!white}28.8 & \cellcolor{cineNarrative!35!white}72.1 & \cellcolor{cineNarrative!36!white}81.5 & \cellcolor{cineNarrative!39!white}\underline{80.5} \\
\href{https://docs.cloud.google.com/gemini-enterprise-agent-platform/models/gemini/3-8-flash}{Gemini 3.8 Flash} & \cellcolor{cineNarrative!43!white}\textbf{40.8} & \cellcolor{cineNarrative!43!white}\textbf{77.7} & \cellcolor{cineNarrative!43!white}\textbf{84.9} & \cellcolor{cineNarrative!43!white}\textbf{83.3} \\
\href{https://platform.claude.com/docs/en/models/haiku-4-5/overview}{Claude Haiku 4.5} & \cellcolor{cineNarrative!15!white}16.9 & \cellcolor{cineNarrative!17!white}60.4 & \cellcolor{cineNarrative!16!white}71.5 & \cellcolor{cineNarrative!23!white}69.8 \\
\href{https://developers.openai.com/api/docs/models/gpt-5-nano}{GPT-5 Nano} & \cellcolor{cineNarrative!9!white}11.5 & \cellcolor{cineNarrative!16!white}59.5 & \cellcolor{cineNarrative!18!white}72.4 & \cellcolor{cineNarrative!24!white}70.4 \\
\href{https://docs.cloud.google.com/gemini-enterprise-agent-platform/models/gemini/3-5-flash-lite}{Gemini 3.5 Flash Lite} & \cellcolor{cineNarrative!31!white}30.7 & \cellcolor{cineNarrative!34!white}71.9 & \cellcolor{cineNarrative!35!white}80.9 & \cellcolor{cineNarrative!35!white}77.8 \\
\href{https://openrouter.ai/deepseek/deepseek-v4-pro-0813}{DeepSeek V4 Pro 1.6T} & \cellcolor{cineNarrative!35!white}\underline{34.0} & \cellcolor{cineNarrative!36!white}\underline{73.2} & \cellcolor{cineNarrative!36!white}\underline{81.6} & \cellcolor{cineNarrative!39!white}\underline{80.5} \\
\href{https://huggingface.co/Qwen/Qwen3-235B-A22B-Instruct-2507}{Qwen3 235B} & \cellcolor{cineNarrative!11!white}13.0 & \cellcolor{cineNarrative!11!white}55.9 & \cellcolor{cineNarrative!10!white}68.3 & \cellcolor{cineNarrative!7!white}59.0 \\
\href{https://huggingface.co/mistralai/Mistral-Small-4-119B-2603}{Mistral 4 119B} & \cellcolor{cineNarrative!7!white}9.7 & \cellcolor{cineNarrative!7!white}53.4 & \cellcolor{cineNarrative!7!white}66.8 & \cellcolor{cineNarrative!14!white}63.8 \\
\href{https://huggingface.co/meta-llama/Llama-4-Scout-17B-16E-Instruct}{Llama 4 Scout 17B} & \cellcolor{cineNarrative!9!white}11.2 & \cellcolor{cineNarrative!8!white}54.4 & \cellcolor{cineNarrative!7!white}66.9 & \cellcolor{cineNarrative!11!white}61.7 \\
\bottomrule
\end{tabular}%
\end{adjustbox}
\end{table}

\begin{table*}[htbp]
\centering
\small
\renewcommand{\arraystretch}{1.13}
\caption{Genre micro-F1 by model and subtitle-input language. Values are percentages; higher is better ($\uparrow$). EN, AR, ID, FA, RO, and VI denote English, Arabic, Indonesian, Persian, Romanian, and Vietnamese, respectively. Within each language column, muted blue shading is normalized across the displayed models in the preferred direction. Best values are bold, and second-best distinct values are underlined at the displayed precision.}
\vspace{5pt}
\label{tab:genre-genre-micro-f1-pct-by-language-appendix}
\begin{adjustbox}{max width=\textwidth}
\begin{tabular}{lrrrrrr}
\toprule
Model & EN & AR & ID & FA & RO & VI \\
\midrule
\href{https://developers.openai.com/api/docs/models/gpt-5.6-sol}{GPT-5.6 Sol} & \cellcolor{cineNarrative!38!white}82.2 & \cellcolor{cineNarrative!38!white}\underline{81.9} & \cellcolor{cineNarrative!35!white}81.7 & \cellcolor{cineNarrative!35!white}\underline{79.9} & \cellcolor{cineNarrative!37!white}\underline{81.4} & \cellcolor{cineNarrative!36!white}82.2 \\
\href{https://docs.cloud.google.com/gemini-enterprise-agent-platform/models/gemini/3-8-flash}{Gemini 3.8 Flash} & \cellcolor{cineNarrative!43!white}\textbf{84.7} & \cellcolor{cineNarrative!43!white}\textbf{85.0} & \cellcolor{cineNarrative!43!white}\textbf{85.5} & \cellcolor{cineNarrative!43!white}\textbf{84.4} & \cellcolor{cineNarrative!43!white}\textbf{84.3} & \cellcolor{cineNarrative!43!white}\textbf{85.7} \\
\href{https://platform.claude.com/docs/en/models/haiku-4-5/overview}{Claude Haiku 4.5} & \cellcolor{cineNarrative!16!white}72.2 & \cellcolor{cineNarrative!19!white}71.2 & \cellcolor{cineNarrative!19!white}73.6 & \cellcolor{cineNarrative!14!white}67.5 & \cellcolor{cineNarrative!17!white}72.0 & \cellcolor{cineNarrative!17!white}72.2 \\
\midrule
\href{https://developers.openai.com/api/docs/models/gpt-5-nano}{GPT-5 Nano} & \cellcolor{cineNarrative!21!white}74.8 & \cellcolor{cineNarrative!19!white}71.2 & \cellcolor{cineNarrative!18!white}73.2 & \cellcolor{cineNarrative!15!white}68.4 & \cellcolor{cineNarrative!19!white}72.7 & \cellcolor{cineNarrative!21!white}74.3 \\
\href{https://docs.cloud.google.com/gemini-enterprise-agent-platform/models/gemini/3-5-flash-lite}{Gemini 3.5 Flash Lite} & \cellcolor{cineNarrative!37!white}81.8 & \cellcolor{cineNarrative!34!white}79.7 & \cellcolor{cineNarrative!36!white}81.9 & \cellcolor{cineNarrative!34!white}79.5 & \cellcolor{cineNarrative!34!white}80.0 & \cellcolor{cineNarrative!37!white}82.6 \\
\midrule
\href{https://openrouter.ai/deepseek/deepseek-v4-pro-0813}{DeepSeek V4 Pro 1.6T} & \cellcolor{cineNarrative!40!white}\underline{83.2} & \cellcolor{cineNarrative!35!white}80.6 & \cellcolor{cineNarrative!38!white}\underline{82.8} & \cellcolor{cineNarrative!34!white}79.4 & \cellcolor{cineNarrative!36!white}81.1 & \cellcolor{cineNarrative!37!white}\underline{82.7} \\
\href{https://huggingface.co/Qwen/Qwen3-235B-A22B-Instruct-2507}{Qwen3 235B} & \cellcolor{cineNarrative!11!white}69.9 & \cellcolor{cineNarrative!12!white}66.8 & \cellcolor{cineNarrative!10!white}69.0 & \cellcolor{cineNarrative!10!white}65.4 & \cellcolor{cineNarrative!11!white}69.1 & \cellcolor{cineNarrative!12!white}69.5 \\
\href{https://huggingface.co/mistralai/Mistral-Small-4-119B-2603}{Mistral 4 119B} & \cellcolor{cineNarrative!11!white}70.0 & \cellcolor{cineNarrative!7!white}64.1 & \cellcolor{cineNarrative!7!white}67.9 & \cellcolor{cineNarrative!7!white}63.7 & \cellcolor{cineNarrative!10!white}68.3 & \cellcolor{cineNarrative!7!white}67.0 \\
\href{https://huggingface.co/meta-llama/Llama-4-Scout-17B-16E-Instruct}{Llama 4 Scout 17B} & \cellcolor{cineNarrative!7!white}68.3 & \cellcolor{cineNarrative!10!white}65.8 & \cellcolor{cineNarrative!7!white}67.7 & \cellcolor{cineNarrative!8!white}64.2 & \cellcolor{cineNarrative!7!white}67.1 & \cellcolor{cineNarrative!10!white}68.4 \\
\bottomrule
\end{tabular}%
\end{adjustbox}
\end{table*}

\begin{table*}[htbp]
\centering
\small
\renewcommand{\arraystretch}{1.13}
\caption{Genre macro-F1 by model and subtitle-input language. Values are percentages; higher is better ($\uparrow$). EN, AR, ID, FA, RO, and VI denote English, Arabic, Indonesian, Persian, Romanian, and Vietnamese, respectively. Within each language column, muted blue shading is normalized across the displayed models in the preferred direction. Best values are bold, and second-best distinct values are underlined at the displayed precision.}
\vspace{5pt}
\label{tab:genre-genre-macro-f1-pct-by-language-appendix}
\begin{adjustbox}{max width=\textwidth}
\begin{tabular}{lrrrrrr}
\toprule
Model & EN & AR & ID & FA & RO & VI \\
\midrule
\href{https://developers.openai.com/api/docs/models/gpt-5.6-sol}{GPT-5.6 Sol} & \cellcolor{cineNarrative!40!white}\underline{80.2} & \cellcolor{cineNarrative!39!white}\underline{80.9} & \cellcolor{cineNarrative!37!white}80.9 & \cellcolor{cineNarrative!38!white}\underline{79.4} & \cellcolor{cineNarrative!40!white}\underline{81.2} & \cellcolor{cineNarrative!37!white}80.5 \\
\href{https://docs.cloud.google.com/gemini-enterprise-agent-platform/models/gemini/3-8-flash}{Gemini 3.8 Flash} & \cellcolor{cineNarrative!39!white}80.0 & \cellcolor{cineNarrative!43!white}\textbf{84.2} & \cellcolor{cineNarrative!43!white}\textbf{84.8} & \cellcolor{cineNarrative!43!white}\textbf{83.2} & \cellcolor{cineNarrative!43!white}\textbf{83.5} & \cellcolor{cineNarrative!43!white}\textbf{84.1} \\
\href{https://platform.claude.com/docs/en/models/haiku-4-5/overview}{Claude Haiku 4.5} & \cellcolor{cineNarrative!25!white}71.3 & \cellcolor{cineNarrative!21!white}68.0 & \cellcolor{cineNarrative!26!white}72.7 & \cellcolor{cineNarrative!19!white}65.5 & \cellcolor{cineNarrative!24!white}70.5 & \cellcolor{cineNarrative!22!white}70.7 \\
\midrule
\href{https://developers.openai.com/api/docs/models/gpt-5-nano}{GPT-5 Nano} & \cellcolor{cineNarrative!28!white}73.3 & \cellcolor{cineNarrative!22!white}68.9 & \cellcolor{cineNarrative!23!white}70.6 & \cellcolor{cineNarrative!20!white}66.0 & \cellcolor{cineNarrative!25!white}71.1 & \cellcolor{cineNarrative!24!white}72.3 \\
\href{https://docs.cloud.google.com/gemini-enterprise-agent-platform/models/gemini/3-5-flash-lite}{Gemini 3.5 Flash Lite} & \cellcolor{cineNarrative!36!white}77.9 & \cellcolor{cineNarrative!33!white}77.0 & \cellcolor{cineNarrative!36!white}79.6 & \cellcolor{cineNarrative!33!white}75.7 & \cellcolor{cineNarrative!35!white}78.1 & \cellcolor{cineNarrative!35!white}78.8 \\
\midrule
\href{https://openrouter.ai/deepseek/deepseek-v4-pro-0813}{DeepSeek V4 Pro 1.6T} & \cellcolor{cineNarrative!43!white}\textbf{82.1} & \cellcolor{cineNarrative!37!white}79.7 & \cellcolor{cineNarrative!39!white}\underline{81.7} & \cellcolor{cineNarrative!36!white}77.9 & \cellcolor{cineNarrative!39!white}80.8 & \cellcolor{cineNarrative!37!white}\underline{80.6} \\
\href{https://huggingface.co/Qwen/Qwen3-235B-A22B-Instruct-2507}{Qwen3 235B} & \cellcolor{cineNarrative!7!white}60.9 & \cellcolor{cineNarrative!7!white}57.5 & \cellcolor{cineNarrative!7!white}59.5 & \cellcolor{cineNarrative!7!white}56.4 & \cellcolor{cineNarrative!7!white}58.8 & \cellcolor{cineNarrative!7!white}61.2 \\
\href{https://huggingface.co/mistralai/Mistral-Small-4-119B-2603}{Mistral 4 119B} & \cellcolor{cineNarrative!18!white}67.6 & \cellcolor{cineNarrative!12!white}61.1 & \cellcolor{cineNarrative!17!white}66.2 & \cellcolor{cineNarrative!7!white}56.7 & \cellcolor{cineNarrative!19!white}67.0 & \cellcolor{cineNarrative!12!white}64.3 \\
\href{https://huggingface.co/meta-llama/Llama-4-Scout-17B-16E-Instruct}{Llama 4 Scout 17B} & \cellcolor{cineNarrative!10!white}62.5 & \cellcolor{cineNarrative!12!white}60.9 & \cellcolor{cineNarrative!10!white}61.8 & \cellcolor{cineNarrative!12!white}60.3 & \cellcolor{cineNarrative!13!white}63.0 & \cellcolor{cineNarrative!8!white}61.9 \\
\bottomrule
\end{tabular}%
\end{adjustbox}
\end{table*}

\begin{table*}[htbp]
\centering
\small
\renewcommand{\arraystretch}{1.13}
\caption{Genre exact match by model and subtitle-input language. Values are percentages; higher is better ($\uparrow$). EN, AR, ID, FA, RO, and VI denote English, Arabic, Indonesian, Persian, Romanian, and Vietnamese, respectively. Within each language column, muted blue shading is normalized across the displayed models in the preferred direction. Best values are bold, and second-best distinct values are underlined at the displayed precision.}
\vspace{5pt}
\label{tab:genre-genre-exact-match-pct-by-language-appendix}
\begin{adjustbox}{max width=\textwidth}
\begin{tabular}{lrrrrrr}
\toprule
Model & EN & AR & ID & FA & RO & VI \\
\midrule
\href{https://developers.openai.com/api/docs/models/gpt-5.6-sol}{GPT-5.6 Sol} & \cellcolor{cineNarrative!30!white}28.0 & \cellcolor{cineNarrative!31!white}29.0 & \cellcolor{cineNarrative!29!white}29.0 & \cellcolor{cineNarrative!28!white}27.5 & \cellcolor{cineNarrative!30!white}30.0 & \cellcolor{cineNarrative!28!white}29.5 \\
\href{https://docs.cloud.google.com/gemini-enterprise-agent-platform/models/gemini/3-8-flash}{Gemini 3.8 Flash} & \cellcolor{cineNarrative!43!white}\textbf{37.0} & \cellcolor{cineNarrative!43!white}\textbf{40.0} & \cellcolor{cineNarrative!43!white}\textbf{40.5} & \cellcolor{cineNarrative!43!white}\textbf{42.5} & \cellcolor{cineNarrative!43!white}\textbf{42.0} & \cellcolor{cineNarrative!43!white}\textbf{43.0} \\
\href{https://platform.claude.com/docs/en/models/haiku-4-5/overview}{Claude Haiku 4.5} & \cellcolor{cineNarrative!15!white}17.5 & \cellcolor{cineNarrative!16!white}16.0 & \cellcolor{cineNarrative!18!white}20.0 & \cellcolor{cineNarrative!14!white}14.0 & \cellcolor{cineNarrative!16!white}17.5 & \cellcolor{cineNarrative!14!white}16.5 \\
\midrule
\href{https://developers.openai.com/api/docs/models/gpt-5-nano}{GPT-5 Nano} & \cellcolor{cineNarrative!10!white}13.5 & \cellcolor{cineNarrative!8!white}9.0 & \cellcolor{cineNarrative!9!white}12.5 & \cellcolor{cineNarrative!10!white}9.0 & \cellcolor{cineNarrative!11!white}13.0 & \cellcolor{cineNarrative!9!white}12.0 \\
\href{https://docs.cloud.google.com/gemini-enterprise-agent-platform/models/gemini/3-5-flash-lite}{Gemini 3.5 Flash Lite} & \cellcolor{cineNarrative!35!white}\underline{31.0} & \cellcolor{cineNarrative!30!white}28.0 & \cellcolor{cineNarrative!33!white}32.5 & \cellcolor{cineNarrative!30!white}29.5 & \cellcolor{cineNarrative!29!white}29.5 & \cellcolor{cineNarrative!32!white}33.5 \\
\midrule
\href{https://openrouter.ai/deepseek/deepseek-v4-pro-0813}{DeepSeek V4 Pro 1.6T} & \cellcolor{cineNarrative!43!white}\textbf{37.0} & \cellcolor{cineNarrative!34!white}\underline{32.0} & \cellcolor{cineNarrative!36!white}\underline{35.0} & \cellcolor{cineNarrative!32!white}\underline{31.0} & \cellcolor{cineNarrative!32!white}\underline{32.5} & \cellcolor{cineNarrative!36!white}\underline{36.5} \\
\href{https://huggingface.co/Qwen/Qwen3-235B-A22B-Instruct-2507}{Qwen3 235B} & \cellcolor{cineNarrative!13!white}16.0 & \cellcolor{cineNarrative!9!white}10.0 & \cellcolor{cineNarrative!11!white}14.0 & \cellcolor{cineNarrative!12!white}11.0 & \cellcolor{cineNarrative!13!white}14.5 & \cellcolor{cineNarrative!9!white}12.5 \\
\href{https://huggingface.co/mistralai/Mistral-Small-4-119B-2603}{Mistral 4 119B} & \cellcolor{cineNarrative!7!white}11.5 & \cellcolor{cineNarrative!7!white}8.0 & \cellcolor{cineNarrative!8!white}12.0 & \cellcolor{cineNarrative!7!white}6.5 & \cellcolor{cineNarrative!7!white}9.5 & \cellcolor{cineNarrative!7!white}10.5 \\
\href{https://huggingface.co/meta-llama/Llama-4-Scout-17B-16E-Instruct}{Llama 4 Scout 17B} & \cellcolor{cineNarrative!8!white}12.5 & \cellcolor{cineNarrative!10!white}10.5 & \cellcolor{cineNarrative!7!white}11.0 & \cellcolor{cineNarrative!9!white}8.5 & \cellcolor{cineNarrative!10!white}12.0 & \cellcolor{cineNarrative!9!white}12.5 \\
\bottomrule
\end{tabular}%
\end{adjustbox}
\end{table*}

\begin{table*}[htbp]
\centering
\scriptsize
\setlength{\tabcolsep}{3pt}
\caption{Genre-wise F1 scores pooled across the six subtitle-input languages for each model. Values are percentages. \emph{Gold} reports label prevalence in the evaluated film set, providing context for less frequent genres. Each genre is evaluated as a one-versus-rest binary classification problem; labels with no positive gold examples are omitted because their F1 is not informative. Within each genre row, blue shading compares performance across models. Best values are bold, and second-best distinct values are underlined at the displayed precision. Model headings link to the corresponding provider or model pages.}
\vspace{5pt}
\label{tab:genre-label-f1-appendix}
\begin{adjustbox}{max width=\textwidth}
\begin{tabular}{lrrrrrrrrrr}
\toprule
Genre & Gold (\%) & \href{https://developers.openai.com/api/docs/models/gpt-5.6-sol}{Sol} & \href{https://docs.cloud.google.com/gemini-enterprise-agent-platform/models/gemini/3-8-flash}{Gemini 3.8} & \href{https://platform.claude.com/docs/en/models/haiku-4-5/overview}{Claude} & \href{https://developers.openai.com/api/docs/models/gpt-5-nano}{Nano} & \href{https://docs.cloud.google.com/gemini-enterprise-agent-platform/models/gemini/3-5-flash-lite}{Gemini 3.5} & \href{https://openrouter.ai/deepseek/deepseek-v4-pro-0813}{DeepSeek} & \href{https://huggingface.co/Qwen/Qwen3-235B-A22B-Instruct-2507}{Qwen3} & \href{https://huggingface.co/mistralai/Mistral-Small-4-119B-2603}{Mistral 4} & \href{https://huggingface.co/meta-llama/Llama-4-Scout-17B-16E-Instruct}{Llama 4} \\
\midrule
Drama & 50.0 & \cellcolor{cineNarrative!33!white}83.0 & \cellcolor{cineNarrative!43!white}\textbf{87.2} & \cellcolor{cineNarrative!25!white}79.5 & \cellcolor{cineNarrative!11!white}73.5 & \cellcolor{cineNarrative!37!white}84.9 & \cellcolor{cineNarrative!40!white}\underline{86.1} & \cellcolor{cineNarrative!19!white}76.8 & \cellcolor{cineNarrative!13!white}74.5 & \cellcolor{cineNarrative!7!white}71.9 \\
Comedy & 38.0 & \cellcolor{cineNarrative!42!white}\underline{90.6} & \cellcolor{cineNarrative!43!white}\textbf{91.3} & \cellcolor{cineNarrative!20!white}77.1 & \cellcolor{cineNarrative!7!white}69.3 & \cellcolor{cineNarrative!39!white}88.9 & \cellcolor{cineNarrative!40!white}89.2 & \cellcolor{cineNarrative!16!white}75.0 & \cellcolor{cineNarrative!10!white}71.3 & \cellcolor{cineNarrative!23!white}79.2 \\
Adventure & 36.5 & \cellcolor{cineNarrative!40!white}82.3 & \cellcolor{cineNarrative!43!white}\textbf{84.7} & \cellcolor{cineNarrative!7!white}55.6 & \cellcolor{cineNarrative!37!white}80.0 & \cellcolor{cineNarrative!41!white}\underline{83.3} & \cellcolor{cineNarrative!35!white}78.3 & \cellcolor{cineNarrative!23!white}68.6 & \cellcolor{cineNarrative!17!white}63.9 & \cellcolor{cineNarrative!34!white}77.7 \\
Action & 34.5 & \cellcolor{cineNarrative!40!white}87.6 & \cellcolor{cineNarrative!43!white}\textbf{90.4} & \cellcolor{cineNarrative!29!white}78.6 & \cellcolor{cineNarrative!31!white}80.3 & \cellcolor{cineNarrative!42!white}\underline{89.9} & \cellcolor{cineNarrative!39!white}87.2 & \cellcolor{cineNarrative!29!white}78.5 & \cellcolor{cineNarrative!7!white}60.8 & \cellcolor{cineNarrative!21!white}72.4 \\
Thriller & 30.0 & \cellcolor{cineNarrative!40!white}\underline{78.8} & \cellcolor{cineNarrative!43!white}\textbf{79.8} & \cellcolor{cineNarrative!25!white}73.4 & \cellcolor{cineNarrative!26!white}73.8 & \cellcolor{cineNarrative!36!white}77.2 & \cellcolor{cineNarrative!39!white}78.2 & \cellcolor{cineNarrative!7!white}67.3 & \cellcolor{cineNarrative!8!white}67.7 & \cellcolor{cineNarrative!10!white}68.3 \\
Fantasy & 22.0 & \cellcolor{cineNarrative!29!white}\underline{72.1} & \cellcolor{cineNarrative!43!white}\textbf{78.9} & \cellcolor{cineNarrative!17!white}65.9 & \cellcolor{cineNarrative!21!white}68.0 & \cellcolor{cineNarrative!20!white}67.6 & \cellcolor{cineNarrative!26!white}70.7 & \cellcolor{cineNarrative!7!white}61.1 & \cellcolor{cineNarrative!11!white}62.9 & \cellcolor{cineNarrative!12!white}63.5 \\
Family & 17.0 & \cellcolor{cineNarrative!40!white}\underline{80.4} & \cellcolor{cineNarrative!43!white}\textbf{82.9} & \cellcolor{cineNarrative!37!white}77.5 & \cellcolor{cineNarrative!22!white}64.6 & \cellcolor{cineNarrative!18!white}60.6 & \cellcolor{cineNarrative!40!white}\underline{80.4} & \cellcolor{cineNarrative!25!white}66.9 & \cellcolor{cineNarrative!32!white}73.2 & \cellcolor{cineNarrative!7!white}51.0 \\
Sci-Fi & 17.0 & \cellcolor{cineNarrative!40!white}83.7 & \cellcolor{cineNarrative!43!white}\textbf{85.9} & \cellcolor{cineNarrative!35!white}79.3 & \cellcolor{cineNarrative!32!white}76.7 & \cellcolor{cineNarrative!37!white}80.4 & \cellcolor{cineNarrative!43!white}\underline{85.5} & \cellcolor{cineNarrative!38!white}81.4 & \cellcolor{cineNarrative!39!white}82.3 & \cellcolor{cineNarrative!7!white}54.7 \\
Mystery & 16.5 & \cellcolor{cineNarrative!39!white}\underline{71.0} & \cellcolor{cineNarrative!43!white}\textbf{76.7} & \cellcolor{cineNarrative!7!white}21.9 & \cellcolor{cineNarrative!34!white}63.4 & \cellcolor{cineNarrative!38!white}69.3 & \cellcolor{cineNarrative!38!white}69.0 & \cellcolor{cineNarrative!7!white}22.4 & \cellcolor{cineNarrative!30!white}57.2 & \cellcolor{cineNarrative!33!white}62.1 \\
Crime & 16.0 & \cellcolor{cineNarrative!38!white}79.1 & \cellcolor{cineNarrative!43!white}\textbf{83.1} & \cellcolor{cineNarrative!32!white}74.0 & \cellcolor{cineNarrative!28!white}70.3 & \cellcolor{cineNarrative!42!white}\underline{82.5} & \cellcolor{cineNarrative!38!white}79.1 & \cellcolor{cineNarrative!25!white}68.0 & \cellcolor{cineNarrative!20!white}63.8 & \cellcolor{cineNarrative!7!white}52.5 \\
Animation & 13.0 & \cellcolor{cineNarrative!43!white}\underline{91.3} & \cellcolor{cineNarrative!43!white}\textbf{91.7} & \cellcolor{cineNarrative!23!white}62.9 & \cellcolor{cineNarrative!37!white}83.4 & \cellcolor{cineNarrative!42!white}91.0 & \cellcolor{cineNarrative!42!white}90.7 & \cellcolor{cineNarrative!34!white}78.2 & \cellcolor{cineNarrative!30!white}72.4 & \cellcolor{cineNarrative!7!white}39.6 \\
Romance & 13.0 & \cellcolor{cineNarrative!26!white}67.5 & \cellcolor{cineNarrative!43!white}\textbf{76.7} & \cellcolor{cineNarrative!30!white}69.8 & \cellcolor{cineNarrative!32!white}70.6 & \cellcolor{cineNarrative!41!white}75.6 & \cellcolor{cineNarrative!41!white}\underline{75.7} & \cellcolor{cineNarrative!28!white}68.3 & \cellcolor{cineNarrative!7!white}56.9 & \cellcolor{cineNarrative!24!white}66.5 \\
Horror & 12.5 & \cellcolor{cineNarrative!43!white}\textbf{88.3} & \cellcolor{cineNarrative!43!white}\underline{87.9} & \cellcolor{cineNarrative!33!white}76.7 & \cellcolor{cineNarrative!30!white}73.2 & \cellcolor{cineNarrative!38!white}82.3 & \cellcolor{cineNarrative!39!white}83.7 & \cellcolor{cineNarrative!13!white}52.8 & \cellcolor{cineNarrative!24!white}66.4 & \cellcolor{cineNarrative!7!white}46.3 \\
Biography & 9.5 & \cellcolor{cineNarrative!38!white}82.5 & \cellcolor{cineNarrative!43!white}\textbf{95.0} & \cellcolor{cineNarrative!32!white}68.6 & \cellcolor{cineNarrative!38!white}82.3 & \cellcolor{cineNarrative!42!white}\underline{93.2} & \cellcolor{cineNarrative!40!white}86.6 & \cellcolor{cineNarrative!7!white}5.1 & \cellcolor{cineNarrative!8!white}6.8 & \cellcolor{cineNarrative!31!white}65.1 \\
Musical & 6.5 & \cellcolor{cineNarrative!40!white}59.5 & \cellcolor{cineNarrative!43!white}\textbf{63.2} & \cellcolor{cineNarrative!13!white}20.7 & \cellcolor{cineNarrative!43!white}\underline{63.0} & \cellcolor{cineNarrative!24!white}35.8 & \cellcolor{cineNarrative!30!white}44.0 & \cellcolor{cineNarrative!25!white}37.1 & \cellcolor{cineNarrative!16!white}25.0 & \cellcolor{cineNarrative!7!white}11.8 \\
History & 5.5 & \cellcolor{cineNarrative!36!white}69.2 & \cellcolor{cineNarrative!34!white}67.2 & \cellcolor{cineNarrative!43!white}\textbf{76.6} & \cellcolor{cineNarrative!24!white}55.6 & \cellcolor{cineNarrative!35!white}67.6 & \cellcolor{cineNarrative!37!white}\underline{69.8} & \cellcolor{cineNarrative!12!white}42.2 & \cellcolor{cineNarrative!15!white}45.8 & \cellcolor{cineNarrative!7!white}37.0 \\
Music & 4.0 & \cellcolor{cineNarrative!26!white}61.5 & \cellcolor{cineNarrative!43!white}\textbf{74.4} & \cellcolor{cineNarrative!23!white}59.5 & \cellcolor{cineNarrative!42!white}\underline{73.6} & \cellcolor{cineNarrative!14!white}52.6 & \cellcolor{cineNarrative!32!white}65.9 & \cellcolor{cineNarrative!7!white}47.5 & \cellcolor{cineNarrative!23!white}59.8 & \cellcolor{cineNarrative!19!white}56.8 \\
War & 3.5 & \cellcolor{cineNarrative!27!white}78.8 & \cellcolor{cineNarrative!32!white}82.0 & \cellcolor{cineNarrative!43!white}\textbf{89.1} & \cellcolor{cineNarrative!7!white}65.6 & \cellcolor{cineNarrative!34!white}83.1 & \cellcolor{cineNarrative!39!white}\underline{86.3} & \cellcolor{cineNarrative!15!white}70.6 & \cellcolor{cineNarrative!28!white}79.1 & \cellcolor{cineNarrative!35!white}84.0 \\
Sport & 2.0 & \cellcolor{cineNarrative!43!white}\textbf{100.0} & \cellcolor{cineNarrative!43!white}\textbf{100.0} & \cellcolor{cineNarrative!38!white}85.7 & \cellcolor{cineNarrative!36!white}81.2 & \cellcolor{cineNarrative!42!white}\underline{97.9} & \cellcolor{cineNarrative!43!white}\textbf{100.0} & \cellcolor{cineNarrative!7!white}0.0 & \cellcolor{cineNarrative!38!white}85.7 & \cellcolor{cineNarrative!30!white}62.9 \\
Western & 2.0 & \cellcolor{cineNarrative!43!white}\textbf{83.7} & \cellcolor{cineNarrative!43!white}\textbf{83.7} & \cellcolor{cineNarrative!20!white}75.0 & \cellcolor{cineNarrative!7!white}69.8 & \cellcolor{cineNarrative!17!white}73.7 & \cellcolor{cineNarrative!43!white}\textbf{83.7} & \cellcolor{cineNarrative!25!white}76.9 & \cellcolor{cineNarrative!33!white}\underline{80.0} & \cellcolor{cineNarrative!33!white}\underline{80.0} \\
Documentary & 0.5 & \cellcolor{cineNarrative!43!white}\textbf{100.0} & \cellcolor{cineNarrative!43!white}\textbf{100.0} & \cellcolor{cineNarrative!43!white}\textbf{100.0} & \cellcolor{cineNarrative!7!white}27.2 & \cellcolor{cineNarrative!43!white}\textbf{100.0} & \cellcolor{cineNarrative!43!white}\textbf{100.0} & \cellcolor{cineNarrative!43!white}\textbf{100.0} & \cellcolor{cineNarrative!36!white}\underline{85.7} & \cellcolor{cineNarrative!43!white}\textbf{100.0} \\
\bottomrule
\end{tabular}%
\end{adjustbox}
\end{table*}

\subsection{Paired Cross-Lingual Uncertainty}
\label{app:rq2-uncertainty}
Table~\ref{tab:paired-language-uncertainty-appendix} resamples matched films across subtitle languages rather than treating translations as independent films. The small narrative declines observed for GPT-5.6 Sol and DeepSeek V4 Pro have intervals that include zero, as do Llama 4 Scout's positive country point estimate and several small positive age-exact changes. Llama's narrative decline remains clearly negative. The paired analysis separates stable directional changes from visually small point-estimate differences.
\begin{table*}[htbp]
\centering
\small
\caption{Paired English-to-cross-lingual differences with percentile confidence intervals. Each $\Delta$ is computed on the same films as the mean over the five non-English subtitle-input languages minus English. Point estimates are calculated before rounding and displayed to at most two decimal places. Confidence intervals are obtained from 2,000 film-level bootstrap resamples that preserve the six-language pairing; genre micro-F1 is recomputed from pooled label decisions within each resample. Intervals containing zero do not indicate a consistently signed change.}
\vspace{5pt}
\label{tab:paired-language-uncertainty-appendix}
\begin{adjustbox}{max width=\textwidth}
\begin{tabular}{lrrrr}
\toprule
Model & Narrative $\Delta$ & Genre F1 $\Delta$ (pp) & Age exact $\Delta$ (pp) & Country EM $\Delta$ (pp) \\
\midrule
\href{https://developers.openai.com/api/docs/models/gpt-5.6-sol}{GPT-5.6 Sol} & $-0.05$ [-0.11, +0.02] & $-0.8$ [-1.71, +0.30] & $-2.9$ [-7.30, +1.40] & $-2.1$ [-3.51, -0.72] \\
\href{https://docs.cloud.google.com/gemini-enterprise-agent-platform/models/gemini/3-8-flash}{Gemini 3.8 Flash} & $-0.09$ [-0.15, -0.02] & $+0.3$ [-0.62, +1.12] & $-2.7$ [-6.90, +1.30] & $-1.1$ [-2.10, -0.02] \\
\href{https://platform.claude.com/docs/en/models/haiku-4-5/overview}{Claude Haiku 4.5} & $-0.14$ [-0.2, -0.07] & $-0.9$ [-2.27, +0.53] & $-2.5$ [-7.30, +2.00] & $-4.4$ [-6.56, -2.25] \\
\midrule
\href{https://developers.openai.com/api/docs/models/gpt-5-nano}{GPT-5 Nano} & $-0.19$ [-0.25, -0.14] & $-2.8$ [-4.51, -1.17] & $+1.4$ [-2.80, +5.50] & $+0.1$ [-2.58, +2.59] \\
\href{https://docs.cloud.google.com/gemini-enterprise-agent-platform/models/gemini/3-5-flash-lite}{Gemini 3.5 Flash Lite} & $-0.13$ [-0.19, -0.06] & $-1.1$ [-2.24, +0.00] & $-0.8$ [-5.10, +3.40] & $-2.0$ [-3.53, -0.42] \\
\midrule
\href{https://openrouter.ai/deepseek/deepseek-v4-pro-0813}{DeepSeek V4 Pro 1.6T} & $-0.06$ [-0.12, +0.01] & $-1.9$ [-3.01, -0.81] & $-4.6$ [-9.60, +0.00] & $-0.6$ [-2.33, +1.37] \\
\href{https://huggingface.co/Qwen/Qwen3-235B-A22B-Instruct-2507}{Qwen3 235B} & $-0.14$ [-0.2, -0.08] & $-1.9$ [-3.52, -0.25] & $+0.6$ [-4.20, +5.30] & $-4.6$ [-7.04, -2.26] \\
\href{https://huggingface.co/mistralai/Mistral-Small-4-119B-2603}{Mistral 4 119B} & $-0.12$ [-0.18, -0.06] & $-3.8$ [-5.76, -1.83] & $+2.3$ [-2.10, +6.50] & $-2.4$ [-4.70, -0.01] \\
\href{https://huggingface.co/meta-llama/Llama-4-Scout-17B-16E-Instruct}{Llama 4 Scout 17B} & $-0.28$ [-0.35, -0.22] & $-1.7$ [-3.45, +0.22] & $+2.5$ [-2.50, +7.30] & $+1.2$ [-1.70, +4.15] \\
\bottomrule
\end{tabular}%
\end{adjustbox}
\end{table*}

\section{Additional Evidence for RQ~\ref{rq:cultural-errors}: Age and National-Rating Calibration}
\label{app:rq3-details}
\subsection{Age Error Magnitude and Direction}
\label{app:rq3-age}
\paragraph{Near-miss accuracy conceals asymmetric age behavior.}
The tables jointly report exact match, one- and two-year tolerance, MAE, and error direction. Across six subtitle languages, Gemini 3.8 Flash matches the reference age exactly in 34.2\% of predictions and falls within one year in 81.4\%; GPT-5.6 Sol reaches 31.1\% and 78.8\%, respectively. Their error directions differ sharply: Sol overpredicts age more often than it underpredicts (48.3\% versus 20.6\%), while Gemini shows the reverse pattern (15.2\% versus 50.6\%).

Large errors also separate models that look similar under near-match metrics. DeepSeek V4 Pro and Sol each have only 5.4\% of predictions more than two years from the reference, compared with 17.0\% for Qwen3, 24.9\% for Mistral 4, and 29.1\% for Llama 4 Scout. Models with similar MAE or within-one-year accuracy can therefore exhibit very different error magnitude and direction.
\begin{table}[htbp]
\centering
\small
\caption{Cultural prediction metrics averaged across all six subtitle languages. Arrows indicate the preferred direction: $\uparrow$ is higher-is-better and $\downarrow$ is lower-is-better. Age and country MAE are computed within their respective ordered label spaces. Within each metric column, bronze shading is normalized across the displayed models in the preferred direction. Best values are bold, and second-best distinct values are underlined at the displayed precision; ties share the same emphasis.}
\vspace{5pt}
\label{tab:cultural-results-appendix}
\begin{adjustbox}{max width=\textwidth}
\begin{tabular}{lrrrrrr}
\toprule
Model & Age exact (\%) $\uparrow$ & Age $\pm$1 (\%) $\uparrow$ & Age $\pm$2 (\%) $\uparrow$ & Age MAE $\downarrow$ & Country EM (\%) $\uparrow$ & Country MAE $\downarrow$ \\
\midrule
\href{https://developers.openai.com/api/docs/models/gpt-5.6-sol}{GPT-5.6 Sol} & \cellcolor{cineCultural!36!white}31.1 & \cellcolor{cineCultural!40!white}78.8 & \cellcolor{cineCultural!42!white}\underline{94.6} & \cellcolor{cineCultural!41!white}\underline{0.98} & \cellcolor{cineCultural!33!white}\underline{65.4} & \cellcolor{cineCultural!35!white}\underline{0.44} \\
\href{https://docs.cloud.google.com/gemini-enterprise-agent-platform/models/gemini/3-8-flash}{Gemini 3.8 Flash} & \cellcolor{cineCultural!43!white}\textbf{34.2} & \cellcolor{cineCultural!43!white}\textbf{81.4} & \cellcolor{cineCultural!43!white}\textbf{95.5} & \cellcolor{cineCultural!43!white}\textbf{0.93} & \cellcolor{cineCultural!43!white}\textbf{77.9} & \cellcolor{cineCultural!43!white}\textbf{0.31} \\
\href{https://platform.claude.com/docs/en/models/haiku-4-5/overview}{Claude Haiku 4.5} & \cellcolor{cineCultural!27!white}27.4 & \cellcolor{cineCultural!33!white}72.8 & \cellcolor{cineCultural!36!white}90.4 & \cellcolor{cineCultural!35!white}1.16 & \cellcolor{cineCultural!24!white}54.3 & \cellcolor{cineCultural!28!white}0.57 \\
\href{https://developers.openai.com/api/docs/models/gpt-5-nano}{GPT-5 Nano} & \cellcolor{cineCultural!12!white}20.7 & \cellcolor{cineCultural!19!white}60.8 & \cellcolor{cineCultural!21!white}80.4 & \cellcolor{cineCultural!20!white}1.56 & \cellcolor{cineCultural!15!white}43.9 & \cellcolor{cineCultural!17!white}0.76 \\
\href{https://docs.cloud.google.com/gemini-enterprise-agent-platform/models/gemini/3-5-flash-lite}{Gemini 3.5 Flash Lite} & \cellcolor{cineCultural!37!white}\underline{31.8} & \cellcolor{cineCultural!32!white}72.2 & \cellcolor{cineCultural!36!white}91.0 & \cellcolor{cineCultural!37!white}1.09 & \cellcolor{cineCultural!32!white}64.7 & \cellcolor{cineCultural!34!white}0.46 \\
\href{https://openrouter.ai/deepseek/deepseek-v4-pro-0813}{DeepSeek V4 Pro 1.6T} & \cellcolor{cineCultural!35!white}30.7 & \cellcolor{cineCultural!41!white}\underline{79.3} & \cellcolor{cineCultural!42!white}\underline{94.6} & \cellcolor{cineCultural!41!white}\underline{0.98} & \cellcolor{cineCultural!27!white}58.0 & \cellcolor{cineCultural!31!white}0.52 \\
\href{https://huggingface.co/Qwen/Qwen3-235B-A22B-Instruct-2507}{Qwen3 235B} & \cellcolor{cineCultural!15!white}22.0 & \cellcolor{cineCultural!17!white}59.2 & \cellcolor{cineCultural!25!white}83.0 & \cellcolor{cineCultural!24!white}1.47 & \cellcolor{cineCultural!18!white}47.9 & \cellcolor{cineCultural!24!white}0.64 \\
\href{https://huggingface.co/mistralai/Mistral-Small-4-119B-2603}{Mistral 4 119B} & \cellcolor{cineCultural!9!white}19.4 & \cellcolor{cineCultural!12!white}54.5 & \cellcolor{cineCultural!13!white}75.1 & \cellcolor{cineCultural!12!white}1.78 & \cellcolor{cineCultural!18!white}47.4 & \cellcolor{cineCultural!21!white}0.68 \\
\href{https://huggingface.co/meta-llama/Llama-4-Scout-17B-16E-Instruct}{Llama 4 Scout 17B} & \cellcolor{cineCultural!7!white}18.6 & \cellcolor{cineCultural!7!white}50.2 & \cellcolor{cineCultural!7!white}70.9 & \cellcolor{cineCultural!7!white}1.93 & \cellcolor{cineCultural!7!white}34.1 & \cellcolor{cineCultural!7!white}0.93 \\
\bottomrule
\end{tabular}%
\end{adjustbox}
\end{table}

\begin{table*}[htbp]
\centering
\small
\renewcommand{\arraystretch}{1.13}
\caption{Age exact match by model and subtitle-input language. Values are percentages; higher is better ($\uparrow$). EN, AR, ID, FA, RO, and VI denote English, Arabic, Indonesian, Persian, Romanian, and Vietnamese, respectively. Within each language column, muted bronze shading is normalized across the displayed models in the preferred direction. Best values are bold, and second-best distinct values are underlined at the displayed precision.}
\vspace{5pt}
\label{tab:cultural-age-exact-match-pct-by-language-appendix}
\begin{adjustbox}{max width=\textwidth}
\begin{tabular}{lrrrrrr}
\toprule
Model & EN & AR & ID & FA & RO & VI \\
\midrule
\href{https://developers.openai.com/api/docs/models/gpt-5.6-sol}{GPT-5.6 Sol} & \cellcolor{cineCultural!38!white}33.5 & \cellcolor{cineCultural!37!white}31.5 & \cellcolor{cineCultural!39!white}\underline{31.5} & \cellcolor{cineCultural!29!white}26.0 & \cellcolor{cineCultural!43!white}\textbf{33.0} & \cellcolor{cineCultural!32!white}31.0 \\
\href{https://docs.cloud.google.com/gemini-enterprise-agent-platform/models/gemini/3-8-flash}{Gemini 3.8 Flash} & \cellcolor{cineCultural!43!white}\textbf{36.5} & \cellcolor{cineCultural!43!white}\textbf{34.5} & \cellcolor{cineCultural!43!white}\textbf{33.0} & \cellcolor{cineCultural!41!white}\underline{32.0} & \cellcolor{cineCultural!42!white}\underline{32.5} & \cellcolor{cineCultural!43!white}\textbf{37.0} \\
\href{https://platform.claude.com/docs/en/models/haiku-4-5/overview}{Claude Haiku 4.5} & \cellcolor{cineCultural!30!white}29.5 & \cellcolor{cineCultural!19!white}23.0 & \cellcolor{cineCultural!33!white}29.0 & \cellcolor{cineCultural!28!white}25.5 & \cellcolor{cineCultural!31!white}28.0 & \cellcolor{cineCultural!29!white}29.5 \\
\midrule
\href{https://developers.openai.com/api/docs/models/gpt-5-nano}{GPT-5 Nano} & \cellcolor{cineCultural!12!white}19.5 & \cellcolor{cineCultural!12!white}19.5 & \cellcolor{cineCultural!15!white}22.0 & \cellcolor{cineCultural!7!white}14.5 & \cellcolor{cineCultural!22!white}24.5 & \cellcolor{cineCultural!19!white}24.0 \\
\href{https://docs.cloud.google.com/gemini-enterprise-agent-platform/models/gemini/3-5-flash-lite}{Gemini 3.5 Flash Lite} & \cellcolor{cineCultural!36!white}32.5 & \cellcolor{cineCultural!40!white}\underline{33.0} & \cellcolor{cineCultural!39!white}\underline{31.5} & \cellcolor{cineCultural!43!white}\textbf{33.0} & \cellcolor{cineCultural!36!white}30.0 & \cellcolor{cineCultural!32!white}31.0 \\
\midrule
\href{https://openrouter.ai/deepseek/deepseek-v4-pro-0813}{DeepSeek V4 Pro 1.6T} & \cellcolor{cineCultural!39!white}\underline{34.5} & \cellcolor{cineCultural!31!white}28.5 & \cellcolor{cineCultural!35!white}30.0 & \cellcolor{cineCultural!30!white}26.5 & \cellcolor{cineCultural!42!white}\underline{32.5} & \cellcolor{cineCultural!34!white}\underline{32.0} \\
\href{https://huggingface.co/Qwen/Qwen3-235B-A22B-Instruct-2507}{Qwen3 235B} & \cellcolor{cineCultural!16!white}21.5 & \cellcolor{cineCultural!14!white}20.5 & \cellcolor{cineCultural!13!white}21.5 & \cellcolor{cineCultural!32!white}27.5 & \cellcolor{cineCultural!16!white}22.0 & \cellcolor{cineCultural!10!white}19.0 \\
\href{https://huggingface.co/mistralai/Mistral-Small-4-119B-2603}{Mistral 4 119B} & \cellcolor{cineCultural!9!white}17.5 & \cellcolor{cineCultural!13!white}20.0 & \cellcolor{cineCultural!7!white}19.0 & \cellcolor{cineCultural!21!white}21.5 & \cellcolor{cineCultural!7!white}18.5 & \cellcolor{cineCultural!12!white}20.0 \\
\href{https://huggingface.co/meta-llama/Llama-4-Scout-17B-16E-Instruct}{Llama 4 Scout 17B} & \cellcolor{cineCultural!7!white}16.5 & \cellcolor{cineCultural!7!white}17.0 & \cellcolor{cineCultural!12!white}21.0 & \cellcolor{cineCultural!20!white}21.0 & \cellcolor{cineCultural!7!white}18.5 & \cellcolor{cineCultural!7!white}17.5 \\
\bottomrule
\end{tabular}%
\end{adjustbox}
\end{table*}

\begin{table*}[htbp]
\centering
\small
\renewcommand{\arraystretch}{1.13}
\caption{Age accuracy within one year by model and subtitle-input language. Values are percentages; higher is better ($\uparrow$). EN, AR, ID, FA, RO, and VI denote English, Arabic, Indonesian, Persian, Romanian, and Vietnamese, respectively. Within each language column, muted bronze shading is normalized across the displayed models in the preferred direction. Best values are bold, and second-best distinct values are underlined at the displayed precision.}
\vspace{5pt}
\label{tab:cultural-age-within-one-year-pct-by-language-appendix}
\begin{adjustbox}{max width=\textwidth}
\begin{tabular}{lrrrrrr}
\toprule
Model & EN & AR & ID & FA & RO & VI \\
\midrule
\href{https://developers.openai.com/api/docs/models/gpt-5.6-sol}{GPT-5.6 Sol} & \cellcolor{cineCultural!42!white}82.0 & \cellcolor{cineCultural!40!white}78.5 & \cellcolor{cineCultural!37!white}77.5 & \cellcolor{cineCultural!41!white}\underline{77.5} & \cellcolor{cineCultural!38!white}78.0 & \cellcolor{cineCultural!42!white}\underline{79.0} \\
\href{https://docs.cloud.google.com/gemini-enterprise-agent-platform/models/gemini/3-8-flash}{Gemini 3.8 Flash} & \cellcolor{cineCultural!43!white}\textbf{83.0} & \cellcolor{cineCultural!43!white}\textbf{81.5} & \cellcolor{cineCultural!43!white}\textbf{82.5} & \cellcolor{cineCultural!43!white}\textbf{79.5} & \cellcolor{cineCultural!43!white}\textbf{82.0} & \cellcolor{cineCultural!43!white}\textbf{80.0} \\
\href{https://platform.claude.com/docs/en/models/haiku-4-5/overview}{Claude Haiku 4.5} & \cellcolor{cineCultural!38!white}78.0 & \cellcolor{cineCultural!32!white}71.5 & \cellcolor{cineCultural!34!white}75.0 & \cellcolor{cineCultural!30!white}68.5 & \cellcolor{cineCultural!30!white}71.0 & \cellcolor{cineCultural!35!white}72.5 \\
\midrule
\href{https://developers.openai.com/api/docs/models/gpt-5-nano}{GPT-5 Nano} & \cellcolor{cineCultural!22!white}60.5 & \cellcolor{cineCultural!18!white}59.0 & \cellcolor{cineCultural!17!white}61.0 & \cellcolor{cineCultural!7!white}48.0 & \cellcolor{cineCultural!28!white}69.5 & \cellcolor{cineCultural!28!white}66.5 \\
\href{https://docs.cloud.google.com/gemini-enterprise-agent-platform/models/gemini/3-5-flash-lite}{Gemini 3.5 Flash Lite} & \cellcolor{cineCultural!36!white}75.5 & \cellcolor{cineCultural!31!white}70.5 & \cellcolor{cineCultural!27!white}69.0 & \cellcolor{cineCultural!35!white}72.5 & \cellcolor{cineCultural!30!white}71.0 & \cellcolor{cineCultural!37!white}74.5 \\
\midrule
\href{https://openrouter.ai/deepseek/deepseek-v4-pro-0813}{DeepSeek V4 Pro 1.6T} & \cellcolor{cineCultural!43!white}\underline{82.5} & \cellcolor{cineCultural!41!white}\underline{79.5} & \cellcolor{cineCultural!39!white}\underline{79.5} & \cellcolor{cineCultural!39!white}76.0 & \cellcolor{cineCultural!40!white}\underline{79.5} & \cellcolor{cineCultural!42!white}\underline{79.0} \\
\href{https://huggingface.co/Qwen/Qwen3-235B-A22B-Instruct-2507}{Qwen3 235B} & \cellcolor{cineCultural!25!white}63.5 & \cellcolor{cineCultural!16!white}57.5 & \cellcolor{cineCultural!12!white}57.0 & \cellcolor{cineCultural!20!white}59.0 & \cellcolor{cineCultural!15!white}57.5 & \cellcolor{cineCultural!22!white}61.0 \\
\href{https://huggingface.co/mistralai/Mistral-Small-4-119B-2603}{Mistral 4 119B} & \cellcolor{cineCultural!18!white}56.0 & \cellcolor{cineCultural!14!white}55.5 & \cellcolor{cineCultural!9!white}54.5 & \cellcolor{cineCultural!12!white}52.0 & \cellcolor{cineCultural!8!white}52.0 & \cellcolor{cineCultural!18!white}57.0 \\
\href{https://huggingface.co/meta-llama/Llama-4-Scout-17B-16E-Instruct}{Llama 4 Scout 17B} & \cellcolor{cineCultural!7!white}44.0 & \cellcolor{cineCultural!7!white}49.5 & \cellcolor{cineCultural!7!white}53.0 & \cellcolor{cineCultural!17!white}56.5 & \cellcolor{cineCultural!7!white}51.0 & \cellcolor{cineCultural!7!white}47.5 \\
\bottomrule
\end{tabular}%
\end{adjustbox}
\end{table*}

\begin{table*}[htbp]
\centering
\small
\renewcommand{\arraystretch}{1.13}
\caption{Age accuracy within two years by model and subtitle-input language. Values are percentages; higher is better ($\uparrow$). EN, AR, ID, FA, RO, and VI denote English, Arabic, Indonesian, Persian, Romanian, and Vietnamese, respectively. Within each language column, muted bronze shading is normalized across the displayed models in the preferred direction. Best values are bold, and second-best distinct values are underlined at the displayed precision.}
\vspace{5pt}
\label{tab:cultural-age-within-two-years-pct-by-language-appendix}
\begin{adjustbox}{max width=\textwidth}
\begin{tabular}{lrrrrrr}
\toprule
Model & EN & AR & ID & FA & RO & VI \\
\midrule
\href{https://developers.openai.com/api/docs/models/gpt-5.6-sol}{GPT-5.6 Sol} & \cellcolor{cineCultural!43!white}\textbf{97.0} & \cellcolor{cineCultural!40!white}94.0 & \cellcolor{cineCultural!42!white}\underline{94.0} & \cellcolor{cineCultural!42!white}\underline{94.0} & \cellcolor{cineCultural!43!white}\textbf{95.0} & \cellcolor{cineCultural!40!white}93.5 \\
\href{https://docs.cloud.google.com/gemini-enterprise-agent-platform/models/gemini/3-8-flash}{Gemini 3.8 Flash} & \cellcolor{cineCultural!43!white}\textbf{97.0} & \cellcolor{cineCultural!43!white}\textbf{96.0} & \cellcolor{cineCultural!43!white}\textbf{94.5} & \cellcolor{cineCultural!43!white}\textbf{95.0} & \cellcolor{cineCultural!43!white}\textbf{95.0} & \cellcolor{cineCultural!43!white}\textbf{95.5} \\
\href{https://platform.claude.com/docs/en/models/haiku-4-5/overview}{Claude Haiku 4.5} & \cellcolor{cineCultural!41!white}\underline{95.5} & \cellcolor{cineCultural!31!white}87.0 & \cellcolor{cineCultural!38!white}91.5 & \cellcolor{cineCultural!33!white}87.5 & \cellcolor{cineCultural!35!white}89.5 & \cellcolor{cineCultural!38!white}91.5 \\
\midrule
\href{https://developers.openai.com/api/docs/models/gpt-5-nano}{GPT-5 Nano} & \cellcolor{cineCultural!23!white}79.5 & \cellcolor{cineCultural!22!white}80.0 & \cellcolor{cineCultural!21!white}81.0 & \cellcolor{cineCultural!7!white}69.0 & \cellcolor{cineCultural!31!white}87.0 & \cellcolor{cineCultural!30!white}86.0 \\
\href{https://docs.cloud.google.com/gemini-enterprise-agent-platform/models/gemini/3-5-flash-lite}{Gemini 3.5 Flash Lite} & \cellcolor{cineCultural!40!white}94.0 & \cellcolor{cineCultural!33!white}88.5 & \cellcolor{cineCultural!36!white}90.0 & \cellcolor{cineCultural!40!white}92.5 & \cellcolor{cineCultural!38!white}92.0 & \cellcolor{cineCultural!34!white}89.0 \\
\midrule
\href{https://openrouter.ai/deepseek/deepseek-v4-pro-0813}{DeepSeek V4 Pro 1.6T} & \cellcolor{cineCultural!41!white}\underline{95.5} & \cellcolor{cineCultural!41!white}\underline{94.5} & \cellcolor{cineCultural!43!white}\textbf{94.5} & \cellcolor{cineCultural!42!white}\underline{94.0} & \cellcolor{cineCultural!42!white}\underline{94.5} & \cellcolor{cineCultural!42!white}\underline{94.5} \\
\href{https://huggingface.co/Qwen/Qwen3-235B-A22B-Instruct-2507}{Qwen3 235B} & \cellcolor{cineCultural!31!white}86.5 & \cellcolor{cineCultural!28!white}84.5 & \cellcolor{cineCultural!22!white}81.5 & \cellcolor{cineCultural!25!white}82.0 & \cellcolor{cineCultural!24!white}82.5 & \cellcolor{cineCultural!23!white}81.0 \\
\href{https://huggingface.co/mistralai/Mistral-Small-4-119B-2603}{Mistral 4 119B} & \cellcolor{cineCultural!17!white}75.0 & \cellcolor{cineCultural!19!white}78.0 & \cellcolor{cineCultural!11!white}75.0 & \cellcolor{cineCultural!13!white}73.5 & \cellcolor{cineCultural!7!white}71.0 & \cellcolor{cineCultural!19!white}78.0 \\
\href{https://huggingface.co/meta-llama/Llama-4-Scout-17B-16E-Instruct}{Llama 4 Scout 17B} & \cellcolor{cineCultural!7!white}66.0 & \cellcolor{cineCultural!7!white}69.0 & \cellcolor{cineCultural!7!white}72.5 & \cellcolor{cineCultural!13!white}73.0 & \cellcolor{cineCultural!14!white}76.0 & \cellcolor{cineCultural!7!white}69.0 \\
\bottomrule
\end{tabular}%
\end{adjustbox}
\end{table*}

\begin{table*}[htbp]
\centering
\small
\renewcommand{\arraystretch}{1.13}
\caption{Age mean absolute error by model and subtitle-input language. Values are reported in years; lower is better ($\downarrow$). EN, AR, ID, FA, RO, and VI denote English, Arabic, Indonesian, Persian, Romanian, and Vietnamese, respectively. Within each language column, muted bronze shading is normalized across the displayed models in the preferred direction. Best values are bold, and second-best distinct values are underlined at the displayed precision.}
\vspace{5pt}
\label{tab:cultural-age-mae-years-by-language-appendix}
\begin{adjustbox}{max width=\textwidth}
\begin{tabular}{lrrrrrr}
\toprule
Model & EN & AR & ID & FA & RO & VI \\
\midrule
\href{https://developers.openai.com/api/docs/models/gpt-5.6-sol}{GPT-5.6 Sol} & \cellcolor{cineCultural!42!white}\underline{0.88} & \cellcolor{cineCultural!41!white}\underline{0.99} & \cellcolor{cineCultural!41!white}0.99 & \cellcolor{cineCultural!40!white}\underline{1.08} & \cellcolor{cineCultural!43!white}\underline{0.98} & \cellcolor{cineCultural!40!white}0.97 \\
\href{https://docs.cloud.google.com/gemini-enterprise-agent-platform/models/gemini/3-8-flash}{Gemini 3.8 Flash} & \cellcolor{cineCultural!43!white}\textbf{0.85} & \cellcolor{cineCultural!43!white}\textbf{0.93} & \cellcolor{cineCultural!43!white}\textbf{0.94} & \cellcolor{cineCultural!43!white}\textbf{0.99} & \cellcolor{cineCultural!43!white}\textbf{0.97} & \cellcolor{cineCultural!43!white}\textbf{0.90} \\
\href{https://platform.claude.com/docs/en/models/haiku-4-5/overview}{Claude Haiku 4.5} & \cellcolor{cineCultural!39!white}0.99 & \cellcolor{cineCultural!32!white}1.25 & \cellcolor{cineCultural!37!white}1.08 & \cellcolor{cineCultural!31!white}1.34 & \cellcolor{cineCultural!35!white}1.18 & \cellcolor{cineCultural!36!white}1.09 \\
\midrule
\href{https://developers.openai.com/api/docs/models/gpt-5-nano}{GPT-5 Nano} & \cellcolor{cineCultural!24!white}1.55 & \cellcolor{cineCultural!19!white}1.60 & \cellcolor{cineCultural!17!white}1.54 & \cellcolor{cineCultural!7!white}2.02 & \cellcolor{cineCultural!28!white}1.33 & \cellcolor{cineCultural!28!white}1.33 \\
\href{https://docs.cloud.google.com/gemini-enterprise-agent-platform/models/gemini/3-5-flash-lite}{Gemini 3.5 Flash Lite} & \cellcolor{cineCultural!39!white}0.99 & \cellcolor{cineCultural!36!white}1.14 & \cellcolor{cineCultural!35!white}1.14 & \cellcolor{cineCultural!39!white}1.09 & \cellcolor{cineCultural!37!white}1.12 & \cellcolor{cineCultural!37!white}1.07 \\
\midrule
\href{https://openrouter.ai/deepseek/deepseek-v4-pro-0813}{DeepSeek V4 Pro 1.6T} & \cellcolor{cineCultural!42!white}\underline{0.88} & \cellcolor{cineCultural!40!white}1.02 & \cellcolor{cineCultural!41!white}\underline{0.98} & \cellcolor{cineCultural!40!white}1.09 & \cellcolor{cineCultural!42!white}\underline{0.98} & \cellcolor{cineCultural!41!white}\underline{0.95} \\
\href{https://huggingface.co/Qwen/Qwen3-235B-A22B-Instruct-2507}{Qwen3 235B} & \cellcolor{cineCultural!30!white}1.36 & \cellcolor{cineCultural!23!white}1.48 & \cellcolor{cineCultural!19!white}1.50 & \cellcolor{cineCultural!26!white}1.48 & \cellcolor{cineCultural!22!white}1.49 & \cellcolor{cineCultural!23!white}1.49 \\
\href{https://huggingface.co/mistralai/Mistral-Small-4-119B-2603}{Mistral 4 119B} & \cellcolor{cineCultural!18!white}1.77 & \cellcolor{cineCultural!15!white}1.72 & \cellcolor{cineCultural!9!white}1.75 & \cellcolor{cineCultural!12!white}1.88 & \cellcolor{cineCultural!7!white}1.87 & \cellcolor{cineCultural!16!white}1.70 \\
\href{https://huggingface.co/meta-llama/Llama-4-Scout-17B-16E-Instruct}{Llama 4 Scout 17B} & \cellcolor{cineCultural!7!white}2.21 & \cellcolor{cineCultural!7!white}1.95 & \cellcolor{cineCultural!7!white}1.78 & \cellcolor{cineCultural!13!white}1.83 & \cellcolor{cineCultural!8!white}1.84 & \cellcolor{cineCultural!7!white}1.95 \\
\bottomrule
\end{tabular}%
\end{adjustbox}
\end{table*}

\begin{table*}[htbp]
\centering
\small
\caption{Age-prediction error magnitude and direction across all six subtitle-input languages. \emph{Under} denotes predictions below the gold suitability age, while \emph{over} denotes predictions above it. Exact, one-year, two-year, and larger-error shares partition all predictions; under- and overprediction rates exclude exact matches. Bronze shading and best/second-best emphasis apply only to Exact (higher is better) and $>2$ years (lower is better); the remaining columns are descriptive.}
\vspace{5pt}
\label{tab:age-error-direction-appendix}
\begin{adjustbox}{max width=\textwidth}
\begin{tabular}{lrrrrrr}
\toprule
Model & Exact (\%) & Off by 1 (\%) & Off by 2 (\%) & >2 years (\%) & Under (\%) & Over (\%) \\
\midrule
\href{https://developers.openai.com/api/docs/models/gpt-5.6-sol}{GPT-5.6 Sol} & \cellcolor{cineCultural!36!white}31.1 & 47.7 & 15.8 & \cellcolor{cineCultural!42!white}\underline{5.4} & 20.6 & 48.3 \\
\href{https://docs.cloud.google.com/gemini-enterprise-agent-platform/models/gemini/3-8-flash}{Gemini 3.8 Flash} & \cellcolor{cineCultural!43!white}\textbf{34.2} & 47.2 & 14.1 & \cellcolor{cineCultural!43!white}\textbf{4.5} & 50.6 & 15.2 \\
\href{https://platform.claude.com/docs/en/models/haiku-4-5/overview}{Claude Haiku 4.5} & \cellcolor{cineCultural!27!white}27.4 & 45.3 & 17.7 & \cellcolor{cineCultural!36!white}9.6 & 31.9 & 40.7 \\
\midrule
\href{https://developers.openai.com/api/docs/models/gpt-5-nano}{GPT-5 Nano} & \cellcolor{cineCultural!12!white}20.7 & 40.1 & 19.7 & \cellcolor{cineCultural!21!white}19.6 & 19.1 & 60.2 \\
\href{https://docs.cloud.google.com/gemini-enterprise-agent-platform/models/gemini/3-5-flash-lite}{Gemini 3.5 Flash Lite} & \cellcolor{cineCultural!37!white}\underline{31.8} & 40.3 & 18.8 & \cellcolor{cineCultural!36!white}9.0 & 50.3 & 17.8 \\
\midrule
\href{https://openrouter.ai/deepseek/deepseek-v4-pro-0813}{DeepSeek V4 Pro 1.6T} & \cellcolor{cineCultural!35!white}30.7 & 48.7 & 15.2 & \cellcolor{cineCultural!42!white}\underline{5.4} & 29.8 & 39.6 \\
\href{https://huggingface.co/Qwen/Qwen3-235B-A22B-Instruct-2507}{Qwen3 235B} & \cellcolor{cineCultural!15!white}22.0 & 37.2 & 23.8 & \cellcolor{cineCultural!25!white}17.0 & 55.7 & 22.3 \\
\href{https://huggingface.co/mistralai/Mistral-Small-4-119B-2603}{Mistral 4 119B} & \cellcolor{cineCultural!9!white}19.4 & 35.1 & 20.6 & \cellcolor{cineCultural!13!white}24.9 & 61.5 & 19.1 \\
\href{https://huggingface.co/meta-llama/Llama-4-Scout-17B-16E-Instruct}{Llama 4 Scout 17B} & \cellcolor{cineCultural!7!white}18.6 & 31.7 & 20.7 & \cellcolor{cineCultural!7!white}29.1 & 48.3 & 33.1 \\
\bottomrule
\end{tabular}%
\end{adjustbox}
\end{table*}

\subsection{Country-Level Calibration and Language Sensitivity}
\label{app:cultural-country-breakdown}
\paragraph{Raw accuracy and value above a national majority baseline can disagree.}
France has the lowest mean model exact match (31.6\%) yet a 72.0\% majority baseline; eight of nine models fall below that baseline, and 95.3\% of nonzero model errors assign a stricter label. Gemini 3.8 Flash reaches 73.8\%, only 1.8 points above the baseline. The United States has a lower 49.0\% baseline but a 41.8-point best-model lift. Country-level interpretation therefore requires both raw exact match and baseline-relative lift.

\paragraph{Country-rating performance does not follow access category.}
DeepSeek's 58.0\% country exact match exceeds Claude's 54.3\% overall and is higher in six of ten systems. It nevertheless trails Gemini in every country and Sol by 7.4 points overall. These reversals reinforce that model access category alone does not characterize cultural-prediction performance.

\paragraph{Sol exhibits different error regimes across national systems.}
Sol's nonzero errors are predominantly stricter in France (90.0\%), Germany (95.3\%), and the Netherlands (88.6\%), but its lift over the respective majority baselines is $-21.1$, $+8.7$, and $+12.9$ points. Brazil is less concentrated: its majority baseline is 31.0\%, and Sol improves it by 25.1 points. The same exact-match score can therefore arise from very different calibration regimes.

\paragraph{Subtitle-language effects are heterogeneous across models and countries.}
Relative to English, Romanian subtitles raise country-rating exact match by 3.7 points for Llama 4 Scout and 3.4 points for GPT-5 Nano. Averaged across models, Romanian input also improves Brazil by 2.3 points and Germany by 1.7 points, whereas all five non-English inputs lower United States accuracy. France is comparatively stable across subtitle languages, with mean model accuracy remaining between 30.0\% and 32.5\%. These patterns describe end-to-end language sensitivity rather than a change in the underlying gold rating.

\begin{table}[htbp]
\centering
\scriptsize
\setlength{\tabcolsep}{3.2pt}
\caption{Country-wise motion-picture rating exact-match percentages by model, averaged across all six subtitle-input languages. Each national classification system retains its own ordered label space; higher is better within every column. Within each model column, bronze shading is normalized across countries to highlight relative country-level difficulty and is therefore not comparable across model columns. Best values are bold, and second-best distinct values are underlined at the displayed precision; ties share the same emphasis.}
\vspace{5pt}
\label{tab:cultural-country-breakdown}
\begin{adjustbox}{max width=\textwidth}
\begin{tabular}{lrrrrrrrrr}
\toprule
Country & \href{https://developers.openai.com/api/docs/models/gpt-5.6-sol}{Sol} & \href{https://docs.cloud.google.com/gemini-enterprise-agent-platform/models/gemini/3-8-flash}{Gemini 3.8} & \href{https://platform.claude.com/docs/en/models/haiku-4-5/overview}{Claude Haiku} & \href{https://developers.openai.com/api/docs/models/gpt-5-nano}{GPT-5 Nano} & \href{https://docs.cloud.google.com/gemini-enterprise-agent-platform/models/gemini/3-5-flash-lite}{Gemini 3.5 Lite} & \href{https://openrouter.ai/deepseek/deepseek-v4-pro-0813}{DeepSeek V4 Pro 1.6T} & \href{https://huggingface.co/Qwen/Qwen3-235B-A22B-Instruct-2507}{Qwen3 235B} & \href{https://huggingface.co/mistralai/Mistral-Small-4-119B-2603}{Mistral 4 119B} & \href{https://huggingface.co/meta-llama/Llama-4-Scout-17B-16E-Instruct}{Llama 4 Scout 17B} \\
\midrule
United States & \cellcolor{cineCultural!43!white}\textbf{87.1} & \cellcolor{cineCultural!43!white}\textbf{90.8} & \cellcolor{cineCultural!41!white}\underline{68.2} & \cellcolor{cineCultural!43!white}\textbf{72.0} & \cellcolor{cineCultural!43!white}\textbf{81.9} & \cellcolor{cineCultural!43!white}\textbf{73.9} & \cellcolor{cineCultural!33!white}56.0 & \cellcolor{cineCultural!43!white}\textbf{63.3} & \cellcolor{cineCultural!43!white}\textbf{53.3} \\
United Kingdom & \cellcolor{cineCultural!33!white}\underline{77.0} & \cellcolor{cineCultural!41!white}\underline{89.2} & \cellcolor{cineCultural!35!white}60.5 & \cellcolor{cineCultural!30!white}53.3 & \cellcolor{cineCultural!38!white}76.0 & \cellcolor{cineCultural!38!white}68.7 & \cellcolor{cineCultural!35!white}\underline{57.7} & \cellcolor{cineCultural!37!white}\underline{56.8} & \cellcolor{cineCultural!36!white}44.8 \\
South Korea & \cellcolor{cineCultural!27!white}70.8 & \cellcolor{cineCultural!24!white}77.0 & \cellcolor{cineCultural!37!white}62.3 & \cellcolor{cineCultural!31!white}\underline{54.8} & \cellcolor{cineCultural!27!white}61.7 & \cellcolor{cineCultural!37!white}66.8 & \cellcolor{cineCultural!26!white}45.2 & \cellcolor{cineCultural!35!white}54.1 & \cellcolor{cineCultural!34!white}41.7 \\
Australia & \cellcolor{cineCultural!32!white}75.6 & \cellcolor{cineCultural!41!white}89.1 & \cellcolor{cineCultural!43!white}\textbf{71.6} & \cellcolor{cineCultural!23!white}44.0 & \cellcolor{cineCultural!42!white}\underline{80.8} & \cellcolor{cineCultural!39!white}\underline{69.5} & \cellcolor{cineCultural!43!white}\textbf{69.2} & \cellcolor{cineCultural!36!white}55.7 & \cellcolor{cineCultural!22!white}26.4 \\
Singapore & \cellcolor{cineCultural!16!white}60.2 & \cellcolor{cineCultural!7!white}64.4 & \cellcolor{cineCultural!24!white}44.2 & \cellcolor{cineCultural!20!white}39.1 & \cellcolor{cineCultural!19!white}52.8 & \cellcolor{cineCultural!23!white}51.0 & \cellcolor{cineCultural!19!white}36.4 & \cellcolor{cineCultural!18!white}35.2 & \cellcolor{cineCultural!21!white}24.8 \\
Sweden & \cellcolor{cineCultural!23!white}66.6 & \cellcolor{cineCultural!15!white}70.6 & \cellcolor{cineCultural!36!white}61.0 & \cellcolor{cineCultural!31!white}54.5 & \cellcolor{cineCultural!26!white}61.0 & \cellcolor{cineCultural!30!white}59.3 & \cellcolor{cineCultural!30!white}51.6 & \cellcolor{cineCultural!32!white}50.8 & \cellcolor{cineCultural!41!white}\underline{50.9} \\
Germany & \cellcolor{cineCultural!10!white}54.2 & \cellcolor{cineCultural!30!white}81.4 & \cellcolor{cineCultural!31!white}54.0 & \cellcolor{cineCultural!18!white}36.6 & \cellcolor{cineCultural!34!white}70.2 & \cellcolor{cineCultural!26!white}53.7 & \cellcolor{cineCultural!33!white}55.7 & \cellcolor{cineCultural!31!white}49.2 & \cellcolor{cineCultural!26!white}32.2 \\
Brazil & \cellcolor{cineCultural!12!white}56.1 & \cellcolor{cineCultural!13!white}68.6 & \cellcolor{cineCultural!25!white}46.3 & \cellcolor{cineCultural!13!white}28.8 & \cellcolor{cineCultural!25!white}59.8 & \cellcolor{cineCultural!25!white}52.6 & \cellcolor{cineCultural!26!white}46.4 & \cellcolor{cineCultural!23!white}40.1 & \cellcolor{cineCultural!23!white}28.1 \\
Netherlands & \cellcolor{cineCultural!11!white}55.4 & \cellcolor{cineCultural!20!white}74.0 & \cellcolor{cineCultural!31!white}54.7 & \cellcolor{cineCultural!17!white}35.4 & \cellcolor{cineCultural!29!white}64.8 & \cellcolor{cineCultural!24!white}52.4 & \cellcolor{cineCultural!23!white}41.1 & \cellcolor{cineCultural!28!white}45.9 & \cellcolor{cineCultural!26!white}31.5 \\
France & \cellcolor{cineCultural!7!white}50.9 & \cellcolor{cineCultural!20!white}73.8 & \cellcolor{cineCultural!7!white}20.3 & \cellcolor{cineCultural!7!white}20.5 & \cellcolor{cineCultural!7!white}37.4 & \cellcolor{cineCultural!7!white}32.2 & \cellcolor{cineCultural!7!white}19.6 & \cellcolor{cineCultural!7!white}22.2 & \cellcolor{cineCultural!7!white}7.4 \\
\bottomrule
\end{tabular}%
\end{adjustbox}
\end{table}

\begin{table*}[htbp]
\centering
\small
\caption{Country-level difficulty, improvement over baseline, and ordinal error direction. The majority baseline always predicts the most frequent gold label. Mean model EM averages the nine comparable models, while best EM is achieved by Gemini 3.8 Flash for every country and is averaged equally across the six subtitle-input languages. Lift is the difference between best EM and majority-baseline EM. Ordinal MAE averages model errors within each country's ordered label space. Stricter errors report the percentage of nonzero ordinal errors that assign a rating above the gold rating; exact predictions and mismatches between labels sharing the same ordinal value are excluded.}
\vspace{5pt}
\label{tab:country-majority-baseline-appendix}
\begin{adjustbox}{max width=\textwidth}
\begin{tabular}{llrrrrrr}
\toprule
Country & Majority label & Baseline EM (\%) & Mean model EM (\%) & Best EM (\%) & Lift (pp) & Ordinal MAE & Stricter errors (\%) \\
\midrule
Australia & M & 42.5 & 64.7 & 89.1 & +46.6 & 0.38 & 65.2 \\
Brazil & 14 & 31.0 & 47.4 & 68.6 & +37.6 & 0.86 & 21.2 \\
France & Tous publics & 72.0 & 31.6 & 73.8 & +1.8 & 0.90 & 95.3 \\
Germany & 12 & 45.5 & 54.1 & 81.4 & +35.9 & 0.57 & 85.7 \\
Netherlands & 12 & 42.5 & 50.6 & 74.0 & +31.5 & 0.89 & 70.4 \\
Singapore & M18 & 26.0 & 45.4 & 64.4 & +38.4 & 0.69 & 55.4 \\
South Korea & 15 & 36.0 & 59.4 & 77.0 & +41.0 & 0.44 & 46.2 \\
Sweden & 15 & 40.0 & 58.5 & 70.6 & +30.6 & 0.53 & 64.0 \\
United Kingdom & 15 & 47.0 & 64.9 & 89.2 & +42.2 & 0.38 & 56.5 \\
United States & R & 49.0 & 71.8 & 90.8 & +41.8 & 0.29 & 39.0 \\
\bottomrule
\end{tabular}%
\end{adjustbox}
\end{table*}

\begin{table*}[htbp]
\centering
\small
\renewcommand{\arraystretch}{1.13}
\caption{Country-rating exact match by model and subtitle-input language. Values are percentages; higher is better ($\uparrow$). EN, AR, ID, FA, RO, and VI denote English, Arabic, Indonesian, Persian, Romanian, and Vietnamese, respectively. Within each language column, muted bronze shading is normalized across the displayed models in the preferred direction. Best values are bold, and second-best distinct values are underlined at the displayed precision.}
\vspace{5pt}
\label{tab:cultural-country-exact-match-pct-by-language-appendix}
\begin{adjustbox}{max width=\textwidth}
\begin{tabular}{lrrrrrr}
\toprule
Model & EN & AR & ID & FA & RO & VI \\
\midrule
\href{https://developers.openai.com/api/docs/models/gpt-5.6-sol}{GPT-5.6 Sol} & \cellcolor{cineCultural!34!white}\underline{67.2} & \cellcolor{cineCultural!33!white}\underline{65.8} & \cellcolor{cineCultural!33!white}\underline{65.7} & \cellcolor{cineCultural!32!white}62.4 & \cellcolor{cineCultural!33!white}\underline{65.8} & \cellcolor{cineCultural!33!white}\underline{65.6} \\
\href{https://docs.cloud.google.com/gemini-enterprise-agent-platform/models/gemini/3-8-flash}{Gemini 3.8 Flash} & \cellcolor{cineCultural!43!white}\textbf{78.8} & \cellcolor{cineCultural!43!white}\textbf{78.6} & \cellcolor{cineCultural!43!white}\textbf{77.5} & \cellcolor{cineCultural!43!white}\textbf{76.5} & \cellcolor{cineCultural!43!white}\textbf{77.2} & \cellcolor{cineCultural!43!white}\textbf{78.6} \\
\href{https://platform.claude.com/docs/en/models/haiku-4-5/overview}{Claude Haiku 4.5} & \cellcolor{cineCultural!27!white}58.0 & \cellcolor{cineCultural!22!white}52.6 & \cellcolor{cineCultural!23!white}53.6 & \cellcolor{cineCultural!23!white}51.3 & \cellcolor{cineCultural!23!white}54.6 & \cellcolor{cineCultural!25!white}55.8 \\
\midrule
\href{https://developers.openai.com/api/docs/models/gpt-5-nano}{GPT-5 Nano} & \cellcolor{cineCultural!15!white}43.8 & \cellcolor{cineCultural!13!white}40.9 & \cellcolor{cineCultural!15!white}44.6 & \cellcolor{cineCultural!13!white}39.2 & \cellcolor{cineCultural!16!white}47.2 & \cellcolor{cineCultural!18!white}47.5 \\
\href{https://docs.cloud.google.com/gemini-enterprise-agent-platform/models/gemini/3-5-flash-lite}{Gemini 3.5 Flash Lite} & \cellcolor{cineCultural!33!white}66.3 & \cellcolor{cineCultural!31!white}63.5 & \cellcolor{cineCultural!32!white}64.4 & \cellcolor{cineCultural!32!white}\underline{63.2} & \cellcolor{cineCultural!33!white}65.5 & \cellcolor{cineCultural!32!white}65.0 \\
\midrule
\href{https://openrouter.ai/deepseek/deepseek-v4-pro-0813}{DeepSeek V4 Pro 1.6T} & \cellcolor{cineCultural!27!white}58.5 & \cellcolor{cineCultural!26!white}57.6 & \cellcolor{cineCultural!28!white}59.6 & \cellcolor{cineCultural!25!white}54.4 & \cellcolor{cineCultural!27!white}59.7 & \cellcolor{cineCultural!27!white}58.4 \\
\href{https://huggingface.co/Qwen/Qwen3-235B-A22B-Instruct-2507}{Qwen3 235B} & \cellcolor{cineCultural!22!white}51.7 & \cellcolor{cineCultural!17!white}46.9 & \cellcolor{cineCultural!18!white}48.4 & \cellcolor{cineCultural!18!white}44.9 & \cellcolor{cineCultural!16!white}47.4 & \cellcolor{cineCultural!18!white}48.0 \\
\href{https://huggingface.co/mistralai/Mistral-Small-4-119B-2603}{Mistral 4 119B} & \cellcolor{cineCultural!20!white}49.4 & \cellcolor{cineCultural!16!white}45.1 & \cellcolor{cineCultural!18!white}47.9 & \cellcolor{cineCultural!18!white}44.9 & \cellcolor{cineCultural!16!white}47.1 & \cellcolor{cineCultural!20!white}49.8 \\
\href{https://huggingface.co/meta-llama/Llama-4-Scout-17B-16E-Instruct}{Llama 4 Scout 17B} & \cellcolor{cineCultural!7!white}33.1 & \cellcolor{cineCultural!7!white}33.9 & \cellcolor{cineCultural!7!white}35.1 & \cellcolor{cineCultural!7!white}31.8 & \cellcolor{cineCultural!7!white}36.8 & \cellcolor{cineCultural!7!white}34.0 \\
\bottomrule
\end{tabular}%
\end{adjustbox}
\end{table*}

\begin{table*}[htbp]
\centering
\small
\renewcommand{\arraystretch}{1.13}
\caption{Country-rating ordinal error by model and subtitle-input language. Values are measured in rating steps; lower is better ($\downarrow$). EN, AR, ID, FA, RO, and VI denote English, Arabic, Indonesian, Persian, Romanian, and Vietnamese, respectively. Within each language column, muted bronze shading is normalized across the displayed models in the preferred direction. Best values are bold, and second-best distinct values are underlined at the displayed precision.}
\vspace{5pt}
\label{tab:cultural-country-ordinal-mae-by-language-appendix}
\begin{adjustbox}{max width=\textwidth}
\begin{tabular}{lrrrrrr}
\toprule
Model & EN & AR & ID & FA & RO & VI \\
\midrule
\href{https://developers.openai.com/api/docs/models/gpt-5.6-sol}{GPT-5.6 Sol} & \cellcolor{cineCultural!36!white}\underline{0.42} & \cellcolor{cineCultural!35!white}\underline{0.44} & \cellcolor{cineCultural!36!white}\underline{0.43} & \cellcolor{cineCultural!35!white}\underline{0.48} & \cellcolor{cineCultural!35!white}\underline{0.45} & \cellcolor{cineCultural!36!white}\underline{0.43} \\
\href{https://docs.cloud.google.com/gemini-enterprise-agent-platform/models/gemini/3-8-flash}{Gemini 3.8 Flash} & \cellcolor{cineCultural!43!white}\textbf{0.29} & \cellcolor{cineCultural!43!white}\textbf{0.30} & \cellcolor{cineCultural!43!white}\textbf{0.31} & \cellcolor{cineCultural!43!white}\textbf{0.33} & \cellcolor{cineCultural!43!white}\textbf{0.32} & \cellcolor{cineCultural!43!white}\textbf{0.30} \\
\href{https://platform.claude.com/docs/en/models/haiku-4-5/overview}{Claude Haiku 4.5} & \cellcolor{cineCultural!29!white}0.53 & \cellcolor{cineCultural!26!white}0.59 & \cellcolor{cineCultural!27!white}0.58 & \cellcolor{cineCultural!27!white}0.61 & \cellcolor{cineCultural!26!white}0.58 & \cellcolor{cineCultural!30!white}0.55 \\
\midrule
\href{https://developers.openai.com/api/docs/models/gpt-5-nano}{GPT-5 Nano} & \cellcolor{cineCultural!17!white}0.75 & \cellcolor{cineCultural!14!white}0.80 & \cellcolor{cineCultural!16!white}0.75 & \cellcolor{cineCultural!12!white}0.88 & \cellcolor{cineCultural!18!white}0.71 & \cellcolor{cineCultural!22!white}0.69 \\
\href{https://docs.cloud.google.com/gemini-enterprise-agent-platform/models/gemini/3-5-flash-lite}{Gemini 3.5 Flash Lite} & \cellcolor{cineCultural!34!white}0.44 & \cellcolor{cineCultural!33!white}0.48 & \cellcolor{cineCultural!34!white}0.47 & \cellcolor{cineCultural!35!white}\underline{0.48} & \cellcolor{cineCultural!34!white}0.46 & \cellcolor{cineCultural!35!white}0.46 \\
\midrule
\href{https://openrouter.ai/deepseek/deepseek-v4-pro-0813}{DeepSeek V4 Pro 1.6T} & \cellcolor{cineCultural!31!white}0.51 & \cellcolor{cineCultural!30!white}0.53 & \cellcolor{cineCultural!31!white}0.50 & \cellcolor{cineCultural!29!white}0.58 & \cellcolor{cineCultural!31!white}0.51 & \cellcolor{cineCultural!32!white}0.51 \\
\href{https://huggingface.co/Qwen/Qwen3-235B-A22B-Instruct-2507}{Qwen3 235B} & \cellcolor{cineCultural!25!white}0.60 & \cellcolor{cineCultural!23!white}0.64 & \cellcolor{cineCultural!23!white}0.64 & \cellcolor{cineCultural!23!white}0.68 & \cellcolor{cineCultural!22!white}0.66 & \cellcolor{cineCultural!25!white}0.64 \\
\href{https://huggingface.co/mistralai/Mistral-Small-4-119B-2603}{Mistral 4 119B} & \cellcolor{cineCultural!22!white}0.65 & \cellcolor{cineCultural!19!white}0.72 & \cellcolor{cineCultural!21!white}0.66 & \cellcolor{cineCultural!21!white}0.73 & \cellcolor{cineCultural!20!white}0.69 & \cellcolor{cineCultural!25!white}0.64 \\
\href{https://huggingface.co/meta-llama/Llama-4-Scout-17B-16E-Instruct}{Llama 4 Scout 17B} & \cellcolor{cineCultural!7!white}0.92 & \cellcolor{cineCultural!7!white}0.92 & \cellcolor{cineCultural!7!white}0.89 & \cellcolor{cineCultural!7!white}0.97 & \cellcolor{cineCultural!7!white}0.89 & \cellcolor{cineCultural!7!white}0.97 \\
\bottomrule
\end{tabular}%
\end{adjustbox}
\end{table*}

\begin{table}[htbp]
\centering
\small
\caption{Country-rating exact-match percentage by national classification system and subtitle-input language, averaged equally across the compared models. Columns correspond to English, Arabic, Indonesian, Persian, Romanian, and Vietnamese; higher is better ($\uparrow$).}
\vspace{5pt}
\label{tab:country-language-appendix}
\begin{adjustbox}{max width=\textwidth}
\begin{tabular}{lrrrrrr}
\toprule
Country & English & Arabic & Indonesian & Persian & Romanian & Vietnamese \\
\midrule
Australia & 65.6 & 63.4 & 65.4 & 62.5 & 65.5 & 65.5 \\
Brazil & 47.7 & 45.1 & 48.4 & 43.9 & 49.9 & 49.4 \\
France & 31.6 & 31.2 & 32.5 & 29.9 & 32.5 & 31.9 \\
Germany & 54.2 & 53.4 & 56.3 & 50.6 & 55.9 & 54.5 \\
Netherlands & 50.8 & 50.1 & 51.5 & 46.6 & 52.6 & 51.9 \\
Singapore & 47.5 & 45.1 & 45.7 & 42.8 & 45.6 & 45.5 \\
South Korea & 61.9 & 59.6 & 58.3 & 57.6 & 59.6 & 59.3 \\
Sweden & 60.2 & 58.9 & 57.5 & 56.9 & 57.6 & 59.7 \\
United Kingdom & 67.9 & 62.0 & 64.8 & 61.3 & 65.9 & 67.3 \\
United States & 75.6 & 69.9 & 71.8 & 68.6 & 71.8 & 73.4 \\
\bottomrule
\end{tabular}%
\end{adjustbox}
\end{table}

\begin{figure}[htbp]
\centering
\includegraphics[width=1.0\linewidth]{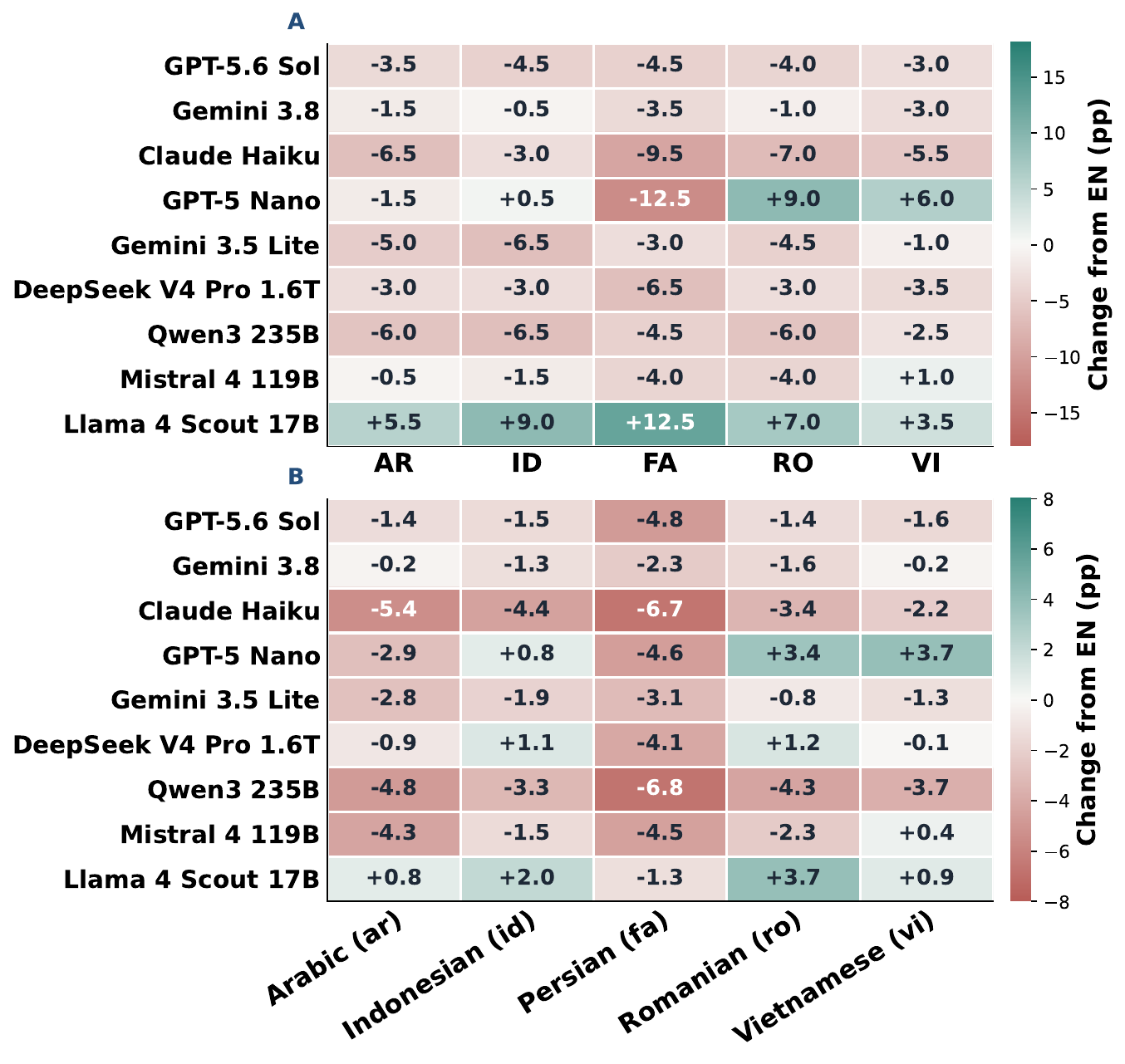}
\caption{Change in cultural prediction relative to English subtitle input. The upper panel reports age accuracy within one year and the lower panel reports country-rating exact match, in percentage points, for each model under Arabic, Indonesian, Persian, Romanian, and Vietnamese subtitles. Absolute language-specific values are reported in the accompanying tables.}
\label{fig:cultural-crosslingual-dashboard}
\end{figure}

\begin{table}[H]
\centering
\small
\renewcommand{\arraystretch}{1.20}
\caption{Country-specific changes in exact-match accuracy under Arabic, Indonesian, Persian, Romanian, and Vietnamese subtitle inputs relative to English. Values are percentage-point differences averaged equally across models. Absolute country-by-language exact-match scores are reported in Table~\ref{tab:country-language-appendix}.}
\vspace{5pt}
\label{tab:country-language-changes-main}
\begin{adjustbox}{max width=\textwidth}
\begin{tabular}{lrrrrr}
\toprule
Country & Arabic $\Delta$ & Indonesian $\Delta$ & Persian $\Delta$ & Romanian $\Delta$ & Vietnamese $\Delta$ \\
\midrule
Australia & \cellcolor{cineDeltaNegative!13!white}\textcolor{cineDeltaNegative}{-2.2} & \cellcolor{cineDeltaNegative!7!white}\textcolor{cineDeltaNegative}{-0.2} & \cellcolor{cineDeltaNegative!16!white}\textcolor{cineDeltaNegative}{-3.1} & \cellcolor{cineDeltaNegative!6!white}\textcolor{cineDeltaNegative}{-0.1} & \cellcolor{cineDeltaNegative!6!white}\textcolor{cineDeltaNegative}{-0.1} \\
Brazil & \cellcolor{cineDeltaNegative!15!white}\textcolor{cineDeltaNegative}{-2.6} & \cellcolor{cineDeltaPositive!8!white}\textcolor{cineDeltaPositive}{+0.7} & \cellcolor{cineDeltaNegative!18!white}\textcolor{cineDeltaNegative}{-3.7} & \cellcolor{cineDeltaPositive!13!white}\textcolor{cineDeltaPositive}{+2.3} & \cellcolor{cineDeltaPositive!12!white}\textcolor{cineDeltaPositive}{+1.8} \\
France & \cellcolor{cineDeltaNegative!7!white}\textcolor{cineDeltaNegative}{-0.4} & \cellcolor{cineDeltaPositive!9!white}\textcolor{cineDeltaPositive}{+0.9} & \cellcolor{cineDeltaNegative!11!white}\textcolor{cineDeltaNegative}{-1.7} & \cellcolor{cineDeltaPositive!9!white}\textcolor{cineDeltaPositive}{+0.9} & \cellcolor{cineDeltaPositive!7!white}\textcolor{cineDeltaPositive}{+0.3} \\
Germany & \cellcolor{cineDeltaNegative!8!white}\textcolor{cineDeltaNegative}{-0.7} & \cellcolor{cineDeltaPositive!13!white}\textcolor{cineDeltaPositive}{+2.1} & \cellcolor{cineDeltaNegative!18!white}\textcolor{cineDeltaNegative}{-3.6} & \cellcolor{cineDeltaPositive!12!white}\textcolor{cineDeltaPositive}{+1.7} & \cellcolor{cineDeltaPositive!7!white}\textcolor{cineDeltaPositive}{+0.3} \\
Netherlands & \cellcolor{cineDeltaNegative!8!white}\textcolor{cineDeltaNegative}{-0.7} & \cellcolor{cineDeltaPositive!8!white}\textcolor{cineDeltaPositive}{+0.7} & \cellcolor{cineDeltaNegative!20!white}\textcolor{cineDeltaNegative}{-4.3} & \cellcolor{cineDeltaPositive!12!white}\textcolor{cineDeltaPositive}{+1.7} & \cellcolor{cineDeltaPositive!10!white}\textcolor{cineDeltaPositive}{+1.1} \\
Singapore & \cellcolor{cineDeltaNegative!14!white}\textcolor{cineDeltaNegative}{-2.4} & \cellcolor{cineDeltaNegative!12!white}\textcolor{cineDeltaNegative}{-1.8} & \cellcolor{cineDeltaNegative!21!white}\textcolor{cineDeltaNegative}{-4.7} & \cellcolor{cineDeltaNegative!12!white}\textcolor{cineDeltaNegative}{-1.9} & \cellcolor{cineDeltaNegative!13!white}\textcolor{cineDeltaNegative}{-2.0} \\
South Korea & \cellcolor{cineDeltaNegative!13!white}\textcolor{cineDeltaNegative}{-2.3} & \cellcolor{cineDeltaNegative!18!white}\textcolor{cineDeltaNegative}{-3.6} & \cellcolor{cineDeltaNegative!20!white}\textcolor{cineDeltaNegative}{-4.3} & \cellcolor{cineDeltaNegative!14!white}\textcolor{cineDeltaNegative}{-2.3} & \cellcolor{cineDeltaNegative!15!white}\textcolor{cineDeltaNegative}{-2.6} \\
Sweden & \cellcolor{cineDeltaNegative!10!white}\textcolor{cineDeltaNegative}{-1.2} & \cellcolor{cineDeltaNegative!15!white}\textcolor{cineDeltaNegative}{-2.7} & \cellcolor{cineDeltaNegative!17!white}\textcolor{cineDeltaNegative}{-3.2} & \cellcolor{cineDeltaNegative!14!white}\textcolor{cineDeltaNegative}{-2.6} & \cellcolor{cineDeltaNegative!7!white}\textcolor{cineDeltaNegative}{-0.4} \\
United Kingdom & \cellcolor{cineDeltaNegative!25!white}\textcolor{cineDeltaNegative}{-5.9} & \cellcolor{cineDeltaNegative!16!white}\textcolor{cineDeltaNegative}{-3.1} & \cellcolor{cineDeltaNegative!28!white}\textcolor{cineDeltaNegative}{-6.6} & \cellcolor{cineDeltaNegative!12!white}\textcolor{cineDeltaNegative}{-1.9} & \cellcolor{cineDeltaNegative!8!white}\textcolor{cineDeltaNegative}{-0.6} \\
United States & \cellcolor{cineDeltaNegative!25!white}\textcolor{cineDeltaNegative}{-5.7} & \cellcolor{cineDeltaNegative!18!white}\textcolor{cineDeltaNegative}{-3.8} & \cellcolor{cineDeltaNegative!29!white}\textcolor{cineDeltaNegative}{-7.0} & \cellcolor{cineDeltaNegative!18!white}\textcolor{cineDeltaNegative}{-3.7} & \cellcolor{cineDeltaNegative!13!white}\textcolor{cineDeltaNegative}{-2.1} \\
\bottomrule
\end{tabular}%
\end{adjustbox}
\end{table}

\section{Additional Evidence for RQ~\ref{rq:evidence-grounding}: Auditable Language-Safety Assessment}
\label{app:rq4-details}
\subsection{Evidence Errors by Category and Model}
\label{app:language-content-error-analysis}
\paragraph{Safety evidence reveals distinct precision--recall trade-offs.}
Sol has the highest category-agnostic recall (96.3\%) but lower precision (70.9\%) than Gemini (83.0\%); DeepSeek reverses the pattern with 77.8\% precision and 45.6\% recall. Gemini is more balanced at 83.0\% precision and 90.9\% recall, yielding the highest agnostic F1 of 86.8\%. These profiles distinguish higher false-positive tendencies from failures to recover relevant evidence.

\paragraph{Category assignment introduces an additional source of error.}
Gemini's category-agnostic F1 of 86.8\% falls to 77.8\% under strict matching, while Sol declines from 81.7\% to 74.5\%. The gap captures cases where a relevant subtitle line is recovered but assigned to the wrong lexical category. Count error provides an additional diagnostic: Gemini has the lowest count MAE among the evaluated models at 5.01 per film and lexical category on average.

\paragraph{Mild obscenity is difficult in both directions.}
Across models, mild obscenity has 26.3\% recall, a 73.7\% miss rate, and an 81.8\% false-positive rate, yielding 20.3\% category-level F1. Strong profanity is more lexically explicit, with 70.0\% precision, 62.6\% recall, and 62.8\% F1. The model-by-category heat map shows how strongly individual systems depart from these averages.
\begin{table}[htbp]
\centering
\small
\setlength{\tabcolsep}{4pt}
\caption{Evidence-grounded language-safety performance by model. Category-agnostic metrics measure recovery of safety-relevant subtitle indices regardless of lexical category, while strict F1 additionally requires the correct category assignment. Count MAE compares predicted and gold category counts and is macro-averaged across the four lexical categories. Best values are bold, and second-best distinct values are underlined.}
\vspace{5pt}
\label{tab:language-content-results-appendix}
\begin{adjustbox}{max width=\textwidth}
\begin{tabular}{lrrrrr}
\toprule
Model & Agnostic P (\%) $\uparrow$ & Agnostic R (\%) $\uparrow$ & Agnostic F1 (\%) $\uparrow$ & Strict F1 (\%) $\uparrow$ & Count MAE $\downarrow$ \\
\midrule
\href{https://developers.openai.com/api/docs/models/gpt-5.6-sol}{GPT-5.6 Sol} & \cellcolor{cineTeal!36!white}70.9 & \cellcolor{cineTeal!43!white}\textbf{96.3} & \cellcolor{cineTeal!40!white}\underline{81.7} & \cellcolor{cineTeal!41!white}\underline{74.5} & \cellcolor{cineTeal!31!white}\underline{7.16} \\
\href{https://docs.cloud.google.com/gemini-enterprise-agent-platform/models/gemini/3-8-flash}{Gemini 3.8 Flash} & \cellcolor{cineTeal!43!white}\textbf{83.0} & \cellcolor{cineTeal!40!white}\underline{90.9} & \cellcolor{cineTeal!43!white}\textbf{86.8} & \cellcolor{cineTeal!43!white}\textbf{77.8} & \cellcolor{cineTeal!43!white}\textbf{5.01} \\
\href{https://platform.claude.com/docs/en/models/haiku-4-5/overview}{Claude Haiku 4.5} & \cellcolor{cineTeal!32!white}61.6 & \cellcolor{cineTeal!20!white}55.0 & \cellcolor{cineTeal!27!white}58.1 & \cellcolor{cineTeal!22!white}37.2 & \cellcolor{cineTeal!21!white}8.95 \\
\midrule
\href{https://developers.openai.com/api/docs/models/gpt-5-nano}{GPT-5 Nano} & \cellcolor{cineTeal!7!white}16.0 & \cellcolor{cineTeal!8!white}33.5 & \cellcolor{cineTeal!7!white}21.7 & \cellcolor{cineTeal!7!white}9.6 & \cellcolor{cineTeal!7!white}11.55 \\
\href{https://docs.cloud.google.com/gemini-enterprise-agent-platform/models/gemini/3-5-flash-lite}{Gemini 3.5 Flash Lite} & \cellcolor{cineTeal!35!white}68.0 & \cellcolor{cineTeal!24!white}62.8 & \cellcolor{cineTeal!31!white}65.3 & \cellcolor{cineTeal!25!white}42.8 & \cellcolor{cineTeal!19!white}9.42 \\
\midrule
\href{https://openrouter.ai/deepseek/deepseek-v4-pro-0813}{DeepSeek V4 Pro 1.6T} & \cellcolor{cineTeal!40!white}\underline{77.8} & \cellcolor{cineTeal!15!white}45.6 & \cellcolor{cineTeal!27!white}57.5 & \cellcolor{cineTeal!27!white}48.3 & \cellcolor{cineTeal!20!white}9.18 \\
\href{https://huggingface.co/Qwen/Qwen3-235B-A22B-Instruct-2507}{Qwen3 235B} & \cellcolor{cineTeal!35!white}68.8 & \cellcolor{cineTeal!8!white}33.4 & \cellcolor{cineTeal!20!white}45.0 & \cellcolor{cineTeal!18!white}29.6 & \cellcolor{cineTeal!14!white}10.27 \\
\href{https://huggingface.co/mistralai/Mistral-Small-4-119B-2603}{Mistral 4 119B} & \cellcolor{cineTeal!30!white}59.7 & \cellcolor{cineTeal!13!white}42.4 & \cellcolor{cineTeal!22!white}49.6 & \cellcolor{cineTeal!19!white}31.7 & \cellcolor{cineTeal!28!white}7.78 \\
\href{https://huggingface.co/meta-llama/Llama-4-Scout-17B-16E-Instruct}{Llama 4 Scout 17B} & \cellcolor{cineTeal!12!white}25.4 & \cellcolor{cineTeal!7!white}32.1 & \cellcolor{cineTeal!11!white}28.4 & \cellcolor{cineTeal!9!white}12.7 & \cellcolor{cineTeal!9!white}11.11 \\
\bottomrule
\end{tabular}%
\end{adjustbox}
\end{table}

\begin{table}[htbp]
\centering
\small
\setlength{\tabcolsep}{5pt}
\caption{Language-safety error profiles by lexical category, averaged across the nine reported English model runs. Arrows indicate the preferred direction: $\uparrow$ is higher-is-better and $\downarrow$ is lower-is-better. Miss rate measures gold safety-relevant evidence that was not recovered, false-positive rate measures unsupported predicted evidence, and cross-category FP measures evidence assigned to an incorrect lexical category. Teal shading is normalized within each metric column across categories. Best values are bold, and second-best distinct values are underlined at the displayed precision; ties share the same emphasis.}
\vspace{5pt}
\label{tab:language-content-error-analysis}
\begin{adjustbox}{max width=\textwidth}
\begin{tabular}{lrrrrrrr}
\toprule
Category & Precision (\%) $\uparrow$ & Recall (\%) $\uparrow$ & F1 (\%) $\uparrow$ & Miss rate (\%) $\downarrow$ & False-positive rate (\%) $\downarrow$ & Cross-category FP (\%) $\downarrow$ & Count MAE $\downarrow$ \\
\midrule
Mild Obscenity & \cellcolor{cineTeal!7!white}18.2 & \cellcolor{cineTeal!7!white}26.3 & \cellcolor{cineTeal!7!white}20.3 & \cellcolor{cineTeal!7!white}73.7 & \cellcolor{cineTeal!7!white}81.8 & \cellcolor{cineTeal!10!white}32.6 & \cellcolor{cineTeal!36!white}8.43 \\
Crude Bodily Language & \cellcolor{cineTeal!32!white}\underline{54.5} & \cellcolor{cineTeal!16!white}35.0 & \cellcolor{cineTeal!24!white}\underline{40.5} & \cellcolor{cineTeal!16!white}65.0 & \cellcolor{cineTeal!32!white}\underline{45.5} & \cellcolor{cineTeal!29!white}\underline{22.1} & \cellcolor{cineTeal!7!white}11.85 \\
Religious profanity/exclamation & \cellcolor{cineTeal!22!white}39.8 & \cellcolor{cineTeal!21!white}\underline{40.2} & \cellcolor{cineTeal!22!white}38.3 & \cellcolor{cineTeal!21!white}\underline{59.8} & \cellcolor{cineTeal!22!white}60.2 & \cellcolor{cineTeal!43!white}\textbf{14.8} & \cellcolor{cineTeal!43!white}\textbf{7.56} \\
Strong Profanity & \cellcolor{cineTeal!43!white}\textbf{70.0} & \cellcolor{cineTeal!43!white}\textbf{62.6} & \cellcolor{cineTeal!43!white}\textbf{62.8} & \cellcolor{cineTeal!43!white}\textbf{37.4} & \cellcolor{cineTeal!43!white}\textbf{30.0} & \cellcolor{cineTeal!7!white}34.0 & \cellcolor{cineTeal!40!white}\underline{7.91} \\
\bottomrule
\end{tabular}%
\end{adjustbox}
\end{table}

\begin{table*}[htbp]
\centering
\small
\caption{English language-safety evidence performance by model and lexical category. Strict F1 requires both the correct subtitle index and lexical category, while recall measures recovery of gold evidence indices within each category. Strong, crude, mild, and religious abbreviate the four language-safety categories defined in the main text. Teal shading compares models within each category and metric. Best values are bold, and second-best distinct values are underlined at the displayed precision.}
\vspace{5pt}
\label{tab:lc-model-category-appendix}
\begin{adjustbox}{max width=\textwidth}
\begin{tabular}{lrrrrrrrr}
\toprule
 & \multicolumn{4}{c}{Strict evidence F1 (\%)} & \multicolumn{4}{c}{Evidence recall (\%)} \\
\cmidrule(lr){2-5}\cmidrule(lr){6-9}
Model & Strong & Crude & Mild & Religious & Strong & Crude & Mild & Religious \\
\midrule
\href{https://developers.openai.com/api/docs/models/gpt-5.6-sol}{GPT-5.6 Sol} & \cellcolor{cineTeal!43!white}\textbf{99.6} & \cellcolor{cineTeal!43!white}\underline{85.6} & \cellcolor{cineTeal!34!white}\underline{39.7} & \cellcolor{cineTeal!43!white}\underline{73.3} & \cellcolor{cineTeal!43!white}\textbf{100.0} & \cellcolor{cineTeal!43!white}\textbf{86.6} & \cellcolor{cineTeal!43!white}\textbf{72.1} & \cellcolor{cineTeal!43!white}\textbf{94.4} \\
\href{https://docs.cloud.google.com/gemini-enterprise-agent-platform/models/gemini/3-8-flash}{Gemini 3.8 Flash} & \cellcolor{cineTeal!43!white}\underline{99.1} & \cellcolor{cineTeal!43!white}\textbf{86.0} & \cellcolor{cineTeal!43!white}\textbf{52.0} & \cellcolor{cineTeal!43!white}\textbf{74.2} & \cellcolor{cineTeal!43!white}\underline{99.2} & \cellcolor{cineTeal!41!white}\underline{82.3} & \cellcolor{cineTeal!36!white}\underline{59.1} & \cellcolor{cineTeal!40!white}\underline{87.8} \\
\href{https://platform.claude.com/docs/en/models/haiku-4-5/overview}{Claude Haiku 4.5} & \cellcolor{cineTeal!28!white}68.5 & \cellcolor{cineTeal!18!white}33.5 & \cellcolor{cineTeal!18!white}17.7 & \cellcolor{cineTeal!20!white}29.1 & \cellcolor{cineTeal!28!white}72.2 & \cellcolor{cineTeal!14!white}23.2 & \cellcolor{cineTeal!19!white}24.9 & \cellcolor{cineTeal!14!white}22.4 \\
\midrule
\href{https://developers.openai.com/api/docs/models/gpt-5-nano}{GPT-5 Nano} & \cellcolor{cineTeal!7!white}22.4 & \cellcolor{cineTeal!7!white}9.4 & \cellcolor{cineTeal!7!white}2.0 & \cellcolor{cineTeal!7!white}4.8 & \cellcolor{cineTeal!7!white}34.9 & \cellcolor{cineTeal!9!white}12.3 & \cellcolor{cineTeal!10!white}8.4 & \cellcolor{cineTeal!7!white}4.1 \\
\href{https://docs.cloud.google.com/gemini-enterprise-agent-platform/models/gemini/3-5-flash-lite}{Gemini 3.5 Flash Lite} & \cellcolor{cineTeal!25!white}61.2 & \cellcolor{cineTeal!21!white}40.2 & \cellcolor{cineTeal!21!white}22.0 & \cellcolor{cineTeal!29!white}47.9 & \cellcolor{cineTeal!13!white}45.9 & \cellcolor{cineTeal!19!white}34.1 & \cellcolor{cineTeal!20!white}28.5 & \cellcolor{cineTeal!28!white}56.7 \\
\midrule
\href{https://openrouter.ai/deepseek/deepseek-v4-pro-0813}{DeepSeek V4 Pro 1.6T} & \cellcolor{cineTeal!28!white}68.2 & \cellcolor{cineTeal!25!white}48.4 & \cellcolor{cineTeal!22!white}22.6 & \cellcolor{cineTeal!33!white}54.1 & \cellcolor{cineTeal!17!white}52.3 & \cellcolor{cineTeal!18!white}33.3 & \cellcolor{cineTeal!16!white}20.5 & \cellcolor{cineTeal!25!white}48.4 \\
\href{https://huggingface.co/Qwen/Qwen3-235B-A22B-Instruct-2507}{Qwen3 235B} & \cellcolor{cineTeal!24!white}58.1 & \cellcolor{cineTeal!13!white}22.9 & \cellcolor{cineTeal!16!white}14.2 & \cellcolor{cineTeal!17!white}23.3 & \cellcolor{cineTeal!13!white}45.5 & \cellcolor{cineTeal!10!white}14.9 & \cellcolor{cineTeal!12!white}12.3 & \cellcolor{cineTeal!13!white}17.9 \\
\href{https://huggingface.co/mistralai/Mistral-Small-4-119B-2603}{Mistral 4 119B} & \cellcolor{cineTeal!26!white}62.4 & \cellcolor{cineTeal!15!white}26.5 & \cellcolor{cineTeal!12!white}8.9 & \cellcolor{cineTeal!20!white}28.9 & \cellcolor{cineTeal!22!white}62.1 & \cellcolor{cineTeal!12!white}19.6 & \cellcolor{cineTeal!10!white}8.1 & \cellcolor{cineTeal!15!white}23.2 \\
\href{https://huggingface.co/meta-llama/Llama-4-Scout-17B-16E-Instruct}{Llama 4 Scout 17B} & \cellcolor{cineTeal!9!white}25.7 & \cellcolor{cineTeal!8!white}12.5 & \cellcolor{cineTeal!8!white}3.5 & \cellcolor{cineTeal!9!white}9.2 & \cellcolor{cineTeal!16!white}51.6 & \cellcolor{cineTeal!7!white}8.9 & \cellcolor{cineTeal!7!white}2.6 & \cellcolor{cineTeal!8!white}7.2 \\
\bottomrule
\end{tabular}%
\end{adjustbox}
\end{table*}

\begin{figure}[htbp]
\centering
\includegraphics[width=0.92\linewidth]{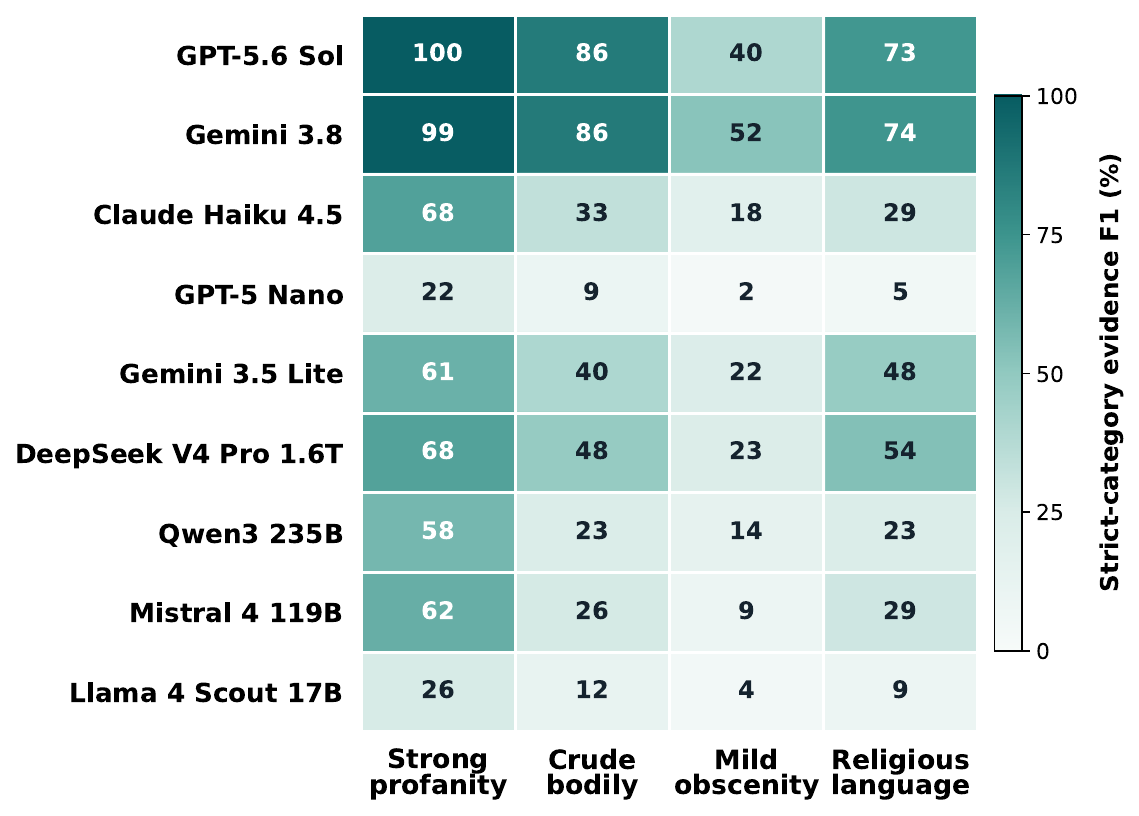}
\caption{Strict-category evidence F1 by model and lexical category. Darker teal indicates higher F1. Strong profanity is easiest overall, while mild obscenity combines low recall with frequent false positives.}
\label{fig:language-content-category-f1}
\end{figure}

\section{Additional Analysis: Within-Family Scaling}
\label{app:scaling-analysis}
\paragraph{Within-family scaling is task dependent.}
Holding model family and English input constant, Ministral narrative overall rises from 2.25 to 2.55 to 2.78 as model size increases from 3B to 8B to 14B, while age MAE falls monotonically from 2.42 to 1.92 to 1.41 years. Classification results are not monotonic: genre micro-F1 increases from 56.7\% to 66.7\% and then falls to 64.8\%, while country-rating exact match moves from 22.5\% to 36.6\% to 35.6\%. Within this family, additional scale therefore benefits narrative reconstruction and age prediction more consistently than genre or national-rating classification.
\begin{table}[htbp]
\centering
\small
\setlength{\tabcolsep}{4pt}
\caption{English-only within-family scaling results for Ministral 3B, 8B, and 14B. Narrative performance improves and age MAE decreases monotonically with model size, while genre micro-F1 and country-rating exact match peak at 8B. Best values are bold, and second-best distinct values are underlined.}
\vspace{5pt}
\label{tab:ministral-scaling-appendix}
\begin{adjustbox}{max width=\textwidth}
\begin{tabular}{lrrrr}
\toprule
Model & Narrative $\uparrow$ & \shortstack{Genre\\F1 (\%) $\uparrow$} & \shortstack{Age\\MAE $\downarrow$} & \shortstack{Country\\EM (\%) $\uparrow$} \\
\midrule
\href{https://huggingface.co/mistralai/Ministral-3-3B-Instruct-2512}{Ministral 3B} & \cellcolor{cineNarrative!7!white}2.25 & \cellcolor{cineNarrative!7!white}56.7 & \cellcolor{cineCultural!7!white}2.42 & \cellcolor{cineCultural!7!white}22.5 \\
\href{https://huggingface.co/mistralai/Ministral-3-8B-Instruct-2512}{Ministral 8B} & \cellcolor{cineNarrative!27!white}\underline{2.55} & \cellcolor{cineNarrative!43!white}\textbf{66.7} & \cellcolor{cineCultural!25!white}\underline{1.92} & \cellcolor{cineCultural!43!white}\textbf{36.6} \\
\href{https://huggingface.co/mistralai/Ministral-3-14B-Instruct-2512}{Ministral 14B} & \cellcolor{cineNarrative!43!white}\textbf{2.78} & \cellcolor{cineNarrative!36!white}\underline{64.8} & \cellcolor{cineCultural!43!white}\textbf{1.41} & \cellcolor{cineCultural!40!white}\underline{35.6} \\
\bottomrule
\end{tabular}%
\end{adjustbox}
\end{table}

\begin{figure}[htbp]
\centering
\includegraphics[width=0.92\linewidth]{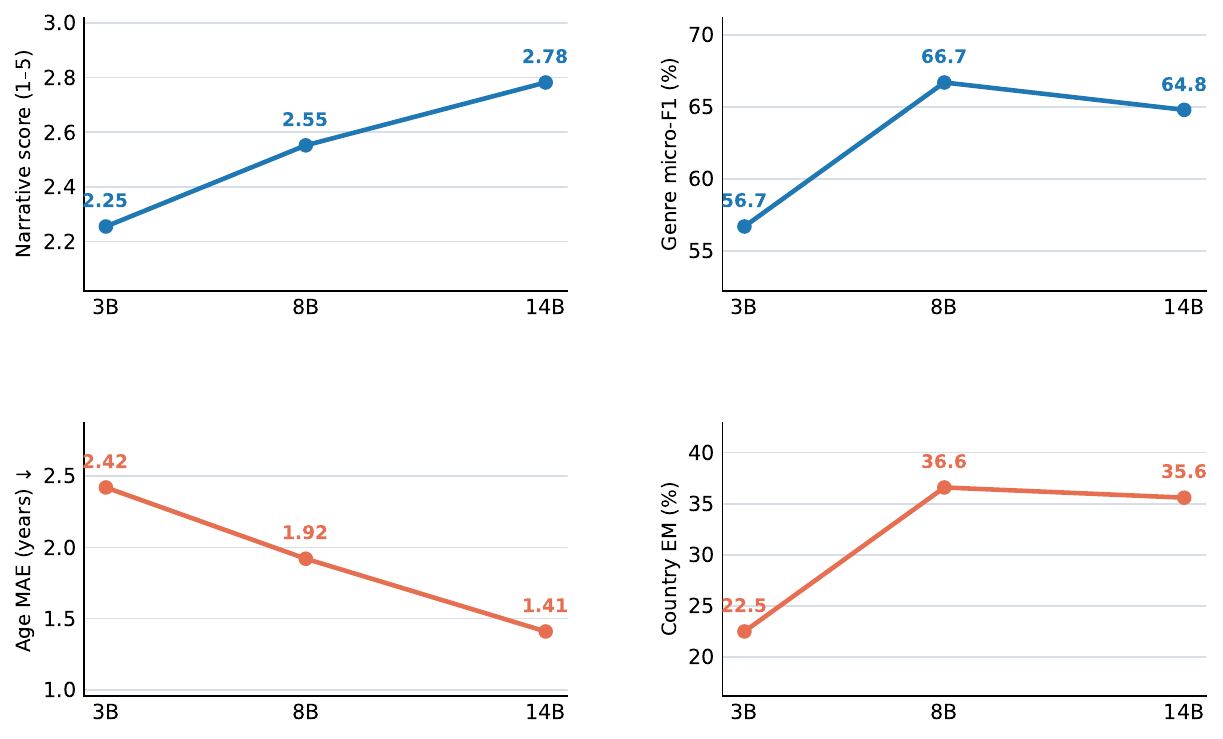}
\caption{English-only Ministral scaling in original metric units. The panels report narrative overall, genre micro-F1, age MAE (lower is better), and national-rating exact match.}
\label{fig:ministral-scaling}
\end{figure}

\section{Qualitative Case Studies}
\label{app:qualitative-cases}
These cases illustrate distinct observable failure mechanisms and one within-film language effect; they are not estimates of population frequency. Text is drawn from saved outputs, and annotations compare it with benchmark gold fields or subtitle-index evidence. Film titles link to IMDb.
\begin{figure}[p]
\centering
\begin{tcolorbox}[
  colback=white,
  colframe=cineNarrative,
  colbacktitle=cineNarrative,
  coltitle=white,
  title={\href{https://www.imdb.com/title/tt1826590/}{\textcolor{white}{\textit{About Last Night}}} (2014) \quad 1h 40m \quad English subtitles},
  fonttitle=\bfseries,
  boxrule=0.65pt,
  arc=1mm,
  left=2mm,right=2mm,top=1mm,bottom=1mm
]
\small
\textbf{Plot---gold.} ``Follow two couples as they journey from the bar to the bedroom and are eventually put to the test in the real world.''

\medskip
\textbf{GPT-5.6 Sol---plot excerpt.} ``After cautious strangers Danny and Debbie meet through their uninhibited friends, a spontaneous hookup grows into love, cohabitation, and a shared dog.''

\textbf{Llama 4 Scout---plot.} ``The film follows Danny, a recently broken-up man, and his friend Bernie, a carefree and promiscuous womanizer, as they navigate their lives, relationships, and personal growth in \caseInvent{Chicago}.''

\medskip
\textbf{Synopsis---gold evidence anchors.} ``Keller is ready to let Danny go, but Danny makes the first move and quits.'' Later, ``Bernie and Joan have rekindled their relationship, even though it is mostly sexual.''

\textbf{Gemini 3.8 Flash---synopsis excerpts.} ``Danny despises his corporate restaurant supply job, eventually \caseInvent{getting fired after defying company policy} to supply a struggling neighborhood bar owned by \caseInvent{his late father's friend}, Casey.'' Later, ``Bernie and Joan reconcile by accepting their unorthodox, fiercely combative dynamic and \caseResolution{move in together}.''

\medskip
\textbf{Genre labels.} Gold: Comedy, Romance. Gemini 3.8 Flash: Comedy, Romance, \caseInvent{Drama}.
\end{tcolorbox}
\caption{A film-level case showing why fluent output is not necessarily faithful. Red marks unsupported location, event, relationship, or genre content; bronze marks an altered relationship outcome. The synopsis lines are verbatim excerpts, not complete synopses; the omitted passages are not scored here. All quotations and labels come from saved English-input records.}
\label{fig:narrative-case}
\end{figure}

\begin{figure}[p]
\centering
\begin{tcolorbox}[
  colback=white,
  colframe=cineCultural,
  colbacktitle=cineCultural,
  coltitle=white,
  title={\href{https://www.imdb.com/title/tt0096754/}{\textcolor{white}{\textit{The Abyss}}} (1989) \quad 2h 20m \quad English subtitles},
  fonttitle=\bfseries,
  boxrule=0.65pt,
  arc=1mm,
  left=2mm,right=2mm,top=1mm,bottom=1mm
]
\small
\textbf{Age suitability and national ratings.} The same subtitle input produces different national label predictions. Teal cells match the gold label; pale red cells do not.

\medskip
\centering
\renewcommand{\arraystretch}{1.17}
\setlength{\tabcolsep}{5pt}
\begin{tabular}{@{}lccc@{}}
\toprule
Target & Gold & GPT-5.6 Sol & GPT-5 Nano \\
\midrule
Age suitability & \textbf{13+} & \cellcolor{cineDeltaNegative!12}15+ & \cellcolor{cineDeltaNegative!17}17+ \\
\midrule
\caseFlag{AU}\, Australia & M & \cellcolor{cineTeal!12}M & \cellcolor{cineDeltaNegative!12}MA15+ \\
\caseFlag{BR}\, Brazil & Livre & \cellcolor{cineDeltaNegative!12}14 & \cellcolor{cineDeltaNegative!17}16 \\
\caseFlag{FR}\, France & Tous publics & \cellcolor{cineDeltaNegative!12}12 & \cellcolor{cineDeltaNegative!17}16 \\
\caseFlag{DE}\, Germany & 12 & \cellcolor{cineTeal!12}12 & \cellcolor{cineDeltaNegative!12}16 \\
\caseFlag{SG}\, Singapore & PG13 & \cellcolor{cineTeal!12}PG13 & \cellcolor{cineDeltaNegative!12}M18 \\
\caseFlag{GB}\, United Kingdom & 15 & \cellcolor{cineTeal!12}15 & \cellcolor{cineTeal!12}15 \\
\caseFlag{US}\, United States & PG-13 & \cellcolor{cineTeal!12}PG-13 & \cellcolor{cineDeltaNegative!12}R \\
\bottomrule
\end{tabular}

\medskip
\raggedright
\textbf{GPT-5 Nano---age rationale excerpt.} ``The content is mature and non-graphic, supporting an \caseInvent{adult-only level}.'' The gold age label is 13+, illustrating a conservative shift in the model's predicted threshold.
\end{tcolorbox}
\caption{An English-input cultural case with seven of the ten country labels shown for legibility. The gold and predicted labels are the exact saved outputs, not translations into a common age scale; red denotes a mismatch within that country's own label space. Flags are rendered at one consistent display size. This case was chosen to illustrate a broad upward shift in age and national classifications, not to estimate its prevalence.}
\label{fig:qualitative-cultural-case}
\end{figure}

\begin{figure}[p]
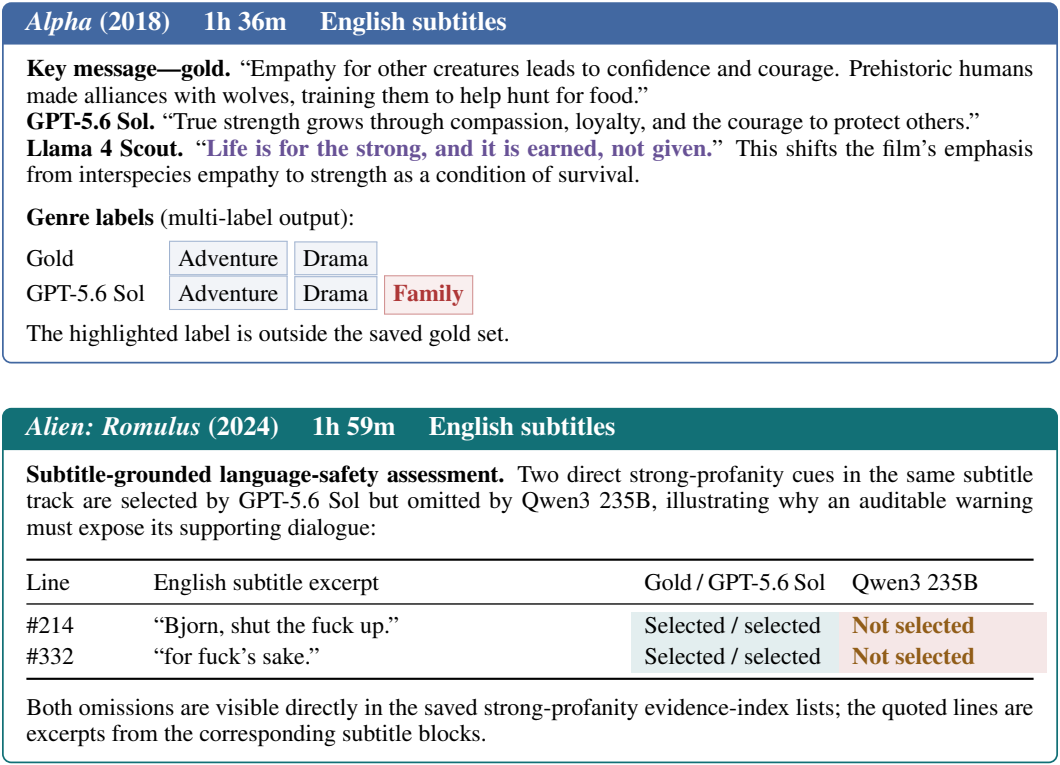

\centering
\begin{tcolorbox}[
  colback=white,colframe=cineNarrative,colbacktitle=cineNarrative,coltitle=white,
  title={\href{https://www.imdb.com/title/tt4244998/}{\textcolor{white}{\textit{Alpha}}} (2018) \quad 1h 36m \quad English subtitles},
  fonttitle=\bfseries,boxrule=0.65pt,arc=1mm,
  left=2mm,right=2mm,top=1mm,bottom=1mm
]
\small
\textbf{Key message---gold.} ``Empathy for other creatures leads to confidence and courage. Prehistoric humans made alliances with wolves, training them to help hunt for food.''

\textbf{GPT-5.6 Sol.} ``True strength grows through compassion, loyalty, and the courage to protect others.''

\textbf{Llama 4 Scout.} ``\caseTheme{Life is for the strong, and it is earned, not given.}'' This shifts the film's emphasis from interspecies empathy to strength as a condition of survival.

\medskip
\textbf{Genre labels} (multi-label output):\par\smallskip
\begin{tabular}{@{}l@{\quad}l@{}}
Gold & \fcolorbox{cineNarrative!50}{cineNarrative!7}{Adventure}\;\fcolorbox{cineNarrative!50}{cineNarrative!7}{Drama} \\
GPT-5.6 Sol & \fcolorbox{cineNarrative!50}{cineNarrative!7}{Adventure}\;\fcolorbox{cineNarrative!50}{cineNarrative!7}{Drama}\;\fcolorbox{cineDeltaNegative!55}{cineDeltaNegative!8}{\caseInvent{Family}} \\
\end{tabular}\par\smallskip
The highlighted label is outside the saved gold set.
\end{tcolorbox}

\vspace{2mm}
\begin{tcolorbox}[
  colback=white,colframe=cineTeal,colbacktitle=cineTeal,coltitle=white,
  title={\href{https://www.imdb.com/title/tt18412256/}{\textcolor{white}{\textit{Alien: Romulus}}} (2024) \quad 1h 59m \quad English subtitles},
  fonttitle=\bfseries,boxrule=0.65pt,arc=1mm,
  left=2mm,right=2mm,top=1mm,bottom=1mm
]
\small
\textbf{Subtitle-grounded language-safety assessment.} Two direct strong-profanity cues in the same subtitle track are selected by GPT-5.6 Sol but omitted by Qwen3 235B, illustrating why an auditable warning must expose its supporting dialogue:

\medskip
\renewcommand{\arraystretch}{1.15}
\setlength{\tabcolsep}{5pt}
\begin{tabularx}{\linewidth}{@{}p{0.1\linewidth}X p{0.18\linewidth}p{0.18\linewidth}@{}}
\toprule
Line & English subtitle excerpt & Gold / GPT-5.6 Sol & Qwen3 235B \\
\midrule
\#214 & ``Bjorn, shut the fuck up.'' & \cellcolor{cineTeal!12}Selected / selected & \cellcolor{cineDeltaNegative!12}\caseResolution{Not selected} \\
\#332 & ``for fuck's sake.'' & \cellcolor{cineTeal!12}Selected / selected & \cellcolor{cineDeltaNegative!12}\caseResolution{Not selected} \\
\bottomrule
\end{tabularx}

\medskip
Both omissions are visible directly in the saved strong-profanity evidence-index lists; the quoted lines are excerpts from the corresponding subtitle blocks.
\end{tcolorbox}
\caption{Two complementary English-input cases. Violet marks a thematic reversal, red a predicted genre outside the saved gold set, and bronze a missed explicit evidence line. Gold labels, subtitle text, and model outputs are taken from saved sample-level records. The examples illustrate mechanisms; they are not estimates of error frequency.}
\label{fig:qualitative-language-case}
\end{figure}

\begin{figure}[p]
\centering
\begin{tcolorbox}[
  colback=white,colframe=cineCultural,colbacktitle=cineCultural,coltitle=white,
  title={\href{https://www.imdb.com/title/tt1560747/}{\textcolor{white}{\textit{The Master}}} (2012) \quad 2h 18m \quad GPT-5.6 Sol},
  fonttitle=\bfseries,boxrule=0.65pt,arc=1mm,
  left=2mm,right=2mm,top=1mm,bottom=1mm
]
\small
\textbf{One film, one model, two subtitle languages.} Gold labels stay fixed while the input changes from English (en) to Vietnamese (vi). Teal marks exact matches; pale red marks mismatches. In this case, Vietnamese input changes four country predictions from mismatches to matches and corrects the age prediction.

\medskip
\centering
\renewcommand{\arraystretch}{1.12}
\setlength{\tabcolsep}{7pt}
\begin{tabular}{@{}lccc@{}}
\toprule
Target & Gold & English (en) & Vietnamese (vi) \\
\midrule
Age suitability & 17+ & \cellcolor{cineDeltaNegative!12}18+ & \cellcolor{cineTeal!12}17+ \\
\midrule
\caseFlag{AU}\, Australia & MA15+ & \cellcolor{cineDeltaNegative!12}R18+ & \cellcolor{cineTeal!12}MA15+ \\
\caseFlag{BR}\, Brazil & 14 & \cellcolor{cineDeltaNegative!12}18 & \cellcolor{cineDeltaNegative!12}16 \\
\caseFlag{FR}\, France & Tous publics & \cellcolor{cineDeltaNegative!12}16 & \cellcolor{cineDeltaNegative!12}16 \\
\caseFlag{DE}\, Germany & 12 & \cellcolor{cineDeltaNegative!12}16 & \cellcolor{cineDeltaNegative!12}16 \\
\caseFlag{NL}\, Netherlands & 12 & \cellcolor{cineDeltaNegative!12}16 & \cellcolor{cineDeltaNegative!12}16 \\
\caseFlag{SG}\, Singapore & M18 & \cellcolor{cineDeltaNegative!12}R21 & \cellcolor{cineTeal!12}M18 \\
\caseFlag{KR}\, South Korea & 19 & \cellcolor{cineTeal!12}19 & \cellcolor{cineTeal!12}19 \\
\caseFlag{SE}\, Sweden & 15 & \cellcolor{cineTeal!12}15 & \cellcolor{cineTeal!12}15 \\
\caseFlag{GB}\, United Kingdom & 15 & \cellcolor{cineDeltaNegative!12}18 & \cellcolor{cineTeal!12}15 \\
\caseFlag{US}\, United States & R & \cellcolor{cineDeltaNegative!12}NC-17 & \cellcolor{cineTeal!12}R \\
\bottomrule
\end{tabular}

\medskip
\raggedright
\textbf{Country exact matches:} English 2/10; Vietnamese 6/10. The four persistent mismatches are retained in the table to show the full ten-country outcome, not only the changed labels.
\end{tcolorbox}
\caption{An illustrative within-film cross-lingual contrast from saved GPT-5.6 Sol outputs. Only the subtitle-input language changes; the gold age and national-rating labels are identical in both runs. This selected case makes a possible cultural benefit of multilingual input tangible, while the aggregate language comparison in the main paper supports the population-level analysis.}
\label{fig:qualitative-crosslingual-case}
\end{figure}

\begin{figure}[p]
\centering
\begin{tcolorbox}[
  colback=white,colframe=cineNarrative,colbacktitle=cineNarrative,coltitle=white,
  title={\href{https://www.imdb.com/title/tt2402927}{\textcolor{white}{\textit{Carol}}} (2015) \quad 1h 58m \quad English subtitles},
  fonttitle=\bfseries,boxrule=0.65pt,arc=1mm,
  left=2mm,right=2mm,top=1mm,bottom=1mm
]
\small
\textbf{Plot---gold.} ``An aspiring photographer develops an intimate relationship with an older woman in 1950s New York.''

\medskip
\textbf{Gemini 3.8 Flash---plot excerpt.} ``In 1950s New York, an aspiring young photographer working at a department store meets an elegant, sophisticated older woman navigating a painful divorce. A magnetic connection develops between them'' [remainder omitted].

\medskip
\textbf{Llama 4 Scout---complete plot.} ``A young woman, Therese, meets \caseInvent{an older married man, Gray, in a taxi}. They strike up a conversation, and Therese is drawn to Gray's world. As they navigate their complicated relationships and desires, Therese becomes embroiled in \caseInvent{Gray's marriage to Carol}, a woman with a troubled past.''

\medskip
\textbf{What changed?} The reference synopsis begins with Therese meeting \emph{Carol} in a department store; Carol is the older woman and Therese's romantic counterpart. The highlighted taxi meeting and marriage invert that central relationship. The narrative evaluator assigns Llama's plot \textbf{1/5} for this sample.
\end{tcolorbox}
\caption{A visually fluent but wrong-premise plot, contrasted with a model that preserves the central relationship. Red spans are exact phrases from Llama's saved output; Gemini's bracketed omission is editorial and not part of its output. The counterexample demonstrates a failure mode, not its prevalence.}
\label{fig:qualitative-plot-case}
\end{figure}

\section{Limitations, Intended Use, and Data Statement}
\label{app:limitations}

\subsection{Limitations}

\paragraph{Subtitle variation and cross-lingual interpretation.}
Subtitle tracks vary across releases and translations. We address this through
structural, temporal, density, language-consistency, and manual checks, and
evaluate all six languages over the same matched films. The resulting
cross-lingual comparisons capture model behavior under each subtitle-language
condition, encompassing translation, tokenization, subtitle density, and
model-language proficiency.

\paragraph{Coverage and benchmark scope.}
\cinesubbench{} prioritizes complete matched coverage across six subtitle
languages and ten national classification systems, yielding 1,012 films for
controlled MultiX comparison. This design supports direct comparison across
tasks, languages, and cultural settings rather than population-level
characterization of global cinema.

\paragraph{Reference labels and normalization.}
Age recommendations and national ratings retain the conventions of their
respective sources and institutions. National systems are preserved as
separate ordered label spaces, with historical or variant labels normalized
only where needed for consistent evaluation. Narrative references provide a
common semantic anchor while permitting valid variation in wording and
emphasis.

\paragraph{Language-safety scope.}
The language-safety task provides auditable lexical-evidence assessment over
four categories in the English subtitle track, enabling line-level evaluation
of both evidence recovery and category assignment. Extending the evidence
taxonomy and multilingual annotations offers a natural direction for future
benchmark development.

\paragraph{Source exposure and prior film knowledge.}
Public film information can appear in model pretraining data. We isolate this
factor by withholding titles, release years, cast, and reference answers from
standard \cinesubbench{} inputs. Appendix~\ref{app:source-exposure-audit}
separately measures the signal available from title/year cues and sensitivity
to character-name perturbations, distinguishing prior film information from
subtitle-conditioned generation.

\paragraph{Modality scope.}
\cinesubbench{} intentionally uses subtitles as a common film-length linguistic
interface that supports consistent evaluation across API-based and open-weight
LLMs. It measures narrative and cultural understanding from temporally ordered
subtitle evidence, while visual action, cinematography, music, gestures, and
other non-speech signals form a complementary multimodal evaluation setting.

\subsection{Intended Use}
\label{app:intended-use}

\cinesubbench{} is intended for research on long-form film understanding,
multilingual subtitle processing, culturally situated audience assessment, and
auditable language-safety evaluation. Its timestamped multilingual tracks can
also support research on subtitle alignment, cross-lingual retrieval, temporal
reasoning, summarization, and accessibility-oriented subtitle systems.
Country-specific ratings should remain associated with their originating
national systems, while language-safety evidence is intended for transparent,
human-reviewable assessment rather than automated censorship.

\subsection{Data Statement}
\label{app:data-statement}

\cinesubbench{} contains structured records for 1,012 films, including source
metadata, normalized task labels, six timestamped subtitle tracks per film,
subtitle-indexed language-safety evidence, and benchmark task definitions.
Source links and provenance fields are retained for traceability, with
construction, verification, sanitization, and field provenance documented in
Appendix~\ref{app:dataset-docs}. Release and redistribution follow the
requirements of the respective source providers.


\end{document}